\pdfoutput=1

\PassOptionsToPackage{capitalise}{cleveref}
\documentclass[
	11pt,           %
	english,        %
	singlespacing,  %
	liststotoc,     %
	headsepline,    %
]{MastersDoctoralThesis}

\makeatletter
\providecommand{\abx@aux@refcontext}[2]{}
\providecommand{\abx@aux@cite}[2]{}
\providecommand{\abx@aux@segm}[3]{}
\providecommand{\abx@aux@page}[2]{}
\providecommand{\abx@aux@defaultrefcontext}[2]{}

\thesistitle{The Geometry of Meaning in Neural Manifolds}
\supervisor{Prof. Emanuele \textsc{Rodol\`a}}
\examiner{}
\degree{Doctor of Philosophy}
\author{Valentino \textsc{Maiorca}}
\addresses{}

\subject{Computer Science}
\keywords{representation learning, latent spaces, neural networks, manifold hypothesis}

\university{\href{https://www.uniroma1.it/}{Sapienza University of Rome}}
\department{\href{https://phd.uniroma1.it/web/COMPUTER-SCIENCE_nD3507_EN.aspx}{Department of Computer Science}}
\faculty{\href{https://i3s.web.uniroma1.it/en}{Faculty of Information Engineering, Informatics, and Statistics}}

\AtBeginDocument{
    \hypersetup{pdftitle=\ttitle}
    \hypersetup{pdfauthor=\authorname}
    \hypersetup{pdfkeywords=\keywordnames}
    \hypersetup{pdfsubject=\subjectname}
}

\usepackage{mathpazo} %
\usepackage[utf8]{inputenc} %
\usepackage[T1]{fontenc} %

\usepackage[autostyle=true]{csquotes} %

\usepackage[bottom]{footmisc} %
\usepackage{soul} %
\usepackage{wrapfig} %
\usepackage{microtype} %
\usepackage{enumitem} %
\usepackage{tcolorbox} %
\tcbuselibrary{skins} %
\usepackage{xspace} %

\usepackage{booktabs} %
\usepackage{nicefrac} %
\usepackage{colortbl} %
\usepackage{multirow} %

\usepackage{algorithm}
\usepackage{algpseudocode}
\providecommand{\theHALG@line}{}
\renewcommand{\theHALG@line}{\thealgorithm.\arabic{ALG@line}}

\usepackage{amsfonts} %
\usepackage{amsmath}
\usepackage{amssymb}
\usepackage{mathtools}
\usepackage{amsthm}
\usepackage{bm} %
\usepackage{bbm} %

\usepackage{graphicx}
\usepackage{grffile} %
\usepackage{xcolor}
\usepackage{overpic}
\usepackage{pgf}
\usepackage{pgfplots}
\usepackage{tikz-3dplot}
\pgfplotsset{compat=1.18} %
\usepackage{subcaption} %
\usepackage{placeins} %

\usepackage[
  backend=biber,
  style=authoryear,
  sorting=ynt,
  sortcites=true,
  natbib=true,        %
  maxbibnames=99,
  maxcitenames=1,     %
  mincitenames=1,     %
  giveninits=false,
  uniquename=init,
  uniquelist=false,   %
  url=true,
]{biblatex}
\appto{\bibsetup}{\sloppy}

\DeclareDelimFormat[cite,parencite,textcite]{multinamedelim}{\addspace\&\addspace}
\DeclareDelimFormat[cite,parencite,textcite]{finalnamedelim}{\addspace\&\addspace}

\DeclareDelimFormat[bib,biblist]{multinamedelim}{\addcomma\space}
\DeclareDelimFormat[bib,biblist]{finalnamedelim}{\addcomma\space}

\DeclareSourcemap{
  \maps[datatype=bibtex]{
    \map{
      \step[fieldsource=author, match=\regexp{(.+)\*}, replace=\regexp{$1\textsuperscript{*}}]
    }
  }
}

\DeclareCiteCommand{\fullcite}
  {\usebibmacro{prenote}}
  {\usedriver
     {\defcounter{minnames}{6}%
      \defcounter{maxnames}{8}%
      \clearfield{year}%
      \clearfield{month}%
      \clearfield{day}%
      \clearfield{endyear}%
      \clearfield{endmonth}%
      \clearfield{endday}%
      \clearfield{labelyear}%
      \clearfield{labelmonth}%
      \clearfield{labelday}%
      \clearfield{labelendyear}%
      \clearfield{labelendmonth}%
      \clearfield{labelendday}%
      \clearlist{location}%
      \clearlist{publisher}%
      \clearfield{note}}
     {cite:\thefield{entrytype}}}
  {\multicitedelim}
  {\usebibmacro{postnote}}

\DeclareBibliographyDriver{cite:article}{%
  \usebibmacro{bibindex}%
  \usebibmacro{begentry}%
  \DeclareNameAlias{author}{default}%
  \printnames{author}%
  \setunit{\addperiod\space}%
  \newunit\newblock
  \printfield{title}%
  \usebibmacro{finentry}}

\DeclareBibliographyDriver{cite:inproceedings}{%
  \usebibmacro{bibindex}%
  \usebibmacro{begentry}%
  \DeclareNameAlias{author}{default}%
  \printnames{author}%
  \setunit{\addperiod\space}%
  \newunit\newblock
  \printfield{title}%
  \usebibmacro{finentry}}

\DeclareBibliographyDriver{cite:incollection}{%
  \usebibmacro{bibindex}%
  \usebibmacro{begentry}%
  \DeclareNameAlias{author}{default}%
  \printnames{author}%
  \setunit{\addperiod\space}%
  \newunit\newblock
  \printfield{title}%
  \usebibmacro{finentry}}

\DeclareBibliographyDriver{cite:misc}{%
  \usebibmacro{bibindex}%
  \usebibmacro{begentry}%
  \DeclareNameAlias{author}{default}%
  \printnames{author}%
  \setunit{\addperiod\space}%
  \newunit\newblock
  \printfield{title}%
  \usebibmacro{finentry}}

\DeclareDelimcontextAlias{pubboxcite}{bib}
\DeclareCiteCommand{\pubboxcite}
  {}
  {\defcounter{maxnames}{99}\defcounter{minnames}{99}%
   {\bfseries\printfield[noformat]{title}}\par\vspace{0.35em}%
   \printnames[default]{author}}
  {}
  {}

\definecolor{sapienzared}{RGB}{130,36,51}
\definecolor{highlight}{HTML}{D2691E}
\colorlet{subtitlecolor}{darkgray}

\definecolor{pltorange}{rgb}{1.0, 0.49803922, 0.05490196}
\definecolor{pltblue}{rgb}{0.12156863, 0.46666667, 0.70588235}

\definecolor{colorxstuff}{HTML}{7B113A}
\definecolor{colorystuff}{HTML}{4F6F52}

\definecolor{ar_i}{rgb}{0.8, 0.2, 0.2}  %
\definecolor{ar_t}{rgb}{0.2, 0.2, 0.8}  %
\definecolor{rr}{rgb}{0.2, 0.8, 0.2}    %
\definecolor{cellblue}{RGB}{200, 220, 240}  %

\definecolor{lightblue}{RGB}{173, 216, 230}    %
\definecolor{lightyellow}{RGB}{255, 255, 224}  %
\definecolor{rdbrown}{HTML}{846358} \definecolor{rdpurple}{HTML}{6741d9} \definecolor{rdorange}{HTML}{f08c00} \definecolor{rdgreen}{HTML}{2f9e44}

\providecommand{\todo}[1]{}
\renewcommand{\todo}[1]{\textcolor{red}{\textbf{[TODO: #1]}}}

\usepackage{amsmath,amsfonts,bm}

\def\1{\bm{1}}

\def\vmu{{\bm{\mu}}}

\def\vxi{{\bm{\xi}}}
\def\veta{{\bm{\eta}}}
\def\va{{\bm{a}}}
\def\vb{{\bm{b}}}
\def\vc{{\bm{c}}}

\def\vh{{\bm{h}}}

\def\vr{{\bm{r}}}
\def\vs{{\bm{s}}}

\def\vu{{\bm{u}}}
\def\vv{{\bm{v}}}

\def\vx{{\bm{x}}}
\def\vy{{\bm{y}}}
\def\vz{{\bm{z}}}

\def\mA{{\bm{A}}}

\def\mC{{\bm{C}}}
\def\mD{{\bm{D}}}
\def\mE{{\bm{E}}}
\def\mF{{\bm{F}}}

\def\mH{{\bm{H}}}
\def\mI{{\bm{I}}}

\def\mL{{\bm{L}}}
\def\mM{{\bm{M}}}
\def\mN{{\bm{N}}}

\def\mP{{\bm{P}}}

\def\mR{{\bm{R}}}
\def\mS{{\bm{S}}}
\def\mT{{\bm{T}}}
\def\mU{{\bm{U}}}
\def\mV{{\bm{V}}}
\def\mW{{\bm{W}}}
\def\mX{{\bm{X}}}
\def\mY{{\bm{Y}}}

\def\mPhi{{\bm{\Phi}}}
\def\mLambda{{\bm{\Lambda}}}

\DeclareMathAlphabet{\mathsfit}{\encodingdefault}{\sfdefault}{m}{sl}
\SetMathAlphabet{\mathsfit}{bold}{\encodingdefault}{\sfdefault}{bx}{n}

\newcommand{\E}{\mathbb{E}}

\newcommand{\R}{\mathbb{R}}

\DeclareMathOperator{\softmax}{softmax}

\DeclareMathOperator*{\argmax}{arg\,max}
\DeclareMathOperator*{\argmin}{arg\,min}

\DeclareMathOperator{\Tr}{Tr}

\newcommand{\STAB}[1]{\begin{tabular}{@{}c@{}}#1\end{tabular}}

\algrenewcommand{\Return}{\State\algorithmicreturn~}

\newcommand{\chapterpublication}[4][]{%
  \begin{publicationinfo}
  {\small\scshape This chapter is based on the following publication:}

  \vspace{0.5em}
  {\raggedright\pubboxcite{#2}\par}

  \vspace{0.25em}
  #3%
  \ifx&#1&%
  \else
    \par\vspace{0.4em}
    \noindent{\small\textbf{Code:} \url{#1}}
  \fi
  \ifx&#4&%
  \else
    \par\vspace{0.5em}
    \noindent{\small\textbf{TL;DR:} \textit{#4}}
  \fi
  \end{publicationinfo}%
}

\newcommand{\sectionpublication}[4][]{%
  \begin{publicationinfo}
  {\small\scshape This section is based on the following publication:}

  \vspace{0.5em}
  {\raggedright\pubboxcite{#2}\par}

  \vspace{0.25em}
  #3%
  \ifx&#1&%
  \else
    \par\vspace{0.4em}
    \noindent{\small\textbf{Code:} \url{#1}}
  \fi
  \ifx&#4&%
  \else
    \par\vspace{0.5em}
    \noindent{\small\textbf{TL;DR:} \textit{#4}}
  \fi
  \end{publicationinfo}%
}

\newcommand{\Corr}{\ensuremath{\Gamma}}                    %

\newcommand{\OperatorParallel}{\ensuremath{\Pi}}           %
\newcommand{\OperatorCoupled}{\ensuremath{T}}              %
\newcommand{\OperatorSameSpace}{\ensuremath{\Sigma}}       %

\newcommand{\hlintrinsic}[1]{\colorbox{blue!5}{#1}}                %
\newcommand{\hlextrinsic}[1]{\colorbox{red!5}{#1}}                 %

\newcommand{\fmnist}{F-MNIST}                                      %
\newcommand{\news}{N24News}                                        %
\newcommand{\ag}{AG News}                                          %
\newcommand{\imdb}{IMDB}                                           %

\newcommand{\ie}{\textit{i.e.}\xspace}
\newcommand{\eg}{\textit{e.g.}\xspace}

\DeclareMathOperator*{\sort}{sort}

\newcommand{\red}[1]{\textcolor{red}{#1}}
\newcommand{\RT}[2][]{\textcolor{red}{T_{#2}^{#1}}}
\newcommand{\Bf}[2][]{\textcolor{blue}{f_{#2}^{#1}}}
\newcommand{\ROmega}[2][]{\textcolor{red}{\vomega_{#2}^{#1}}}
\newcommand{\Rb}[2][]{\textcolor{red}{\vb_{#2}^{#1}}}
\newcommand{\allw}{\red{\vomega}}   %
\newcommand{\allb}{\red{\vb}}       %

\newcommand{\iti}{{ITI}\xspace}
\newcommand{\linearact}{Lin-\textsc{AcT}\xspace}
\newcommand{\caa}{{CAA}\xspace}
\newcommand{\reft}{{ReFT}\xspace}
\newcommand{\prompt}{{Prompt}\xspace}
\newcommand{\lorartc}{LRTC-RL\xspace}
\newcommand{\lofitrl}{LoFIT-RL\xspace}
\newcommand{\RTC}{{RTC}\xspace}
\newcommand{\toxclsrtp}{Tox$_{\mathrm{RTC}}^{\mathrm{RTP}}$\xspace}
\newcommand{\toxclstet}{Tox$_{\mathrm{RTC}}^{\mathrm{TET}}$\xspace}
\newcommand{\pplwik}{PPL$_{\mathrm{WIK}}$\xspace}
\newcommand{\imgscore}{IMGScore\xspace}
\newcommand{\clipscore}{CLIPScore\xspace}

\newcommand{\qwensevenb}{Qwen2.5-7B\xspace}
\newcommand{\qwenonefiveb}{Qwen2.5-1.5B\xspace}
\newcommand{\gemmatwob}{Gemma2-2B\xspace}

\newcommand{\Obs}[1][]{\if\relax\detokenize{#1}\relax\mathcal{O}\else\mathcal{O}_{\mathcal{#1}}\fi}     %
\newcommand{\obs}{o}                                                               %
\newcommand{\Data}{\mathcal{D}}                                                    %
\newcommand{\dimof}[1]{d_{\mathcal{#1}}}                                           %
\newcommand{\Enc}[1][]{\if\relax\detokenize{#1}\relax E\else E_{\mathcal{#1}}\fi}                         %
\newcommand{\Dec}[1][]{\if\relax\detokenize{#1}\relax D\else D_{\mathcal{#1}}\fi}                         %
\newcommand{\anchors}[1][]{\if\relax\detokenize{#1}\relax\mathcal{A}\else\mathcal{A}_{\mathcal{#1}}\fi}  %
\newcommand{\anchmat}[1][]{\if\relax\detokenize{#1}\relax\mA\else\mA_{\mathcal{#1}}\fi}                  %
\newcommand{\nanch}{K}                                                             %
\DeclareMathOperator{\simf}{sim}                                                   %
\DeclareMathOperator{\relproj}{rel}                                                %
\newcommand{\CorrSide}[1]{\Corr_{\mathcal{#1}}}                                    %
\newcommand{\0}{\bm{0}}

\newcommand{\vlambda}{{\bm{\lambda}}}
\newcommand{\vomega}{{\bm{\omega}}}

\begin{document}

\frontmatter

\begin{titlepage}
	\begin{center}

		\vspace*{.06\textheight}
		{\scshape\LARGE \univname\par}\vspace{1.5cm}
		\textsc{\Large Doctoral Thesis}\\[0.5cm]

		\HRule \\[0.4cm]
		{\huge \bfseries \ttitle\par}\vspace{0.4cm}
		\HRule \\[1.5cm]

		\begin{minipage}[t]{0.4\textwidth}
			\begin{flushleft} \large
				\emph{Author:}\\
				\authorname\\
			\end{flushleft}
		\end{minipage}
		\begin{minipage}[t]{0.4\textwidth}
			\begin{flushright} \large
				\emph{Advisor:} \\
				\supname \\[0.5cm]
				\emph{ELLIS Co-Advisor:} \\
				Prof. Francesco \textsc{Locatello} \\
				\textit{Institute of Science and Technology Austria (ISTA)}
			\end{flushright}
		\end{minipage}\\[3cm]

		\vfill

		\large \textit{A thesis submitted in fulfillment of the requirements\\
		for the degree of \degreename}\\[0.3cm]
		\textit{in the}\\[0.4cm]
		\deptname\\
		\facname\\
		\textit{Sapienza University of Rome}\\[0.4cm]
		\textit{and carried out as part of the ELLIS Ph.D. Program}\\[2cm]

		\vfill

		{\large 31 October 2025}\\[4cm]

		\vfill
	\end{center}
\end{titlepage}

\thispagestyle{empty}
\null\vspace*{1cm}
\begin{flushleft}
	\large
	Thesis defended on 26 January 2026\\
	in front of a Board of Examiners composed of:\\[3mm]
	\begin{tabular}{@{}l}
		Prof.\ Barbara \textsc{Re}\\
		{\small\textit{University of Camerino}}\\[1.5mm]
		Prof.\ Chiara \textsc{Di Francescomarino}\\
		{\small\textit{University of Trento}}\\[1.5mm]
		Prof.\ Han \textsc{van der Aa}\\
		{\small\textit{University of Vienna}}
	\end{tabular}

	\vspace{1.2cm}

	External reviewers:\\[3mm]
	\begin{tabular}{@{}l}
		Prof.\ Marco \textsc{Baroni}\\
		{\small\textit{ICREA and Universitat Pompeu Fabra, Barcelona}}\\[1.5mm]
		Prof.\ Alessandro \textsc{Laio}\\
		{\small\textit{SISSA -- International School for Advanced Studies, Trieste}}
	\end{tabular}
\end{flushleft}
\vfill
\clearpage

\thispagestyle{plain}
\begin{center}
	{\Large \textbf{Abstract}}
\end{center}
\addchaptertocentry{Abstract}

\noindent \textit{Just as different languages express the same concept through different words, neural networks express similar functions through incompatible coordinate systems.} Different architectures, random initializations, and optimization trajectories produce incompatible parameterizations in \textbf{extrinsic} coordinates, obscuring functional relationships between independently trained models. The question is: when models independently solve similar tasks on related data, do their learned representations share exploitable structure despite incompatible coordinates? \textbf{This thesis shows that semantic correspondence is a fundamental primitive revealing shared geometric structure beneath incompatible parameterizations.}
By identifying which elements refer to the same semantic content, correspondence breaks the symmetry of arbitrary implementation choices, anchoring incompatible systems to a shared semantic frame and revealing the underlying \textbf{intrinsic} geometry that ultimately enables \textbf{zero-shot} transformation between independently trained models.

Building on this principle, we walk through fourteen scientific contributions exploring correspondence-based alignment across diverse settings: zero-shot model stitching, cross-modal translation, policy transfer, activation steering, and neural decoding. These methods span domains (vision, language, reinforcement learning, neuroscience), architectures (CNNs, Transformers, graph networks), and scales (from small supervised models to billion-parameter generative systems). Each exploits different geometric properties---angular, metric, spectral, sparse, or distributional---yet all instantiate the same underlying principle.
\textbf{Remarkably, correspondence-based alignment extends beyond artificial systems}: fMRI data show that different human brains processing identical stimuli exhibit alignable geometric structure, suggesting a general principle of learning systems.

From a practical standpoint, correspondence-based alignment enables model reuse without expensive joint training, knowledge transfer between independently developed systems, efficient transformation by exploiting low-dimensional semantic structure, and direct evaluation of representations in latent space without requiring downstream task deployment.
The methods we developed have achieved top-tier recognition (oral and spotlight presentations) in ML conferences and widespread adoption~\citep{moroni2025optimizing,wang2023understanding,wu2023zero,kiefer2023a}, contributing to weakly supervised pretraining~\citep{chen2023}, out-of-distribution detection~\citep{demirel2025ood}, federated learning~\citep{crisostomi2024charts}, and policy transfer in RL~\citep{jian2023policy}, and formed the conceptual foundation for the UniReps workshops at NeurIPS (2023--2025).

\textit{By revealing intrinsic geometric structure beneath extrinsic parameterization, semantic correspondence transforms model isolation into interoperability.}
\vspace{1em}

\cleardoublepage

\chapter*{Publications}
\addcontentsline{toc}{chapter}{Publications}

This thesis is based on the following publications:

$^{\ast}$\,Equal contribution.

\section*{Main Authored Publications}

\begin{refsection}
\emergencystretch=2em

\subsection*{Core Methods}

\begin{enumerate}
    \item \fullcite{moschella2023relative}

    \textbf{ICLR 2023} \textit{(\textcolor{blue}{Oral, notable-top-5\%})}

    \item \fullcite{maiorca2023semantic}

    \textbf{NeurIPS 2023}

    \item \fullcite{norelli2023asif}

    \textbf{NeurIPS 2023}

    \item \fullcite{cannistraci2024bricks}

    \textbf{ICLR 2024} \textit{(\textcolor{blue}{Spotlight})}
    \\[2pt] \textit{Preliminary version:} \fullcite{cannistraci2023infusing} --- \textbf{ICML TAG-ML Workshop 2023}

    \item \fullcite{fumero2024latent}

    \textbf{NeurIPS 2024}

    \item \fullcite{basile2025residual}

    \textbf{TMLR 2025}
\end{enumerate}

\subsection*{Extensions and Applications}

\begin{enumerate}[resume]
    \item \fullcite{cannistraci2023bootstrapping}

    \textbf{ICLR Tiny Papers Track 2023}

    \item \fullcite{vedula2024scalable}

    \textbf{ICML AI4Science Workshop 2024}

    \item \fullcite{maiorca2024inverse}

    \textbf{arXiv 2024}

    \item \fullcite{mager2024s}

    \textbf{NeurIPS UniReps Workshop 2024}

    \item \fullcite{ricciardi2023r3l}

    \textbf{EWRL 2023}
    \\[2pt] \textit{Extended version:} \fullcite{ricciardi2024latent} --- \textbf{arXiv 2025}

    \item \fullcite{maiorca2024multi}

    \textbf{Cosyne 2024}

    \item \fullcite{headpursuit}

    \textbf{NeurIPS 2025} \textit{(\textcolor{blue}{Spotlight})}

    \item \fullcite{lineas}

    \textbf{NeurIPS 2025}
\end{enumerate}

\end{refsection}

\section*{Other Authored Publications}

The following publications were authored or co-authored during the PhD but are not part of this thesis:

\begin{refsection}

\begin{enumerate}
    \item \fullcite{crisostomi2022metric}

    \textbf{LoG 2022}

    \item \fullcite{trappolini2023sparse}

    \textbf{IEEE Transactions on Artificial Intelligence 2023}

    \item \fullcite{ruscio2023attention}

    \textbf{ICLR Tiny Papers Track 2023}

    \item \fullcite{santilli2023accelerating}

    \textbf{ACL 2023}

    \item \fullcite{Setti2025.02.16.638500}

    \textbf{Nature Communications 2026}
\end{enumerate}

\end{refsection}

\cleardoublepage
\tableofcontents

\mainmatter

\makeatletter
\let\old@endpart\@endpart
\def\@endpart{\vspace{3cm}}
\makeatother

\part{Foundations: The Principle of Correspondence}

\begin{center}
\includegraphics[width=0.6\textwidth]{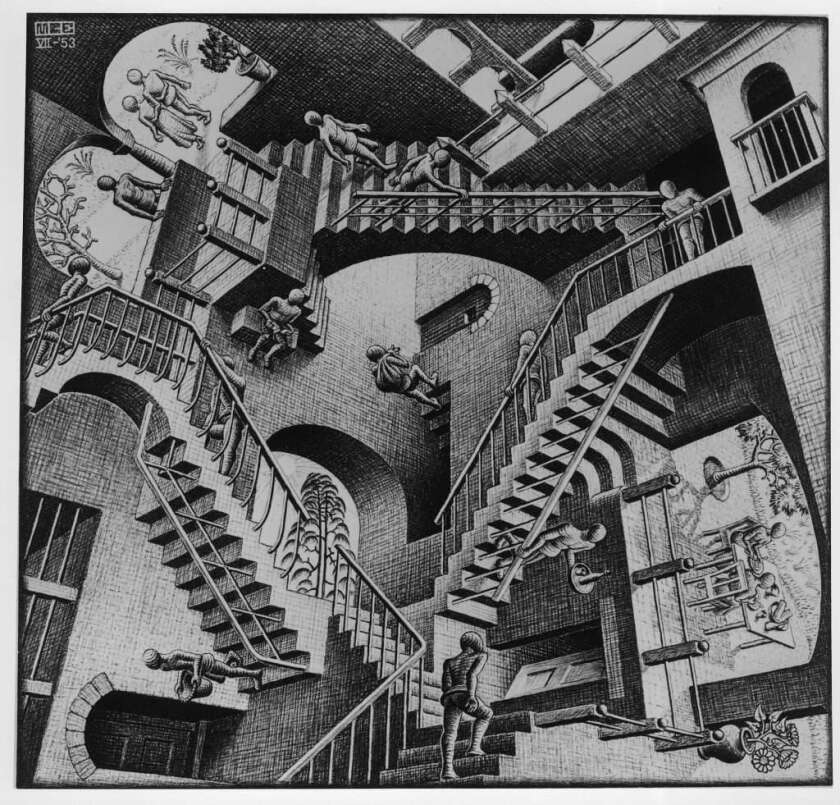}
\captionof*{figure}{\textbf{Relativity} by M.C. Escher (1953). Different reference frames, each internally consistent, describing the same underlying structure. Like the figures in Escher's lithograph who experience gravity from different orientations yet inhabit the same architectural space, neural networks may express similar semantic relationships through incompatible coordinate systems.}
\end{center}
\vfill
\begin{center}\footnotesize\textcopyright{} The M.C. Escher Company, The Netherlands. All rights reserved. Reproduced for academic, non-commercial use.\end{center}

\makeatletter
\let\@endpart\old@endpart
\makeatother

\clearpage

\chapter{Introduction}
\label{ch:introduction}

Consider two neural networks trained independently on the same image classification dataset. They use different architectures---one a \gls{cnn}, the other a transformer. They start from different random initializations. They follow different optimization trajectories through parameter space. When training completes, they achieve comparable accuracy, but their internal representations appear radically incompatible: different dimensionalities, different coordinate systems, different numerical scales. Model A encodes ``fluffy Persian cat'' at coordinates $[0.82, -0.34, 0.91, \dots]$ in 512 dimensions; Model B represents it at $[-0.47, 0.65, -0.23, \dots]$ in 2048 dimensions.

Here is the question that motivates this thesis: \textit{Are these representations fundamentally different, or merely different expressions of similar structure?}

The distinction matters. Different architectures impose different inductive biases. Different initialization seeds lead to different minima in the loss landscape. Different optimization paths traverse different regions of parameter space. Each choice---architecture, initialization, learning rate, batch size---constitutes an \gls{extrinsic-factor} that affects how the learned function gets parameterized. Yet both models face identical \glspl{intrinsic-factor}: they see the same data, they optimize the same objective, they must extract the same semantic distinctions to succeed. They must learn that cats cluster with dogs but separate from vehicles. They must preserve fine-grained distinctions between breeds. They must develop invariance to lighting and pose transformations.

Could it be that these intrinsic constraints---the semantic structure in data and task---induce similar geometric organization across models, despite the divergent extrinsic factors? That the incompatible coordinates merely obscure a shared underlying geometry?

\section{Motivation: Why Correspondence Exposes Intrinsic Geometry}

This thesis establishes that \gls{correspondence-concept} systematically reveals geometric structure shared across independently trained models, enabling zero-shot transformation between their representations.

When we identify semantically corresponding entities across representation spaces---shared examples, parallel observations, analogous concepts---this correspondence exposes which geometric relationships persist despite different parameterizations. The raw coordinates diverge arbitrarily: absolute positions, vector norms, axis orientations all depend on extrinsic initialization and architectural choices. But deeper geometric properties converge: angular relationships between semantically related concepts, spectral structure of representation manifolds, hierarchical organization of clusters, patterns of invariance to transformations. These intrinsic geometric properties reflect the semantic structure both models must extract to solve their tasks.

By systematically identifying which structures persist through correspondence, we can construct transformations that change coordinate systems while preserving semantic relationships. This enables practical transformation without retraining---we need not train models jointly, match their architectures, or share their parameters. We need only establish correspondence that reveals preserved geometric structure, then construct transformations respecting that structure. Like alchemists who sought to transform matter by understanding its fundamental nature, we transform representation forms by understanding their geometrical essence---but where alchemy relied on mystical principles, this work relies on correspondence that systematically reveals which geometric properties encode meaning and which merely reflect arbitrary parameterization choices.

\section{A Unified Principle Guiding Fourteen Methods}

This thesis demonstrates the correspondence principle through fourteen methods that exploit different geometric structures while instantiating the same underlying mechanism. We can understand these methods through the geometric properties they rely on, independent of their technical implementations:

\paragraph{Angular structure.} Several methods exploit angular relationships, which remain invariant to the rotations and scalings introduced by different parameterizations. The \gls{relrep} method~\citep[][\cref{ch:relative-representations}]{moschella2023relative}---presented with an oral at ICLR 2023---establishes the foundational \gls{correspondence-parallel} approach by anchoring representations to shared examples and measuring angular distances. This produces \glspl{semantic-coordinate} that enable direct comparison across arbitrary architectures. In~\citet{cannistraci2024bricks}, with a preliminary workshop version~\citep{cannistraci2023infusing}, we extend this by combining multiple geometric invariances for richer representations (\cref{ch:product-invariances}). In \cref{ch:domain-applications}, we apply similar principles to policy networks~\citep{ricciardi2023r3l}, later extended in~\citet{ricciardi2024latent}, and to the latent uncertainty of model ensembles~\citep{mager2024s}. In ASIF~\citep{norelli2023asif}, we apply them to cross-modal architectures (\cref{ch:asif}).

\paragraph{Spectral structure.} Other methods use spectral properties of representation manifolds, since eigenstructure reveals intrinsic geometry independent of coordinate choice. In our latent functional maps work~\citep{fumero2024latent}, we adapt techniques from geometric correspondence to neural representations, enabling transformation through spectral operators (\cref{ch:latent_functional_maps}). In ResiDual~\citep{basile2025residual}, we apply spectral decomposition within single models to isolate task-relevant components (\cref{ch:residual}).

\paragraph{Orthogonal alignment.} When correspondence provides sufficient paired points, direct orthogonal transformations can align representation spaces while preserving geometric structure. In our semantic alignment work~\citep{maiorca2023semantic}, we construct these transformations between independently trained models, obtaining cross-model translation (\cref{ch:semantic_alignment}). In~\citet{maiorca2024inverse}, we do not estimate such a map: we chain the relative projection of \gls{relrep} (\cref{ch:relative-representations}) with a pseudo-inverse that returns to the target's absolute coordinates, so that its decoder can be reused (\cref{ch:inverse-relative-projection}). This demonstrates that chaining correspondence-based extraction with reconstruction achieves direct transformation through modular components.

\paragraph{Distributional properties.} Some methods establish correspondence between learned representations and desired distributional characteristics. In LinEAS~\citep{lineas}, we align the intermediate activations of a model with target concept distributions for controlled generation (\cref{ch:lineas}). In our bootstrapping work~\citep{cannistraci2023bootstrapping}, we iteratively refine a small seed correspondence with optimal transport across word embeddings, visual encoders and languages (\cref{ch:bootstrapping}). In~\citet{vedula2024scalable}, we learn embeddings that amortize Gromov-Wasserstein optimal transport, so that correspondence is found from unpaired samples and new samples are aligned with a single transport solve (\cref{ch:learning_to_align}).

\paragraph{Component specialization.} Within complex models, different components often specialize for different aspects of the task. Identifying this functional correspondence enables targeted intervention. In HeadPursuit~\citep{headpursuit}, we discover correspondence between attention heads and semantic concepts in language and vision-language models (\cref{ch:headpursuit}).

\paragraph{Applications and scope.} Correspondence-based alignment eliminates the need for expensive joint training or model retraining, allowing practitioners to repurpose existing models for ensemble uncertainty estimation, cross-architecture composition, and direct representation analysis without downstream task deployment.

The principles validate across computer vision (\glspl{cnn}, ResNets, Vision Transformers), natural language processing (BERT, \glsxtrshort{robertab}, \gls{llm} families), reinforcement learning (policy transfer), and cross-modal settings (text-image translation). We presented outcomes of this line of research at top-tier venues including ICLR, NeurIPS and TMLR, with oral and spotlight presentations recognizing their contributions. The community has widely adopted these techniques~\citep{moroni2025optimizing,wang2023understanding,wu2023zero,kiefer2023a}: our methods have been used for merging the latent spaces of models trained on different tasks~\citep{crisostomi2024charts}, and the \gls{correspondence-concept} principle formed the conceptual foundation for the UniReps workshop at NeurIPS 2023, 2024, and 2025, whose proceedings appeared in PMLR~\citep{unireps2023,unireps2024,unireps2025}.

\begin{sloppypar}
The principle extends to biological systems: different human brains processing identical visual stimuli develop alignable geometric structure in fMRI responses~\citep[][\cref{ch:domain-applications}]{maiorca2024multi}, suggesting that geometric convergence from shared semantic constraints may be a general principle of learning systems beyond artificial neural networks.
\end{sloppypar}

\chapter{Background and Related Work}
\label{ch:background}

Modern deep learning relies on learned representations---internal encodings that capture semantic structure from data and task objectives. Consider a neural network processing images: before mapping pixels to class labels, the network first transforms inputs into intermediate representations that compress high-dimensional data while preserving semantically relevant structure. A representation is a vector encoding where similar inputs (cats versus cats) map to nearby points, while dissimilar inputs (cats versus cars) map to distant points. This intermediate encoding can be reused across multiple downstream tasks, making representation learning a general paradigm~\citep{bengio2013representation}.

Formally, a layer $f_\ell: \R^{d_{\mathrm{in}}} \to \R^d$\glsadd{Layer} (the $\ell$-th of the $L$ layers of a neural network) maps inputs to $d$-dimensional representations. Architectures, at any point in their computational graph, can be factored into \gls{encoder}-\gls{decoder} structure: the whole model $F_\theta$, with parameters $\theta$, factors as $F_\theta = \Dec \circ \Enc$, where \gls{encoder} $\Enc: \Obs \to \R^d$ extracts latent representations from input space $\Obs$ (images, text, sensor data), and \gls{decoder} $\Dec$ maps representations to the task output (class labels $c \in \mathcal{C}$, reconstructed observations in $\Obs$, predictions). For a dataset $\Data = \{\obs_1, \dots, \obs_n\}$, the encoder produces representations $\{\Enc(\obs_i)\}$ forming a point cloud in $\R^d$---a geometric object we can analyze using distances, angles, and neighborhoods.

This geometric perspective treats representations as points in a structured space where proximity reflects semantic similarity. When we train neural networks, we hope they discover geometric structure that mirrors semantic relationships in the data: images of cats cluster together, images of cars cluster separately, and smooth transformations in input space (rotating an object) correspond to continuous paths in representation space.

These representations often exhibit similarities across independently trained models, yet remain numerically incompatible due to arbitrary choices in architecture and optimization. This chapter establishes the empirical evidence for this phenomenon (\cref{sec:background:similarity-puzzle}), introduces geometric vocabulary for reasoning about representational structure (\cref{sec:manifolds-geometry}), reviews existing approaches to representation alignment and their limitations (\cref{sec:latent-alignment}), and examines structural properties of learned representations (\cref{sec:structural-analysis}).

\section{The Representation Similarity Puzzle}
\label{sec:background:similarity-puzzle}

When two models are trained independently on similar data, do they discover similar geometric structure? Empirical evidence suggests they do---but with a critical caveat that makes this observation both intriguing and problematic.

Within and across architecture families, representations converge toward shared semantic structure. Networks trained from different random seeds develop aligned representational subspaces despite numerical differences in their coordinates~\citep{morcos2018insights,raghu2017svcca}, and in \glspl{cnn} individual features partly match one to one across seeds~\citep{Li2015-jo}. The same phenomenon holds when varying hyperparameters, data and architecture: \gls{vae} encoders trained with different hyperparameters and learning objectives learn similar representations in all layers except the mean and variance layers~\citep{Bonheme2022-tk}, word-embedding spaces of different languages can be aligned by an orthogonal map~\citep{conneau2017word}, and an intrinsic multi-scale distance, which compares data manifolds even across spaces of different dimensionality, suggests that networks of different architectures (VGG-16 and ResNet-20) reach similar final representation structure~\citep{tsitsulin2020shape}.

Joint training on paired data extends this convergence across modalities. Vision-language models trained on image-text pairs learn cross-modal representations that align through shared optimization~\citep{clip,align}, though a persistent modality gap remains even after alignment~\citep{liang2022mind}. These successes demonstrate that when optimization explicitly enforces correspondence or when architectural constraints restrict variation, semantic structure reliably converges.

\begin{figure}[t]
\centering
\includegraphics[width=\linewidth]{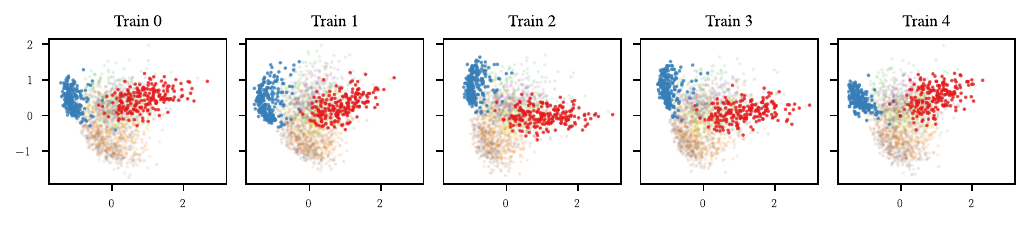}\\[0.5em]
\includegraphics[width=\linewidth]{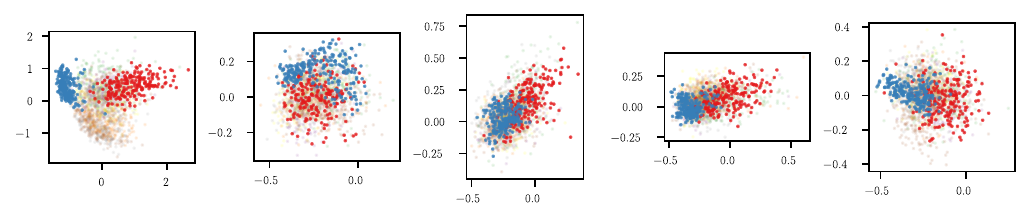}
\caption{\textbf{High-dimensional evidence of coordinate mismatch.} Five independent trainings of a high-dimensional \gls{mnist} autoencoder. \textbf{Top}: \gls{pca} fitted independently on each space produces highly similar projections---intrinsic geometry is shared. \textbf{Bottom}: \gls{pca} fitted on the leftmost space and applied to all five reveals severe misalignment---extrinsic coordinates differ. The incompatibility is purely optimization-induced. Colors: \gls{mnist} classes.}
\label{fig:background-pca-evidence}
\end{figure}

Yet this convergence of semantic structure coexists with divergence of numerical coordinates. Mode connectivity literature shows that different optimization trajectories reach functionally similar solutions in distinct regions of parameter space that a straight line cannot join at low loss, but a simple curve of near-constant accuracy can~\citep{garipov2018loss}. Building on the lottery ticket hypothesis~\citep{frankle2018lottery}, \citet{frankle2020linear} further show that networks trained from the same checkpoint under different SGD noise (data order and augmentation) end in minima joined by a linear path of low error once the shared checkpoint is taken early enough in training: only the smallest \gls{mnist} network has this property from initialization, while ResNet-20 on \gls{cifart} and ResNet-50 on ImageNet acquire it after 3\% and 20\% of training. Direct numerical comparison fails: the same input maps to different coordinates in each run (\cref{fig:background-pca-evidence}).

The mismatch between shared intrinsic structure and divergent extrinsic coordinates is visible in high-dimensional latent spaces. Consider five independent trainings of a high-dimensional \gls{mnist} autoencoder (\cref{fig:background-pca-evidence}). When we fit \gls{pca} independently on each space, the resulting projections are similar: the same clusters appear in virtually identical configurations, revealing that intrinsic geometry is shared across runs. Yet when we fit a single \gls{pca} on the first space and apply it to the others, severe misalignment emerges---different \gls{pca} projections from the leftmost space fail to reveal similar structure in others. The incompatibility is purely optimization-induced: the models have learned the same intrinsic structure and parameterized it through different extrinsic coordinates. This high-dimensional evidence demonstrates concretely what mode connectivity shows in parameter space: independent optimization of the same learning problem discovers functionally equivalent but coordinatewise incompatible solutions.

This creates a practical challenge for model reuse and composition. Combining modules from different models---connecting an encoder from a source model to a decoder from a target model---requires compatible representations, yet independent training produces incompatible coordinate systems. Existing transfer learning approaches solve this through retraining on target tasks~\citep{pan2009survey,yosinski2014transferable}, while model stitching introduces trainable adaptation layers and works best under architectural similarity~\citep{Bansal2021-oj,Csiszarik2021-yi}. Zero-shot transformation across independently trained models with different architectures remained an open problem: how to exploit semantic convergence while overcoming coordinate divergence.

\section{Geometric Foundations}
\label{sec:manifolds-geometry}

Understanding how independently trained models can learn compatible representations requires geometric vocabulary for describing structure that persists across coordinate transformations. This section introduces the manifold hypothesis---the observation that high-dimensional data concentrates on low-dimensional structure---and distinguishes intrinsic properties (coordinate-invariant geometry) from extrinsic properties (coordinate-dependent artifacts). These concepts provide the mathematical language for reasoning about what remains invariant when two models learn similar semantic structure in different coordinate systems.

\subsection{The Manifold Hypothesis}

A foundational assumption in modern machine learning holds that high-dimensional data---natural images, text sequences, sensor measurements---does not uniformly populate its ambient space. Instead, data concentrates near low-dimensional manifolds embedded in high-dimensional spaces. A $256 \times 256$ RGB image formally lives in $\R^{196608}$, yet the space of natural photographs occupies a tiny fraction of this vast volume. Images vary smoothly along semantic dimensions (object pose, lighting conditions, scene composition) that parameterize a much lower-dimensional manifold.

This manifold hypothesis explains why neural networks can learn effective representations despite the apparent curse of dimensionality. Networks discover and exploit the low-dimensional structure where data actually concentrates, so they do not need to model all possible configurations of high-dimensional inputs. Empirical measurements support this view: intrinsic dimensionality estimates~\citep{ansuini2019intrinsic,pope2021intrinsic,valeriani2022geometry,cheng2024evidence,cheng2025emergence} consistently reveal effective dimensionality far smaller than ambient dimensionality. Testing procedures have been proposed to verify whether real datasets exhibit manifold structure~\citep{https://doi.org/10.48550/arxiv.1310.0425}.

Three mechanisms drive this low-dimensional concentration. First, natural data exhibits inherent regularities---spatial coherence in images, syntactic and semantic constraints in language---that reduce effective degrees of freedom. Second, discriminative tasks require only separating categories, without modeling full generative distributions, allowing learned representations to discard task-irrelevant variation. Third, architectural choices directly enforce compression through bottleneck layers in encoder-decoder networks~\citep{Hinton2006-sb}, low-rank adaptation like LoRA~\citep{hu2021lora}, and regularization objectives that collectively push solutions toward low-dimensional manifolds.

Learned representations reflect this structure. Visualization of ImageNet-trained networks reveals hierarchical organization where nearby points share semantic features~\citep{zeiler2014visualizing,doimo2020hierarchical}. Large-scale pretraining objectives induce structured manifolds where semantic relationships organize geometric neighborhoods~\citep{bert,clip}. We can summarize this as: latent representations are points on \textbf{manifolds} $\mathcal{M} \subset \R^d$ whose intrinsic dimensionality $\dim(\mathcal{M})$ is substantially smaller than ambient dimensionality $d$, which itself is vastly smaller than the dimensionality of the original data.

\subsection{Intrinsic vs. Extrinsic Geometry}

The same manifold can be described using infinitely many coordinate systems, all equally valid yet numerically distinct. This coordinate ambiguity creates a fundamental distinction between two types of geometric properties.

\textbf{Intrinsic properties} depend only on the manifold's internal structure, independent of how it sits in ambient space. Distances measured along the manifold surface (geodesics), angular relationships between directions, local neighborhoods, and topological features like connected components all remain invariant under coordinate transformations. These properties are coordinate-free: they describe relationships \textit{within} the manifold that persist regardless of external embedding choices.

\textbf{Extrinsic properties} depend on how the manifold is embedded in ambient space. Absolute coordinate values, axis orientations, and the specific parameterization used to represent points all change under coordinate transformations, even when the underlying manifold remains unchanged. These properties are coordinate-dependent artifacts of our choice of representation.

Consider a concrete example: a sphere embedded in three-dimensional space. Intrinsic properties---surface area, radius, the existence of an arc connecting any two points---remain constant whether we rotate the sphere, translate it, or describe it using Cartesian versus spherical coordinates. Extrinsic coordinates change with each transformation, but the sphere's essential geometric structure persists.

This distinction connects directly to neural network representations. When two models are trained on similar data with similar objectives, they discover similar \textit{intrinsic structure}: semantic relationships that determine which concepts cluster together, how categories separate, and what hierarchical organization emerges. This structure is determined by the data distribution and task requirements---factors intrinsic to the learning problem.

However, the \textit{extrinsic coordinates} used to represent this structure vary arbitrarily based on parameterization choices. Network architecture (layer dimensions, activation functions, normalization schemes) determines the ambient space structure. Optimization dynamics (initialization seed, learning rate schedule, batch composition) determine which specific point in parameter space is reached, breaking coordinate symmetry in arbitrary ways. Two networks can learn to represent ``dog'' versus ``cat'' using the same geometric relationship (angular separation, clustering distance) while assigning these concepts to completely different numerical coordinates.

The puzzle documented in \cref{sec:background:similarity-puzzle}---that models learn similar semantic structure in incompatible coordinate systems---is precisely this intrinsic-extrinsic separation. Models converge to similar intrinsic manifold geometry (semantic relationships) but parameterize it using different extrinsic coordinates (numerical representations). Understanding alignment requires exploiting intrinsic structure while factoring out extrinsic variation.

\subsection{Connections to Dimensionality Reduction}

The problem of factoring intrinsic structure from extrinsic coordinates is not new to machine learning. Classical dimensionality reduction methods address precisely this challenge for single datasets: given high-dimensional observations, recover the underlying low-dimensional manifold structure. \Gls{pca}~\citep{pearson1901on} identifies linear subspaces of maximal variance, assuming intrinsic structure aligns with these directions. Nonlinear methods like Isomap~\citep{Tenenbaum2000-xh} and Locally Linear Embedding~\citep{Roweis2000-ft} extend this principle to curved manifolds by preserving geodesic distances or local reconstruction relationships. More recent techniques like t-SNE~\citep{van2008visualizing} prioritize local neighborhood preservation for visualization, though they distort global geometry.

These methods share a common principle: factor out arbitrary embedding choices (extrinsic coordinates) to reveal coordinate-invariant structure (intrinsic geometry). However, they address a simpler setting than neural network alignment. Dimensionality reduction assumes a single dataset observed in one coordinate system and seeks a low-dimensional embedding. Neural network alignment confronts a harder problem: relating multiple independently discovered manifolds that live in different ambient spaces, potentially with different dimensionalities, where there are no shared observations and the only bridge is semantic correspondence.

\section{Existing Alignment Approaches}
\label{sec:latent-alignment}

The challenge of aligning or combining representations from different models has motivated diverse approaches across machine learning. This section reviews major paradigms and their limitations.

\subsection{Transfer Learning and Fine-Tuning}

The dominant approach for reusing pretrained representations is transfer learning: a model pretrained on a source task is adapted to a target task through continued training~\citep{pan2009survey,bert,clip}. Feature transferability depends on architectural depth and task similarity---lower layers capture general features that transfer broadly, while higher layers specialize to source tasks~\citep{yosinski2014transferable}. Continual adaptation risks catastrophic forgetting, where performance on previous tasks degrades as parameters update~\citep{kirkpatrick2017overcoming}. While effective, this paradigm fundamentally requires retraining: target task data, labels, and computational resources for gradient descent. It offers no mechanism for zero-shot transformation or combining representations from multiple independently trained models without modification.

\subsection{Procrustes Analysis and Linear Alignment}

A more geometric approach comes from natural language processing, where Procrustes analysis has successfully aligned word embedding spaces. Orthogonal Procrustes finds the orthogonal matrix $\mR$ minimizing $\lVert \mX\mR - \mY \rVert_F$, where the rows of $\mX$ and $\mY$ are paired representations, preserving distances and angles while aligning coordinate systems. A linear map learned from a seed dictionary enabled early successes in cross-lingual word embeddings~\citep{DBLP:journals/corr/MikolovLS13}, with subsequent refinements incorporating normalization and an orthogonality constraint~\citep{xing-etal-2015-normalized}, unsupervised alignment without seed dictionaries~\citep{conneau2017word}, analyses of when the isomorphism assumption behind these maps fails~\citep{vulic-etal-2020-good}, and subword-enriched representations~\citep{bojanowski2016enriching}. Beyond language, Procrustes has aligned the later layers of convolutional networks trained with different random initializations as accurately as a general linear map~\citep{Csiszarik2021-yi}, revealing that optimization-induced coordinate differences can be factored out through linear transformations when intrinsic geometry is shared.

However, Procrustes faces significant limitations when extending beyond word embeddings: it requires matched dimensionality between source and target, assumes relationships are linear (orthogonal transformations exactly preserve distances), and needs correspondences, given as paired examples or seed dictionaries or induced from the data~\citep{conneau2017word}. These restrictions become prohibitive when aligning models with different architectures, dimensionalities, or tasks where natural correspondences may not exist.

\subsection{Model Stitching}

Another paradigm directly composes layers from different networks---connecting an encoder from a source model to a decoder from a target model. Trainable stitching layers can learn transformations enabling composition~\citep{Lenc2014-gy}, but this reintroduces the retraining requirement that transfer learning already imposes. A single linear or affine stitching layer often suffices when networks share their architecture, both across random initializations~\citep{Csiszarik2021-yi} and across training methods such as supervised and self-supervised learning~\citep{Bansal2021-oj}. Model stitching thus requires training adaptation layers (sacrificing zero-shot capability) and works best under architectural similarity.

\subsection{Representation Similarity Measures}

Complementary to transformation methods are techniques for quantifying similarity between representation spaces without enabling transformation. \Gls{cca}~\citep{hotelling1992relations} finds maximally correlated linear projections between two variable sets. Modern variants like \gls{svcca}~\citep{raghu2017svcca} incorporate dimensionality reduction for neural network comparison, while \gls{pwcca}~\citep{morcos2018insights} refines weighting schemes. \Gls{cka}~\citep{Kornblith2019-sz} offers an alternative that compares representational Gram matrices with invariance to orthogonal transformations and isotropic scaling, though recent work reveals sensitivity to sample subset selection~\citep{davari2022reliability}. \Gls{rsa}~\citep{Kriegeskorte2008Nov}, imported from neuroscience, compares second-order similarity structure through pairwise distance matrices, enabling comparison between neural networks and brain recordings despite radically different coordinate systems. More recently, \citet{basile2025intrinsic} directly related representation similarity to the intrinsic dimensionality, proposing a nonlinear similarity estimator based on the latter.

These measures provide tools for quantifying statistical correlation between spaces, but they share a fundamental limitation: they measure similarity without enabling transformation between the original spaces. CCA jointly estimates projections from both spaces into new low-dimensional subspaces where projections are maximally correlated, but doesn't produce operators for mapping between the original ambient representations. CKA and RSA quantify agreement through second-order statistics without revealing how to transform representations. They answer ``how similar?'' but not ``how to transform?''

\subsection{Geometric Alignment Methods}

More geometrically principled approaches come from optimal transport theory. The Gromov-Wasserstein distance~\citep{memoli2011gromov,alvarez2018gromov} measures dissimilarity between metric spaces by finding couplings that minimally distort pairwise distances, operating purely on intrinsic distance structure without requiring common coordinate systems. This makes it applicable to cross-modal and cross-architecture settings where no natural embedding exists. However, geometric rigor comes at computational cost: computing Gromov-Wasserstein distances requires solving nonconvex optimization with $O(n^3)$ complexity, limiting scalability. Functional maps~\citep{ovsjanikov2012fmap}, a spectral approach from shape analysis, represent correspondences as linear operators on low-dimensional spectral bases, achieving $O(k^3)$ complexity where $k \ll n$ is the number of basis functions. While more efficient, functional maps assume spectral structure amenable to this representation.

\subsection*{Existing Approaches and Their Gap}

Existing approaches have complementary limitations. Transfer learning enables effective adaptation but requires retraining and offers no mechanism for zero-shot combination. Procrustes provides geometrically principled alignment but demands matched dimensionality, linearity, and explicit correspondence. Model stitching demonstrates compatibility within architecture families but needs training and works best under restricted conditions. Similarity measures quantify correlation without enabling transformation. Geometric methods offer principled frameworks but face computational barriers or structural assumptions. Zero-shot transformation across independently trained models with arbitrary architectures remained an open challenge.

\section{Structural Properties of Representations}
\label{sec:structural-analysis}

Beyond global manifold geometry, neural networks exhibit analyzable internal structure. Mechanistic interpretability reveals that architectural components specialize for distinct semantic functions, while linear representation structure shows that semantic concepts often correspond to interpretable directions. These properties suggest that representations are structured compositions amenable to systematic analysis and manipulation.

\subsection{Mechanistic Interpretability: Assigning Meaning to Components}

Mechanistic interpretability approaches representations by decomposing models along architectural boundaries and assigning semantic meaning to the resulting components. Transformer architectures process information through additive accumulation: each layer and attention head contributes updates to a shared residual stream~\citep{elhage2021mathematical,vaswani2017attention}, creating a decomposable structure where the final representation is a sum of contributions from individual components. This architectural choice enables component-level analysis.

The logit lens~\citep{nostalgebraist2020interpreting} exploits this structure by projecting intermediate residual stream states directly through the final output layer, revealing that representations at an intermediate layer $\ell$ of an $L$-layer model already contain information about final predictions. The tuned lens~\citep{belrose2023eliciting} refines this idea by learning lightweight affine transformations that more accurately decode intermediate states. These methods demonstrate that representations remain interpretable throughout network depth, at intermediate layers as well as at the output, suggesting that semantic content accumulates progressively through processing.

Individual components exhibit specialization: attention heads cluster into functional groups performing distinct operations. Some heads track positional information, others capture syntactic relationships, and still others attend preferentially to rare words~\citep{voita2019analyzing}. Pruning experiments reveal that many heads can be removed without performance degradation~\citep{michel2019sixteen}, while the heads that remain critical are the specialized ones~\citep{voita2019analyzing}. Induction heads~\citep{olsson2022context} are argued to implement much of in-context learning by copying patterns from earlier in the sequence, demonstrating that specific algorithmic computations localize to identifiable components.

Beyond attention, feedforward layers take part in factual recall, which proceeds in three steps: early \gls{mlp} layers enrich the subject representation with attributes, information about the relation propagates to the final position, and attention heads then extract the queried attribute from the enriched subject~\citep{geva2023dissecting}. Factual knowledge (``Paris is the capital of France'') localizes to specific layer ranges and can be edited by modifying targeted component weights~\citep{meng2022locating}. This structured localization extends across modalities. CLIP attention heads specialize for properties such as shape, location, color or letters, while also being decomposable into combinations of text-based concept directions~\citep{gandelsman2024interpreting}. The same decomposition extends to vision transformers beyond CLIP, such as DeiT, DINO and Swin, by mapping their component contributions into CLIP space~\citep{balasubramanian2024decomposing}. The last layer of CLIP has further been hypothesized to encode compositional attributes such as color, size and counting in independent, nearly orthogonal subspaces~\citep{wolff2023the}. Architectural components have interpretable semantic functions, suggesting that representations organize according to principled decompositions.

\subsection{Linear Representation of Semantic Meaning}

A recurring empirical pattern emerges across diverse architectures: semantic concepts correspond to linear directions in representation space~\citep{linearrep_hypothesis}. The classic example ``king $-$ man $+$ woman $\approx$ queen''~\citep{mikolov2013word2vec} demonstrates semantic composition through vector arithmetic, revealing a deeper principle that semantic operations can be performed linearly. Individual directions encode interpretable concepts like ``male versus female'' or ``English versus French''~\citep{linearrep_hypothesis}, and \citet{jiang2024on} argue that the next-token prediction objective, together with the implicit bias of gradient descent, promotes such linear representations.

Sparse autoencoders~\citep{huben2024sparse,templeton2024scaling} learn overcomplete dictionaries that decompose representations into interpretable features, revealing that semantic meaning often factors into sparse linear combinations, with little entanglement across dimensions. \citet{thasarathan2025universal} extend this principle to cross-model decomposition with a single sparse autoencoder.

This linear structure \citep{linearrep_hypothesis,nanda2023emergent,jiang2024on} matters because it makes concepts tractable and ``interpretable'' (we can assign a simple, direct meaning to each of them). When semantic meaning is expressed as linear directions, it becomes possible to identify specific concepts, manipulate them through vector arithmetic, and transfer them between models. Linear structure provides a bridge between geometric analysis (angles, distances, subspaces) and semantic interpretability (what concepts are represented, how they combine). Combined with component specialization from mechanistic interpretability, linear representation structure suggests that neural networks develop analyzable, manipulable geometric organizations.

\chapter{Framework: Correspondence as a Structural Primitive}
\label{ch:methodology}

\section{Neural Representations as Conditioned Manifolds}
\label{sec:neural-representations}

\paragraph{Semantic Structure: What Should Be Learned.}

Before discussing how neural networks learn, consider what they should learn. When we train an image classifier on ImageNet, there exists an abstract notion of the semantic structure that should emerge: cats should be representationally closer to dogs than to cars; fine-grained distinctions within bird species should be preserved; invariances to lighting and pose should emerge. This is about semantic relationships. It does not depend on specific coordinates in $\R^{512}$ or $\R^{768}$.

Data $\Data$ and task objective $\mathcal{L}_{\mathrm{task}}$\glsadd{Loss} jointly determine this abstract {\em semantic structure} $\mathcal{I}(\Data, \mathcal{L}_{\mathrm{task}})$---which concepts should be near versus far, what hierarchical organization should exist, which features distinguish classes, what invariances matter. The task objective plays a critical role: classification demands in-class indistinguishability and cross-class separation; contrastive learning requires positive pairs close together, negatives apart; regression imposes continuity. Data provides the semantic material; task determines which structure must be extracted to succeed.

\paragraph{Grounding Abstract Structure Through Optimization.}

We cannot train models in this abstract world---gradient descent requires differentiable functions in finite-dimensional spaces. To learn $\mathcal{I}(\Data, \mathcal{L}_{\mathrm{task}})$, we must instantiate it through parameterized optimization:
\begin{equation}
\boxed{\theta^* \in \argmin_{\theta \in \Theta(\mathcal{H})} \hlintrinsic{$\mathcal{L}_{\mathrm{task}}(F_\theta; \Data)$} + \hlextrinsic{$\mathcal{R}(\theta)$}},
\label{eq:learning}
\end{equation}
\begingroup\sloppy%
where $\theta$ ranges over the parameter space $\Theta(\mathcal{H})$ allowed by the architecture $\mathcal{H}$, and $\mathcal{R}$ is the explicit regularizer. This equation realizes abstract semantic structure $\mathcal{I}(\Data, \mathcal{L}_{\mathrm{task}})$ through concrete parameterization. The \hlintrinsic{task loss $\mathcal{L}_{\mathrm{task}}$} measures how well function $F_\theta$ captures the required semantic relationships---whether cats are near dogs, whether class boundaries are well-separated, whether required invariances emerge. These are \hlintrinsic{\glspl{intrinsic-factor}} that determine {\em what} should be learned. In contrast, \hlextrinsic{architecture $\mathcal{H}$} and \hlextrinsic{regularization $\mathcal{R}$} provide the differentiable substrate and optimization biases---\glspl{cnn} with convolutional inductive biases, Transformers with attention mechanisms, ResNets with skip connections. These are \hlextrinsic{\glspl{extrinsic-factor}} determining {\em how} that semantic structure is parameterized. \hlextrinsic{Optimization dynamics} (initialization, learning rate, batch size, optimizer choice) further bias which particular solution training reaches within the space of functions that satisfy task requirements.
\par\endgroup

\begin{figure}[t]
\centering
\includegraphics[width=\textwidth]{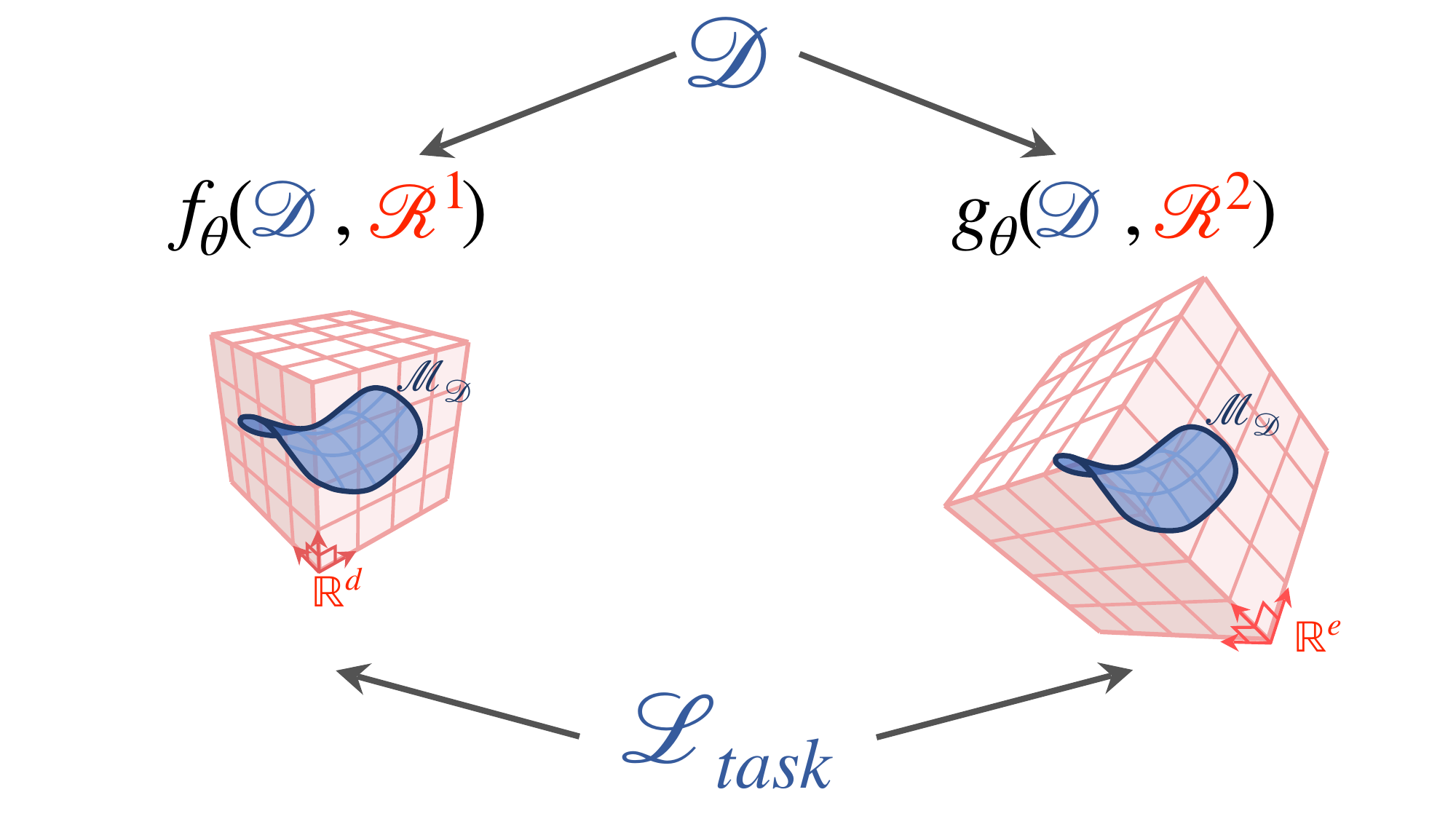}
\caption{\textbf{Realizing abstract structure through parameterized optimization.} Two models optimized with the same \hlintrinsic{data $\Data$} and \hlintrinsic{task loss $\mathcal{L}_{\mathrm{task}}$} but different \hlextrinsic{extrinsic factors} (architecture and regularization $\mathcal{R}^1$ vs. $\mathcal{R}^2$) produce learned functions that encode similar intrinsic semantic relationships yet through incompatible coordinate systems ($\R^{\dimof{X}}$ vs. $\R^{\dimof{Y}}$, with $\dimof{X}$, $\dimof{Y}$ the latent dimensions). The resulting manifolds $\mathcal{M}(\Enc_{\theta^*})$ exhibit similar geometry embedded in extrinsic coordinates that differ by arbitrary transformations; $f_\theta(\Data,\mathcal{R}^1)$ and $g_\theta(\Data,\mathcal{R}^2)$ in the figure denote the encoders of the two models (trained with regularizers $\mathcal{R}^1$, $\mathcal{R}^2$) applied to $\Data$, $\mathcal{M}_{\Data}$ denotes each resulting manifold, and $\R^d$, $\R^e$ denote $\R^{\dimof{X}}$, $\R^{\dimof{Y}}$.}
\label{fig:manifold-thesis}
\end{figure}

This realization is illustrated in \cref{fig:manifold-thesis}: two models with different extrinsic configurations produce geometrically similar manifolds in incompatible coordinates.

\paragraph{From Function Parameterization to Data Encoding.}

As shown in \cref{fig:manifold-thesis},\footnote{Figure adapted from \textit{Autoencoder Diagrams} by Keenan Crane (2025), used under CC0 1.0 Universal. Adaptation by Baoying Wang.} the encoder $\Enc_{\theta^*}$ of the learned model produces representations forming the manifold $\mathcal{M}(\Enc_{\theta^*})$. These representations are themselves a form of parameterization: they encode data through coordinate choices determined by architecture and optimization. Different architectures can realize similar semantic structure $\mathcal{I}$ through incompatible parameterizations---a CNN and Transformer trained on ImageNet may both learn that cats cluster with dogs and separate from vehicles, yet encode these relationships through coordinate systems with different dimensionalities, orientations, and scales. Just as $\theta^*$ parameterizes the learned function, the representations $\Enc_{\theta^*}(\obs)$ of observations $\obs$ parameterize the data's semantic content. Architecture is the vehicle for instantiating abstract structure; it does not determine what structure exists.

\paragraph{Terminology: Encoders and Decoders.}

Throughout this thesis, we use {\em \gls{encoder}} to denote whatever function precedes the representation space of interest---the computational pathway from input to latent representation. Conversely, {\em \gls{decoder}} denotes whatever function follows the representation space---the pathway from latent representation to output. For a classification network, the \gls{encoder} usually comprises convolutional layers and pooling operations that produce feature representations, while the \gls{decoder} is the classification head mapping features to class logits. For a generative model, the encoder maps inputs to a latent bottleneck, while the decoder reconstructs outputs from latent codes. For a language model, intermediate layer activations are the representations, with earlier layers forming the encoder and later layers the decoder. This functional decomposition (input $\to$ encoder $\to$ representation $\to$ decoder $\to$ output) applies regardless of architectural details, allowing us to discuss representation transformation in a general way across diverse model types.

This separation of intrinsic structure from extrinsic parameterization is empirically observed across the thesis: models trained on related data with discriminative objectives consistently learn similar semantic relationships despite architectural differences. The alignment methods presented rely on architecture parameterizing semantic structure; they would fail if architecture determined it. Architecture provides inductive biases affecting learning efficiency, but when capacity is sufficient and objectives discriminative, different architectures converge to functionally similar solutions.

\section{Correspondence as Primitive}
\label{sec:correspondence}

\paragraph{Terminology and Glossary.} The framework concepts introduced in this chapter---correspondence-parallel operators, correspondence-coupled operators, semantic coordinates, intrinsic factors, extrinsic factors, and correspondence itself---receive comprehensive definitions in the List of Symbols (Glossary), where they are referenced throughout subsequent chapters. Each glossary entry provides a formal definition, geometric interpretation, and pointers to relevant sections, allowing readers to consult precise specifications when encountering these terms in later chapters. This design enables method chapters to reference framework concepts concisely through glossary lookups, without repeating definitions, maintaining conceptual clarity while avoiding redundancy.

\Cref{sec:neural-representations} established that models can learn similar intrinsic semantic structure through different extrinsic parameterizations, yet these parameterization differences create incompatible coordinate systems. How do we reveal the hidden functional similarity? The answer is {\em \gls{correspondence-concept}}: establishing semantic relationships between observation spaces that anchor arbitrary coordinates to shared semantic content. \Gls{correspondence-concept} is the fundamental primitive that enables all transformation methods in this thesis.

\paragraph{Defining Correspondence.}

In its most general form, a correspondence $\gls{Corr} \subseteq \mathcal{P}(\Obs[X]) \times \mathcal{P}(\Obs[Y])$ establishes semantic relationships between observation spaces $\Obs[X]$ and $\Obs[Y]$, operating on power sets to relate collections of elements. This general definition accommodates sophisticated structures: set-based correspondences that relate groups of observations, probabilistic correspondences that assign confidence weights to pairings (a weighted extension of this relation), or hierarchical correspondences that organize semantic relationships at multiple scales. While such sophisticated structures arise in certain applications, most methods in this thesis employ a simpler element-wise form: $\Corr \subseteq \Obs[X] \times \Obs[Y]$, which pairs individual observations $(\obs_i, \obs'_i) \in \Corr$ based on shared semantic meaning. Encoders $\Enc[X]: \Obs[X] \to \mathcal{X}$ and $\Enc[Y]: \Obs[Y] \to \mathcal{Y}$ map these observations into the representation spaces $\mathcal{X}$ and $\mathcal{Y}$ of the source and target models. The image of $\Corr$ under the encoders gives the paired representations $(\vx_i, \vy_i) = (\Enc[X](\obs_i), \Enc[Y](\obs'_i))$ that the operators of \cref{sec:operator-families} consume; these operators act on $\mathcal{X}$ and $\mathcal{Y}$. The observations enter only through the encoders.

\paragraph{Notation.}
Throughout the thesis, observations $\obs \in \Obs$\glsadd{obs}\glsadd{Obs} are collected in finite datasets $\Data \subset \Obs$\glsadd{Data}. An encoder $\Enc[X]: \Obs[X] \to \mathcal{X} \subseteq \R^{\dimof{X}}$\glsadd{Enc}\glsadd{Space}\glsadd{DimLatent} is followed by a decoder or task head $\Dec[X]$\glsadd{Dec}, and the whole model is $F_\theta = \Dec \circ \Enc$\glsadd{Model}. A representation $\vx = \Enc[X](\obs)$\glsadd{RepVec} is one row of the matrix $\mX \in \R^{n \times \dimof{X}}$\glsadd{RepMat} that stacks $n$\glsadd{NumSamples} samples. Anchors $\anchors = \{a_j\}_{j=1}^{\nanch} \subset \Obs[X]$\glsadd{Anchors}\glsadd{NumAnchors} are observations whose embeddings $\va_j = \Enc[X](a_j)$\glsadd{AnchorEmb} form the rows of $\anchmat[X] \in \R^{\nanch \times \dimof{X}}$\glsadd{AnchorMat}; a similarity $\simf$\glsadd{Sim} and the relative projection $\relproj$\glsadd{RelProj} built on it are defined in \cref{ch:relative-representations}. Concrete maps between spaces are written $\mT$ (general linear)\glsadd{LinMap}, $\mR$ (orthogonal), $\mP$ (permutation or coupling), and $\vb$ (bias), and $\tau \in \mathcal{T}$\glsadd{TransfClass} denotes a generic transformation from a class $\mathcal{T}$. Scalars are italic, vectors bold lowercase, matrices bold capitals, and spaces and sets calligraphic.

What does ``shared semantic meaning'' mean concretely? We establish correspondence when we recognize that elements from different spaces refer to the same concepts or to semantically similar concepts within a defined relationship. An image $\obs_i$ from the source model's training set and image $\obs'_i$ from the target model's training set both depict cats from the same ImageNet class (identical concept). Two sentences are translations of each other in different languages, performing equivalent semantic functions despite different surface forms (translation equivalence). Two images show the same scene in different artistic styles (instance pairing with varying attributes). These are all instances of correspondence---identifying when elements from different spaces should be semantically aligned, despite living in incompatible coordinate systems.

\paragraph{What Correspondence Reveals.}

To understand why correspondence is fundamental, consider what happens when we lack it. Take two models trained on identical data but with different random initialization seeds. Their learned representations encode the same semantic structure---both know cats are near dogs, both separate vehicles from animals---yet this structure lives in coordinate systems that differ by an unknown, arbitrary transformation. Without correspondence, we cannot determine which dimensions in the source model encode which concepts, or how the source model's encoding of ``cat'' relates to the target model's encoding of the same concept. The coordinate systems remain incomparable despite underlying functional similarity.

\textit{\Gls{correspondence-concept} breaks this symmetry by anchoring arbitrary coordinates to shared semantics.} When we identify pairs $(\obs_i, \obs'_i)$ with shared semantic meaning---both occurrences of the word ``cat,'' or the same ImageNet image seen by both models---we constrain the arbitrary degrees of freedom that independent optimization introduced. The models' representations of $\obs_i$ must correspond to their representations of $\obs'_i$, providing semantic anchors that fix coordinate alignment. These anchors reveal which geometric properties reflect \hlintrinsic{\glslink{intrinsic-factor}{intrinsic functional structure}} preserved across models, distinguishing them from \hlextrinsic{\glslink{extrinsic-factor}{extrinsic implementation artifacts}} that depend merely on coordinate choice.

The type of correspondence determines what geometric structure can be exploited. {\em Angular correspondence} relates directions in representation space, revealing cosine similarity patterns and directional alignment. {\em Metric correspondence} preserves pairwise distances, exposing quasi-isometric structure and enabling distance-preserving transformations. {\em Spectral correspondence} derives from eigendecompositions of manifold operators, revealing functional bases and frequency content. {\em Component-level correspondence} identifies which neurons or attention heads encode which concepts, enabling surgical interventions on individual units. {\em Distributional correspondence} operates on statistical properties of collections of examples, capturing covariance structure and higher-order moments. Each correspondence type unveils different intrinsic geometry, and correspondence strength determines how reliably that geometry can be exploited for transformation.

\paragraph{Informal Taxonomy.}

Correspondences vary considerably in strength and specificity, and understanding this spectrum proves essential for selecting appropriate methods and setting realistic expectations for transformation quality.

{\em Shared semantic anchors} provide the strongest correspondence: identical observations encoded by both spaces. The same CIFAR-10 images processed by different vision models, identical sentences embedded by different language models, the same visual stimuli recorded in the fMRI responses of different subjects (\cref{sec:cosyne})---each pair $(\obs_i, \obs'_i)$ provides a direct semantic touchpoint where representations refer to exactly the same entity. This cleanest form of correspondence enables precise geometric comparisons across models.

{\em Cross-modal pairs} occupy the next tier: semantically related but structurally different observations. Image-caption pairs for vision-language models, translation equivalents for multilingual systems---these correspondences bridge different representational structures that encode shared semantic content. They exceed shared anchors in an important sense: they unite images with text, or different languages, through semantic equivalence. An image is not a caption, yet both refer to the same concept, demanding methods that handle this structural heterogeneity while respecting semantic unity.

{\em Task structure} provides weaker but still useful correspondence through shared objectives or label spaces, without requiring instance-level pairing. Two models trained to classify cats versus dogs develop similar decision boundaries even if trained on completely different image sets, because the task itself provides implicit correspondence: the cat-versus-dog semantic axis should align across models. Methods exploit this by using task labels to identify relevant representational structure without requiring paired examples, though the resulting correspondence is less precise than explicit pairing.

{\em Distributional patterns} provide the weakest but most general correspondence through statistical properties inferred from activation patterns: means, covariances, spectral decay rates, activation sparsity. Without explicit pairing, methods rely on distributional assumptions---similar concepts cluster, different concepts separate---enabling unsupervised alignment when no explicit correspondence exists. This generality comes at a cost: success requires stronger geometric assumptions about representation structure.

\subsection{Correspondence Encodes Task Intent}
\label{sec:correspondence-task-intent}

\textit{Correspondence choice is not neutral: it implicitly encodes a task.} The same pair of representation spaces admits multiple different correspondences, each appropriate for different tasks, and this task-conditional nature determines method selection while predicting success or failure.

Consider two contrasting scenarios, both involving image models but requiring different correspondences. In the first scenario---cross-modal retrieval in the style of CLIP---the goal is to find images that match text queries. The appropriate correspondence $\Corr$ pairs concepts across modalities: $(\obs_{\mathrm{img},i}, \obs'_{\mathrm{text},i})$ where both refer to the same concept (for instance, both representing ``cat''). If we arrange these pairs in a matrix, different concepts appear in different rows: (cat-image, cat-text), (dog-image, dog-text), and so on, while the same concept appears across columns in different modalities. This correspondence structure encodes a specific semantic intent: cross-modal representations should be similar. An image of a cat should retrieve the word ``cat'' because they share semantic content despite living in different modalities.

The second scenario---style transfer toward Van Gogh's artistic style---requires completely different correspondence structure despite potentially using similar generative models. Here, the appropriate correspondence $\Corr'$ pairs instances with and without the target property: $(\obs_{\mathrm{regular},i}, \obs'_{\mathrm{VanGogh},i})$ where both represent the same scene or object, but one is rendered in regular style while the other exhibits Van Gogh's characteristic brushwork and color palette. Arranged as a matrix, each row holds one instance (a house in regular style, the same house in Van Gogh style), while the columns are the two style conditions (without target style, with target style). This correspondence encodes distributional intent: learn the transformation that adds ``Van Gogh-ness'' to any image, capturing stylistic difference as a shift in representation space.

The contrast is evident. Visualized as matrices, both have one pair per row, but the pairs share different things: in $\Corr$ the two columns share the concept and differ in modality, while in $\Corr'$ they share the instance and differ in style. Correspondence structure directly reflects task structure: using the wrong correspondence for a given task fails regardless of transformation method sophistication. Style transfer cannot be learned from cross-modal pairs, nor can retrieval succeed with style-paired instances.

\paragraph{Standard Correspondences and Transfer.}

Not all correspondences transfer equally. Some follow standard semantics: they align with the general semantic structure models learned during pretraining, before task-specific fine-tuning. Linguistic structure correspondences derived from Wikipedia---capturing relationships between words, grammatical patterns, and semantic fields---transfer successfully to Amazon product reviews because both domains share underlying grammatical and semantic patterns. Such correspondences align with the geometry models discover during training on general objectives, making them robust across domains.

Why does this transfer work so reliably? Models trained with sufficiently general objectives---language modeling over diverse text, image classification on datasets with thousands of categories---learn broad semantic structure extending well beyond the specific training distribution. They discover that certain concepts cluster together, that particular relationships recur across contexts, that specific transformations preserve meaning. Correspondences respecting this learned structure, anchoring to the same semantic axes the model already discovered, transfer successfully across domains by building on structure the model already internalized. The model has decomposed its representations along meaningful dimensions; correspondence merely identifies which dimensions correspond across different contexts or different models.

In contrast, task-specific correspondences are optimized for particular applications and transfer only within domains sharing the specific task structure. The Van Gogh style correspondence is excellent for style transfer but irrelevant for object classification tasks. The image-caption correspondence is perfect for cross-modal retrieval but unhelpful for fine-grained texture discrimination. The distributional correspondence that enables one type of steering may fail completely when applied to a different downstream task. This highlights that correspondence design is task-conditional: we must choose $\Corr$ to align with both the geometric structure the model learned and the specific application we have in mind.

\section{Two Operator Families Enabled by Correspondence}
\label{sec:operator-families}

We now examine how to exploit a given correspondence for transformation. Methods in this thesis instantiate two operator families distinguished by how they use correspondence: independently per space, or jointly across both sides. The distinction determines critical trade-offs between modularity and decoder compatibility.

\Glspl{correspondence-parallel} ($\gls{OperatorParallel}_{\Corr}^{\mathcal{X}}$) transform each space independently using only its side of the correspondence, indexed consistently with the other side. Space $\mathcal{X}$ uses only the encodings $\{\vx_i\}$ of $\{\obs_i\}$ from pairs $(\obs_i, \obs'_i) \in \Corr$, while space $\mathcal{Y}$ independently uses only $\{\vy_i\}$. Outputs align because they reference semantically corresponding elements, producing representations in shared auxiliary coordinates. \Glspl{correspondence-coupled} ($\gls{OperatorCoupled}_{\Corr}: \mathcal{X} \to \mathcal{Y}$) use both sides together, exploiting pairing relationships to construct explicit transformations from source to target coordinates, producing outputs in the target's native space. This choice involves inherent trade-offs, summarized in \cref{tab:operator-families}: parallel operators offer modularity but require auxiliary decoders; coupled operators enable direct decoder reuse but demand pairwise estimation.

\subsection{Correspondence-Parallel Operators \texorpdfstring{$\OperatorParallel_{\Corr}^{\mathcal{X}}$}{}}
\label{sec:operator-parallel}

\Glspl{correspondence-parallel} exploit correspondence through independent transformations that align because they reference semantically corresponding elements. Given a correspondence $\Corr \subseteq \Obs[X] \times \Obs[Y]$ with paired elements $\{(\obs_i, \obs'_i)\} \subseteq \Corr$, we construct two separate operators. Space $\mathcal{X}$ transforms using only $\CorrSide{X} = \{\obs_i : (\obs_i, \obs'_i) \in \Corr\}$, encoded as $\{\vx_i\}$---its side of the correspondence---while space $\mathcal{Y}$ independently transforms using only $\CorrSide{Y} = \{\obs'_i : (\obs_i, \obs'_i) \in \Corr\}$, encoded as $\{\vy_i\}$. These transformations occur in complete isolation: space $\mathcal{X}$ never accesses representations from $\mathcal{Y}$ and vice versa. Compatibility comes from the semantic correspondence between the elements each space uses; the two transformations never interact directly.

We denote these operators as $\OperatorParallel_{\Corr}^{\mathcal{X}}$ and $\OperatorParallel_{\Corr}^{\mathcal{Y}}$, with the superscript emphasizing that each space uses its correspondence side independently. Outputs $\OperatorParallel_{\Corr}^{\mathcal{X}}(\vx)$ and $\OperatorParallel_{\Corr}^{\mathcal{Y}}(\vy)$ live in a shared coordinate system because the correspondence structure itself defines their common coordinate frame; the transformations do not interact during computation. When $(\obs_i, \obs'_i)$ refer to the same semantic entity---the same image, the same concept, the same linguistic structure---the transformation applied using $\vx_i$ in space $\mathcal{X}$ becomes semantically analogous to the transformation applied using $\vy_i$ in space $\mathcal{Y}$, despite their computational independence.

\paragraph{Output Space.}

\Glspl{correspondence-parallel} produce outputs in an auxiliary coordinate system defined by the correspondence structure itself. This coordinate system differs from the extrinsic coordinates of both original spaces. The transformed representations live in a shared space whose axes are defined by the correspondence domain---semantic anchors, task labels, spectral frequencies, or other structure captured by the correspondence. This can be thought of as living in \glspl{semantic-coordinate}.

\paragraph{Key Property: Modularity.}

The primary advantage of correspondence-parallel operators is one-to-many modularity, making them attractive for model ecosystems. Once we compute $\OperatorParallel_{\Corr}^{\mathcal{X}}(\vx)$ for the source model, this transformed representation becomes immediately compatible with any other model transformed using the same correspondence $\Corr$. Adding a new model to the ecosystem requires only applying $\OperatorParallel_{\Corr}^{\mathcal{Z}}$ to its representations $\vz \in \mathcal{Z}$, using the side of $\Corr$ that lives in its input domain; no modification of existing transformations is necessary, and no pairwise mappings need to be learned. This modularity allows the approach to scale efficiently to an effective ``nexus'' of models and representation spaces, where the cost of pairwise methods would become prohibitive. Multi-modal retrieval systems that must align vision, language, and audio representations; model zoos where new models are continually added; continual learning scenarios where new tasks introduce new models---all benefit from this one-to-many scaling.

\paragraph{Trade-off.}

The fundamental trade-off is that outputs live in auxiliary coordinates, outside any model's original extrinsic space, which makes them incompatible with pretrained decoders. To generate outputs---images, text, actions---we must either train auxiliary decoders specific to the semantic coordinate system, learn inverse mappings to reconstruct original coordinates, or restrict ourselves to tasks like retrieval where auxiliary coordinates suffice. We gain modularity at the cost of direct decoder compatibility and space-specific nuances, a trade-off that determines when parallel operators are appropriate.

\subsection{Correspondence-Coupled Operators \texorpdfstring{$\OperatorCoupled_{\Corr}$}{}}
\label{sec:operator-coupled}

\Glspl{correspondence-coupled} take a different approach: they use both sides of the correspondence together to construct explicit transformations from source to target. Given a correspondence $\Corr \subseteq \Obs[X] \times \Obs[Y]$ with paired elements $\{(\obs_i, \obs'_i)\} \subseteq \Corr$, these operators require simultaneous access to both paired elements and their relationships, using the joint structure of $(\vx_i, \vy_i)$ pairs to estimate transformation parameters that map directly from source coordinates to target coordinates. We denote these operators as $\OperatorCoupled_{\Corr}: \mathcal{X} \to \mathcal{Y}$, emphasizing that they exploit the full correspondence structure to construct mappings between spaces.

The key conceptual difference from correspondence-parallel operators is that we exploit the pairing relationship itself: both that $\vx_i$ and $\vy_i$ refer to the same semantic entity and how they relate within their respective coordinate systems. This allows us to estimate coordinate transformations that map from source to target while preserving relevant geometric structure. The crucial advantage is that outputs live in the target's native extrinsic coordinate system. The transformed representation $\OperatorCoupled_{\Corr}(\vx)$ lives in the same space as $\vy$, with the same dimensionality, scale, and structure, making it immediately ready for use with the target model's pretrained decoder.

\paragraph{Decoder Reuse and Zero-Shot Stitching.}

This direct compatibility with target decoders enables powerful applications. Transformed representations $\OperatorCoupled_{\Corr}(\vx)$ are immediately usable with the target's pretrained decoder---no auxiliary coordinate conversion, no inverse mappings, no new decoder training. This enables {\em zero-shot stitching}: composing encoder-transformation-decoder pipelines without joint training. We chain a source encoder, the learned transformation, and a target decoder into a complete pipeline, where the transformation preserves the metric properties and structural relationships the target decoder expects. Downstream tasks proceed as if the representations had been computed natively by the target model.

When is this essential? Generation tasks requiring target decoders to produce images or text; policy execution in reinforcement learning where actions must be generated in a specific format; classification with fixed decoders where we cannot modify the final layer---all scenarios demanding exact preservation of the target's coordinate system, making correspondence-coupled operators the natural choice.

\paragraph{Limitations.}

The main trade-off is reduced modularity compared to correspondence-parallel operators. Each source-target pair $(\mathcal{X}, \mathcal{Y})$ requires estimating a separate transformation $\OperatorCoupled_{\Corr}$, and adding a new target model with representation space $\mathcal{Z}$ demands estimating an entirely new transformation from scratch, limiting scalability to many-model ecosystems. Additionally, the target decoder must remain compatible with transformed representations: distributional shifts or artifacts can degrade performance even when representations live in the correct coordinate system. We gain direct decoder compatibility at the cost of modularity and scalability, a trade-off that becomes prohibitive when many target models must be supported.

\begin{table}[t]
\centering
\begin{tabular}{@{}>{\raggedright\arraybackslash}p{0.24\textwidth}>{\raggedright\arraybackslash}p{0.34\textwidth}>{\raggedright\arraybackslash}p{0.34\textwidth}@{}}
\toprule
\textbf{Property} & \textbf{Parallel ($\OperatorParallel_{\Corr}^{\mathcal{X}}$)} & \textbf{Coupled ($\OperatorCoupled_{\Corr}$)} \\
\midrule
\textbf{How $\Corr$ is used} & Independently per space & Both sides together \\
& ($\CorrSide{X}$ and $\CorrSide{Y}$ separate) & (paired) \\ \addlinespace[0.5em]
\textbf{Output space} & Auxiliary coordinates & Target extrinsic coordinates \\
& (defined by $\Corr$ structure) & (native to $\mathcal{Y}$) \\ \addlinespace[0.5em]
\textbf{Decoder compatibility} & Requires auxiliary decoder & Direct reuse of target decoder \\ \addlinespace[0.5em]
\textbf{Modularity} & High (one-to-many) & Lower (pairwise) \\ \addlinespace[0.5em]
\textbf{Primary advantage} & Scales to many models & Immediate decoder compatibility \\ \addlinespace[0.5em]
\textbf{Primary trade-off} & Cannot reuse decoders & Requires separate transform per pair \\ \addlinespace[0.5em]
\textbf{Example notation} & $\OperatorParallel_{\Corr}^{\mathcal{X}}(\vx)$ & $\OperatorCoupled_{\Corr}(\vx)$ \\
\bottomrule
\end{tabular}
\caption{\textbf{Comparison of Operator Families.} The families differ in how they exploit correspondence structure, where outputs live, and the resulting trade-offs between modularity and decoder compatibility.}
\label{tab:operator-families}
\end{table}

\paragraph{Notational Shorthand for Same-Space Coupling.}

When correspondence-coupled operators apply within a single space ($\OperatorCoupled_{\Corr}: \mathcal{X} \to \mathcal{X}$)---linking internal model components to semantic structure within one model---we use the shorthand $\gls{OperatorSameSpace}_{\Corr}: \mathcal{X} \to \mathcal{X}$ for notational convenience; for same-space operators the correspondence $\Corr$ pairs elements of one representation space with an external reference (text encodings, unembedding directions, a target distribution) rather than elements of two spaces. This pattern appears in methods like ResiDual (see \cref{ch:residual}), which aligns residual stream components to task attributes, HeadPursuit (see \cref{ch:headpursuit}), which discovers head-to-concept correspondences, and LinEAS (see \cref{ch:lineas}), which aligns activation distributions within a single model. The underlying principle remains correspondence-coupled transformation; $\OperatorSameSpace_{\Corr}$ simply denotes the special case where domain and codomain coincide.

\section{Empirical Framework and Thesis Scope}
\label{sec:empirical-framework}

\subsection*{Hypothesis}

The correspondence framework presented in this chapter rests on a central hypothesis validated empirically across fourteen research works spanning diverse domains and architectures:

\begin{quote}
\textit{When neural networks learn from semantically related data with sufficiently general training objectives, their intrinsic geometric structures converge. Establishing appropriate semantic correspondence between observation spaces reveals this convergence, exposing geometric properties that enable structure-preserving transformation.}
\end{quote}

This is an empirically robust phenomenon observed consistently across settings. It is not a formal theorem with provable sufficiency conditions. We demonstrate this principle across domains including vision, language, reinforcement learning, multimodal learning, and neuroscience; architectures ranging from CNNs and ResNets to Transformers, Vision Transformers, and graph networks; scales from small supervised models trained on modest datasets to large foundation models with billions of parameters; and tasks spanning classification, generation, retrieval, policy learning, and neural decoding. The consistency of this phenomenon across such diverse settings suggests a deep principle governing how neural networks represent semantic structure.

\subsection*{What We Demonstrate}

Across these diverse settings, the thesis establishes several key empirical findings that validate the correspondence framework.

First, correspondence reliably reveals exploitable geometric structure. Semantic correspondence consistently exposes geometric properties---angular relationships, metric structure, spectral frequencies, sparse patterns, distributional characteristics---that enable transformation between models without retraining or joint optimization. This reliability across different types of geometric structure suggests that correspondence acts as a fundamental primitive for representation manipulation.

Second, the two operator families introduced in this chapter cover a broad spectrum of use cases while instantiating the same underlying principle. Correspondence-parallel operators provide modularity and efficient scaling to many models, making them ideal for model ecosystems and multi-target scenarios. Correspondence-coupled operators provide direct decoder compatibility, enabling generation and policy execution. Despite their different trade-offs, both families exploit correspondence-revealed structure, demonstrating that the framework accommodates diverse application requirements within a unified conceptual lens.

Third, task-conditional correspondences extend the framework's reach to diverse applications. By choosing correspondence structure to match task intent---whether cross-modal retrieval requires concept alignment across modalities, style transfer requires instance pairing with and without target properties, or policy stitching uses shared anchor observations---we apply the same principle to qualitatively different problems. The correspondence choice encodes the task objective, making the framework adaptable across tasks.

Fourth, exploiting correspondence-revealed structure yields sample efficiency gains. Methods that use correspondence typically require few pairs (\cref{ch:bootstrapping,ch:lfm}) or few unpaired samples (\cref{ch:lineas}), showing how semantic structure helps the optimization process when applied or lets it be sidestepped entirely.

Fifth, we identify boundary conditions and failure modes, providing practical guidance for when methods succeed and when they fail. Insufficient correspondence---too few pairs, weak semantic relationships, noisy alignments---produces unreliable transformations (on \gls{cifart}, retrieval with only the seed anchors of \cref{ch:bootstrapping} falls far below retrieval with the full anchor set, \cref{tab:bootstrapping-cifar}). Violated geometric assumptions---assuming metric preservation when only angular structure aligns, or vice versa---yield poor results (across architectures, the zero-shot product of invariances can score below a single metric; \cref{sec:product-invariances:conclusions}). Architectural incompatibility---attempting to align models that learned fundamentally different semantic structures due to divergent task objectives---cannot be overcome by correspondence alone. By characterizing these failure modes empirically (see also the limits stated in \cref{sec:uncertainty:discussion}), we provide practitioners with realistic expectations and diagnostic tools.

\subsection*{What We Do NOT Claim}

This thesis does not provide formal sufficiency conditions guaranteeing when correspondence will reveal alignable structure. We do not prove theorems of the form: ``If architecture satisfies a given constraint, data satisfies a given property, and objective satisfies a given condition, then correspondence $\Corr$ reveals alignable structure with probability $p$.'' Such formal guarantees would require theoretical machinery that does not yet exist for understanding the geometric properties of learned neural representations.

What we provide is mathematical precision in defining the framework itself. We rigorously define correspondence $\Corr$ and characterize what semantic relationships it establishes between observation spaces. We precisely specify the operator families $\OperatorParallel_{\Corr}^{\mathcal{X}}$ and $\OperatorCoupled_{\Corr}: \mathcal{X} \to \mathcal{Y}$, their domain-codomain structure, and their composition patterns. We mathematically formulate the transformations that exploit correspondence to map between representation spaces. We characterize the geometric invariances each method preserves, whether angular, metric, spectral, sparse, or distributional. The framework components are mathematically well-defined.

However, determining whether the framework applies successfully to a specific model pair remains empirical. We cannot predict a priori whether two given models will have sufficiently similar intrinsic geometry for correspondence to enable useful transformation. Each method chapter therefore follows consistent empirical methodology: stating correspondence assumptions explicitly, validating those assumptions on real problems with authentic experimental results, discussing the failure modes and boundary conditions it observes (for example \cref{sec:product-invariances:conclusions,sec:uncertainty:discussion}), and providing practical guidance for when to use or avoid the method based on observed performance patterns.

The contribution of this thesis is systematic empirical demonstration. It does not offer formal proof. We show that correspondence-enabled transformation works reliably enough, across enough diverse domains and architectures, with sufficient sample efficiency, to constitute a general principle for representation manipulation. But ``reliably enough'' is validated through extensive experimentation per application domain; theoretical analysis does not formally guarantee it. This empirical stance is both honest about the current state of understanding and pragmatically useful for practitioners needing methods that work in practice.

\makeatletter
\let\old@endpart\@endpart
\def\@endpart{\vspace{3cm}}
\makeatother

\part{Semantic Coordinates: Parallel Operators}

\begin{center}
\includegraphics[width=0.6\textwidth]{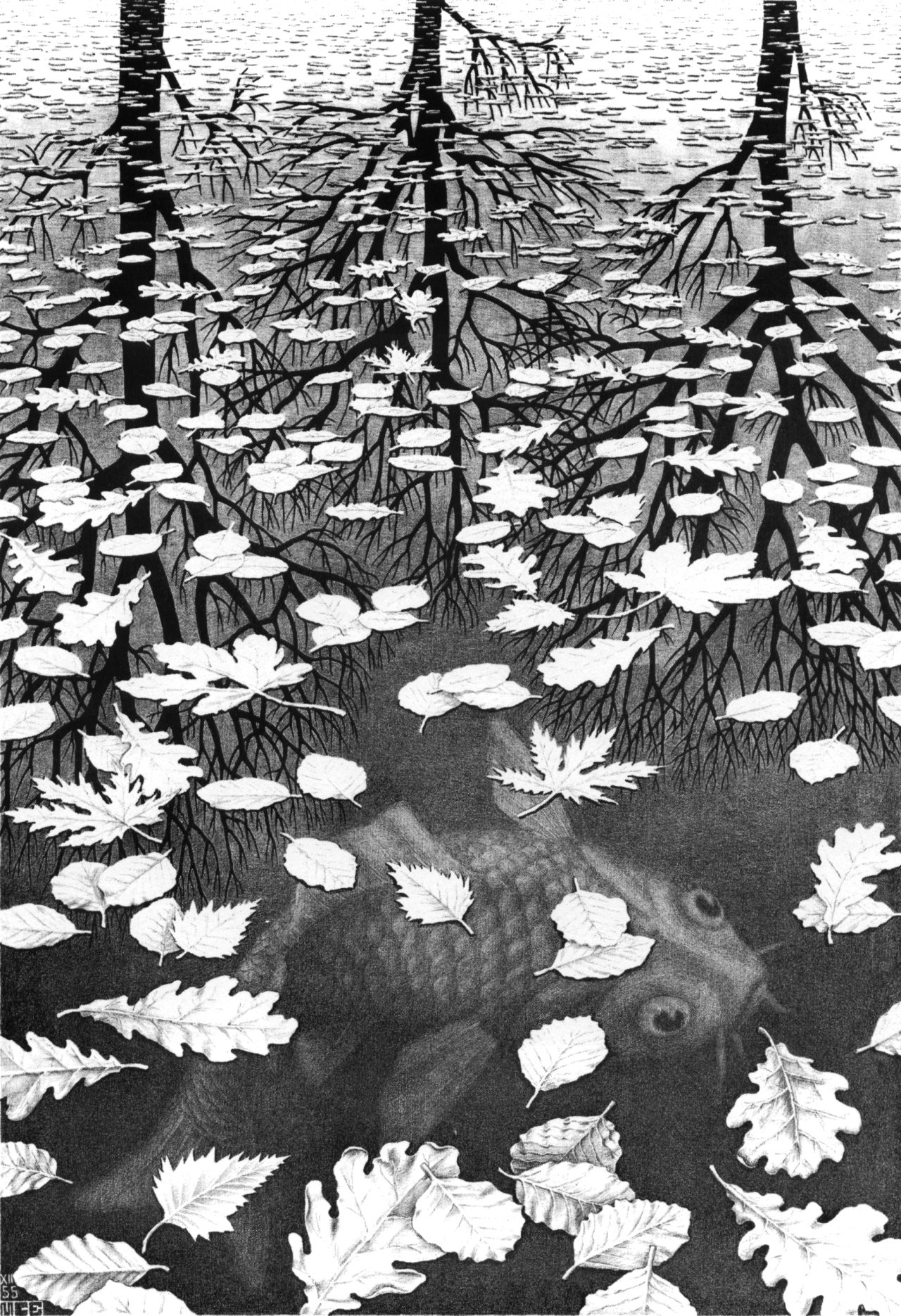}
\captionof*{figure}{\textbf{Three Worlds} by M.C. Escher (1955). Three representations---reflection, surface, and depth---coexist in a single manifold, illustrating independent projections onto shared semantics.}
\end{center}
\vfill
\begin{center}\footnotesize\textcopyright{} The M.C. Escher Company, The Netherlands. All rights reserved. Reproduced for academic, non-commercial use.\end{center}

\makeatletter
\let\@endpart\old@endpart
\makeatother

\clearpage

Two networks trained apart describe the same concepts in coordinates that have nothing in common, much as two languages name the same things with different words. What the two share is the relations among the things they represent, and a correspondence, a set of elements known to mean the same in both, is enough to expose those relations. The operators of this part do exactly that and nothing more: each model describes its own space relative to the shared elements, and the descriptions can then be compared, composed and transferred across models, architectures and modalities, without any model ever seeing another and without training anything jointly.

Neural representations come from optimization in arbitrary coordinate systems chosen by random initialization, stochastic gradient descent trajectories, and architectural constraints. These extrinsic coordinates obscure the intrinsic geometric structure determined by semantic content in the data. Correspondence-parallel operators transform extrinsic coordinates to semantic coordinates---intrinsic representations anchored to shared semantic relationships, independent of the coordinate frame that optimization happened to produce.

These operators transform spaces independently: each model uses only its side of the correspondence, projecting onto its own anchors or extracting its own geometric structure without accessing other spaces. The outputs live in auxiliary semantic space where geometric relationships become directly comparable across independently trained systems. This design offers modularity---one semantic space can interface with many models, requiring a number of transformations linear in the number of models; aligning every pair would require a quadratic number. The cost is decoder incompatibility: semantic coordinates live in auxiliary space, separate from the target models' native coordinates, requiring decoders trained explicitly on semantic representations.

\Cref{ch:relrep} introduces the foundational operator, relative representations, which describes each sample by its angles to a set of anchors in correspondence. \cref{ch:product-invariances} composes several such projections, each invariant to a different class of transformations, into a product space in which the task selects the relevant geometry. \cref{ch:bootstrapping,ch:asif} turn to the correspondence itself: the former expands a small seed of anchor pairs into a full correspondence, the latter scales it to millions of image--text pairs to assemble multimodal models without training. \cref{ch:learning_to_align} removes the need for paired samples altogether by learning the alignment through optimal transport, and \cref{ch:domain-applications} carries the same operators to neuroscience, ensemble uncertainty, and reinforcement learning.

\chapter[Relative Representations]{Relative Representations Enable Zero-Shot Latent Space Communication}
\label{ch:04}
\label{ch:relrep}
\label{ch:relative-representations}

\chapterpublication[https://github.com/lucmos/relreps]{moschella2023relative}{\textbf{ICLR 2023} \textit{(\textcolor{blue}{Oral, notable-top-5\%})}}{A data correspondence across latent spaces can condition a transformation that enables zero-shot alignment between them.}

\bigskip

\section{Introduction}

This chapter presents \emph{relative representations} (\gls{relrep}), the foundational \gls{correspondence-parallel} operator that enables zero-shot communication between independently trained models.

\begin{figure}[t]
 \centering
 \begin{overpic}[trim=0cm .3cm 0cm -.1cm,clip,width=.95\linewidth]{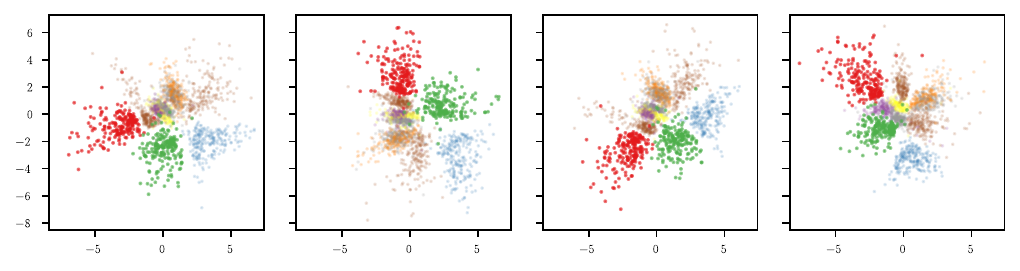}
   \put(12, 24.25){\small Train 1}
   \put(37, 24.25){\small Train 2}
   \put(61, 24.25){\small Train 3}
   \put(85, 24.25){\small Train 4}
 \end{overpic}
 \caption{Latent spaces learned by distinct trainings of the same AE on the \gls{mnist} dataset.
 The bottleneck has size $2$, thus there is no dimensionality reduction involved in the visualization of the latent space.
 The stochasticity in the training phase induces intrinsically similar representations.}
 \label{fig:latent-rotation}
\end{figure}

As established in \cref{ch:methodology}, neural networks trained on similar data with similar objectives discover similar semantic relationships but parameterize them through incompatible coordinate systems due to \glspl{extrinsic-factor}. Training the same autoencoder with a 2-dimensional bottleneck on \gls{mnist} several times from scratch yields latent spaces that differ from one another (\cref{fig:latent-rotation}). Absolute coordinates shift with each training run, yet angular relationships between points remain approximately the same across all realizations: the spaces are equal only up to a nearly isometric transformation.

The core assumption of the method explains why angular relationships persist when absolute coordinates vary: optimization stochasticity---random initialization, data shuffling, hyperparameters and other stochastic factors---induces angle-preserving transformations (rotations, reflections and rescaling) on the learned representation space. These transformations preserve angles between vectors but scramble absolute positions. Cosine similarity factors out these transformations by construction, measuring only the angle between representations regardless of coordinate frame.

However, cosine similarity compares two vectors of the same space, and direct comparison between independently trained spaces remains notoriously challenging: which neurons in model A correspond to which neurons in model B? The solution is to establish a shared reference frame using anchors in correspondence (\gls{Corr}) across the two spaces---the same samples, or parallel ones such as translations. By computing angular relationships to these anchors, \glspl{semantic-coordinate} are obtained that are directly comparable across models.

Relative representations implement this through angular projection onto $\nanch$ shared anchors: $\vr = \relproj(\vx) = [\cos(\vx, \va_1), \dots, \cos(\vx, \va_\nanch)]$, where $\vx = \Enc[X](\obs)$ is the embedding of an observation $\obs$ (all defined in \cref{sec:relative-representations} below). This \gls{correspondence-parallel} operator $\gls{OperatorParallel}_{\Corr}^{\mathcal{X}}$ inherits this invariance, revealing the intrinsic angular geometry. Each dimension measures semantic proximity to a shared reference point, yielding coordinates that generalize across architectures, random seeds, languages and datasets, and, with parallel data, even modalities (\cref{ch:asif}).
The method's power lies in its simplicity: no optimization, no learned parameters, no iterative refinement---just matrix multiplication computing cosine similarities.

Building on the alignment methods reviewed in \cref{ch:background}, \gls{relrep} demonstrates that \gls{correspondence-parallel} transformation using angular invariances enables: (1) zero-shot stitching between models trained with different seeds or architectures, (2) zero-shot stitching across languages and datasets through parallel and out-of-distribution anchors, and (3) a quantitative differentiable measure of model performance correlated with standard downstream metrics.

These capabilities establish the foundation for extensions in \crefrange{ch:05}{ch:08} (\gls{correspondence-parallel} variants) and enable the bridge to \gls{correspondence-coupled} operators (\cref{ch:10}).

\subsection{Relative Representations}\label{sec:relative-representations}
Given a training set $\Data \subset \Obs$, a neural network learns an embedding function $\Enc[X]: \Obs \to \mathcal{X} \subseteq \R^d$ with parameters $\theta$, which maps each sample $\obs_i \in \Data$ to its latent representation, or \textbf{absolute representation}, $\vx_i = \Enc[X](\obs_i)$.

\begin{figure}[h]
 \centering
 \begin{overpic}[trim=0cm 1.7cm 0cm 1.8cm,clip,width=0.9\linewidth, ]{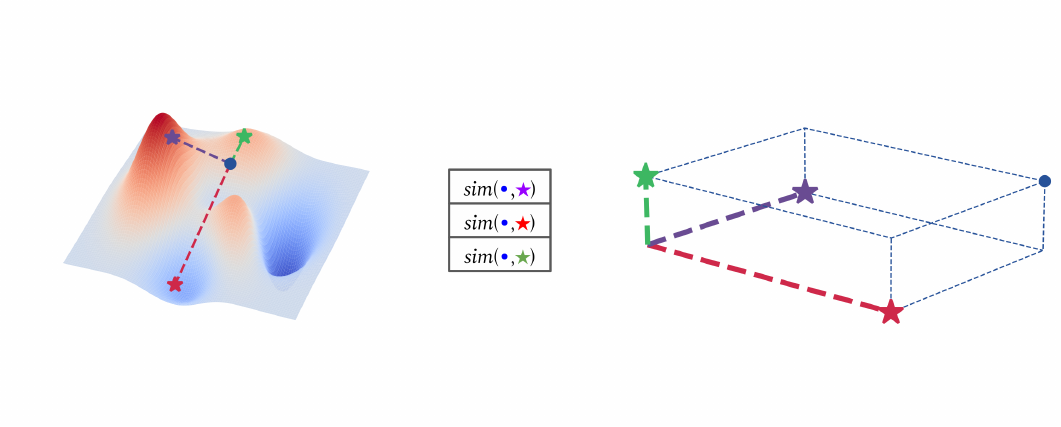}
  \put(35, 8){\huge $\Rightarrow$}
  \put(53, 8){\huge $\Rightarrow$}
 \end{overpic}
 \caption{{\em Left}: Three anchors (colored stars) are selected on the data manifold; given a point on the manifold (blue dot), its similarity w.r.t. the three anchors is computed, yielding a vector of dimensionality 3 (middle). {\em Right}: Each dimension is treated as coefficients in a coordinate system defined by the anchors. Anchors are orthogonal in this example only for visualization purposes.}
 \label{fig:relative-distances}
\end{figure}

\looseness=-1To build the representation, a subset $\anchors$ of the training data $\Data$ is selected, denoted as \textbf{anchor} samples.
Every sample in the training distribution will be represented with respect to the embedded anchors
$\va_j = \Enc[X](a_j) \text{ with } a_j \in \anchors$.
As a measure capturing the relation between the anchors and the other samples, we consider a generic similarity function $\simf:\R^d \times \R^d \to \R$, yielding a scalar score $\simf(\vx_i, \vx_{i'})$ between two absolute representations.
Given the $\nanch = \lvert\anchors\rvert$ anchors in an arbitrary ordering $a_1, \dots, a_\nanch$, we define the \textbf{relative representation} of $\obs_i \in \Data$ as:
\begin{align}
\vr_i = \relproj(\vx_i) =
[
\simf(\vx_i, \va_1),
\simf(\vx_i, \va_2),
\dots,
\simf(\vx_i, \va_\nanch)
] \,.
\end{align}
\Cref{fig:relative-distances} illustrates the key differences between absolute and relative representations.

\vspace{1ex}\noindent\textbf{Choice of the anchors.}
Anchors directly affect the expressivity of the relative representation space, and are related to the task at hand. For example, in a classification task, we should sample anchors from each class in the training set, in order to well represent each data sample in $\Data$.

One case of interest arises when the data comes from different domains or modalities $\Obs[X]$, $\Obs[Y]$, and a partial correspondence $\Corr \subseteq \Obs[X] \times \Obs[Y]$ relates a subset $\CorrSide{X}$ of $\Obs[X]$ to a subset $\CorrSide{Y}$ of $\Obs[Y]$. In this case, anchors $\anchors[X] \subseteq \CorrSide{X}$ can be sampled, and corresponding anchors on the other domain obtained directly as $\anchors[Y] = \{\obs' : (\obs, \obs') \in \Corr,\ \obs \in \anchors[X]\}$ \citep{norelli2023asif}. These are referred to as \emph{parallel anchors}. An example of parallel anchors appears in \cref{sec:nlp-app}, where $\Obs[X]$ and $\Obs[Y]$ are Amazon reviews in two different languages.

The choice of the anchors is not restricted to elements in the training distribution. Given an encoder pretrained on a fixed training distribution, elements from a set $\anchors^{\mathrm{ood}}$ that is out-of-domain w.r.t. $\Data$ can be picked, and the relative representations built on top of $\anchors^{\mathrm{ood}}$. These are referred to as \emph{OOD anchors} and exploited, e.g., to solve domain adaptation tasks where access to correspondence is unavailable and data labels are scarce. Real-world examples appear in \cref{sec:nlp-app,sec:cv-app}. For a preliminary analysis of different anchor selection strategies, see the ``Anchors analysis'' section of the appendix of \citet{moschella2023relative}.

\vspace{1ex}\noindent\textbf{Achieving latent invariance.}
The cosine similarity is chosen as the similarity function due to the properties it induces on the relative representation.
The cosine similarity $\cos(\cdot,\cdot)$ is the dot product of unit vectors, corresponding to the cosine of the angle $\alpha$ between the two:
\begin{align}
 \cos(\vu, \vv) = \frac{\vu^\top \vv}{\lVert \vu \rVert \lVert \vv \rVert} = \cos\alpha \,.
\end{align}
The value $\cos\alpha$ does not change if we apply the same angle-preserving transformation $\mT$ to two vectors $\vu$ and $\vv$,
 i.e., the cosine similarity is invariant to rotations, reflections, and rescaling.
While this is not true for translations, NNs commonly employ normalization techniques (e.g., InstanceNorm \citep{instance-norm}) to center the latent spaces. Under this assumption, cosine similarity guarantees a relative representation $\vr_i$ invariant to angle-preserving transformations.

This means the encoder $\Enc[X]$ can be replaced by any other encoder $\Enc[Y]$ that produces different representations with same angles, i.e.:
\begin{align}
[
\cos(\vx_i, \va_1),
\dots,
\cos(\vx_i, \va_\nanch)]=[
\cos(\vy_i, \va'_1),
\dots,
\cos(\vy_i, \va'_\nanch)
] \,,
\end{align}
where $\vy_i = \Enc[Y](\obs_i) = \mT \vx_i$ and $\va'_j = \Enc[Y](a_j) = \mT\va_j$, and $\mT$ is an arbitrary angle-preserving transformation.
A practical application of this invariance is the possibility of comparing latent spaces across multiple trainings, and re-using models as demonstrated in \cref{sec:latent-communication,sec:model-reusability}.

We remark that other choices of similarity function can be made to enforce different invariances into the representation. For example, one may impose invariance to non-isometric deformations with bounded distortion.
This was not necessary in the experiments of this thesis, as NNs that generalize sufficiently well can handle small perturbations. Nevertheless, this invariance can be enforced by design with vector quantization algorithms, leaving further exploration to future work.

\section{Latent Space Communication}\label{sec:latent-communication}
This section demonstrates how relative representations can effectively be used to produce latent spaces that are stable under a variety of factors. The central question is: Given two different learning models that are trained separately on different data, can their latent embeddings be compared? This assumes that the two models are trained on a similar phenomenon, e.g., on two different samplings of the English language or on two different modalities.

\looseness=-1The answer is yes: the angular invariance enables effective communication between different but semantically equivalent latent spaces. %
The analysis shows how different word embedding spaces, once projected onto relative representations, are intrinsically the same (\cref{sec:word-embeddings}); how the relative projection preserves the metric of each space (\cref{sec:metric-preservation}); how the similarity between the relative counterparts of two or more embedding spaces is a good predictor of model performance (\cref{sec:manifold-performance}); and confirms that relative representations in the training phase are not detrimental to performance (\cref{sec:abs-vs-rel}).

\subsection{Word Embeddings}\label{sec:word-embeddings}
\vspace{1ex}\noindent\textbf{Experimental setting.}
Two different word embeddings on the English language are selected, namely \texttt{FastText} \citep{bojanowski2016enriching} and \texttt{Word2Vec} \citep{mikolov2013word2vec}. Both models are pretrained on different data, but partly share a vocabulary from which 20K words are extracted. Using $300$ randomly drawn parallel anchors, each embedding space is converted to a relative one.
\Cref{fig:latent-rotation-comparison} (left) shows the original and the relative embeddings.

For each word $\obs$, its corresponding encodings $\vx \in \mathcal{X}$ and $\vy \in \mathcal{Y}$ in the source and target space are considered; the metrics are computed on $\vx, \vy$ in the absolute setting and on $\relproj_{\anchors[X]}(\vx), \relproj_{\anchors[Y]}(\vy)$ in the relative setting. Three different metrics measure their similarity (in a setting similar to \citet{vulic-etal-2020-good}):
(i) \emph{Jaccard}: the discrete Jaccard similarity between the set of neighbors of $\vx$ in $\mathcal{X}$ and of $\vy$ in $\mathcal{Y}$;
(ii) \emph{Mean Reciprocal Rank}: measures the (reciprocal) ranking of $\obs$ among the top-$k$ neighbors of $\vx$ in the target space;
(iii) \emph{Cosine}: measures the cosine similarity between $\vx$ and $\vy$.
For implementation details, see the ``Implementation Details'' section of the appendix of \citet{moschella2023relative}.

\begin{table}[h]

\noindent\makebox[\textwidth][c]{%
 \begin{minipage}{.30\textwidth}
  \begin{overpic}[trim=-1cm 1cm -0.5cm 0cm,width=1\linewidth]{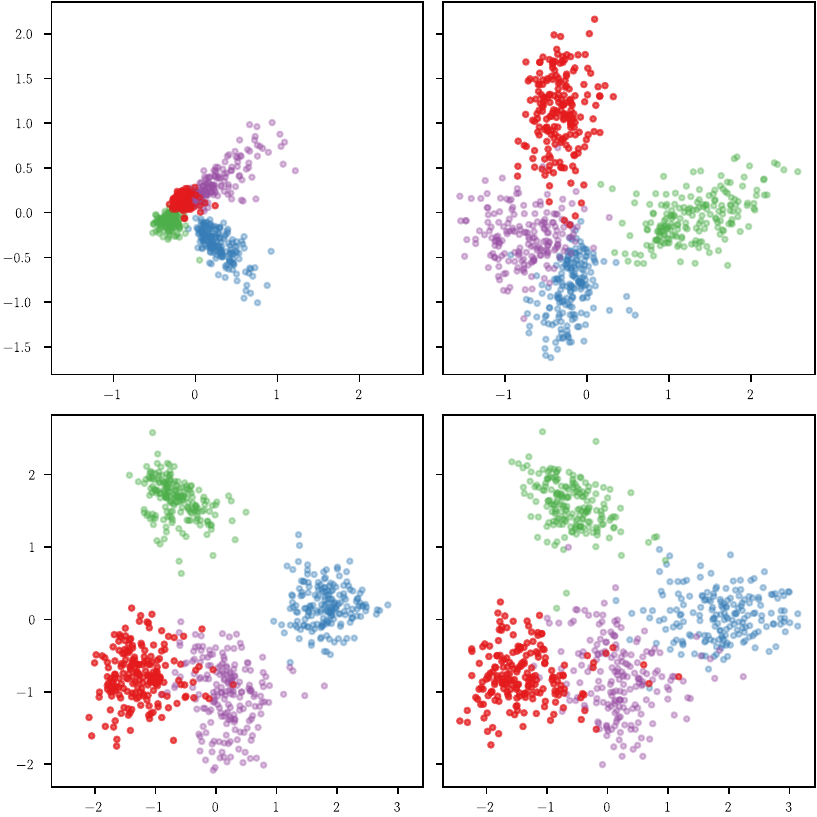}
   \put(11.5, 88){\texttt{FastText}}
   \put(55.5, 88){\texttt{Word2Vec}}
   \put(-1, 50){\rotatebox{90}{\small Absolute}}
   \put(-1, 5){\rotatebox{90}{\small Relative}}
  \end{overpic}
 \end{minipage}%
 \hspace{0.1cm}%
 \begin{minipage}{.65\textwidth}
  \small
  \setlength{\tabcolsep}{4pt}
  \begin{tabular}{llllll}
  \toprule
   & Source & Target & \multicolumn{1}{c}{Jaccard ↑} & \multicolumn{1}{c}{MRR ↑} & \multicolumn{1}{c}{Cosine ↑} \\
  \midrule
  \multirow{4}{*}{{\STAB{\rotatebox[origin=c]{90}{Absolute}}}} & \multirow{2}{*}{\texttt{{FT}}} & \texttt{{FT}} & $1.00 \pm 0.00$ & $1.00 \pm 0.00$ & $1.00 \pm 0.00$ \\
       &       & \texttt{{W2V}} & $0.00 \pm 0.00$ & $0.00 \pm 0.00$ & $0.01 \pm 0.00$ \\[0.5ex]
       & \multirow{2}{*}{\texttt{{W2V}}} & \texttt{{FT}} & $0.00 \pm 0.00$ & $0.00 \pm 0.00$ & $0.01 \pm 0.00$ \\
       &       & \texttt{{W2V}} & $1.00 \pm 0.00$ & $1.00 \pm 0.00$ & $1.00 \pm 0.00$ \\
  \cmidrule(lr){2-6}
     \multirow{4}{*}{\STAB{\rotatebox[origin=c]{90}{Relative}}} & \multirow{2}{*}{\texttt{{FT}}} & \texttt{{FT}} & $1.00 \pm 0.00$ & $1.00 \pm 0.00$ & $1.00 \pm 0.00$ \\
       &       & \texttt{{W2V}} & $0.34 \pm 0.01$ & $0.94 \pm 0.00$ & $0.86 \pm 0.00$ \\[0.5ex]
       & \multirow{2}{*}{\texttt{{W2V}}} & \texttt{{FT}} & $0.39 \pm 0.00$ & $0.98 \pm 0.00$ & $0.86 \pm 0.00$ \\
       &       & \texttt{{W2V}} & $1.00 \pm 0.00$ & $1.00 \pm 0.00$ & $1.00 \pm 0.00$ \\
  \bottomrule
  \end{tabular}

  \vspace{0.4cm}
 \end{minipage}%
}
\caption{Qualitative \emph{(left)} and quantitative \emph{(right)} comparisons of English word embeddings using absolute and relative representations. PCA is applied only for visualization. All metrics are calculated with $k=10$ averaged over 20K words and across 10 different random seeds. For visualizations with other dimensionality reductions and similar experiments on \gls{cifart}, see the ``Anchors analysis'' and ``Additional results'' sections of the appendix of \citet{moschella2023relative}.}
\label{fig:latent-rotation-comparison}
\end{table}

\vspace{1ex}\noindent\textbf{Result analysis.}
\Cref{fig:latent-rotation-comparison} \emph{(left)} highlights clusters of semantically similar words and shows that the absolute representations are incoherent across the two latent spaces, while the relative embeddings are highly similar.
The average Jaccard similarity reported in \cref{fig:latent-rotation-comparison} \emph{(right)}, says that the word neighborhoods of the relative representations overlap by 34\% on average in one direction and 39\% in the other (the remaining 61--66\% is due to semantic differences, that are not taken into account by the discrete nature of the Jaccard metric). By contrast, the absolute neighborhoods share no words (Jaccard score equal to zero); for a match to happen, it would mean that the \texttt{FastText} and \texttt{Word2Vec} embeddings of a given English word are almost the same, which is highly unlikely.
MRR, close to a perfect score for the relative representations, shows that the most-similar word to a given one is usually itself, even if their cosine similarity doesn't reach 1.

Overall, these results show that relative representations are preserved across different word embedding models, validating the core assumption of the method.

\subsection{Preservation of Metric Structure}\label{sec:metric-preservation}
The previous subsection demonstrated that relative representations enable direct comparison across different word embedding models. This subsection analyzes how the transformation preserves the underlying metric structure of the absolute spaces.

Let $\mX \in \R^{n\times \dimof{X}}$ and $\mY \in \R^{n\times \dimof{Y}}$ stack the absolute embeddings of two spaces $\mathcal{X}$, $\mathcal{Y}$ (\texttt{FastText} and \texttt{Word2Vec}), with corresponding relative embeddings $\mX_{\mathrm{rel}} \in \R^{n\times \nanch}$ and $\mY_{\mathrm{rel}} \in \R^{n\times \nanch}$, where $\nanch$ is the number of anchors. While \cref{fig:latent-rotation-comparison} showed that $\mathcal{X}$ and $\mathcal{Y}$ become well-aligned in the relative space, it remains to verify whether self-similarities within each space are preserved by the relative transformation.

\begin{figure}[h]
    \centering
    \includegraphics[width=.6\linewidth]{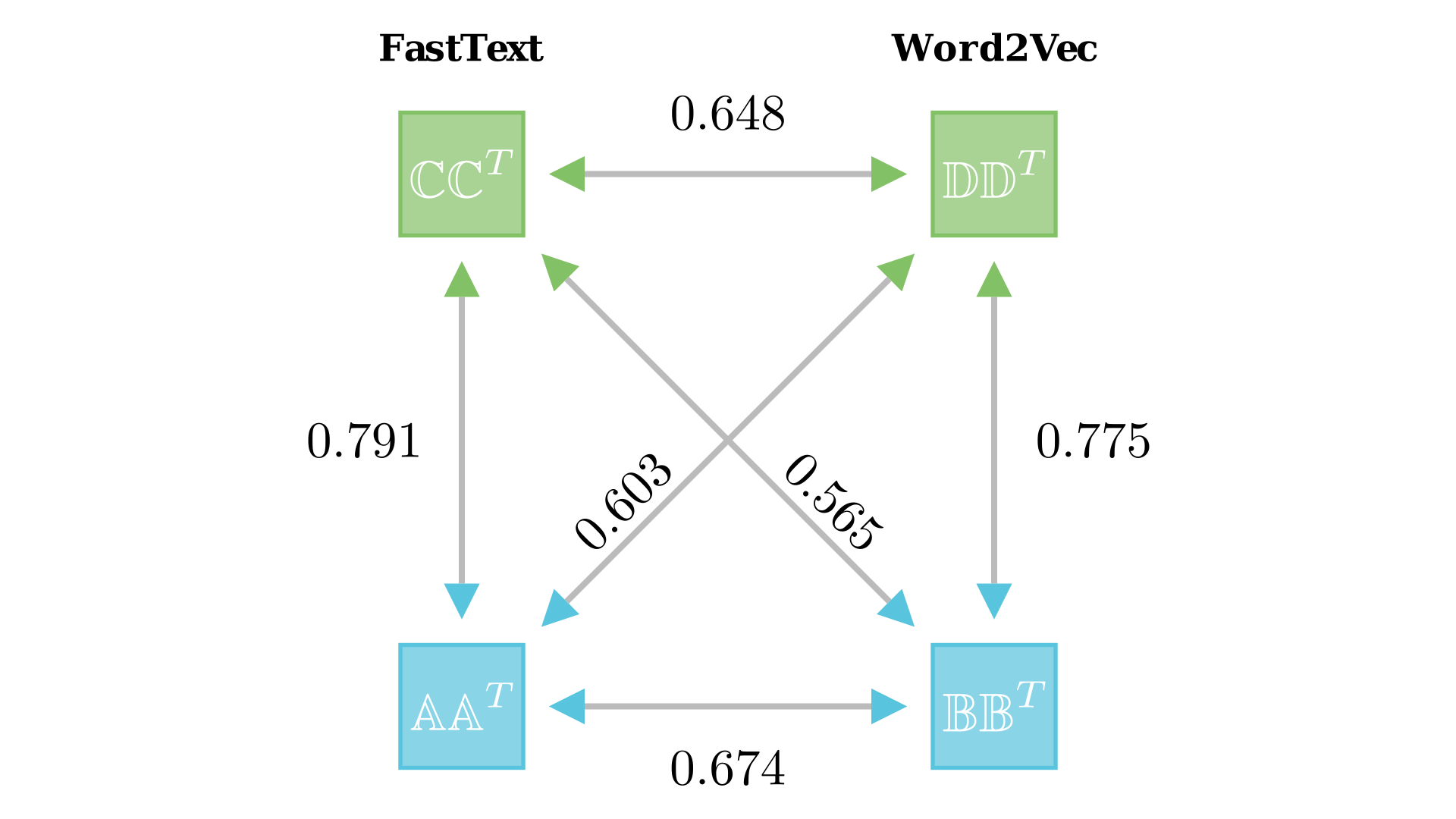}
    \caption{Self-similarity correlations between absolute and relative spaces for \texttt{FastText} and \texttt{Word2Vec}. Blue nodes represent self-similarities in the absolute embeddings $\mX$, $\mY$; green nodes represent relative spaces $\mX_{\mathrm{rel}}$, $\mY_{\mathrm{rel}}$ (in the figure, $\mathbb{A}$, $\mathbb{B}$ are $\mX$, $\mY$ and $\mathbb{C}$, $\mathbb{D}$ are $\mX_{\mathrm{rel}}$, $\mY_{\mathrm{rel}}$). Vertical arrows indicate how much the underlying metric in the absolute space is preserved by the relative coordinate transformation, measured via Pearson correlation coefficient. Horizontal and diagonal arrows give the same correlation between the two models' (absolute or relative) self-similarities.}
    \label{fig:rel-space-correlations}
\end{figure}

\Cref{fig:rel-space-correlations} quantifies this preservation by measuring correlations between pairwise similarities within each space before and after the relative transformation. The vertical arrows in the figure show Pearson correlation coefficients between absolute and relative self-similarities, indicating how much of the geometric structure is retained through the coordinate transformation. Besides enabling latent space communication across models, relative representations preserve the underlying absolute latent space metric up to a certain degree (Pearson correlation 0.79 for \texttt{FastText} and 0.78 for \texttt{Word2Vec}).

This preservation property matters for practical applications: the correlations suggest that semantic relationships encoded in the absolute geometry remain accessible after the relative projection.

\subsection{Latent distance as a performance proxy}\label{sec:manifold-performance}
\looseness=-1\vspace{1ex}\noindent\textbf{Experimental setting.}
This experiment considers a node classification task on the \gls{cora} graph dataset~\citep{cora}. First, a \emph{reference} model that achieves good accuracy on a validation set is trained. Then, $\approx2000$ models are trained with various combinations of seed, number of epochs, number of layers, dropout probability, activation functions, optimizer type, learning rate or type of graph embedder.
All the models are classically trained using absolute representations, which are converted to relative post-training
by projecting the embeddings onto $300$ randomly drawn but fixed anchors. For each model, its classification accuracy is measured and the similarity of its space with the reference one computed. This similarity is computed as the average cosine similarity between the node embeddings produced by a given model and the corresponding embeddings in the reference.

\begin{figure}[h]
 \centering
 \begin{overpic}[trim=-.5cm -.6cm -.5cm 0cm,clip,width=.73\linewidth]{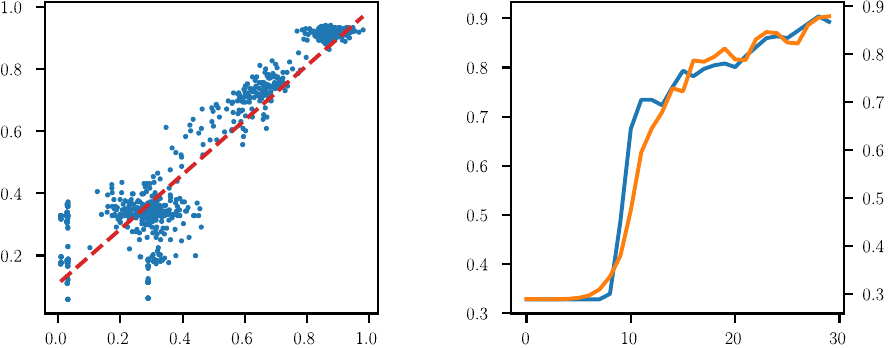}
 \put(-1, 14){\rotatebox{90}{Performance}}
 \put(19, -2){Similarity}

 \put(49, 14){\rotatebox{90}{\color{pltblue}Performance}}
 \put(71, -2){Epochs}
 \put(97.5, 16){\rotatebox{90}{\color{pltorange}Similarity}}
 \end{overpic}
 \caption{Graph node classification task on \gls{cora}. \emph{Left:} Correlation between the performance of $\approx2000$ models and the similarity of their latent spaces with respect to a well-performing reference model.
 \emph{Right:} The same correlation plotted over time. The mean Pearson correlation over all models is $0.955$, after filtering out the models having best validation accuracy below $0.5$.}
 \label{fig:data-manifold}
\end{figure}

\vspace{1ex}\noindent\textbf{Result analysis.}
The scatter plot in \cref{fig:data-manifold} \emph{(left)} shows that better-performing models tend to be the ones with the latent spaces most similar to the reference. The performance-similarity correlation also holds over time, as shown in \cref{fig:data-manifold} \emph{(right)}. %
For additional correlation examples, see the ``Additional results'' section of the appendix of \citet{moschella2023relative}.
This metric is differentiable, enabling an explicit supervision signal on the latent space, which does not require labeled data and could be readily exploited in a teacher-student framework.

Overall, these results suggest that the similarity between the relative representations of latent spaces is a good proxy to evaluate model performance.

\subsection{Training with Absolute vs. Relative representations}\label{sec:abs-vs-rel}
\vspace{1ex}\noindent\textbf{Experimental setting.}
Architectures that do or do not employ the relative representation while training are compared. In these experiments, the models vary slightly according to the dataset; however, the relative and absolute versions are always comparable in terms of architecture, number of learnable parameters and hyperparameters. The ``Implementation Details'' section of the appendix of \citet{moschella2023relative} and the open-source code provide further implementation details.
Classification tasks on several datasets are considered, spanning the image domain \citep{Lecun1998-we,Xiao2017-er,Krizhevsky2009-hv} and the graph domain \citep{Planetoid}.

\begin{table}[h]
\caption{Performance comparison between relative and absolute representations on several image and graph datasets. The metric is the classification weighted F1 score ($\pm$ std), over 6 seeds (image datasets in \%, graph datasets as fractions).}
\label{tab:abs-rel-performance-comparison}
\tiny
\begin{center}
\begin{tabular}{ccccccccccccc}
\toprule
&\multicolumn{4}{c}{\bf Image Classification} & \multicolumn{3}{c}{\bf Graph Node Classification} \\
\cmidrule(lr){2-5}\cmidrule(lr){6-8}
\multicolumn{1}{c}{}
& \multicolumn{1}{c}{\bf \gls{mnist}}
& \multicolumn{1}{c}{\bf \glsxtrshort{fmnist}}
& \multicolumn{1}{c}{\bf \gls{cifart}}
& \multicolumn{1}{c}{\bf \gls{cifarh}}
& \multicolumn{1}{c}{\bf \gls{cora}}
& \multicolumn{1}{c}{\bf CiteSeer}
& \multicolumn{1}{c}{\bf PubMed}
\\ \midrule
\textbf{Relative} & $97.91 \pm 0.07$ & $90.19 \pm 0.27$ & $87.70 \pm 0.09$ & $66.72 \pm 0.35$ & $0.89 \pm 0.02$ & $0.77 \pm 0.03$ & $0.91 \pm 0.01$ \\
\textbf{Absolute} & $97.95 \pm 0.10$ & $90.32 \pm 0.21$ & $87.85 \pm 0.06$ & $68.88 \pm 0.14$ & $0.90 \pm 0.01$ & $0.78 \pm 0.03$ & $0.91 \pm 0.01$ \\
\bottomrule

\end{tabular}
\end{center}
\end{table}

\vspace{1ex}\noindent\textbf{Result analysis.}
The results, reported in \cref{tab:abs-rel-performance-comparison}, show that relative representations, when used at training time, are not detrimental to performance in general. This is further shown in
\Cref{%
tab:cross-model-monolingual,%
tab:cifar100-imagenet,%
tab:multilingual-en-coarse-grained},
where a subset of the results compares the absolute and relative representations on a variety of domains, datasets and tasks. For comparative results across more datasets and tasks, see the ``Additional results'' section of the appendix of \citet{moschella2023relative}.

Overall, these results show that relative representations are effective when involved in end-to-end training, without large performance drops.

\section{Zero-Shot Model Stitching}\label{sec:model-reusability}
This section illustrates how the latent space communication demonstrated in \cref{sec:latent-communication} enables zero-shot interoperability of pretrained neural components. In previous works, such as \citet{Lenc2014-gy,Bansal2021-oj}, stitching layers are \emph{trainable} linear projections that allow swapping parts of different networks.
In contrast, relative representations enable \emph{zero-shot stitching} of different neural components, treating them as frozen black-box modules.

\begin{figure}[h]
 \centering
 \begin{overpic}[trim=-0.27cm -0.15cm -.3cm 0cm,clip,width=1\linewidth]{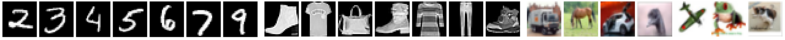}
  \put(0, 2.75){\rotatebox{90}{S}}
 \end{overpic}
 \begin{overpic}[trim=-0.27cm 0cm -.3cm 0cm,clip,width=1\linewidth]{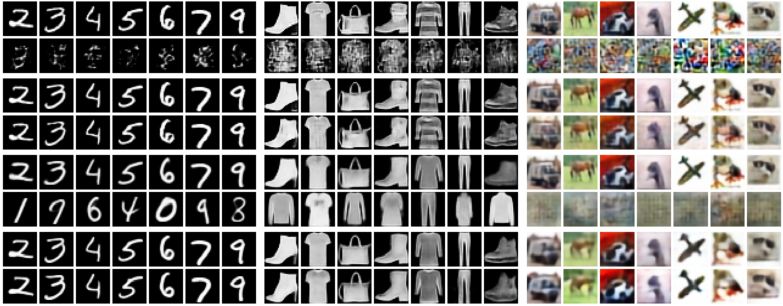}
  \put(0, 31){\rotatebox{90}{\small \textit{Abs.}}}
  \put(0, 21.75){\rotatebox{90}{\small \textit{Rel.}}}
  \put(0, 12.25){\rotatebox{90}{\small \textit{Abs.}}}
  \put(0, 3){\rotatebox{90}{\small \textit{Rel.}}}
  \put(98.5, 26.5){\rotatebox{90}{\small \textbf{AE}}}
  \put(98.5, 7.25){\rotatebox{90}{\small \textbf{VAE}}}
  \put(98, 19.25){\color{black}\line(0,1){17.75}}
  \put(98, .75){\color{black}\line(0,1){17.5}}
 \end{overpic}
 \caption{Reconstruction examples. Each column is a different image. The first row (S) shows the source images. Each block of rows is an architecture (AE, VAE); within it, each pair of rows is a representation (Abs., Rel.), non-stitched first, then stitched.}
 \label{fig:encoder-decoder-swap}
\end{figure}

A generic \emph{stitched model} is defined as the composition of an encoder that embeds data, plus a \emph{relative} decoder specialized in a downstream task (classification, reconstruction).
The stitching operation is always performed without training or fine-tuning, in a zero-shot fashion.
Stitching capabilities are shown across combinations of
different stochasticity sources (\cref{fig:encoder-decoder-swap,tab:cross-train-performance-comparison}),
neural architectures (\cref{tab:cross-model-monolingual,tab:multilingual-en-coarse-grained})
or datasets (\cref{tab:cifar100-imagenet}).
Strong real-world applications in NLP (\cref{sec:nlp-app}) and CV (\cref{sec:cv-app}), e.g., zero-shot predictions on novel languages, are presented.
For implementation details, see the ``Implementation Details'' section of the appendix of \citet{moschella2023relative}.

\subsection{Image Reconstruction}\label{sec:model-reusability-ae}
\looseness=-1\vspace{1ex}\noindent\textbf{Experimental setting.}
Zero-shot stitching is performed with AEs and VAEs trained end-to-end with absolute or relative representations on several datasets.
For each combination of model and dataset, 5 trainings with different seeds are performed, and the resulting encoders and decoders are zero-shot stitched together.

\vspace{1ex}\noindent\textbf{Result analysis.}
In \cref{fig:encoder-decoder-swap} the stitched models that employ absolute representations (\emph{Abs.}) produce erroneous predictions, since the latent spaces obtained from distinct trainings are incompatible.
Although the absolute VAE does not produce compatible latent spaces, it is regularized, thus all embeddings produced by the encoders correspond to wrong but semantically meaningful reconstructions.
Relative representations (\emph{Rel.}) exhibit almost indistinguishable reconstructions between the models trained end-to-end and the stitched ones. Quantitative results are in \cref{tab:cross-train-performance-comparison}.

These results support the claim that relative representations are empirically invariant to training stochasticity.

\begin{table}[h]
\caption{Stitching performance.
The MSE ($\times10^{-3}$, $\pm$ std) between the input images and their reconstructions is computed over 5 different seeds; the last column averages the four datasets. Stitching with relative representations yields an error between 5 and 35 times lower than the absolute counterpart.}
\label{tab:cross-train-performance-comparison}
\scriptsize
\begin{center}
\begin{tabular}{cccrrrr|r}
\toprule
& \multicolumn{1}{c}{\bf }
& \multicolumn{1}{c}{\bf }
& \multicolumn{1}{c}{\bf \gls{mnist}}
& \multicolumn{1}{c}{\bf \glsxtrshort{fmnist}}
& \multicolumn{1}{c}{\bf \gls{cifart}}
& \multicolumn{1}{c}{\bf \gls{cifarh}}
& \multicolumn{1}{c}{\textbf{MSE ↓}}\\
\midrule
\multirow{4}{*}{\STAB{\rotatebox[origin=c]{90}{\textbf{AE}}}} & \multirow{2}{*}{\STAB{\rotatebox[origin=c]{90}{\textit{Abs.}}}} & Non-Stitch. &
$0.66 \pm 0.02$ & $1.57 \pm 0.03$ & $1.94 \pm 0.08$ & $2.13 \pm 0.08$ & $1.58 \pm 0.05$ \\
& & Stitch. &
$97.79 \pm 2.48$ & $120.54 \pm 6.81$ & $86.74 \pm 4.37$ & $97.17 \pm 3.50$ & $100.56 \pm 4.29$ \\[1ex]
& \multirow{2}{*}{\STAB{\rotatebox[origin=c]{90}{\textit{Rel.}}}} & Non-Stitch. &
$1.18 \pm 0.02$ & $3.59 \pm 0.04$ & $2.83 \pm 0.13$ & $3.50 \pm 0.08$ & $2.78 \pm 0.07$ \\
& & Stitch. &
$2.83 \pm 0.20$ & $6.37 \pm 0.29$ & $5.39 \pm 1.18$ & $18.03 \pm 12.46$ & $8.16 \pm 3.53$ \\
\midrule
\multirow{4}{*}{\STAB{\rotatebox[origin=c]{90}{\textbf{VAE}}}} & \multirow{2}{*}{\STAB{\rotatebox[origin=c]{90}{\textit{Abs.}}}} & Non-Stitch. &
$1.31 \pm 0.04$ & $4.38 \pm 0.03$ & $2.68 \pm 0.06$ & $3.00 \pm 0.03$ & $2.84 \pm 0.04$ \\
& & Stitch. &
$98.51 \pm 1.49$ & $118.96 \pm 2.96$ & $69.02 \pm 1.54$ & $78.57 \pm 1.88$ & $91.27 \pm 1.97$\\[1ex]
& \multirow{2}{*}{\STAB{\rotatebox[origin=c]{90}{\textit{Rel.}}}} & Non-Stitch. &
$2.97 \pm 0.14$ & $6.81 \pm 0.06$ & $5.18 \pm 0.22$ & $5.93 \pm 0.14$ & $5.22 \pm 0.14$ \\
& & Stitch. &
$13.43 \pm 6.79$ & $24.03 \pm 13.15$ & $11.20 \pm 3.15$ & $11.23 \pm 2.38$ & $14.97 \pm 6.37$ \\
\bottomrule
\end{tabular}
\end{center}
\end{table}

\subsection{Text Classification}\label{sec:nlp-app}
This section shows practical examples of the use of parallel anchors (\cref{sec:relative-representations}).

\begin{table}[h]
\scriptsize
\centering
\caption{
Cross-lingual stitching performance comparison.
The table reports the mean weighted F1 ($\pm$ std) and MAE on Amazon Reviews coarse-grained, across 5 seeds.}
\label{tab:multilingual-en-coarse-grained}
\begin{tabular}{clrrrrrr}
\toprule
 & & \multicolumn{2}{c}{Absolute} & \multicolumn{4}{c}{Relative} \\
 \cmidrule(lr){3-4}
 \cmidrule(l){5-8}
 & & \multicolumn{2}{c}{} & \multicolumn{2}{c}{Translated} & \multicolumn{2}{c}{Wikipedia} \\
 \cmidrule(lr){5-6}
 \cmidrule(l){7-8}
 Decoder & Encoder &   \multicolumn{1}{c}{FScore} &    \multicolumn{1}{c}{MAE} &   \multicolumn{1}{c}{FScore} &    \multicolumn{1}{c}{MAE} &   \multicolumn{1}{c}{FScore} &    \multicolumn{1}{c}{MAE} \\
\midrule
\multirow{4}{*}{en} & en & $91.54 \pm 0.58$ & $0.08 \pm 0.01$ & $90.06 \pm 0.60$ & $0.10 \pm 0.01$ & $90.45 \pm 0.52$ & $0.10 \pm 0.01$ \\
 & es & $43.67 \pm 1.09$ & $0.56 \pm 0.01$ & $82.78 \pm 0.81$ & $0.17 \pm 0.01$ & $78.53 \pm 0.30$ & $0.21 \pm 0.00$ \\
 & fr & $54.41 \pm 1.61$ & $0.45 \pm 0.02$ & $78.49 \pm 0.66$ & $0.21 \pm 0.01$ & $70.41 \pm 0.57$ & $0.29 \pm 0.01$ \\
 & ja & $48.72 \pm 0.90$ & $0.51 \pm 0.01$ & $65.72 \pm 0.55$ & $0.34 \pm 0.01$ & $66.31 \pm 0.80$ & $0.34 \pm 0.01$ \\
\bottomrule
\end{tabular}
\end{table}

\begin{table}[h]
\scriptsize
\caption{Cross-architecture stitching performance comparison. The table reports the mean weighted F1 ($\pm$ std) for each dataset, across 5 different seeds.}
\label{tab:cross-model-monolingual}
\begin{center}
\begin{tabular}{llrrrr}
\toprule
& & \multicolumn{1}{c}{\gls{trec}} & \multicolumn{1}{c}{\gls{dbpedia}} & \multicolumn{2}{c}{\gls{amazon}} \\ \cmidrule{5-6}
 & & & & \multicolumn{1}{c}{\textit{Coarse}} & \multicolumn{1}{c}{\textit{Fine}} \\
\midrule
\multirow{2}{*}{{\STAB{\rotatebox[origin=c]{90}{Abs.}}}} & Non-Stitch & $91.70 \pm 1.39$ & $98.62 \pm 0.58$ & $87.81 \pm 1.58$ & $55.35 \pm 3.19$ \\
 & Stitch & $21.49 \pm 3.64$ & $6.96 \pm 1.46$ & $49.58 \pm 2.95$ & $19.01 \pm 2.04$ \\
\midrule
\multirow{2}{*}{{\STAB{\rotatebox[origin=c]{90}{Rel.}}}} & Non-Stitch & $88.08 \pm 1.37$ & $97.42 \pm 2.05$ & $85.08 \pm 1.93$ & $48.92 \pm 3.57$ \\
 & Stitch & $75.89 \pm 5.38$ & $80.47 \pm 21.14$ & $72.37 \pm 7.32$ & $33.24 \pm 7.21$ \\
 \bottomrule
\end{tabular}
\end{center}
\end{table}

\vspace{1ex}\noindent\textbf{Experimental setting.}
Two different text classification settings are considered.

\emph{Cross-lingual}: given a review, the associated star rating is predicted, using multi-lingual data from the Amazon Reviews dataset \citep{amazon-reviews}.
Following \citet{amazon-reviews}, a binarized version of the task is used, with FScore and MAE as metrics. The ``Additional results'' section of the appendix of \citet{moschella2023relative} reports results on the fine-grained formulation.
Four different pretrained language-specific RoBERTa transformers \citep{roberta} are adopted, and their zero-shot stitching performance on languages never seen by the classifier is evaluated.
Parallel anchors are used in two variants: i) \emph{Translated}: English reviews translated\footnote{\texttt{=GOOGLETRANSLATE} function available in Google Sheets.} into the other languages; ii) \emph{Wikipedia}: an external corpus, WikiMatrix \citep{wikimatrix}, providing parallel sentences extracted from Wikipedia.

\emph{Cross-architecture}: assessed on three different datasets: \gls{trec} (coarse) \citep{hovy-etal-2001-toward}, \gls{dbpedia} \citep{zhang2015AGNews}, Amazon Reviews (English split). Two different pretrained BERT \citep{bert} transformers (cased and uncased version), ELECTRA \citep{electra} and RoBERTa are adopted.

\vspace{1ex}\noindent\textbf{Result analysis.}
\Cref{tab:multilingual-en-coarse-grained} and~\cref{tab:cross-model-monolingual} show for the first time that it is possible to learn to solve a downstream task on a specific language or transformer and perform predictions on another.

Stitching with absolute representations yields performances comparable to random guessing across the board, proving that relative representations are a key element for the success of this kind of zero-shot stitching.
Moreover, \cref{tab:multilingual-en-coarse-grained} shows the robustness that relative representations have on the choice of anchors, even when they are noisy (\emph{Translated} case), or their distribution differs from the one of the downstream task (\emph{Wikipedia} case), as long as their encoding can be handled correctly by the encoder. In this case, the encoder is pretrained to represent a variety of texts in a specific language, thus, even if WikiMatrix has a completely different domain from Amazon Reviews, the transformer still computes a meaningful and comparable representation with those of the reviews.
\looseness=-1Complete results on all language combinations and the performance obtained by a multi-lingual transformer are reported in the ``Additional results'' section of the appendix of \citet{moschella2023relative}.
To the best of the authors' knowledge, multi-lingual transformers were the only alternative for obtaining compatible representations across languages.

According to these results, relative representations show invariance to different architectures and data distribution shifts (e.g., different train languages).

\subsection{Image Classification}\label{sec:cv-app}
This section shows practical examples of the use of OOD anchors (\cref{sec:relative-representations}).

\begin{table}[h]
\scriptsize
\centering
\caption{
Stitching performance comparison with different encoding techniques. The table reports the mean weighted F1 ($\pm$ std) on \gls{cifarh} coarse-grained and \gls{imagenet1k}, across 5 seeds.}
\label{tab:cifar100-imagenet}
\begin{tabular}{llrrrr}
\toprule
      &      & \multicolumn{2}{c}{\gls{cifarh}} & \multicolumn{2}{c}{\gls{imagenet1k}} \\\cmidrule(lr){3-4}\cmidrule(lr){5-6}
Decoder & Encoder &  \multicolumn{1}{c}{Absolute}  &   \multicolumn{1}{c}{Relative} &   \multicolumn{1}{c}{Absolute} &   \multicolumn{1}{c}{Relative} \\
\midrule
\multirow{4}{*}{rexnet-100} & rexnet-100 & $82.06 \pm 0.15$ & $80.22 \pm 0.28$ & $73.78 \pm 0.29$ & $72.61 \pm 0.16$ \\
      & vit-base-patch16-224 & \multicolumn{1}{c}{-} & $54.98 \pm 0.44$ & \multicolumn{1}{c}{-} & $37.39 \pm 0.36$ \\
      & vit-base-resnet50-384 & \multicolumn{1}{c}{-} & $53.33 \pm 0.37$ & \multicolumn{1}{c}{-} & $42.36 \pm 0.36$ \\
      & vit-small-patch16-224 & \multicolumn{1}{c}{-} & $59.82 \pm 0.32$ & \multicolumn{1}{c}{-} & $43.75 \pm 0.27$ \\
\cmidrule{1-6}
\multirow{4}{*}{vit-base-patch16-224} & rexnet-100 & \multicolumn{1}{c}{-} & $76.81 \pm 0.49$ & \multicolumn{1}{c}{-} & $30.78 \pm 0.81$ \\
      & vit-base-patch16-224 & $93.15 \pm 0.05$ & $91.94 \pm 0.10$ & $80.91 \pm 0.29$ & $78.86 \pm 0.33$ \\
      & vit-base-resnet50-384 & $6.21 \pm 0.33$ & $81.42 \pm 0.38$ & $0.07 \pm 0.05$ & $44.72 \pm 0.57$ \\
      & vit-small-patch16-224 & \multicolumn{1}{c}{-} & $84.29 \pm 0.86$ & \multicolumn{1}{c}{-} & $48.31 \pm 0.72$ \\
\cmidrule{1-6}
\multirow{4}{*}{vit-base-resnet50-384} & rexnet-100 & \multicolumn{1}{c}{-} & $79.79 \pm 0.43$ & \multicolumn{1}{c}{-} & $53.46 \pm 0.68$ \\
      & vit-base-patch16-224 & $4.69 \pm 0.07$ & $84.46 \pm 0.19$ & $0.08 \pm 0.04$ & $62.21 \pm 0.54$ \\
      & vit-base-resnet50-384 & $91.41 \pm 0.09$ & $90.77 \pm 0.16$ & $82.55 \pm 0.30$ & $81.88 \pm 0.16$ \\
      & vit-small-patch16-224 & \multicolumn{1}{c}{-} & $84.66 \pm 0.16$ & \multicolumn{1}{c}{-} & $61.32 \pm 0.36$ \\
\cmidrule{1-6}
\multirow{4}{*}{vit-small-patch16-224} & rexnet-100 & \multicolumn{1}{c}{-} & $75.35 \pm 0.41$ & \multicolumn{1}{c}{-} & $37.58 \pm 0.44$ \\
      & vit-base-patch16-224 & \multicolumn{1}{c}{-} & $81.23 \pm 0.31$ & \multicolumn{1}{c}{-} & $50.08 \pm 0.63$ \\
      & vit-base-resnet50-384 & \multicolumn{1}{c}{-} & $78.35 \pm 0.69$ & \multicolumn{1}{c}{-} & $45.45 \pm 1.41$ \\
      & vit-small-patch16-224 & $90.07 \pm 0.19$ & $88.85 \pm 0.44$ & $77.73 \pm 0.41$ & $76.36 \pm 0.40$ \\
\bottomrule
\end{tabular}
\end{table}

\vspace{1ex}\noindent\textbf{Experimental setting.}
A classification task on \gls{imagenet1k} and \gls{cifarh} with coarse labels (20) is considered, using 4 different pretrained image encoders:
three variants of the ViT transformer \citep{dosovitskiy2021an} and RexNet \citep{rexnet}.

\vspace{1ex}\noindent\textbf{Result analysis.}
The results in \cref{tab:cifar100-imagenet} highlight how the relative representations allow stitching modules with different encoding dimensionality, since the decoder receives a relative representation with guaranteed equal size. Further, the results demonstrate the ability to generalize and perform zero-shot stitching on \gls{cifarh}, although that data was never seen by the encoder since it is frozen and pretrained on \gls{imagenet1k}.
\texttt{rexnet-100} is the only encoder whose latent dimensionality is higher than the number of anchors, and the biggest drop in stitching performance happens on \gls{cifarh} when the decoder is trained on it. This suggests the number of anchors is an important hyperparameter; for a deeper analysis, see the ``Anchors analysis'' section of the appendix of \citet{moschella2023relative}.

Overall, these results prove that relative representations can bridge general-purpose encoders and pretrained task-specific decoders.

\chapter[Product of Invariances]{Product of Invariances: Composition of Correspondence-Parallel Operators}
\label{ch:05}
\label{ch:product-invariances}

\chapterpublication[https://github.com/icannistraci/latent-invariances]{cannistraci2024bricks}{\textbf{ICLR 2024} \textit{(\textcolor{blue}{Spotlight})}}{Uncertain which metric reveals structure? Compute all simultaneously---each product component carries one invariance, and the task selects the relevant geometry.}

\bigskip

\begin{figure}[t]
 \centering
 \includegraphics[width=0.9\linewidth]{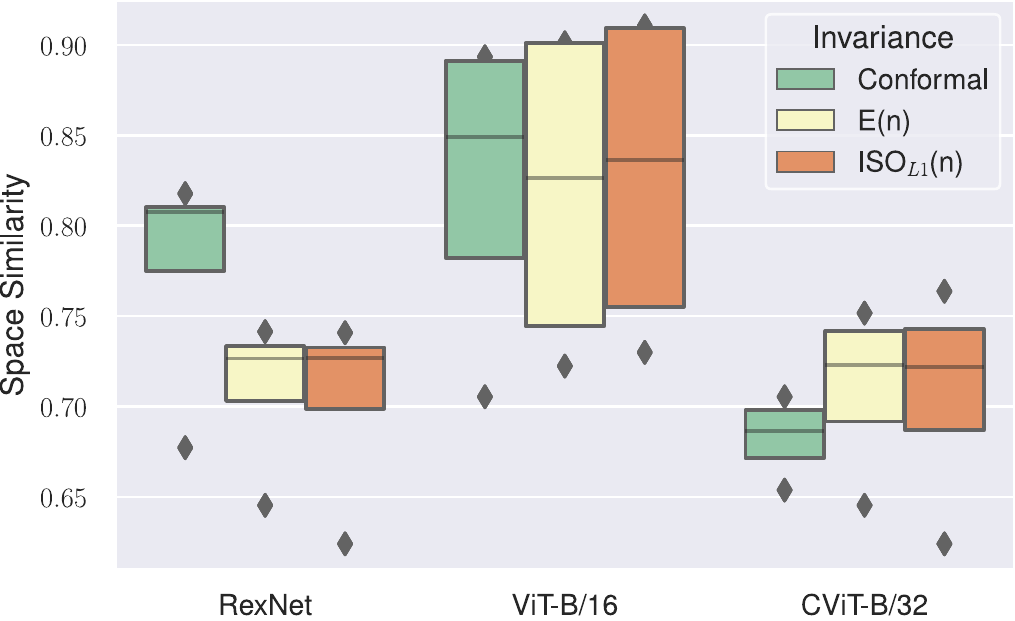}
 \caption{\textbf{Motivation for product of invariances}. Different transformations can affect latent spaces in complex ways: the CKA similarity between pretrained models on \gls{fmnist}, measured after infusing invariance to one class of transformations at a time (conformal, Euclidean, $L_1$ isometries; legend: Conformal, E(n), ISO$_{L_1}$(n), with $n$ the latent dimension), changes with the model, and no single transformation connects all latent spaces. Combining multiple distance functions, each invariant to specific transformations, captures these complex relationships between latent spaces.}
 \label{fig:ch05:teaser}
\end{figure}

\section{Introduction}

This chapter demonstrates how to compose \glspl{correspondence-parallel} to construct richer \glspl{semantic-coordinate}. The fundamental question is simple: \cref{ch:relrep} introduced \gls{relrep} using cosine similarity---but which distance function should determine the correspondence-parallel transformation? Each metric encodes geometric invariances: cosine similarity preserves angular relationships under orthogonal transformations and isotropic scaling; Euclidean distance ($L_2$ norm) remains invariant to rigid motions; Manhattan distance (\glsxtrshort{l1}) and Chebyshev distance (\glsxtrshort{linf})\glsunset{l1}\glsunset{linf} share the same guaranteed invariances, to translations and permutations. \cref{tab:product-invariances} formalizes these invariance profiles across seven transformation classes---no single metric dominates universally.

As established in \cref{ch:methodology}, the \gls{correspondence-parallel} family ($\gls{OperatorParallel}_{\gls{Corr}}^{\mathcal{X}}$) transforms spaces independently using their respective sides of correspondence. \gls{relrep} instantiates this using angular distance to shared anchors. The product of invariances extends this principle: since the relevant invariance is not known in advance, it computes \gls{relrep} transformations under multiple metrics simultaneously, constructing a product space in which each component carries one invariance. No single distance metric has to be selected.

\Cref{fig:ch05:teaser} demonstrates the core challenge: no universal transformation class connects all pretrained models. The figure shows CKA similarity distributions between pretrained encoders evaluated on \gls{fmnist} after infusing invariances to three transformation classes: conformal transformations (global orthogonal transformations composed with local rescaling), Euclidean isometries (distance-preserving rigid motions), and isometries of the \gls{l1} distance (translations and coordinate permutations, \cref{tab:product-invariances}). Each bar shows the distribution of similarity scores to other models under that invariance. The score diversity reveals that different encoders align best under different transformation classes. Some encoders achieve their highest similarity under conformal invariance, others under Euclidean, while \gls{l1} invariance scores close to Euclidean. This variability stems from \hlextrinsic{\glspl{extrinsic-factor}}: random initialization, neural architecture, and data modality (\cref{ch:methodology}). Selecting a single metric a priori risks alignment failure when model pairs relate through transformation classes that do not match that metric's invariance profile.

The solution: use all metrics simultaneously. Computing \gls{correspondence-parallel} transformations under cosine, Euclidean, Manhattan, and Chebyshev constructs a product space in which each component carries one invariance. Learned aggregation then extracts task-relevant structure. This approach treats metric choice as fundamental ambiguity to embrace: when the transformation group relating two representations cannot be known a priori, preserve structure under all candidate groups.

\section{Method: Product Space Construction}
\label{sec:product:method}

\begin{figure}[H]
 \centering
 \begin{overpic}[trim=0 0 0 0,clip,tics=5,width=.625\linewidth]{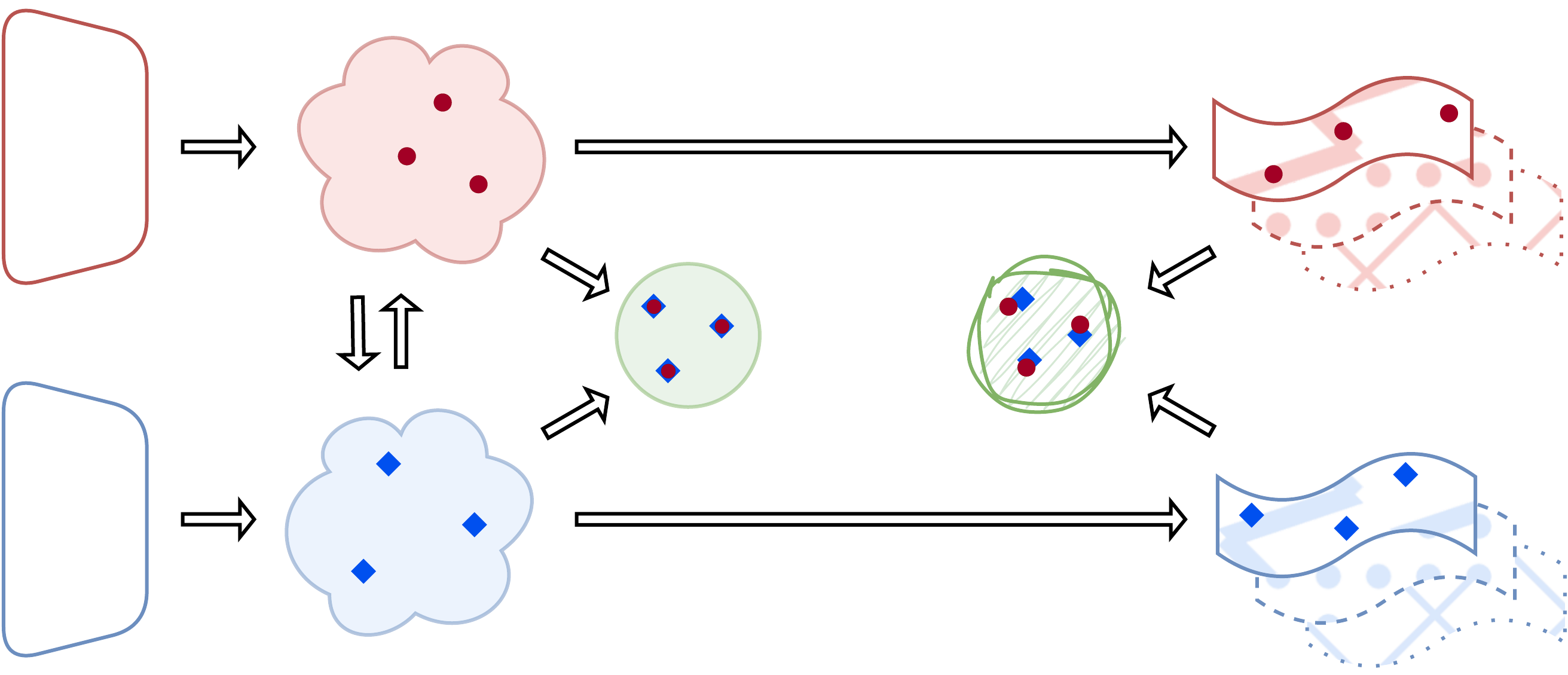}
  \put(-5,33.5){$\Obs$}
  \put(-5,9.5){$\Obs$}
  \put(3, 33.5){$\Enc[X]$}
  \put(3, 9.5){$\Enc[Y]$}
  \put(27, 20.5){\scriptsize $\tau$}
  \put(16, 20.5){\scriptsize $\tau^{-1}$}
  \put(16.75,38){$\mathcal{X}$}
  \put(16,2){$\mathcal{Y}$}
  \put(49.5,20.5){\tiny $\mathcal{M} \approx \tilde{\mathcal{M}}$}
  \put(75.5,20.5){$\rho \approx \pi_\mathcal{M}$}
  \put(32,20.5){$\pi_\mathcal{M}$}
  \put(38,37){\small Relative representations}
  \put(36,4){\small Relative representations}
 \end{overpic}
 \caption{\textbf{Product space framework}. The intrinsic manifold $\mathcal{M}$ shared by two latent spaces $\mathcal{X},\mathcal{Y}$ related by an unknown transformation is approximated through the product space $\tilde{\mathcal{M}}$. Each component is a \gls{relrep} space computed using a similarity function $\simf_i$ invariant to a specific transformation class. Combining these spaces yields a representation $\rho$ that approximates projection onto the unknown intrinsic manifold $\mathcal{M}$.}
 \label{fig:ch05:method}
\end{figure}

\subsection{Setting and Background}

Following the encoder-decoder decomposition from \cref{ch:methodology}, consider neural networks as compositions $F_\theta = \Dec \circ \Enc$, where encoder $\Enc: \Obs \to \mathcal{X}$ computes latent representations $\vx = \Enc(\obs)$ for $\obs \in \Obs$, with $\mathcal{X} \subseteq \R^d$ and $d$ much smaller than the input dimension, and decoder $\Dec$ solves the task at hand (reconstruction, classification). The parameters $\theta$ are left implicit in $\Enc$ and $\Dec$ for notational convenience. The encoders are written $\Enc[X]$, $\Enc[Y]$ when two are involved.

\gls{relrep} (\cref{ch:relrep}) established angular invariance through cosine similarity, demonstrating how a single correspondence-parallel operator constructs \glspl{semantic-coordinate}. The product of invariances extends this by using multiple similarity functions simultaneously, each inducing different invariances, to capture complex transformations between latent spaces.

\subsection{Geometric Framework}

Consider different neural networks $F,F'$ with latent spaces $\mathcal{X},\mathcal{Y}$. The goal is modeling the transformation class $\mathcal{T}$ that relates these spaces. This could be a known group (e.g., rotations) or a complex, nontrivial class. Networks may differ by initialization seeds (training dynamics), architecture, or even domain ($\Obs[X] \neq \Obs[Y]$), inducing different effects on latent geometry (\cref{fig:ch05:teaser}).

The fundamental geometric hypothesis: these variations induce changes in the \hlextrinsic{extrinsic parameterization} of representations, but there exists an underlying \hlintrinsic{intrinsic} manifold $\mathcal{M}$ where the semantic structure coincides (\cref{fig:ch05:method}).

\begin{assumption}
Given multiple models $F_1, \dots, F_k$, there exists a manifold $\mathcal{M}$ identifying an equivalence class of encoders $\mathcal{E}_\mathcal{T}$ induced by transformation class $\mathcal{T}$, defined as
\[
\mathcal{E}_\mathcal{T} := \left\{ \Enc \mid \pi_\mathcal{M} \circ \tau \circ \Enc = \pi_\mathcal{M} \circ \Enc, \quad \forall \tau \in \mathcal{T} \right\},
\]
where $\pi_\mathcal{M}$ denotes projection onto $\mathcal{M}$. The manifold $\mathcal{M}$ is equipped with metric $\operatorname{dist}_{\mathcal{M}}$ preserved under $\mathcal{T}$:
\[
\operatorname{dist}_{\mathcal{M}}(\pi_\mathcal{M}(\vx),\pi_\mathcal{M}(\vx')) = \operatorname{dist}_{\mathcal{M}}(\pi_\mathcal{M}(\tau(\vx)),\pi_\mathcal{M}(\tau(\vx'))), \quad \forall \vx,\vx' \in \mathcal{X},\ \forall \tau \in \mathcal{T}.
\]
\end{assumption}

The objective is constructing a function $\rho$ that independently projects latent spaces $\mathcal{X}_1, \dots, \mathcal{X}_k$ into $\mathcal{M}$ while remaining invariant to $\mathcal{T}$: $\rho(\tau(\vx)) = \rho(\vx)$ for all $\tau \in \mathcal{T}$ and all $\vx \in \mathcal{X}_1 \cup \dots \cup \mathcal{X}_k$, when the anchor embeddings are transformed by the same $\tau$. Generalizing \gls{relrep} (\cref{ch:relrep}) to arbitrary distance metrics provides a straightforward construction for representations $\rho$ invariant to specific transformation classes.

\subsection{Product Projection}

The metric $\operatorname{dist}_{\mathcal{M}}$ is typically unknown a priori. Moreover, capturing $\mathcal{T}$ with a single transformation class proves challenging (\cref{fig:ch05:teaser} demonstrates this empirically). The solution: approximate $\mathcal{M}$ through product space $\tilde{\mathcal{M}} := \prod_{i=1}^m \mathcal{M}_i$, where each component results from projecting $\mathcal{X}$ into a \gls{relrep} space equipped with a different similarity function $\simf_i$.

Each $\mathcal{M}_i$ inherits properties from similarity function $\simf_i$, which is invariant to a specific, known transformation class $\tilde{\mathcal{T}}_i$ (e.g., dilations). Combining this set of invariances yields representation $\rho$ that approximates $\pi_\mathcal{M}$ (\cref{fig:ch05:method}). $\relproj_{\simf_i}(\vx)$ denotes the relative representation of $\vx$ with respect to the anchor set of the space containing $\vx$ ($\anchors[X]$ for $\vx \in \mathcal{X}$) under similarity $\simf_i$ (\cref{ch:relrep}).

\begin{definition}[Product projection]
Given latent spaces $\mathcal{X}_1, \dots, \mathcal{X}_k$ related by unknown transformation class $\mathcal{T}$, and a set of similarity functions $\{\simf_i\}_{i=1}^m$ where each $\simf_i$ is invariant to known transformation class $\tilde{\mathcal{T}}_i$ (i.e., $\relproj_{\simf_i}(\tau(\vx)) = \relproj_{\simf_i}(\vx)$ for all $\tau \in \tilde{\mathcal{T}}_i$, with the anchor embeddings transformed alike), the product projection $\rho: \mathcal{X}_1 \cup \dots \cup \mathcal{X}_k \to \tilde{\mathcal{M}}$ is defined as:
\[
 \rho(\vx) = \phi( \relproj_{\simf_1}(\vx), \dots, \relproj_{\simf_m}(\vx) ),
\]
where $\phi$ is an aggregation function (e.g., concatenation) merging the relative spaces induced by each $\simf_i$.
\end{definition}

Aggregation strategies for implementing $\phi$ are detailed in \cref{sec:merging_subspaces}.

\subsection{Distance-Induced Invariances}

The \gls{relrep} framework admits multiple similarity or distance functions $\simf$: \glsentrylong{cos}, \glsentrylong{eu}, \gls{l1}, \gls{linf}, each inducing invariance to a specific transformation class. \cref{tab:product-invariances} formalizes which transformations each metric remains invariant to. Formal definitions, synthetic examples, and visualizations appear in the appendix of \citet{cannistraci2024bricks}.

\begin{table}[t]
\centering
\small
\caption{\textbf{Distance-induced invariances.} No single metric provides universal coverage.}
\label{tab:product-invariances}
\resizebox{\textwidth}{!}{%
\begin{tabular}{lccccccc}
\toprule
\textbf{Similarity} & \textbf{Isotropic} & \textbf{Orthogonal} & \textbf{Translation} & \textbf{Permutation} & \textbf{Affine} & \textbf{Linear} & \textbf{Manifold} \\
\textbf{Function} & \textbf{Scaling} & \textbf{Transf.} & & & \textbf{Transf.} & \textbf{Transf.} & \textbf{Isometry} \\
\midrule
Absolute & $\times$ & $\times$ & $\times$ & $\times$ & $\times$ & $\times$ & $\times$ \\
Cosine & $\checkmark$ & $\checkmark$ & $\times$ & $\checkmark$ & $\times$ & $\times$ & $\times$ \\
Euclidean ($L_2$) & $\times$ & $\checkmark$ & $\checkmark$ & $\checkmark$ & $\times$ & $\times$ & $\times$ \\
Manhattan (\gls{l1}) & $\times$ & $\times$ & $\checkmark$ & $\checkmark$ & $\times$ & $\times$ & $\times$ \\
Chebyshev (\gls{linf}) & $\times$ & $\times$ & $\checkmark$ & $\checkmark$ & $\times$ & $\times$ & $\times$ \\
\bottomrule
\end{tabular}
}
\end{table}

\subsection{Aggregation Strategies}
\label{sec:merging_subspaces}

Constructing product space $\tilde{\mathcal{M}}$ requires aggregating invariant components. Consider latent space $\mathcal{X}$ produced by encoder $\Enc: \Obs \to \mathcal{X}$, and set of similarity functions $\{\simf_i\}_{i=1}^m$. With one space per $\simf_i$, the method produces $m$ relative latent spaces, where each space enforces invariance to a specific transformation class through its similarity function (\glsentrylong{cos}, \glsentrylong{eu}, \gls{l1}, or \gls{linf}).

These spaces admit diverse aggregation strategies, corresponding to different choices of $\phi$:
\begin{itemize}
 \item \emph{Concatenation} (Concat): Subspaces are independently normalized and concatenated, yielding cartesian product structure for $\tilde{\mathcal{M}}$.
 \item \emph{Aggregation by sum} (MLP+Sum): Similar to DeepSets \citep{zaheer2017deep}, spaces are independently normalized and non-linearly projected, then summed.
 \item \emph{Self-Attention} (SelfAttention): Spaces are independently normalized and aggregated via self-attention.
\end{itemize}
Before the Sum or SelfAttention aggregation, each subspace is first normalized by a LayerNorm, optionally followed by a small \gls{mlp} (a linear layer and a tanh activation); the tables label these variants ``MLP+''.
Unless otherwise specified, experiments use \emph{Aggregation by sum}. Implementation details appear in the appendix of \citet{cannistraci2024bricks}.
The resulting product space yields robust latent representations: invariant components combine to capture complex transformations, improving downstream task performance.

\section{Experiments}
\label{sec:ch05:exp}

Experiments analyze the effectiveness of product space construction for capturing complex transformations between latent spaces. \cref{sec:ch05:exp-spaces-similarity} examines latent space similarity for models trained from scratch (\gls{ae} and \gls{vae} architectures) and pretrained foundation models, analyzing transformation classes that emerge when enforcing different invariances. \cref{sec:ch05:exp-downstream-task} assesses zero-shot stitching performance across text, vision, and graph modalities. \cref{sec:ch05:exp-ablation} ablates aggregation functions. \cref{sec:ch05:exp-subspace-selection} investigates attention weights for optimal space selection.

\subsection{Latent Space Analysis} \label{sec:ch05:exp-spaces-similarity}
\subsubsection*{Training from Scratch}

\textbf{Experimental setting.} Image reconstruction experiments use \gls{cifart}, \gls{cifarh} \citep{Krizhevsky2009-hv}, \gls{mnist}, and \fmnist{} \citep{Xiao2017-er} datasets with comparable convolutional architectures: \glspl{ae} and \glspl{vae}. Models include architectures with unflattened latent image bottlenecks (\gls{ae} and \gls{vae}) preserving spatial structure, and architectures with linear projections into flattened latent spaces (\gls{linae} and \gls{linvae}). Five instances of each model are trained to convergence with different random seeds. Each latent space is then projected into its relative counterpart using distinct projection functions, infusing different invariances. Cross-seed latent space similarity is measured for each combination.

\begin{figure}[ht]
 \centering
 \begin{overpic}[trim=0 0 0 0,clip,width=.475\linewidth]{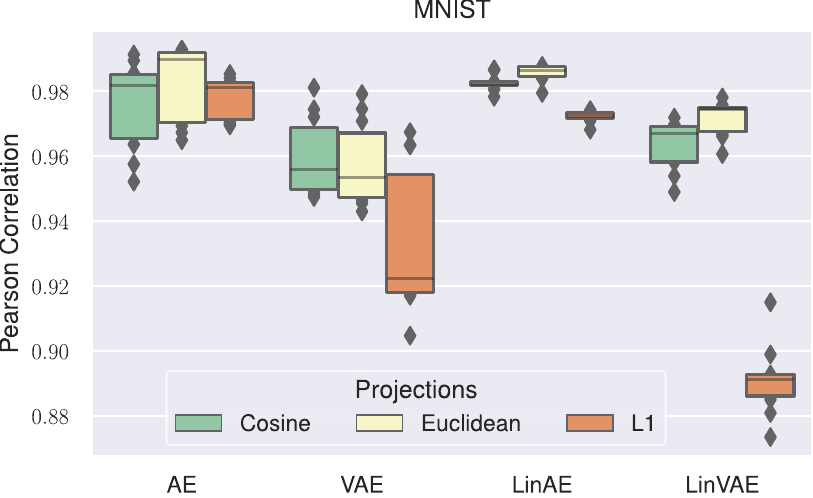}
 \end{overpic}
 \hfill
 \begin{overpic}[trim=0 0 0 0,clip,width=.475\linewidth]{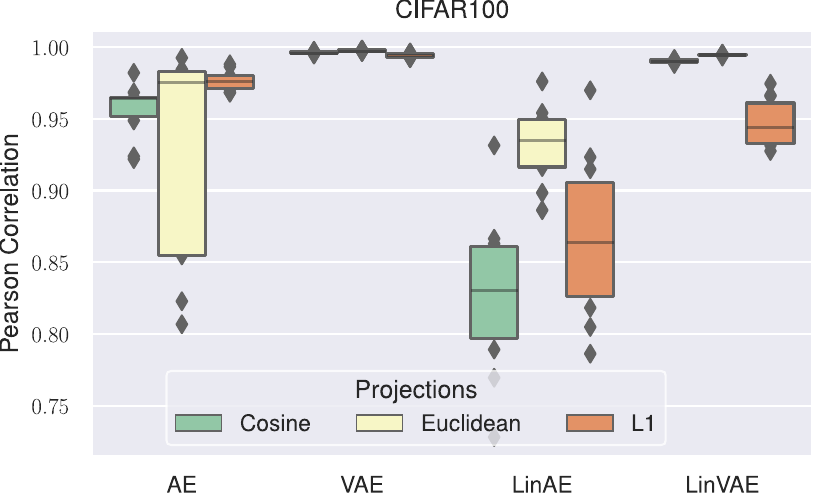}
 \end{overpic}
 \vspace{-.25cm}
 \caption{\textbf{Latent Spaces Cross-Seed Similarity.} Cross-seed Pearson correlation of latent spaces for \glspl{ae} and \glspl{vae}, with and without linear bottleneck, trained on \gls{mnist} (\emph{left}) and \gls{cifarh} (\emph{right}) until convergence. No single projection consistently outperforms others across all settings. The \gls{linf} projection is not displayed to improve visualization.}
 \label{fig:ch05:aes-main-pearson}
\end{figure}

\textbf{Results.} \cref{fig:ch05:aes-main-pearson} shows Pearson cross-seed correlations for various architectures across datasets. The results demonstrate that unifying latent spaces obtained with different initializations through a single invariance is impossible. For instance, \cref{fig:ch05:aes-main-pearson} shows highest cross-seed similarity achieved using different projection types for \gls{vae} versus \gls{linvae} architectures, even when all other parameters remain fixed. Dataset variations alter behavioral trends across architectures. This challenges the \gls{relrep} assumption \citep{moschella2023relative} that angle-preserving transformations primarily drive correlation among latent spaces of models trained with different seeds.

\subsubsection*{Pretrained Models} \label{sec:ch05:exp-pretrained-similarity}

\textbf{Experimental setting.} Experiments analyze latent space similarity for pretrained foundation models in vision and text domains. Vision experiments evaluate five distinct foundation models (convolutional or transformer-based) on \gls{cifart}, \gls{cifarh}, \gls{mnist}, and \gls{fmnist}. Text experiments assess seven foundation models on \gls{dbpedia} \citep{zhang2015AGNews}, \gls{trec} \citep{hovy-etal-2001-toward}, and \gls{n24news} \citep{wang-EtAl:2022:LREC3}.

\begin{figure}[ht]
 \centering
 \begin{overpic}[trim=0 0 0 0,clip,width=.475\linewidth]{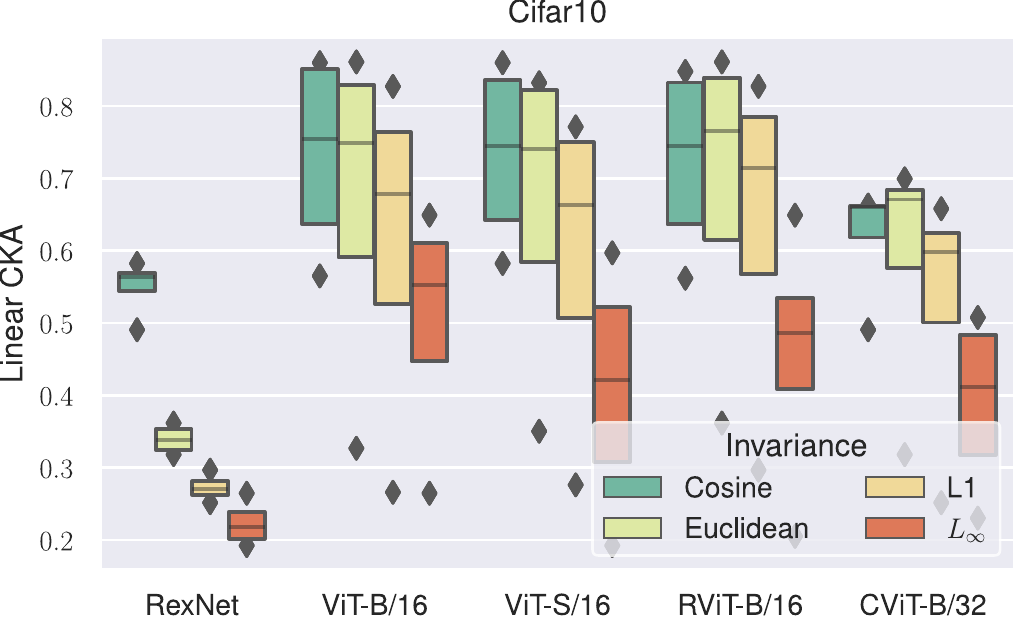}
 \end{overpic}
 \hfill
 \begin{overpic}[trim=0 0 0 0,clip,width=.475\linewidth]{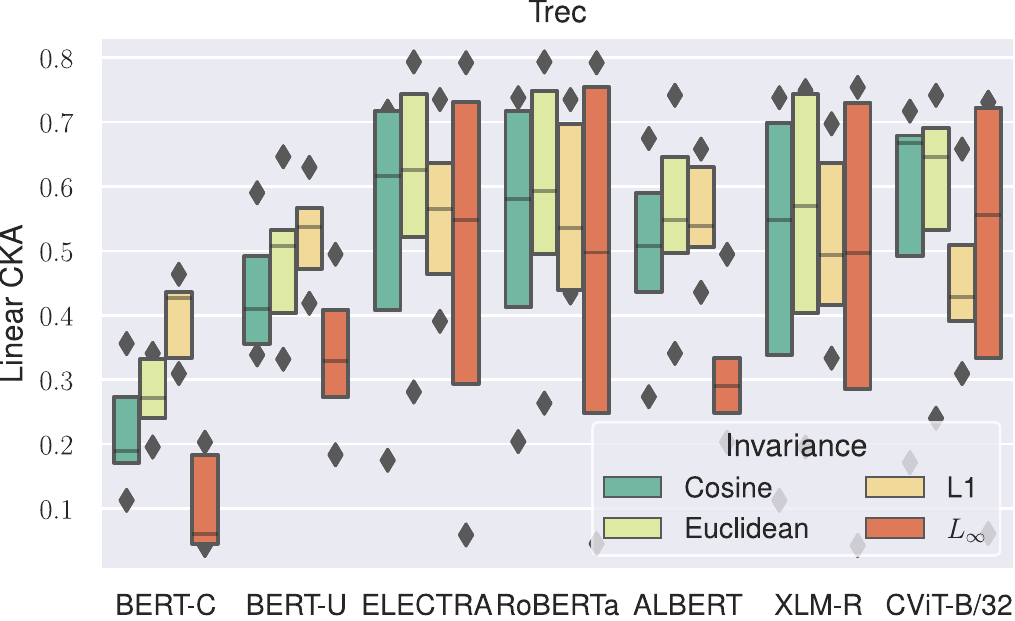}
 \end{overpic}
 \vspace{-.25cm}
 \caption{\textbf{Latent Spaces Cross-Architecture Similarity.} Linear \gls{cka} similarity of latent spaces across several pretrained models on \gls{cifart} (\emph{left}) and \gls{trec} (\emph{right}). In each bar, we report the space similarities distribution to the other models while infusing a specific invariance. There is no singular projection that consistently outperforms others across all configurations.}
 \label{fig:ch05:pretrained-main-cka}
\end{figure}

\textbf{Results.} \cref{fig:ch05:pretrained-main-cka} shows Linear \gls{cka} correlations for pretrained models across vision and text modalities. The analysis reveals no universally shared transformation class connecting latent spaces of foundation models across distinct conditions. On \gls{cifart}, highest similarity is achieved with different projection functions for different architectures. Similar architectures (ViT-based models) exhibit similar trends on the same dataset.

\textbf{Takeaway.} The transformation class correlating latent spaces---whether from pretrained or trained-from-scratch models---depends on datasets, architectures, tasks, and other \hlextrinsic{\glspl{extrinsic-factor}}.

\subsection{Downstream Task: Zero-Shot Stitching} \label{sec:ch05:exp-downstream-task}

\textbf{Experimental setting.} Zero-shot stitching experiments use text, vision, and graph modalities. Vision and text domains employ the same datasets and pretrained models from \cref{sec:ch05:exp-pretrained-similarity}. The graph domain uses \gls{cora} \citep{cora} with \gls{gcn} trained from scratch. Additionally, zero-shot stitching reconstruction uses \glspl{ae} on \gls{cifarh}.

Relative decoders are trained with three seed values. Representations are transformed into \glspl{semantic-coordinate} by projecting encodings onto 1280 randomly selected but fixed anchors. Varying the number of anchors results in different emerging transformations, indicating a single projection function cannot capture desired invariances. Nevertheless, the product approach achieves highest scores regardless of anchor count~\citep[see the appendix of][]{cannistraci2024bricks}.

\textbf{Zero-shot model stitching.} Following \cref{ch:relrep}, stitching combines components from different networks: frozen encoders handle data embedding while relative decoders manage downstream tasks. The process requires no training or fine-tuning, hence \textit{zero-shot}.

This chapter introduces the \textit{Stitching Index}: the ratio between stitching score and end-to-end score. This measures how closely stitching accuracy aligns with the original; a stitching index of one indicates no performance drop.
\begin{table}[ht]
 \caption{\textbf{Graph and Text Classification Stitching Performance.} Zero-shot accuracy scores (mean $\pm$ standard deviation over seeds) across various decoders, seeds, and datasets. For the text domain, we use pretrained models, while for the graph domain, we train \texttt{GCN} models from scratch and evaluate the stitching across seeds. Additionally, we calculate the stitching index for the graph modality, showing that composing different projections using the \emph{Aggregation by sum} enables zero-shot stitching \textit{without} any performance drop in this setting, ensuring competitive end-to-end performance.
 }
 \label{tab:ch05:text-graph-partial-stitching}
 \centering
 \footnotesize
 \resizebox{0.8\textwidth}{!}{%
  \begin{tabular}{lcccc}
   \toprule
    & \multicolumn{2}{c}{Text} & \multicolumn{2}{c}{Graph} \\
   \cmidrule(l){2-3}
   \cmidrule(l){4-5}
    & \multicolumn{2}{c}{\texttt{ALBERT}} & \multicolumn{2}{c}{\texttt{GCN}} \\
   \cmidrule(l){2-3}
   \cmidrule(l){4-5}
    Projection & \gls{dbpedia}↑ & \gls{trec}↑ & \gls{cora} ↑ & Stitching Index ↑ \\
   \midrule
    \glsentrylong{cos} & $0.50\pm 0.02$ & $0.54 \pm 0.03$ & $\textbf{0.53} \pm 0.06$ & 0.71 \\
    \glsentrylong{eu} & $0.50 \pm 0.00$ & $0.60 \pm 0.03$ & $0.27 \pm 0.06$ & 0.58 \\
    \gls{l1} & $\textbf{0.52} \pm 0.01$ & $\textbf{0.65} \pm 0.02$ & $0.26 \pm 0.06$ & 0.58 \\
    \gls{linf} & $0.18 \pm 0.02$ & $0.29 \pm 0.06$ & $0.12 \pm 0.03$ & \textbf{1.00}\\
   \midrule
    \glsentrylong{cos}, \glsentrylong{eu}, \gls{l1}, \gls{linf} & $\textbf{0.53} \pm 0.01$ & $\textbf{0.65} \pm 0.02$ & $\textbf{0.77} \pm 0.01$ & \textbf{1.00}\\
   \bottomrule
  \end{tabular}
 }
\end{table}

\begin{table}[ht]
 \caption{\textbf{Image Classification Stitching Performance Cross-Architecture and Cross-Seed.} Zero-shot accuracy score (mean $\pm$ standard deviation over seeds) across different pretrained models, seeds, and datasets. The proposed method using the \emph{Aggregation by sum} consistently achieves the highest accuracy score or comparable results, without prior knowledge of the optimal projection to employ.
 }
 \label{tab:ch05:images-stitching}
 \centering
 \footnotesize
 \resizebox{.8\textwidth}{!}{%
  \begin{tabular}{llcccc}
   \toprule
             &         & \multicolumn{4}{c}{Accuracy ↑}                     \\
   \cmidrule(l){3-6}
   Encoder        & Projection      & \gls{cifarh}    & \gls{cifart}   & \gls{mnist}   & \gls{fmnist}   \\
   \midrule
   \multirow[t]{5}{*}{\texttt{CViT-B/32}} & \glsentrylong{cos}       & $0.52 \pm 0.03$    & $\textbf{0.87} \pm 0.02$ & $0.61 \pm 0.06$   & $0.68 \pm 0.02$   \\
             & \glsentrylong{eu}      & $0.53 \pm 0.02$    & $\textbf{0.87} \pm 0.02$ & $\textbf{0.66} \pm 0.05$ & $\textbf{0.70} \pm 0.03$ \\
             & \gls{l1}        & $\textbf{0.53} \pm 0.04$  & $\textbf{0.87} \pm 0.02$ & $\textbf{0.66} \pm 0.05$ & $\textbf{0.70} \pm 0.03$ \\
             & \gls{linf}     & $0.27 \pm 0.04$    & $0.52 \pm 0.04$   & $0.57 \pm 0.03$   & $0.55 \pm 0.01$   \\
   \cmidrule(l){2-6}
             & \glsentrylong{cos}, \glsentrylong{eu}, \gls{l1}, \gls{linf}& $\textbf{0.58} \pm 0.03$  & $\textbf{0.88} \pm 0.02$ & $\textbf{0.68} \pm 0.05$ & $\textbf{0.70} \pm 0.01$ \\
   \midrule
   \multirow[t]{5}{*}{\texttt{RViT-B/16}} & \glsentrylong{cos}       & $\textbf{0.79} \pm 0.03$  & $0.94 \pm 0.01$   & $0.69 \pm 0.04$   & $0.76 \pm 0.03$   \\
             & \glsentrylong{eu}      & $\textbf{0.79} \pm 0.03$  & $0.94 \pm 0.01$   & $\textbf{0.71} \pm 0.04$ & $0.77 \pm 0.03$   \\
             & \gls{l1}        & $0.77 \pm 0.04$    & $\textbf{0.95} \pm 0.01$ & $\textbf{0.71} \pm 0.04$ & $\textbf{0.79} \pm 0.03$ \\
             & \gls{linf}     & $0.31 \pm 0.03$    & $0.75 \pm 0.04$   & $0.61 \pm 0.05$   & $0.60 \pm 0.03$   \\
   \cmidrule(l){2-6}
             & \glsentrylong{cos}, \glsentrylong{eu}, \gls{l1}, \gls{linf}& $\textbf{0.81} \pm 0.04$  & $\textbf{0.95} \pm 0.01$ & $\textbf{0.72} \pm 0.04$ & $0.76 \pm 0.04$   \\
   \bottomrule
  \end{tabular}
 }
\end{table}

\begin{figure}[ht]
 \hfill \begin{overpic}[trim=0cm 0cm 0cm -1cm,tics=5,clip,width=.875\linewidth,]{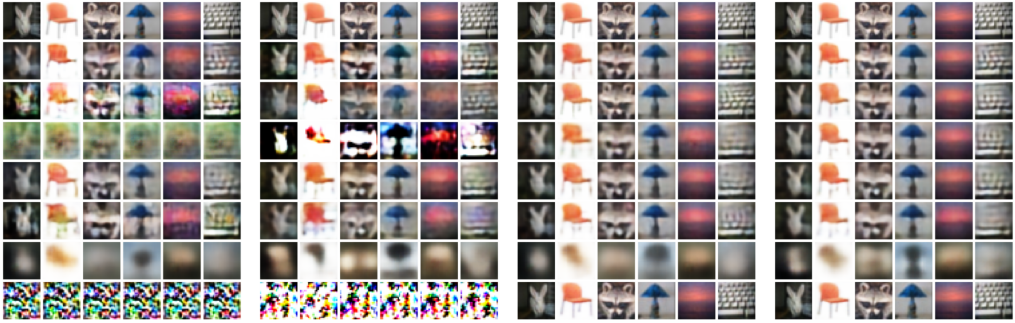}
  \put(-5.5, 29.1){\tiny Source}
  \put(-14.5, 25.25){\tiny Cos, Euc, \gls{l1}, \gls{linf}}
  \put(-10, 21.25){\tiny Cos, Euc, \gls{l1}}
  \put(-5.5, 17.25){\tiny \glsentrylong{cos}}
  \put(-8.5, 13.3){\tiny \glsentrylong{eu}}
  \put(-3, 9.5){\tiny \gls{l1}}
  \put(-4, 5.5){\tiny \gls{linf}}
  \put(-7, 1.5){\tiny Absolute}
   \put(6.9, 32.5){$\Dec[X] \circ \Enc[Y] $}
   \put(32.3, 32.5){$\Dec[Y] \circ \Enc[X] $}
   \put(57.725, 32.5){$\Dec[X] \circ \Enc[X] $}
   \put(83.2, 32.5){$\Dec[Y] \circ \Enc[Y] $}
 \end{overpic}
 \caption{\textbf{Reconstruction Qualitative Results on \gls{cifarh} using \glspl{ae}.}
 The last two sets of columns ($\Dec[X] \circ \Enc[X]$ and $\Dec[Y] \circ \Enc[Y]$) are the end-to-end \glspl{ae} with unique initialization seeds, while the first two ($\Dec[X] \circ \Enc[Y]$ and $\Dec[Y] \circ \Enc[X]$) illustrate the outputs of the zero-shot stitching of these independently trained models (in every column $\Dec \circ \Enc$ abbreviates $\Dec \circ \rho \circ \Enc$, with $\rho$ the product projection built on each encoder's own anchors; in the Absolute row no projection is applied).
 The first row displays the source images, the two subsequent rows show the results from distinct combinations of projection aggregated through the \emph{Aggregation by sum}, while the last one shows the outputs from a baseline model that does not incorporate this methodology.
 When using the \gls{linf} projection (\emph{second to last row}), the reconstructed images are blurred, but when removing the \gls{linf} space from the aggregation (\emph{third row}) the reconstruction drops in performance, meaning that it captures information not captured by the others.}
 \label{fig:ch05:aes-qualitative}
\end{figure}

\textbf{Results.} \cref{tab:ch05:images-stitching,tab:ch05:text-graph-partial-stitching} present classification performance for various projection functions across modalities. \cref{fig:ch05:aes-qualitative} shows qualitative reconstruction results.

As observed in \cref{sec:ch05:exp-spaces-similarity}, no single optimal projection function exists across architectures, modalities, or even within individual datasets. The product space framework---combining multiple invariances with trainable aggregation---achieves superior accuracy across most scenarios. Each independent projection and the aggregated product space maintain constant dimensionality, ensuring fair comparison. Additional results appear in the appendix of \citet{cannistraci2024bricks}.

\Cref{tab:ch05:text-graph-partial-stitching} demonstrates that the method enables zero-shot stitching without performance degradation on graphs while ensuring competitive end-to-end performance.

\textbf{Takeaway.} A product space of invariant components can improve the zero-shot stitching performance without any prior knowledge of the class of transformation that relates different spaces.

\subsection{Aggregation Functions: Ablation} \label{sec:ch05:exp-ablation}

\textbf{Experimental setup.} Ablation experiments evaluate merging strategies from \cref{sec:merging_subspaces}. Zero-shot stitching classification is performed on three modalities using the same datasets and models from previous sections. Relative decoders are trained with three seeds, and accuracy is assessed on each assembled model.

\begin{table}[ht]
 \caption{\textbf{Ablation Study on the Aggregation Functions.} Zero-shot classification accuracy score (mean $\pm$ standard deviation over seeds) across different architectures, seeds, and datasets. The \textit{Aggregation by sum} (\textit{third row}) obtains consistently the best accuracy score.
 }
 \label{tab:ch05:partial-ablation}
 \centering
 \footnotesize
 \begin{tabular}{lccc}
  \toprule
       & Vision     & Text      & Graph     \\
  \cmidrule(l){2-4}
  Aggregation  & \texttt{RViT-B/16}  & \texttt{BERT-C}   & \texttt{GCN}    \\
  \midrule
  Concat*   & $0.81 \pm 0.00$   & $0.54 \pm 0.03$   & $0.75 \pm 0.02$   \\
  MLP+SelfAttention & $0.84 \pm 0.01$   & $0.51 \pm 0.03$   & $0.63 \pm 0.13$      \\
  MLP+Sum   & $\textbf{0.85} \pm 0.00$ & $\textbf{0.55} \pm 0.04$ & $\textbf{0.77} \pm 0.01$ \\
  SelfAttention  & $0.76 \pm 0.03$   & $0.36 \pm 0.22$   & $0.76 \pm 0.02$   \\
  \bottomrule
 \end{tabular}
\end{table}

\textbf{Results.} \cref{tab:ch05:partial-ablation} presents ablation results across aggregation methodologies. MLP+Sum consistently outperforms alternatives. This method preprocesses each subspace with an independent \gls{mlp} (normalization, linear layer, tanh activation), then sums the resulting representations. The Concat method (marked with an asterisk) is not directly comparable since it increases dimensionality linearly with the number of subspaces.

\textbf{Takeaway.} Aggregation by sum consistently achieves highest performance among merging strategies without increasing dimensionality.

\subsection{Space Selection} \label{sec:ch05:exp-subspace-selection}

Preceding sections demonstrated integrating multiple invariances through projection functions and aggregation strategies. This section analyzes whether tuning only the aggregation strategy at stitching time provides cost-effective optimal space selection. Experiments focus on \textit{Self-attention} aggregation (\cref{sec:merging_subspaces}), fine-tuning only the query, key and value (\glsxtrshort{qkv}) parameters responsible for space blending. Each space is generated by its own projection function. Note: stitching-time fine-tuning is exclusive to this experiment.

\textbf{Experimental setup.} Two crucial components exist within the stitched model: (1) linear projections associated with \gls{qkv} in the attention mechanism, responsible for selecting and blending spaces, and (2) the \gls{mlp} classification head following the attention mechanism, which classifies aggregated embeddings.

Two approaches are examined: \texttt{QKV opt} fine-tunes only the first component, while \texttt{MLP opt} fine-tunes the second. Experiments use \gls{cifarh} with \texttt{RexNet} encoder and \texttt{ViT-B/16} decoder.

\textbf{Results.} \cref{table:ch05:attention-opt-table} summarizes downstream classification accuracy using various projection functions and aggregation strategies. Incorporating multiple invariances with \textit{Self-attention} aggregation performs poorly; \textit{MLP+Sum} or \glsentrylong{cos} projection alone proves more effective. This is expected: the attention mechanism is trained to improve end-to-end performance; its objective does not reward compatibility between spaces.

Incorporating adaptation strategies at stitching time boosts performance, whether focusing on optimal space selection and blending (\texttt{QKV opt}) or the classification head (\texttt{MLP opt}). Informed fine-tuning of space blending parameters (only \gls{qkv} projections) impacts performance more than tuning the entire classifier.

\begin{figure}[ht]
 \centering
 \begin{overpic}[trim=0 0 0 -.4cm,clip,width=.85\linewidth]{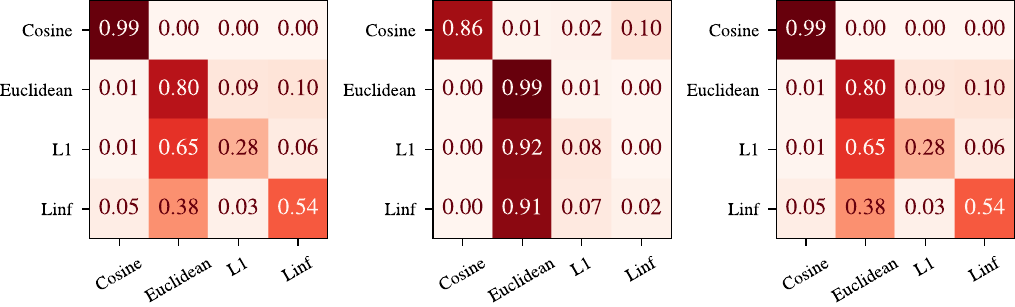}
  \put(14, 31.15){\texttt{Original}}
  \put(48.25, 31.15){\texttt{QKV opt}}
  \put(82.5, 31.15){\texttt{MLP opt}}
 \end{overpic}
 \vspace{-.25cm}
 \caption{\textbf{Attention Weights Visualization.} Comparison of attention weights for the stitched model with \texttt{RexNet} as encoder and \texttt{ViT-B/16} as decoder on \gls{cifarh}, before and after fine-tuning. \textit{(left)} the attention weights of the initial zero-shot stitched model, which remain unchanged when fine-tuning the classifier (\texttt{MLP opt}) \textit{(right)}. Conversely, fine-tuning the \gls{qkv} projections (\texttt{QKV opt}) \textit{(center)} leads to a notable shift in attention weights, assigning lower importance to the space that performs worst individually. L1 and Linf in the axes are $L_1$ and $L_\infty$.}
 \label{fig:ch05:attention-opt-cm}
\end{figure}

\Cref{fig:ch05:attention-opt-cm} illustrates attention weights averaged over the test dataset. Zero-shot stitched model attention weights (\textit{left}) remain unchanged when fine-tuning only the classifier (\textit{right}). Fine-tuning \gls{qkv} projections (\textit{center}) shifts attention weights to allocate less importance to worse-performing projections (\gls{linf}).

\begin{table}[ht]
 \caption{\textbf{Optimal Space Selection}. Classification accuracy for the stitched model with \texttt{RexNet} as encoder and \texttt{ViT-B/16} as decoder on \gls{cifarh}, using different projection functions and aggregation strategies. Fine-tuning the space selection and blending module (\texttt{QKV opt}) has a larger effect on performance improvement than fine-tuning the \gls{mlp} head (\texttt{MLP opt}). Bold: best per block.}
 \label{table:ch05:attention-opt-table}
 \centering
 \footnotesize
 \begin{tabular}{llc}
  \toprule
  Projection        & Aggregation      & Accuracy ↑  \\
  \midrule
  \glsentrylong{cos}         & -        & $\textbf{0.50}$ \\
  \glsentrylong{eu}        & -        & $0.38$   \\
  \gls{l1}         & -        & $0.24$   \\
  \gls{linf}       & -        & $0.21$   \\
  \midrule
  \glsentrylong{cos}, \glsentrylong{eu}, \gls{l1}, \gls{linf} & SelfAttention     & $0.17$   \\
  \glsentrylong{cos}, \glsentrylong{eu}, \gls{l1}, \gls{linf} & MLP+Sum       & $\textbf{0.45}$ \\
  \midrule
  \glsentrylong{cos}, \glsentrylong{eu}, \gls{l1}, \gls{linf} & SelfAttention + \texttt{QKV opt} & $\textbf{0.75}$ \\
  \glsentrylong{cos}, \glsentrylong{eu}, \gls{l1}, \gls{linf} & SelfAttention + \texttt{MLP opt} & $0.52$   \\
  \bottomrule
 \end{tabular}
\end{table}

\textbf{Takeaway.} Appropriate selection and aggregation of the optimal invariant spaces are crucial in enhancing latent communication between neural models.

\section{Conclusions}
\label{sec:product-invariances:conclusions}

The experiments of this chapter do not single out a best projection, and the image stitching results in \cref{tab:ch05:images-stitching} show why one should not expect them to: the single metric that performs best changes from one encoder to the next and from one dataset to the next, so that any metric fixed in advance would be the right choice for some pairs of spaces and the wrong one for others. This is precisely the situation the product of invariances is designed for. By combining the four metrics and aggregating them by sum, the product removes the need to choose beforehand, and it does so at little cost, since it matches or beats the best single projection in seven of the eight image cells; the only exception is \texttt{RViT-B/16} on \gls{fmnist}, where the product scores 0.76 against 0.79 for \gls{l1}. The size of the gain, however, varies with the modality. It is largest on graphs, where the product reaches 0.77 on \gls{cora} against 0.53 for the best single projection (\cref{tab:ch05:text-graph-partial-stitching}), and it is marginal on text, where with \texttt{ALBERT} the product scores 0.53 on \gls{dbpedia} and 0.65 on \gls{trec} against 0.52 and 0.65 for \gls{l1} alone.

These observations translate into a simple recommendation: whenever the invariance relating two latent spaces is unknown, we suggest aggregation by sum over all four metrics as the default, because it is the best aggregation in all three modalities of the ablation (\cref{tab:ch05:partial-ablation}). What should not be done, on the other hand, is to prune components on the basis of their individual scores. \gls{linf} is the clearest case: it has the lowest accuracy among the single projections in every stitching table of this chapter, and yet removing it from the product hurts, raising the reconstruction mean squared error from 0.06 to 0.10~\citep[see the appendix of][]{cannistraci2024bricks}, a degradation that \cref{fig:ch05:aes-qualitative} makes visible in the reconstructions themselves. A component that is weak on its own, then, is not thereby a component that can be dropped from the product.

The approach has its limits, and the experiments expose three of them. The first is that on a given encoder--decoder pair the zero-shot product can lose to a single metric: with a \texttt{RexNet} encoder and a \texttt{ViT-B/16} decoder on \gls{cifarh}, \glsentrylong{cos} alone reaches 0.50 while the product with aggregation by sum stops at 0.45, and fine-tuning at stitching time is needed to close the gap: tuning the \gls{qkv} projections of the self-attention aggregation brings the accuracy to 0.75, against 0.52 for tuning the classification head (\cref{table:ch05:attention-opt-table}). The second limit concerns cost, because fewer components can suffice: on graphs, aggregating \glsentrylong{cos} and \glsentrylong{eu} alone already reaches 0.78 against 0.77 with all four components~\citep[see the appendix of][]{cannistraci2024bricks}, so a strategy for selecting the subspaces would make the product cheaper without making it worse. The third limit is structural: every similarity function entering the product must have a closed form, so invariances whose similarity function has no closed form are out of reach unless the similarity functions themselves are learned.

Even where the product wins, its margin can be thin: on \gls{cifarh}, stitching \texttt{CViT-B/32} with \texttt{ViT-B/16}, it is present at every anchor count from 2 to 2000, but it stays between 0.01 and 0.08, with 0.28 against 0.24 at 10 anchors and 0.83 against 0.80 at 1280~\citep[see the appendix of][]{cannistraci2024bricks}.

\begin{sloppypar}
Seen through the framework of this thesis, the product of invariances is a \gls{correspondence-parallel} operator in which the invariance class is chosen separately for each component. Each component is a relative projection onto the same set of anchors, so every space is transformed using only its own side of the correspondence, which is what makes the operator parallel; the product then keeps one component per candidate class and leaves it to the aggregation to learn which of them matter for the task at hand. What the operator does not relax is the assumption that the correspondence is given in full, as a complete set of parallel anchors. The next two chapters turn to the correspondence itself: \cref{ch:bootstrapping} expands a small seed of anchor pairs into a full correspondence, and \cref{ch:asif} scales it to 1.6 million image--text pairs.
\end{sloppypar}

\chapter{Bootstrapping Parallel Anchors}
\label{ch:bootstrapping}

\chapterpublication[https://github.com/icannistraci/bootstrapping-ao]{cannistraci2023bootstrapping}{\textbf{ICLR Tiny Papers Track 2023}}{Amplify 15 seed pairs to 300 via Optimal Transport---geometry can discover semantic correspondence.}

\section{Introduction}
\label{sec:bootstrapping:introduction}
The \gls{correspondence-parallel} framework (\cref{ch:methodology}) requires parallel anchors to construct \glspl{semantic-coordinate}. Obtaining such anchors can be expensive in specialized domains or with proprietary datasets. This chapter explores whether a \emph{minimal} correspondence can be geometrically amplified via iterative refinement. Can a small seed $\gls{Corr}^{\mathrm{seed}}$ combined with geometric structure---revealed by that seed---enable discovery of expanded correspondence $\Corr^{\mathrm{boot}}$, without requiring the complete correspondence $\Corr$ upfront?

The method employs optimal transport as a geometric prior, expanding a seed correspondence $\Corr^{\mathrm{seed}}$ of 15 pairs to $\Corr^{\mathrm{boot}}$ with 300 pairs via Sinkhorn-based optimization~\citep{cuturi2013sinkhorn}---requiring only 5\% of the full correspondence as supervision. The approach alternates between estimating correspondence geometrically and refining anchors to improve that alignment. This bootstrapping enables the \gls{correspondence-parallel} operators $\gls{OperatorParallel}_\Corr$ from \cref{ch:methodology} to function with drastically reduced supervision.

Results reveal domain dependence: in \gls{nlp} word embeddings, bootstrapped anchors \emph{outperform ground truth} (0.52 versus 0.34 Jaccard similarity, \cref{fig:w2v-latent-rotation-comparison}). This counterintuitive result suggests that vocabulary-aligned word pairs, while semantically valid, do not constitute geometrically optimal anchors for retrieval. Sinkhorn optimization discovers anchors that better preserve the manifold structure learned during training, capturing latent geometric relationships that surface-level word matching misses. On \gls{cifart}, however, the bootstrapped anchors at best match ground truth, while still far exceeding the seed baseline (\cref{tab:bootstrapping-cifar}). This divergence suggests that geometry is a proxy for semantics when manifold structure exhibits sufficient semantic organization. Word embeddings satisfy this condition---their angular structure encodes semantic relationships that geometric optimization can exploit---while CIFAR representations gain less from bootstrapping, potentially due to rotational symmetries or weaker semantic clustering. Characterizing which geometric properties enable bootstrapping remains an open question.

The geometric approach explored here provides weak supervision for correspondence discovery. \cref{ch:learning_to_align} extends this principle to unsupervised correspondence discovery via Gromov-Wasserstein optimal transport, removing the need for seed pairs entirely.

\begin{table}[ht]
\caption{
 Qualitative (\textit{left}) and quantitative (\textit{right}) evaluation of the \gls{ao} method in the retrieval task between \texttt{FastText} (FT) and \texttt{Word2Vec} (W2V) embeddings of about 20k English words (\cref{ch:relrep}); mean and standard deviation across five random seeds.
}
\label{fig:w2v-latent-rotation-comparison}
\noindent\makebox[\textwidth][c]{%
 \begin{minipage}{.25\textwidth}
  \begin{overpic}[trim=0cm 0cm -1.5cm -2cm,width=1\linewidth,angle=90]{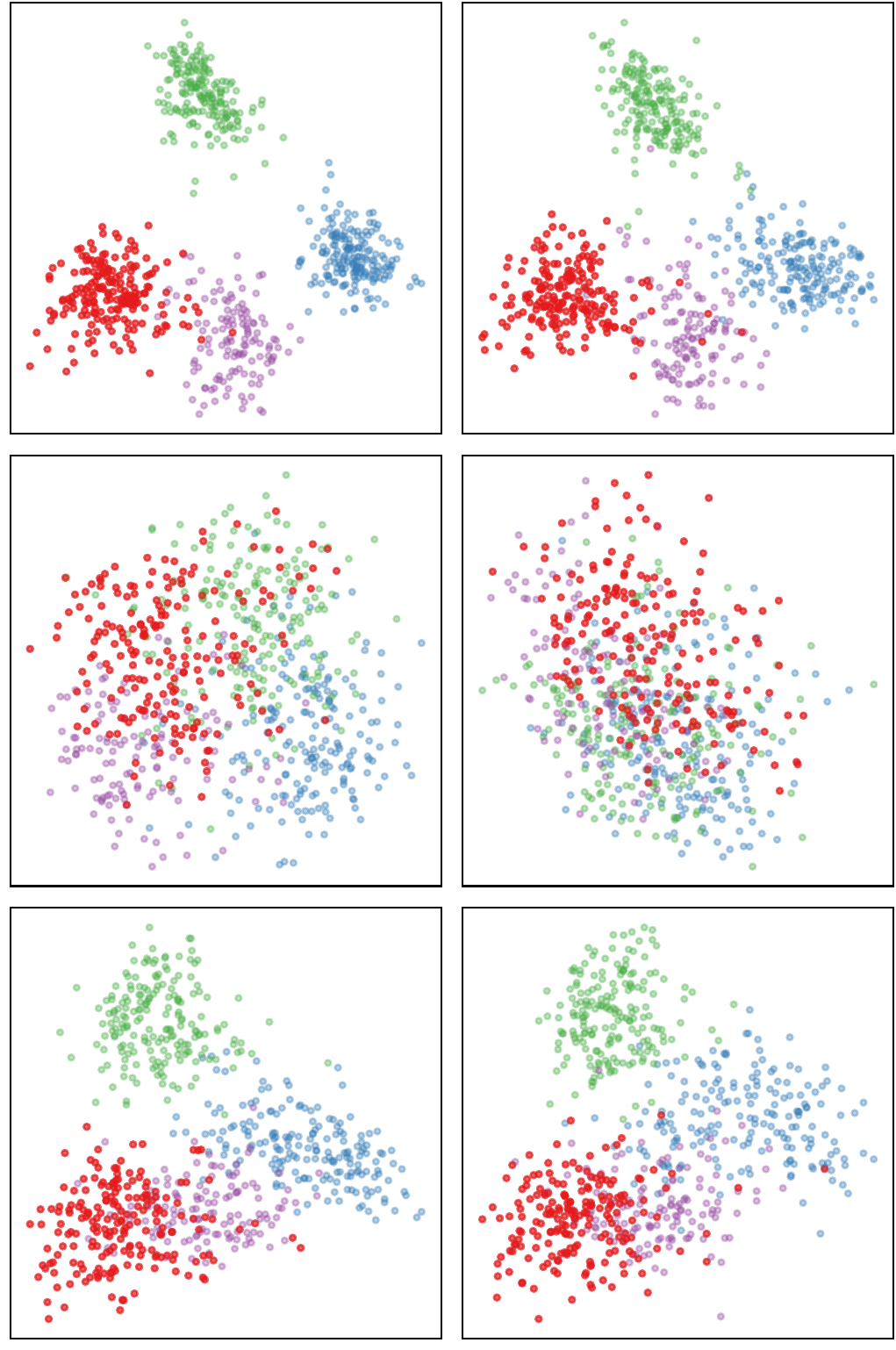}
   \put(17, 63){\small GT}
   \put(46.5, 63){\small Seed}
   \put(79, 63){\small AO}
   \put(-0, 39.8){\rotatebox{90}{\texttt{W2V}}}
   \put(-0, 12){\rotatebox{90}{\texttt{FT}}}
  \end{overpic}
 \end{minipage}
 \hfill{}
 \begin{minipage}{.61\textwidth}
  \small
  \resizebox{\linewidth}{!}{%
  \begin{tabular}{llllll}
  \toprule
   & Src & Tgt & \multicolumn{1}{c}{Jaccard $\uparrow$} & \multicolumn{1}{c}{MRR $\uparrow$} & \multicolumn{1}{c}{Cosine $\uparrow$} \\
  \midrule
   \multirow{2}{*}{{\STAB{\rotatebox[origin=c]{90}{GT{}}}}} & \multirow{1}{*}{\texttt{{FT}}} & \texttt{{W2V}} & $0.34 \pm 0.01$ & $0.94 \pm 0.00$ & $0.86 \pm 0.00$ \\
        & \multirow{1}{*}{\texttt{{W2V}}} & \texttt{{FT}} & $0.39 \pm 0.00$ & $0.98 \pm 0.00$ & $0.86 \pm 0.00$ \\
  \midrule
   \multirow{2}{*}{{\STAB{\rotatebox[origin=c]{90}{Seed{}}}}} & \multirow{1}{*}{\texttt{{FT}}} & \texttt{{W2V}} & $0.06 \pm 0.01$ & $0.11 \pm 0.01$ & $0.85 \pm 0.01$ \\
        & \multirow{1}{*}{\texttt{{W2V}}} & \texttt{{FT}} & $0.06 \pm 0.01$ & $0.15 \pm 0.02$ & $0.85 \pm 0.01$ \\
  \midrule
   \multirow{2}{*}{{\STAB{\rotatebox[origin=c]{90}{AO{}}}}} & \multirow{1}{*}{\texttt{{FT}}} & \texttt{{W2V}} & $0.52 \pm 0.00$ & $0.99 \pm 0.00$ & $0.94 \pm 0.00$ \\
        & \multirow{1}{*}{\texttt{{W2V}}} & \texttt{{FT}} & $0.50 \pm 0.01$ & $0.99 \pm 0.00$ & $0.94 \pm 0.00$ \\
  \bottomrule
  \end{tabular}}
  \vspace{-0.25cm}
 \end{minipage}
}
\end{table}

\begin{table}[ht]
\centering
\caption{Retrieval on about 20k \gls{cifart} images encoded by \texttt{ViT-base} (768 dimensions) and \texttt{ViT-small} (384 dimensions), comparing the \gls{ao} anchors with the GT and Seed baselines and extending the word-embedding results of \cref{fig:w2v-latent-rotation-comparison} to images. All three anchor sets are used to build relative representations; mean and standard deviation across five random seeds.}
\label{tab:bootstrapping-cifar}
\small
\begin{tabular}{llllll}
\toprule
 & Src & Tgt & \multicolumn{1}{c}{Jaccard $\uparrow$} & \multicolumn{1}{c}{MRR $\uparrow$} & \multicolumn{1}{c}{Cosine $\uparrow$} \\
\midrule
 \multirow{2}{*}{\STAB{\rotatebox[origin=c]{90}{GT}}} & \texttt{ViT-base} & \texttt{ViT-small} & $0.11 \pm 0.00$ & $0.27 \pm 0.01$ & $0.97 \pm 0.00$ \\
  & \texttt{ViT-small} & \texttt{ViT-base} & $0.10 \pm 0.00$ & $0.28 \pm 0.01$ & $0.97 \pm 0.00$ \\
\midrule
 \multirow{2}{*}{\STAB{\rotatebox[origin=c]{90}{Seed}}} & \texttt{ViT-base} & \texttt{ViT-small} & $0.03 \pm 0.00$ & $0.03 \pm 0.01$ & $0.97 \pm 0.00$ \\
  & \texttt{ViT-small} & \texttt{ViT-base} & $0.03 \pm 0.00$ & $0.04 \pm 0.01$ & $0.96 \pm 0.00$ \\
\midrule
 \multirow{2}{*}{\STAB{\rotatebox[origin=c]{90}{AO}}} & \texttt{ViT-base} & \texttt{ViT-small} & $0.10 \pm 0.01$ & $0.23 \pm 0.03$ & $0.97 \pm 0.00$ \\
  & \texttt{ViT-small} & \texttt{ViT-base} & $0.10 \pm 0.00$ & $0.28 \pm 0.01$ & $0.97 \pm 0.00$ \\
\bottomrule
\end{tabular}
\end{table}

\section{Method}
\label{sec:bootstrapping:method}

Consider two domains $\Obs[X]$ and $\Obs[Y]$, observed through finite samples $\Data_{\mathcal{X}} \subset \Obs[X]$ and $\Data_{\mathcal{Y}} \subset \Obs[Y]$, with learned embedding functions $\Enc[X]: \Obs[X] \to \mathcal{X} = \R^{\dimof{X}}$ and $\Enc[Y]: \Obs[Y] \to \mathcal{Y} = \R^{\dimof{Y}}$, where possibly $\dimof{X} \neq \dimof{Y}$. Given two sets of anchors $\anchors[X]\subset\Obs[X]$ and $\anchors[Y]\subset\Obs[Y]$, where $\anchors[X]$ is known while $\anchors[Y]$ is the unknown set of counterparts to recover, of which only $\anchors[Y]^{\mathrm{seed}}$ (defined below) is observed, {\em parallel anchors} are defined as a subset of pairs $\Corr \subseteq \anchors[X] \times \anchors[Y]$ in semantic correspondence, with $\nanch = \lvert\Corr\rvert$ (e.g., images and captions as in \citet{norelli2023asif}). Index the anchors as $\anchors[X] = \{a_j\}_{j=1}^{\nanch}$ and $\Corr = \{(a_j, a'_j)\}_{j=1}^{\nanch}$, so that row $j$ of every anchor matrix refers to the pair $j$; the seed pairs are the first $\lvert\Corr^{\mathrm{seed}}\rvert$ indices. Following the correspondence framework in \cref{ch:methodology}, parallel anchors constitute correspondence $\Corr$, with seed correspondence $\Corr^{\mathrm{seed}} \subseteq \Corr$, the set of seed anchors defined below.

The relative representation for a sample $\obs \in \Obs[X]$ with embedding $\vx = \Enc[X](\obs)$ (see \cref{ch:04} for full development; same for $\Obs[Y]$) is computed as:
$
\relproj_{\anchors[X]}(\vx) = \vx \anchmat[X]^\top$, where $\anchmat[X]=\bigoplus_{a_j \in\anchors[X]} \va_j$ with $\va_j = \Enc[X](a_j)$, and $\bigoplus$ denotes the row-wise concatenation operator. All embeddings are rescaled to unit norm, i.e., $\lVert\vx\rVert = \lVert\vy\rVert = 1$, corresponding to the choice of cosine similarity as a similarity function~\citep{moschella2023relative}.

\looseness=-1 The method requires only few initial \emph{seed} anchors, denoted as $\Corr^{\mathrm{seed}} \subseteq \Corr$, with $\anchors[X]^{\mathrm{seed}}$ and $\anchors[Y]^{\mathrm{seed}}$ its projections on the two domains, where $\lvert\Corr^{\mathrm{seed}}\rvert\ll\nanch$.
With no prior knowledge of $\anchors[Y]$, the optimization process is initialized by approximating $\anchmat[Y] \approx \widetilde{\anchmat}_{\mathcal{Y}}$ with the known seed $\anchmat[Y]^{\mathrm{seed}}=\bigoplus_{a'_j \in \anchors[Y]^{\mathrm{seed}}} \Enc[Y](a'_j)$ concatenated with $\nanch - \lvert\Corr^{\mathrm{seed}}\rvert$ random embeddings, i.e. $\widetilde{\anchmat}_{\mathcal{Y}}= \anchmat[Y]^{\mathrm{seed}}\oplus \mN$, with $\mN \in \R^{(\nanch - \lvert\Corr^{\mathrm{seed}}\rvert)\times\dimof{Y}}$ with rows drawn from $\mathcal{N}(\0, \mI)$.
The optimization objective over $\widetilde{\anchmat}_{\mathcal{Y}}$ is:
\begin{align}
\argmin_{\substack{ \widetilde{\anchmat}_{\mathcal{Y}}
\text{ s.t. }
\lVert (\widetilde{\anchmat}_{\mathcal{Y}})_{j,:} \rVert = 1,\ j = 1,\dots,\nanch
}} \; \sum_{\obs' \in \Data_{\mathcal{Y}}}
 \operatorname{MSE}(
 \relproj_{\anchors[X]}(\Enc[X](\pi(\obs'))),
\relproj_{\widetilde{\anchmat}_{\mathcal{Y}}}(\Enc[Y](\obs'))
 )
\end{align}
where $\relproj_{\widetilde{\anchmat}_{\mathcal{Y}}}(\vy) = \vy \widetilde{\anchmat}_{\mathcal{Y}}^\top$, with $\vy = \Enc[Y](\obs')$, is the relative representation with respect to the current anchor estimate, $\operatorname{MSE}$ is the mean squared error, and $\mP \in \R^{\lvert\Data_{\mathcal{X}}\rvert \times \lvert\Data_{\mathcal{Y}}\rvert}$ is a soft matching matrix (an entropic transport plan) between the two sample sets (rows indexed by the samples of $\Data_{\mathcal{X}}$, columns by the samples of $\Data_{\mathcal{Y}}$; in the experiments $\lvert\Data_{\mathcal{X}}\rvert = \lvert\Data_{\mathcal{Y}}\rvert$ (the same words in both spaces, shuffled), so $\mP$ approximates a permutation), and $\pi(\obs') := \argmax_{\obs \in \Data_{\mathcal{X}}} \mP_{\obs,\obs'}$ denotes the sample of $\Data_{\mathcal{X}}$ with the largest mass in the column of $\obs'$, estimated at each optimization step by the Sinkhorn \citep{cuturi2013sinkhorn} algorithm exploiting the initial seed and the current approximation of the remaining anchors:
\[
\mP = \operatorname{sinkhorn}_{(\obs,\obs') \in \Data_{\mathcal{X}} \times \Data_{\mathcal{Y}}}\bigl(
\relproj_{\anchors[X]}(\Enc[X](\obs)),\,
\relproj_{\widetilde{\anchmat}_{\mathcal{Y}}}(\Enc[Y](\obs'))
\bigr).
\]
After convergence, $\widetilde{\anchmat}_{\mathcal{Y}}$ is discretized into $\widetilde{\anchors}_{\mathcal{Y}} \subseteq \Data_{\mathcal{Y}}$, the samples of $\Data_{\mathcal{Y}}$ whose embeddings are nearest to the rows of $\widetilde{\anchmat}_{\mathcal{Y}}$.
The bootstrapped correspondence is $\Corr^{\mathrm{boot}} = \{(a_j, \tilde{a}'_j)\}_{j=1}^{\nanch}$, with $\tilde{a}'_j \in \widetilde{\anchors}_{\mathcal{Y}}$ the sample nearest to row $j$ of $\widetilde{\anchmat}_{\mathcal{Y}}$.
In summary: known are $\anchors[X]$ and $\Corr^{\mathrm{seed}}$; optimised is $\widetilde{\anchmat}_{\mathcal{Y}}$; returned are $\widetilde{\anchors}_{\mathcal{Y}}$ and $\Corr^{\mathrm{boot}}$.

\section{Algorithm and Hyperparameters}
\label{sec:bootstrapping:algorithm}

\Cref{alg:anchor-optimization} outlines the pseudocode for the optimization procedure described in \cref{sec:bootstrapping:method}, while \cref{tab:app-hyperparameters} details the hyperparameters.
The method initializes $\widetilde{\anchmat}_{\mathcal{Y}}$ and optimizes it iteratively for $n_{\mathrm{it}}$ steps (``Number of optimization steps'' in \cref{tab:app-hyperparameters}). At each step, the Sinkhorn algorithm computes a rough estimate of the permutation between the two relative spaces. The loss function minimized is the MSE, with particular emphasis placed on ensuring that the optimized parameters $\widetilde{\anchmat}_{\mathcal{Y}}$ adhere to unit norm using the parametrization of \citet{lezcano2019trivializations}. This ensures the effectiveness of the optimization and reduces the search space.

\begin{algorithm}[ht]
	\caption{\glsxtrlong{ao}}
  \label{alg:anchor-optimization}
	\begin{algorithmic}[1]
   \State Initialize $\widetilde{\anchmat}_{\mathcal{Y}} = \anchmat[Y]^{\mathrm{seed}}\oplus \mN$, with $\mN \sim \mathcal{N}(\0,\mI)$ and $\mN \in \R^{(\nanch - \lvert\Corr^{\mathrm{seed}}\rvert) \times \dimof{Y}}$
   \State Compute $\mX_{\mathrm{rel}} = \bigoplus_{\obs \in \Data_{\mathcal{X}}} \relproj_{\anchors[X]}(\Enc[X](\obs))$, the relative representations of the samples in $\Data_{\mathcal{X}}$
		\For {$n_{\mathrm{it}}$ steps}
   \State Compute $\mY_{\mathrm{rel}} = \bigoplus_{\obs' \in \Data_{\mathcal{Y}}} \relproj_{\widetilde{\anchmat}_{\mathcal{Y}}}(\Enc[Y](\obs'))$, the relative representations of the samples in $\Data_{\mathcal{Y}}$
   \State Estimate the soft matching between the rows of $\mX_{\mathrm{rel}}$ and the rows of $\mY_{\mathrm{rel}}$ with
$
\mP = \operatorname{sinkhorn}(
\mX_{\mathrm{rel}},
\mY_{\mathrm{rel}}
)$
   \State Reorder the rows of $\mX_{\mathrm{rel}}$ according to $\mP$, so that row $i$ matches row $i$ of $\mY_{\mathrm{rel}}$
   \State Compute the error $\operatorname{MSE}(\mX_{\mathrm{rel}}, \mY_{\mathrm{rel}})$
   \State Optimize $\widetilde{\anchmat}_{\mathcal{Y}}$ to minimize the error, while abiding by the constraint that every row of $\widetilde{\anchmat}_{\mathcal{Y}}$ has unit norm
		\EndFor
   \Return the samples of $\Data_{\mathcal{Y}}$ whose embeddings are nearest to the rows of $\widetilde{\anchmat}_{\mathcal{Y}}$
	\end{algorithmic}
\end{algorithm}

\begin{table}[ht]
\centering
\caption{Hyperparameters for the \gls{ao} method in \textit{retrieval} and \textit{zero-shot stitching} tasks.}
\label{tab:app-hyperparameters}
\begin{tabular}{lll}
\toprule
Hyperparameter & Retrieval & Zero-shot stitching \\
\midrule
Random seed & \texttt{0}, \texttt{1}, \texttt{2}, \texttt{3}, \texttt{4} & \texttt{0}, \texttt{1}, \texttt{2}, \texttt{3}, \texttt{4} \\
Number of anchors to approximate & \texttt{300} & \texttt{300} \\
Number of seed anchors & \texttt{15} & \texttt{15} \\
Number of optimization steps & \texttt{250} & \texttt{125} \\
Learning Rate & \texttt{0.02} & \texttt{0.05} \\
Optimizer & \texttt{Adam} & \texttt{Adam}\\
Loss & \texttt{MSE} & \texttt{MSE}\\
Sinkhorn eps & \texttt{1e-4} & \texttt{1e-4} \\
Sinkhorn stop error & \texttt{1e-5} & \texttt{1e-5} \\
Number of Sinkhorn steps & \texttt{1} & \texttt{1} \\
\bottomrule
\end{tabular}
\end{table}

\section{Experiments} \label{sec:bootstrapping:experiments}

\looseness=-1 This section assesses the effectiveness of the \gls{ao} method. Experiments utilize 15 anchors to approximate 300 parallel anchors that are the ground truth in all downstream tasks. The performance is compared against two baselines: (1) \textit{GT}, the Ground Truth employing all anchors that the method aims to semantically approximate, (2) \textit{Seed}, exploiting only the seed anchors.

Anchor quality is evaluated using retrieval metrics: \gls{jaccard-similarity}, \gls{mrr}, and \gls{cosine-similarity}.

The method effectively discovers parallel anchors in the \gls{nlp} and \gls{cv} domains, as demonstrated in \cref{fig:w2v-latent-rotation-comparison,tab:bootstrapping-cifar}. Experiments explore different word embeddings and pretrained foundational visual encoders, assessing the quality of the discovered anchors through a retrieval task.
Results demonstrate that, when given the same number of starting anchors, the method outperforms the approach that relies solely on those without optimizing.
Moreover, results are \textit{comparable or superior} to those obtained with all the ground truth parallel anchors.

Furthermore, \cref{tab:stitching-multilingual-summary} demonstrates that the method can discover parallel anchors across different domains: the approach finds aligned Amazon reviews in different languages with unavailable ground truth. Using only 15 out-of-domain (\textit{OOD}, \cref{ch:relrep}) parallel anchors~\citep{moschella2023relative}, the method enables zero-shot stitching, allowing a classifier trained on one language to perform predictions on another.

\begin{table}[ht]
\centering
\caption{Cross-lingual zero-shot stitching performance evaluation on Amazon Reviews, mean $\pm$ std across five seeds; Fscore $\uparrow$, MAE $\downarrow$.}
\label{tab:stitching-multilingual-summary}
\small
\setlength{\tabcolsep}{4pt}
\begin{tabular}{cccccccc}
\cmidrule(l){3-8}
     &   & \multicolumn{2}{c}{GT} & \multicolumn{2}{c}{Seed} & \multicolumn{2}{c}{AO} \\ \cmidrule(l){3-8}
Dec.    & Enc. & Fscore   & MAE   & Fscore  & MAE  & Fscore   & MAE   \\
\midrule
  en & es & 0.51 $\pm$ 0.01 & 0.67 $\pm$ 0.02 & 0.44 $\pm$ 0.01 & 0.80 $\pm$ 0.01 & 0.48 $\pm$ 0.01 & 0.70 $\pm$ 0.02 \\
  es & en & 0.50 $\pm$ 0.02 & 0.72 $\pm$ 0.04 & 0.41 $\pm$ 0.01 & 0.92 $\pm$ 0.02 & 0.46 $\pm$ 0.01 & 0.76 $\pm$ 0.02 \\
\bottomrule
\end{tabular}
\end{table}

\chapter[ASIF]{ASIF: Coupled Data Turns Unimodal Models to Multimodal Without Training}
\label{ch:07}
\label{ch:asif}

\chapterpublication[https://github.com/noranta4/ASIF]{norelli2023asif}{\textbf{NeurIPS 2023}}{Sparse \gls{relrep} over a cross-modal correspondence at scale (1.6M image-text pairs) yield a CLIP-like model from frozen unimodal encoders, with no training.}

\bigskip

\section{Introduction}

\begin{wrapfigure}[8]{r}
{0.59\linewidth}
\centering
		\vspace{-0.99cm}
   \begin{overpic}
		[trim=0cm 0cm 0cm 0cm,clip,width=0.999\linewidth]{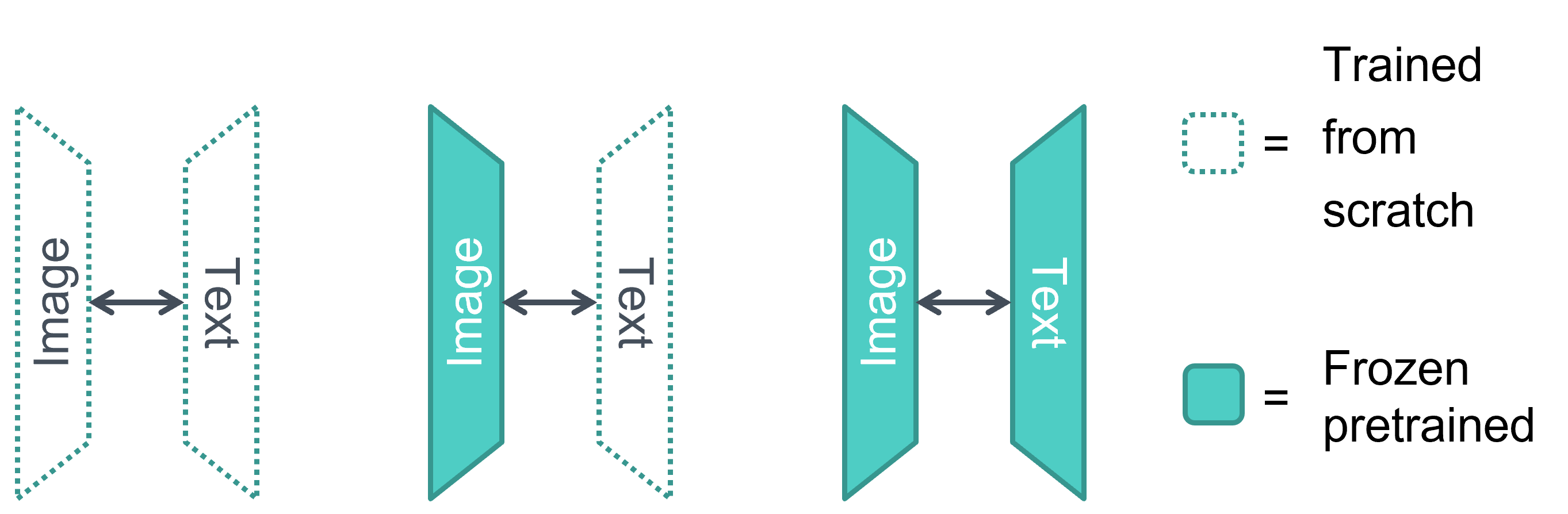}
\put(4,28){\small CLIP}
\put(31.8,28){\small LiT}
\put(56.5,28){\small ASIF}
 		\end{overpic}
 		 		 \vspace{-0.55cm}
 \caption{ASIF is a simple recipe to align the representations of two frozen pretrained encoders.}
 \label{fig-models}
\end{wrapfigure}

This chapter demonstrates how \glspl{correspondence-parallel} (\gls{OperatorParallel}, introduced in~\cref{ch:methodology}) extend to cross-modal settings through semantic correspondence and massive-scale anchor refinement. While relative representations~(\cref{ch:relrep}) typically employ hundreds of correspondence anchors \gls{Corr}, ASIF scales to correspondence cardinality $\nanch=1.6$M through local, context-dependent sparsification that constructs adaptive \glspl{semantic-coordinate} per query. The \gls{correspondence-parallel} approach is essential here: cross-modal zero-shot classification requires comparing images and text in a shared coordinate system. Transforming images directly to text coordinates would require \glspl{correspondence-coupled}. By transforming both modalities independently to semantic coordinates, ASIF enables cross-modal zero-shot classification using only frozen encoders.

\medskip

The key challenge in cross-modal alignment is that correspondence must be semantic: an image of a golden retriever and the caption ``a photo of a golden retriever'' are conceptually related but not identical samples. ASIF addresses this by treating image-text pairs as correspondence anchors, then applying the correspondence-parallel \gls{relrep} transformation (\cref{ch:relrep}) to find the best cross-modal match. The method employs two treatments (sparsification and exponentiation), followed by normalization: \textbf{sparsification} retains only the top-$k$ most relevant anchors per query; \textbf{exponentiation} amplifies similarity differences to focus on the most relevant semantic concepts; \textbf{normalization} ensures consistent similarity computations in the transformed space. These operations are applied independently per query, creating local coordinate systems where each query selects its own contextually relevant subset from the full correspondence set.

\begin{wrapfigure}[28]{r}
{0.2815\linewidth}
\centering
		\vspace{-0.5cm}
 \hspace{-1.79cm}
 \hspace{0.8cm}
  \begin{overpic}
				[trim=0cm 0cm 0cm 0cm,clip,width=1.06\linewidth]{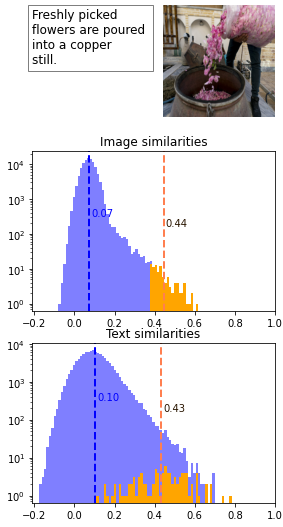}
 		\end{overpic}
 \vspace{-0.30cm}
 \hspace{-0.79cm}
 \caption{\textbf{Captions of similar images are themselves similar.} Distribution of similarities of 100k embedded pairs in the training set versus the above image and caption embeddings. Highlighted in orange are the 1000 pairs with the highest image similarity. Dashed lines: mean similarity over all pairs (blue) and over the highlighted 1000 (orange); y-axis: count, log scale.}
 		 		\label{fig-similar}
\end{wrapfigure}

The efficacy of cross-modal correspondence rests on semantic coherence across modalities: captions of similar images are themselves similar~(\cref{fig-similar}). This structural alignment means that representations constructed from similarities to correspondence anchors become quasi mode-invariant---an image and its caption share nearly identical \glspl{semantic-coordinate} even though they originate from different representation spaces. The correspondence-parallel transformation thus enables zero-shot multimodal alignment using only two frozen pretrained encoders and \textbf{no additional training}~(\cref{fig-models}), achieving performance comparable to CLIP~\citep{clip} while requiring up to 250$\times$ fewer image-text pairs than prior contrastively trained models.

\medskip

Beyond zero-shot performance, ASIF inherits the interpretability and modularity of \glspl{correspondence-parallel}. Each dimension corresponds to similarity with a specific anchor pair, enabling direct attribution of predictions to training data. The non-parametric structure enables trivial model editing---adding or removing correspondence anchors requires only encoding or deleting embeddings, with no retraining. This chapter demonstrates that the \gls{correspondence-parallel} framework scales to massive anchor sets when combined with principled sparsification, which can be read through the semantic horizon principle (\cref{sec-method}).

\section[Scaling Correspondence-Parallel Operators to Cross-Modal Settings]{Scaling Correspondence-Parallel\\Operators to Cross-Modal Settings}
\label{sec-method}

ASIF extends the relative representations framework~(\cref{ch:relrep}) from $\nanch$ in the hundreds to $\nanch=1.6$M image-text pairs as correspondence anchors. This massive scaling addresses the cross-modal challenge: unlike same-modality alignment where correspondence links identical samples across models, cross-modal correspondence must be semantic. Large correspondence sets increase the likelihood that semantically relevant anchors exist for any query, but introduce computational and statistical challenges that require principled refinement.

The core operation is straightforward: to find the best caption for an image $\obs^*$ from $m$ candidates $\{\hat{\obs}'_1, \dots, \hat{\obs}'_m\}$, ASIF computes the relative representation of $\obs^*$ with respect to the image anchors $\anchors[X] = \{a_1, \dots, a_\nanch\} \subset \Obs[X]$ (images, $\nanch = \lvert\Corr\rvert = \lvert\anchors[X]\rvert$) and treats this \textit{as if} it were the relative representation of the ideal caption $\obs'^*$ with respect to the corresponding text anchors $\anchors[Y] = \{a'_1, \dots, a'_\nanch\} \subset \Obs[Y]$ (texts). The candidate whose relative representation with respect to $\anchors[Y]$ is most similar to this projected representation is selected. This is simply a nearest neighbor search in semantic coordinates, enabled by the shared relative space $\R^\nanch$.

Dense representations comparing every query to 1.6M anchors accumulate noise from irrelevant dimensions while being less efficient to compare than sparse vectors~\citep{norelli2023asif}. ASIF addresses this through local, context-dependent \glspl{semantic-coordinate} where each query defines its own coordinate system by selecting contextually relevant anchors. The transformation applies two treatments (sparsification and exponentiation) independently per query, followed by normalization. \textbf{Sparsification} retains only the top-$k$ highest similarity dimensions ($k=800$, or 0.05\% of $\nanch=1.6$M). This can be read as a semantic horizon: below a certain similarity, geometric operations still return precise values, but these values reflect noise and carry no semantic content~\citep{karlgren2021semantic}. \textbf{Exponentiation} raises non-zero similarities to power $p=8$, amplifying differences---similarity 0.8 becomes $0.8^8 \approx 0.168$ while 0.6 becomes $0.6^8 \approx 0.017$, a ratio of about 10$\times$ from the original 1.33$\times$. Finally, the vectors are \textbf{normalized} so that they can be compared consistently. The combined effect creates local coordinate systems adapted per query: for example, a satellite image selects different anchors than a dog image, a vehicle query different anchors than an architecture query.

\subsection{Properties of ASIF Models}

Building on top of \glspl{correspondence-parallel} like \gls{relrep}~(\cref{ch:relrep}) yields several intriguing properties:

\looseness=-1\textit{No training and data-centric design.}
ASIF spaces are built on top of two independently pretrained encoders and the embeddings of a multimodal dataset, without training or fine-tuning any neuron.
Being deployable or updatable modularly, ASIF is radically ``data centric''~\citep{datacentric}. Adjusting the model by adding or forgetting specific samples is trivial.
This is particularly important as the right to use specific assets may change over time, and removing the effect of specific samples from a trained network requires sophisticated forgetting techniques~\citep{golatkar2021mixed,golatkar2020forgetting,golatkar2020eternal,ginart2019making,guo2020certified}. In ASIF, the encoders are pretrained with established datasets that do not change over time, while removing the effect of a multimodal pair is as simple as deleting its embeddings.

\textit{Data efficiency: }By exploiting the shared geometry of two pretrained encoders, ASIF models require far fewer ground-truth multimodal pairs to become effective. Experiments confirm that ASIF reaches competitive zero-shot performance on diverse classification benchmarks using a fraction of the multimodal data of its predecessors, reaching respectable accuracy even with thousands of pairs. This aligns with classical work in computer vision, where prototypical networks~\citep{proto} are a strong baseline in the extremely few-shot regime.

\looseness=-1\textit{Interpretability:} Sparse relative representations make ASIF interpretable classifiers. Each prediction can be traced back to a small set of data points in the multimodal dataset---corresponding to the dimensions that are nonzero in both the image and the label relative representations---accountable for the outcome (at most $k$), see \cref{fig-deepdive}. This enables visualizations of the relevant samples contributing to a correct or incorrect prediction at no extra cost, in stark contrast with other approaches that are often costly~\citep{Achille_2021_CVPR} and fragile~\citep{koh2017understanding,basu2020influence}.

\subsection{Implementation That Scales}

ASIF pays the price of avoiding training with a larger memory footprint and increased computation at inference time, requiring computation of embeddings and of cosine similarities against the multimodal dataset.
As such, ASIF should not be considered a general one-stop replacement for CLIP, although experiments scaled ASIF to 1.6M pairs while maintaining reasonable inference speed. The non-optimized implementation of ASIF is less than 2$\times$ slower than CLIP.
On a positive note, there are two well-established techniques that could radically enhance the efficiency of ASIF and potentially enable scalability to billions of entries.
The memory footprint required to store all embeddings can be greatly reduced using product quantization~\citep{product-quantization}, while inverse indexing~\citep{inverse-indexing} can circumvent the need for computing cosine similarities against the entire dataset at test time. These techniques are both implemented in the FAISS library~\citep{faiss}.
Finally, the distribution of pairs chosen during inference is rather short-tailed, presenting opportunities to massively prune the model even \textit{at deployment time}, deleting from memory data that is never used. However, assessment of large-scale optimized ASIF models is beyond the scope of these experiments. While an interesting direction for future research, this chapter focuses on establishing the potential of the ASIF method.

\subsection{Implementation Details}

\begin{sloppypar}
The experiments employ vision transformers (DEIT~\citep{deit} and DINO~\citep{dino}) as image encoders and SentenceT~\citep{sentencet} for text encoding, with the Conceptual 12M dataset (CC12M)~\citep{cc12m} providing 1.6M image-text pairs as correspondence anchors. Complete implementation details and design choice rationale are provided in \citet{norelli2023asif}.
\end{sloppypar}

\begin{table*}[t]
\centering
\resizebox{\textwidth}{!}{%
\begin{tabular}{lccccc}
\textbf{Method} & \textbf{Dataset size} & \textbf{ImageNet} & \textbf{CIFAR-100} & \textbf{Pets} & \textbf{ImageNet v2} \\
\hline
CLIP~\citep{clip}  & 400M (private)     & 68.6      & 68.7      & 88.9     & -        \\
CLIP~\citep{clip}  & 15M (public)     & 31.3      & -       & -      & -        \\
LiT~\citep{lit}   & 10M (public)     & 66.9      & -       & -      & -        \\
CLIP$^\dagger$  & 901M (private)     & 50.6      & 47.9      & 70.3     & 43.3       \\
LiT~\citep{lit}   & 901M (private)     & 70.1      & 70.9      & 88.1     & 61.7       \\
\hline
ASIF (sup vis. encoder)     & 1.6M (public)     & 60.9$^*$      & 50.2      & 81.5     & 52.2 \\
ASIF (unsup vis. encoder)     & 1.6M (public)     & 53.0$^*$      & 46.5      & 74.7     & 45.9
\end{tabular}}
\caption[Zero-shot classification accuracy of different multimodal designs.]{ \textbf{Zero-shot classification accuracy of different multimodal designs.} CLIP and LiT implementations vary by dataset and the visual transformer used as image encoder. The first CLIP and LiT entries use a VITb16 as ASIF, the last CLIP and LiT entries use a VITb32 (larger patch size). The public dataset of CLIP is a curated subset of YFCC100m~\citep{thomee2016yfcc100m}, while LiT and ASIF use CC12M. $^\dagger$CLIP as reported by \citet{lit}, setting ``uu'' (both towers trained from scratch). $^*$A subset of the ImageNet validation set was used to tune the two hyperparameters of ASIF which were then applied to the other datasets. The reported ImageNet accuracy is measured on a separate test set. When tuning on different datasets, accuracies remain consistent (see \citet{norelli2023asif} for additional details).
}
\label{tab-main}
\vspace{-0.4cm}
\end{table*}

\section{Empirical Evidence}
\label{sec-experiments}
This section compares ASIF to traditional multimodal models based on contrastive training, CLIP and LiT, then examines the classification of a single image in detail, unpacking the relative representations and following the classification algorithm step by step.
The analysis begins by discussing the pretrained encoders and dataset forming the tested ASIF models.

\subsection{Pretrained Encoders and Multimodal Dataset}

\looseness=-1The primary experiments utilized vision transformers as image encoders, pretrained either in a supervised manner (DEIT base~\citep{deit}) or in an unsupervised manner (DINO VITs8~\citep{dino}), on ImageNet 1k~\citep{deng2009imagenet} and 21k~\citep{imagenet21k} respectively.
The embedding size was 768 for DEIT and 384 for DINO. This configuration aligns with that used in LiT~\citep{lit}, with the sole distinction being that, unlike LiT, ASIF uses a pretrained, frozen text encoder. The text encoder is the SentenceT transformer~\citep{sentencet}, trained on a dataset comprising more than 1 billion sentences obtained from the internet.
The experiments employed the first 1.6M entries of the Conceptual 12M dataset (CC12M)~\citep{cc12m} as the multimodal dataset. This dataset amasses images and filtered alt-texts collected from the internet. To optimize performance on a single Tesla T4 GPU, the analysis was limited to the initial 1.6M pairs. The ``scaling-laws'' experiments also utilized DEIT tiny and small vision transformers~\citep{deit} and two smaller SentenceT encoders.

\subsection{Zero-Shot Performance}

The quality of the ASIF multimodal model was assessed by comparing its zero-shot classification performance against CLIP and LiT on four datasets: CIFAR-100, ImageNet, ImageNet v2, and Pets~\citep{Krizhevsky2009-hv,deng2009imagenet,imagenetv2,pets}; see \cref{tab-main}.
Label prompts were crafted as in \citet[Table 11]{lit}. ASIF achieves competitive performance with CLIP and LiT using two frozen pretrained encoders and a fraction of the image-text pairs.

The two ASIF models reported in \cref{tab-main} differ in the vision encoder, which is respectively supervisedly (DEIT) and unsupervisedly pretrained (DINO). Both hyperparameters $k$ and $p$ were tuned on a subset of the ImageNet validation set, yielding $k=800$ and $p=8$.
The comparison between the two versions is not completely fair since the visual transformer architecture of DINO is smaller (e.g., its codings are 384-dimensional, those of DEIT 768-dimensional) but corroborates the effectiveness of ASIF with encoders pretrained using no supervision. Additional results with a wider collection of ASIF models based on different visual and textual backbones are reported in the original work~\citep{norelli2023asif}.

\underline{\textit{Summary:}} Overall, the ASIF procedure achieves competitive zero-shot results with a fraction of the image-text pairs used in prior work (assuming access to other pretrained, even unsupervisedly, unimodal models).

\begin{figure}[t]
\centering
 \begin{overpic}
		[trim=0cm 0cm 0cm 0.55cm,clip,width=0.65\linewidth]{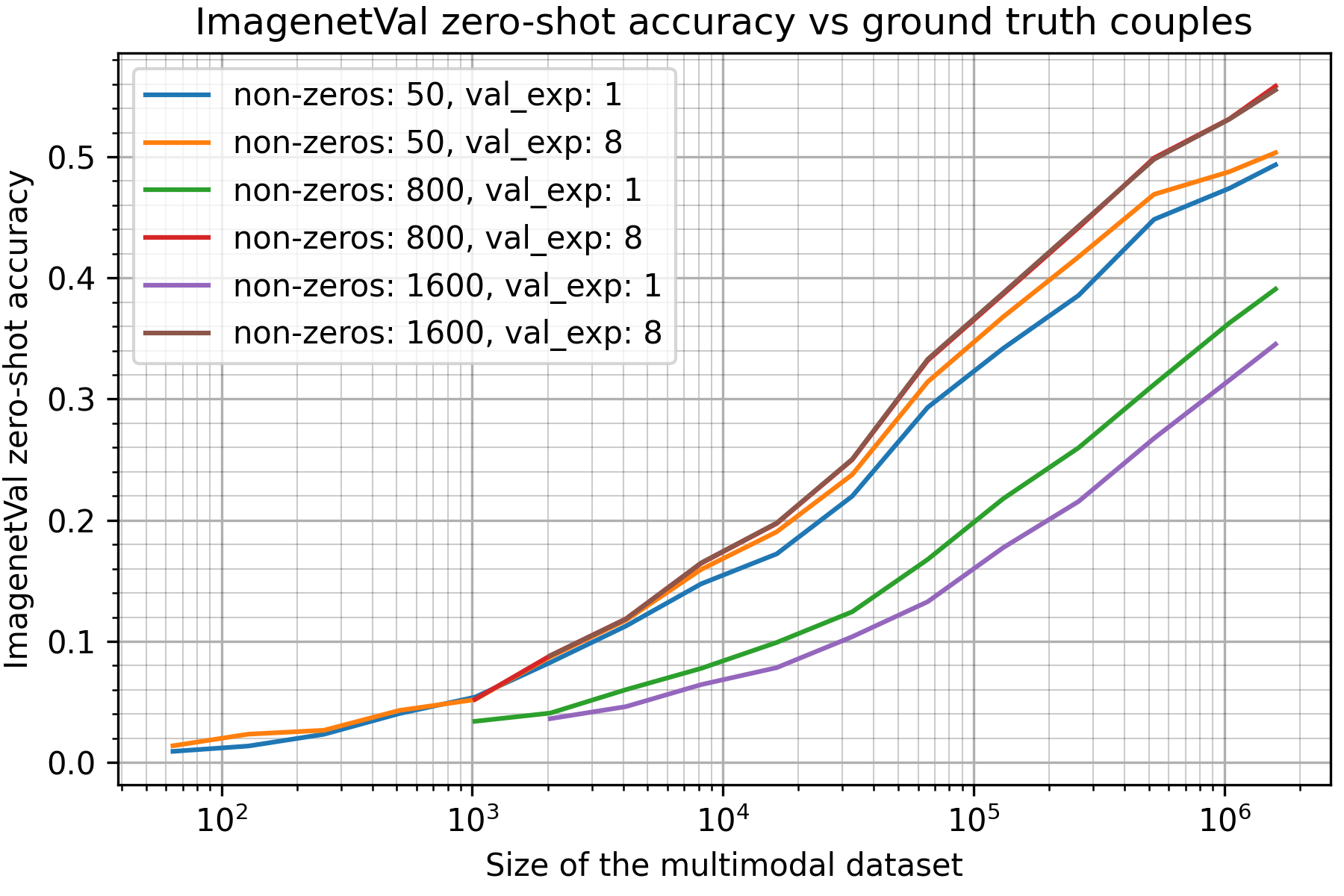}
		\end{overpic}

		\caption{\textbf{ASIF is a learning algorithm:} ImageNet validation-set accuracy improves smoothly as the multimodal dataset size increases. Colors show the impact of $k$ and $p$ (non-zeros, val exp).}\label{fig-hyper}
\end{figure}

\subsection{ASIF Scaling Laws}

\Cref{fig-hyper} shows the full zero-shot accuracy trend on ImageNet as the size of the multimodal dataset grows for different choices of $k$ and $p$. ASIF models become effective very quickly: a non-trivial $18\%$ classification accuracy is reached using just the first 10,000 pairs in the multimodal dataset.
Building an ASIF model from scratch still requires a lot of unimodal data to pretrain the encoders, but this data may come untied and even without any label.

Additional tests employed smaller image and text encoders on multimodal datasets of different sizes ($10^2$ to $10^6$ image-text pairs from CC12M) to provide early evidence about ASIF scaling laws. The experiments used DEIT tiny, small, and base vision transformers~\citep{deit}, along with two smaller SentenceT encoders. As shown in \cref{fig-hyper}, ImageNet classification accuracy grows smoothly with the size of the multimodal dataset; \citet{norelli2023asif} report that performance does not saturate earlier with smaller encoders. These results are promising but still preliminary, further experiments with larger multimodal datasets are left for future work.

\underline{\textit{Summary:}} The experiments show a steady improvement in ASIF's performance as the size of the multimodal dataset increases. \citet{norelli2023asif} report that performance deteriorates with smaller encoders, but even then there is no sign of saturation or plateau.

\begin{figure}[t]
 \centering
  \begin{overpic}[trim=0cm 0cm 0cm 0cm,clip,width=0.99\linewidth]{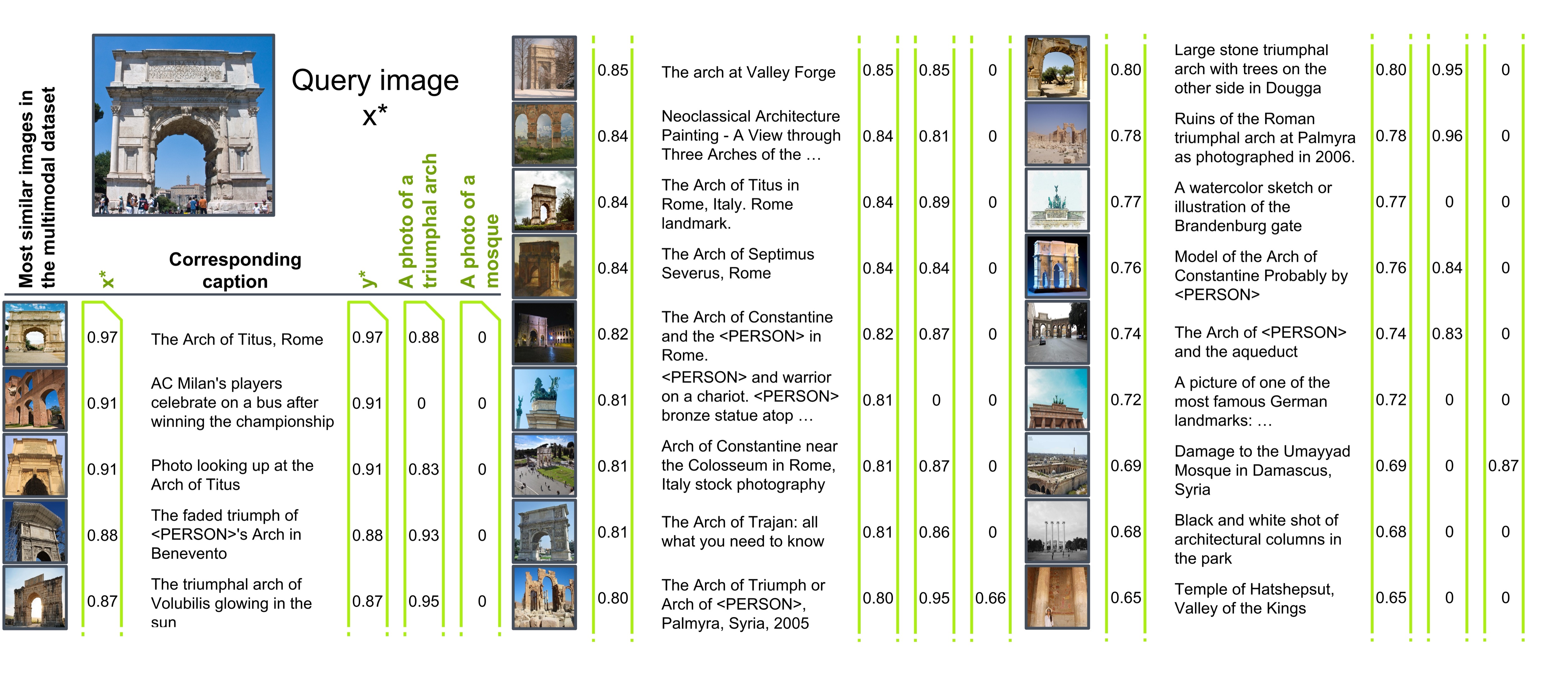}
         \end{overpic}
 \vspace{-0.3cm}
  \caption{\textbf{ASIF representations are interpretable and amendable.} Thorough analysis of the relative representations---the four vectors in green---produced by ASIF to assign the best caption to a given image $\obs^*$ ($x^*$ and $y^*$ in the figure are $\obs^*$ and the ideal caption $\obs'^*$). The values shown are the raw similarities of the retained entries, before exponentiation and normalization. Every dimension corresponds to the similarity of the input image (text) to a unique image (text) entry in the multimodal dataset. The visualization shows training samples leading to correct or incorrect predictions, which is useful to curate better data. For example, the second training sample is broken; its removal enables production of an updated ASIF model in seconds.
  }
 \label{fig-deepdive}
 \end{figure}

\subsection{Dynamically Updating an ASIF Model}

Consider classifying EuroSAT~\citep{helber2019eurosat} satellite images using ASIF. Initially, the zero-shot performance outperforms random chance but is not impressive ($29.4\%$ unsup. configuration, 10 classes). CLIP, while better, also falls short with a $54.1\%$ performance rate.

Consider acquiring 100 new image-text pairs from EuroSAT with the goal of developing a new multimodal model based on an updated training set, anticipating the need to process more satellite images in the future. With CLIP, this would require fine-tuning or retraining the model from scratch using the updated dataset.
In contrast, ASIF requires only obtaining and storing the new pairs' embeddings. The model retains its usability for all the previous images, but it now also makes accurate predictions for satellite images, achieving an $82.2 \pm 2.0\%$ accuracy on EuroSAT.\footnote{Why does CLIP perform better zero-shot? Likely, its 400M pairs have more samples close to satellite images than ASIF's 1.6M samples from CC12M, evidenced by ASIF's drastic improvement with just 100 new samples.}

Similarly, when samples must be removed from the multimodal training dataset---either because they are faulty (as shown in \cref{fig-deepdive}) or because the license to use them has been lost---the process is as simple as deleting the corresponding embeddings to get a new model.

\underline{\textit{Summary:}}
ASIF enables quick model adjustments and data handling without the need for retraining, unlike traditional models such as CLIP. This demonstrates its practicality in real-world scenarios, such as the emergence of a new setting or the loss of rights for assets used during training.

\subsection{Deep Dive into a Classification}

To better understand why and how the ASIF procedure works, this section follows step by step the single classification depicted in \cref{fig-deepdive}, showing the entries of the multimodal dataset used to build the relative representations. For simplicity, assume $k=23$.

The task is to find the ImageNet label of the upper left image in \cref{fig-deepdive}. The first step is to compute its relative representation with respect to the multimodal dataset, which selects the 23 most similar images in the collection. The most similar are indeed triumphal arches that---as similarity decreases---turn to generic structures with arches or ancient monuments. No other image in the multimodal dataset is involved in the classification; all the entries in the collection except for these 23 could be replaced, and as long as no new image is more similar to the input image, every computation in the classification process would remain the same. This consistency of the ASIF procedure is in contrast to traditional approaches like CLIP and LiT, where the parameters defining the encoder computations are shaped by the whole multimodal dataset through contrastive learning.

The central step of the ASIF procedure follows: jumping into the text space by treating the relative representation of the input image \textit{as if} it were the relative representation of its ideal caption. The vector remains the same, but its meaning changed: now it signifies how much the ideal caption should be similar to 23 captions.

The efficacy of this jump hinges entirely on the quality of the multimodal dataset. Looking at the 23 captions in \cref{fig-deepdive}, the most fitting candidate caption corresponds to one of the prompts associated with the \texttt{triumphal\_arch} label in ImageNet, such as \textit{``a photo of a triumphal arch''}.
Conversely, a dataset featuring 23 non-informative captions, such as file names \textit{``IMG\_20180823.jpg''} or camera settings \textit{``D90 18.0-70.0 mm f/3.5-4.5''}, would not allow the model to recognize the correct class even with the best image and text encoders.
This limitation is actually a defining feature of the ASIF procedure, as the meaning attributed to the input is ultimately determined by the multimodal dataset and not by the encoders: by acting just on the image-text pairs, full control over the model's output is maintained.

\chapter[Learning to Align]{Learning to Align: Scalable Unsupervised Alignment via Learned Embeddings}
\label{ch:learning_to_align}

\chapterpublication{vedula2024scalable}{\textbf{ICML AI4Science Workshop 2024}}{An amortized Gromov-Wasserstein solver learns correspondence without paired samples, and the learned alignment generalizes to new samples with a single optimal transport solve.}

\section{Introduction}

This chapter presents a scalable approach to unsupervised alignment that learns the correspondence structure through gradient-based optimization. Unlike the correspondence-parallel methods examined in the preceding chapters (\cref{ch:relrep,ch:product-invariances,ch:bootstrapping,ch:asif}), which extract \glspl{semantic-coordinate} via predefined geometric transformations, this method employs learned neural embeddings to implicitly parametrize the transport cost between representation spaces.

Within the correspondence framework established in \cref{ch:methodology}, this approach explores \emph{learned correspondence} operators. While the method constructs a correspondence $\gls{Corr}$ between metric spaces by optimizing embeddings that minimize metric disagreement (the optimal coupling $\mP^*$ of \cref{sec:learned_gw_approach} acts as a soft correspondence $\Corr$), it differs from explicit correspondence-parallel operators in its treatment of the optimization problem. The embeddings $f_\theta$ and $g_\phi$, one per space, parametrize the transport cost, and the plan obtained from that cost aligns the pairwise distance structures. Like the operators of the preceding chapters, $f_\theta$ and $g_\phi$ map each space into a common domain. Here the maps are learned; in those chapters they were fixed in advance.

The central contribution addresses scalability: standard Gromov-Wasserstein (GW) solvers require solving a sequence of entropic optimal transport problems at inference time, limiting applicability to large-scale settings. By amortizing this computational burden into a training phase that learns embeddings, the method reduces inference to a single optimal transport problem. This inductive capability enables zero-shot alignment of new samples without recomputing the full GW solution.

The chapter is organized as follows. \cref{sec:learned_gw_approach} introduces the core formulation as a bilevel optimization problem that learns embeddings minimizing metric disagreement. \cref{sec:extensions} presents extensions including simulated annealing of entropic regularization, spectral-geometric regularization exploiting graph structure, and a ranking-based formulation achieving scale invariance. \cref{sec:learning_experiments} validates the approach on synthetic isometric transformations and real-world single-cell multiomics alignment tasks, demonstrating both computational efficiency and alignment quality improvements over standard GW solvers.

\section{Learning transport cost via embeddings}
\label{sec:learned_gw_approach}
\begin{figure}[t]
 \centering
 \includegraphics[width=0.5\textwidth]{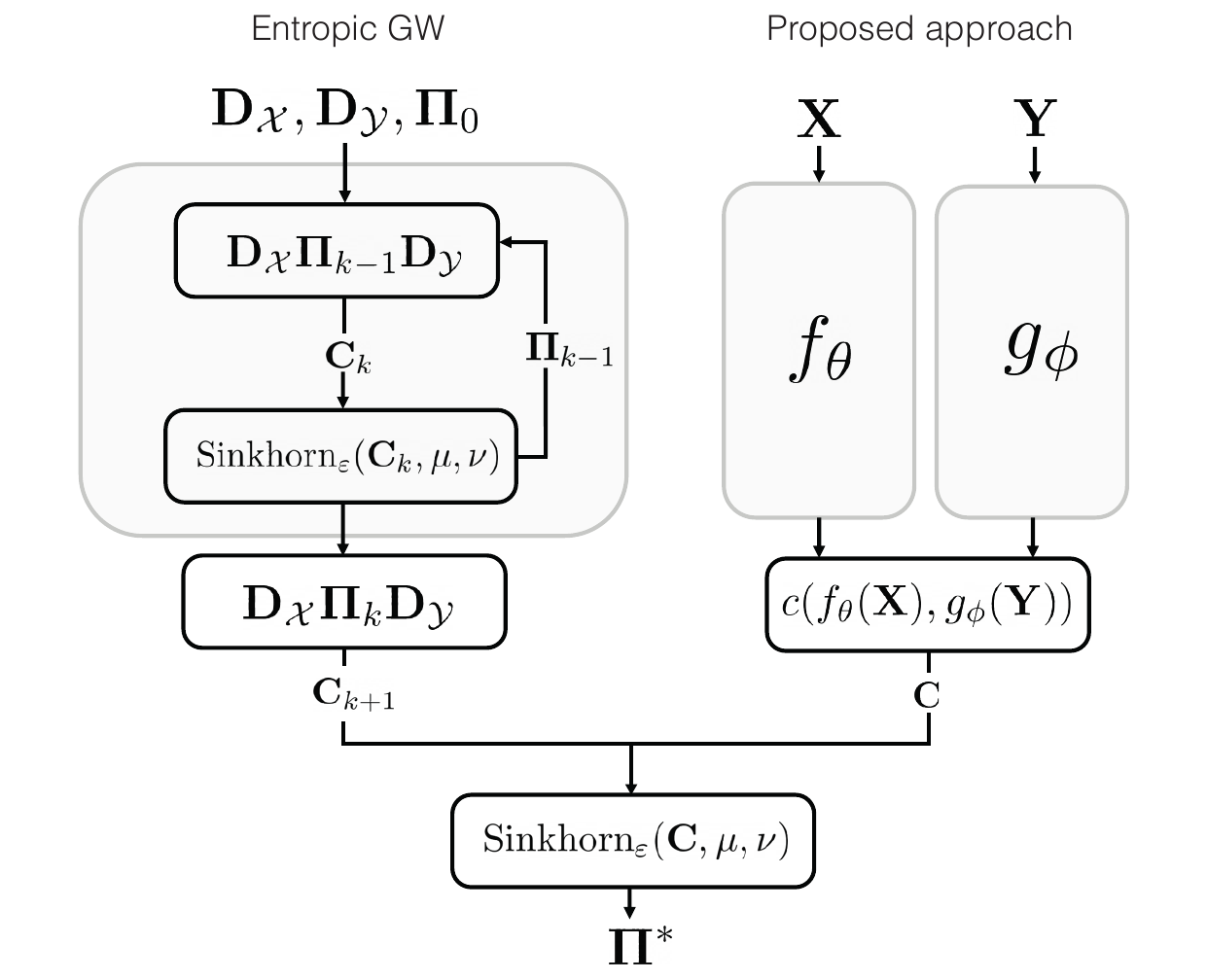}
 \caption{Entropic Gromov-Wasserstein solver (left) solves a sequence of regularized optimal transport ($\epsilon$-OT) problems using the Sinkhorn algorithm. In contrast, the proposed approach learns, via a pair of embeddings, $f_\theta$ and $g_\phi$, the transport cost that directly produces the sought alignment $\mP^*$ by solving a single $\epsilon$-OT problem. While the learning of the embeddings still requires multiple calls to the $\epsilon$-OT solver, their cost is amortized at inference time. In the figure, $\Pi$ denotes the coupling $\mP$, $k$ the index of the solver iteration $t$, and $\varepsilon$ the regularization strength $\epsilon$.}
 \label{fig:our_approach_vs_gw}
\end{figure}

Consider $n$ samples drawn from a space $\mathcal{X}$ equipped with a metric $\operatorname{dist}_{\mathcal{X}}$, and $n$ samples from a space $\mathcal{Y}$ with metric $\operatorname{dist}_{\mathcal{Y}}$. Two descriptions of each set matter in what follows: the pointwise features of its samples, stacked in the matrices $\mX \in \R^{n \times \dimof{X}}$ and $\mY \in \R^{n \times \dimof{Y}}$, and the pairwise distances between them, collected in $\mD_{\mathcal{X}}, \mD_{\mathcal{Y}} \in \R^{n \times n}$. The weight vectors $\mu, \nu \in \Delta_n$, points of the probability simplex, assign a weight to every sample: uniform or, depending on what the spaces represent, given by discrete volume elements. Because the two sets live in different spaces, their samples cannot be compared feature by feature, and aligning them amounts to finding a permutation that pairs each sample of one set with a sample of the other. The Gromov-Wasserstein distance~\citep{memoli2011gromov} relaxes this search from permutations to couplings, that is, to the set $U(\mu, \nu) = \{ \mP \in \R_+^{n \times n} \,|\, \mP \1_n = \mu,\ \mP^\top \1_n = \nu \}$ of nonnegative matrices with the prescribed marginals. Of a coupling, GW asks only that it preserve the pairwise distances as well as possible:
\begin{equation}
 \label{eq:gw}
 \operatorname{dist}_{\mathrm{GW}}^2 = \min_{\mP \in U(\mu, \nu)} \left\| \mD_{\mathcal{X}} - \mP \mD_{\mathcal{Y}} \mP^\top \right\|_F^2.
\end{equation}
The price of this formulation is that the coupling appears on both sides of $\mD_{\mathcal{Y}}$, so the objective is quadratic in $\mP$ and \cref{eq:gw} is a (relaxed) quadratic assignment problem (QAP). The contrast with the case in which both sets lie in the same space is instructive, because there alignment reduces to a linear optimal transport (OT) problem, $\min_{\mP \in U(\mu, \nu)} \langle \mP, \mC \rangle$ for a cost matrix $\mC$~\citep{peyre2019computational}, and adding the entropic term $\epsilon \langle \mP, \log \mP \rangle$ turns it into the $\epsilon$-OT problem, which the Sinkhorn algorithm solves efficiently~\citep{cuturi2013sinkhorn}. The entropic GW solver~\citep{solomon_entropic_QAP, peyre2019computational}, which adds the same entropic term to \cref{eq:gw}, exploits exactly this contrast: it linearizes \cref{eq:gw} around the current coupling $\mP_t$, reads a cost matrix $\mC_{t+1}$ off that linearization, and projects back onto $U(\mu, \nu)$ by solving one $\epsilon$-OT problem, so that a single quadratic problem is replaced by a sequence of linear ones.

To scale GW solvers to large sample sizes, the approach begins with the following question: can an entropic OT problem be identified whose solution coincides with that of the entropic GW problem? The rationale is that, given an unpaired set of samples, if an equivalent entropic OT problem can be determined, fast entropic OT solvers can be employed to calculate the assignment. One problem fitting this criterion, by construction, is the entropic OT problem solved in the last iterate of the entropic GW solver. However, computing this problem would require iterating through the GW solver, making it impractical. Phrasing this question as an optimization problem yields:
\begin{equation}
\begin{aligned}
 \label{eq:ideal_problem_ours}
 \mC^* =& \argmin_{\mC} \, \left\| \mD_{\mathcal{X}} - \mP(\mC) \mD_{\mathcal{Y}} \mP^\top(\mC) \right\|_F^2 \\
 & \,\, \text{ s.t. } \mP(\mC) = \argmin_{\mP \in U(\mu, \nu)} \langle \mP, \mC\rangle.
\end{aligned}
\end{equation}
It is a bilevel optimization problem whose solution yields the alignment $\mP^* = \mP(\mC^*)$: the inner problem is linear OT and it produces an assignment that is optimal with respect to the cost $\mC$, and the outer problem demands that the resulting $\mP(\mC)$ is GW-optimal, i.e., it aligns the metrics $\mD_{\mathcal{X}}$ and $\mD_{\mathcal{Y}}$. While seemingly elegant, \cref{eq:ideal_problem_ours} has two major problems: (i) because $\mC$ is unbounded, this objective is very unstable and difficult to optimize; (ii) more practically, \cref{eq:ideal_problem_ours} results in a \textit{transductive} approach; given a new set of unpaired samples, this problem needs to be solved anew, which is not scalable.

To mitigate this, instead of optimizing the cost matrix $\mC$ (in \cref{eq:ideal_problem_ours}), the method \textit{implicitly} parametrizes it as a pairwise cost measured on \textit{learned embeddings} of pointwise features $\mX$ and $\mY$. This leads to the following modified objective,
\begin{equation}
\begin{aligned}
 \label{eq:final_objective_ours}
 (\theta^*, \phi^*) =& \argmin_{\theta, \phi} \, \left\| \mD_{\mathcal{X}} - \mP(\theta, \phi) \mD_{\mathcal{Y}} \mP^\top(\theta, \phi) \right\|_F^2 \\
 & \,\, \text{ s.t. } \mP(\theta, \phi) = \argmin_{\mP \in U(\mu, \nu)} \langle \mP, {c}(f_\theta(\mX), g_\phi(\mY))\rangle,
\end{aligned}
\end{equation}
where $f, g$ are learnable functions, modeled via neural networks, embedding $\mX$ and $\mY$, respectively, into a common domain $\mathcal{Z}$, and the alignment is $\mP^* = \mP(\theta^*, \phi^*)$. The cost is realized through the embedding, while the ground cost $c: \mathcal{Z} \times \mathcal{Z} \to \R$, evaluated on $\vz_i = f_\theta(\vx_i)$ and $\vz'_j = g_\phi(\vy_j)$ with $\vx_i$, $\vy_j$ the rows of $\mX$, $\mY$, is fixed to the simple Euclidean ($c(\vz, \vz') = \| \vz - \vz' \|^2$) or cosine ($c(\vz, \vz') = 1 - \cos(\vz, \vz')$) form; $c(f_\theta(\mX), g_\phi(\mY))$ denotes the $n \times n$ matrix of these values. The problem is solved via gradient descent. To backpropagate gradients to $f$ and $g$, the inner problem is first relaxed to an $\epsilon$-OT problem, then implicit differentiation~\citep{amos2017optnet} is employed to calculate $\frac{\partial{\mP}}{\partial{\mC}}$~\citep{eisenberger2022unified}, which is backpropagated to update the weights of $f$ and $g$.

From a geometric perspective, the samples from $\mathcal{X}$ and $\mathcal{Y}$ are embedded into a common domain $\mathcal{Z}$, where the samples are \textit{OT-aligned} with the same assignment that makes the metric spaces $(\mathcal{X}, \operatorname{dist}_{\mathcal{X}})$ and $(\mathcal{Y}, \operatorname{dist}_{\mathcal{Y}})$ \textit{GW-aligned}.
From a computational standpoint, this framework can be viewed as an \textit{amortized entropic GW solver}. \cref{fig:our_approach_vs_gw} presents the parallels between this solver and the entropic GW solver~\citep{solomon_entropic_QAP}.
The ground cost measured on the embeddings $c(f_\theta(\mX), g_\phi(\mY))$ can be interpreted as the cost matrix $\mC_{t+1} = \mD_{\mathcal{X}} \mP_t \mD_{\mathcal{Y}}$ (as depicted in \cref{fig:our_approach_vs_gw}) produced by running the entropic GW solver for $t$ iterations. Post training, the neural networks can be viewed as amortizing the GW iterations, in similar spirit to recent amortized optimization techniques proposed for fast calculation of convex conjugates~\citep{amos2023tutorial, amos2022amortizing}.

From a practical standpoint, this results in an \textit{inductive} GW solver. At inference, when a new set of unpaired samples from $\mathcal{X}$ and $\mathcal{Y}$ is encountered, only a single entropic OT problem must be solved, which is highly scalable. Moreover, since the solver is gradient-descent-based, it allows flexibility to induce domain knowledge, additional regularization, and inductive biases on the assignment, the cost, and the neural networks $f, g$, respectively. Several such examples are discussed in \cref{sec:extensions}.

Finally, while this approach may resemble the \textit{inverse OT} (iOT) problem~\citep{inv_ot1, inv_ot2} in that it involves learning the transport cost, it differs in the minimized objective. While iOT targets finding a cost realizing a given assignment (hence, requiring coupled data), this learning problem does not assume a known target permutation. It finds one minimizing the pairwise distance disagreement on unaligned data. From this perspective, the approach can be seen as a variational analog of the iOT problem.

\section{Extensions: regularization and scale invariance}\label{sec:extensions}
While the objective function in \cref{eq:final_objective_ours} is straightforward to evaluate, the resulting optimization problem remains an NP-hard QAP. In practice, reaching good local minima consistently without imposing further inductive biases is challenging. This is also true for the entropic GW solver~\citep{peyre2019computational}---while sequential linearization and projection via Sinkhorn algorithm works reasonably in practice, no guarantees exist on its global convergence. In fact, many scenarios exist where it fails to recover a meaningful local minimum. This section introduces several regularization techniques to remediate poor convergence: (i) \textit{simulated annealing} of the entropic regularization strength $\epsilon$, and (ii) \textit{spectral-geometric regularization of the OT cost}. A new objective matching distance ranks instead of distances themselves is also presented, which can be employed as an alternative to \cref{eq:final_objective_ours}.

\begin{figure}[t]
 \centering
 \includegraphics[width=0.25\textwidth]{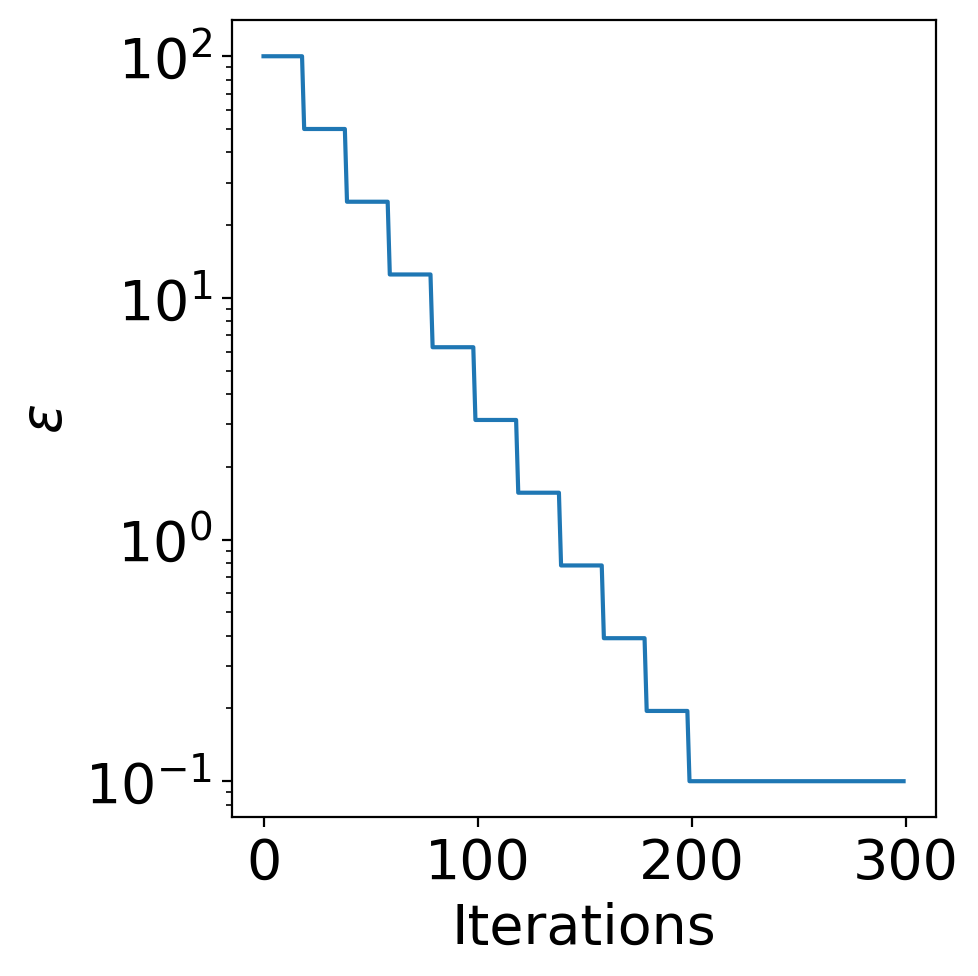}
 \includegraphics[width=0.25\textwidth]{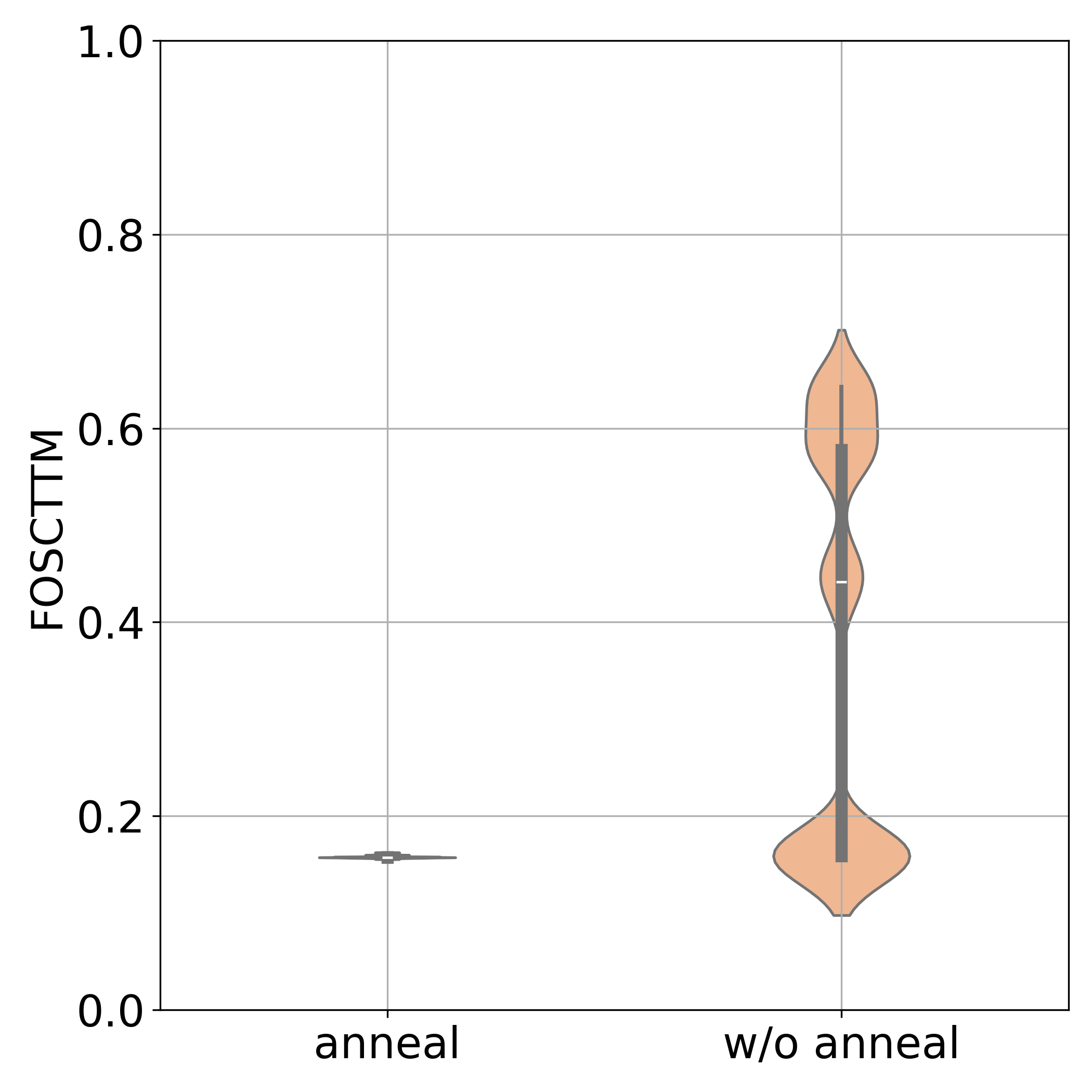} \includegraphics[width=0.25\textwidth]{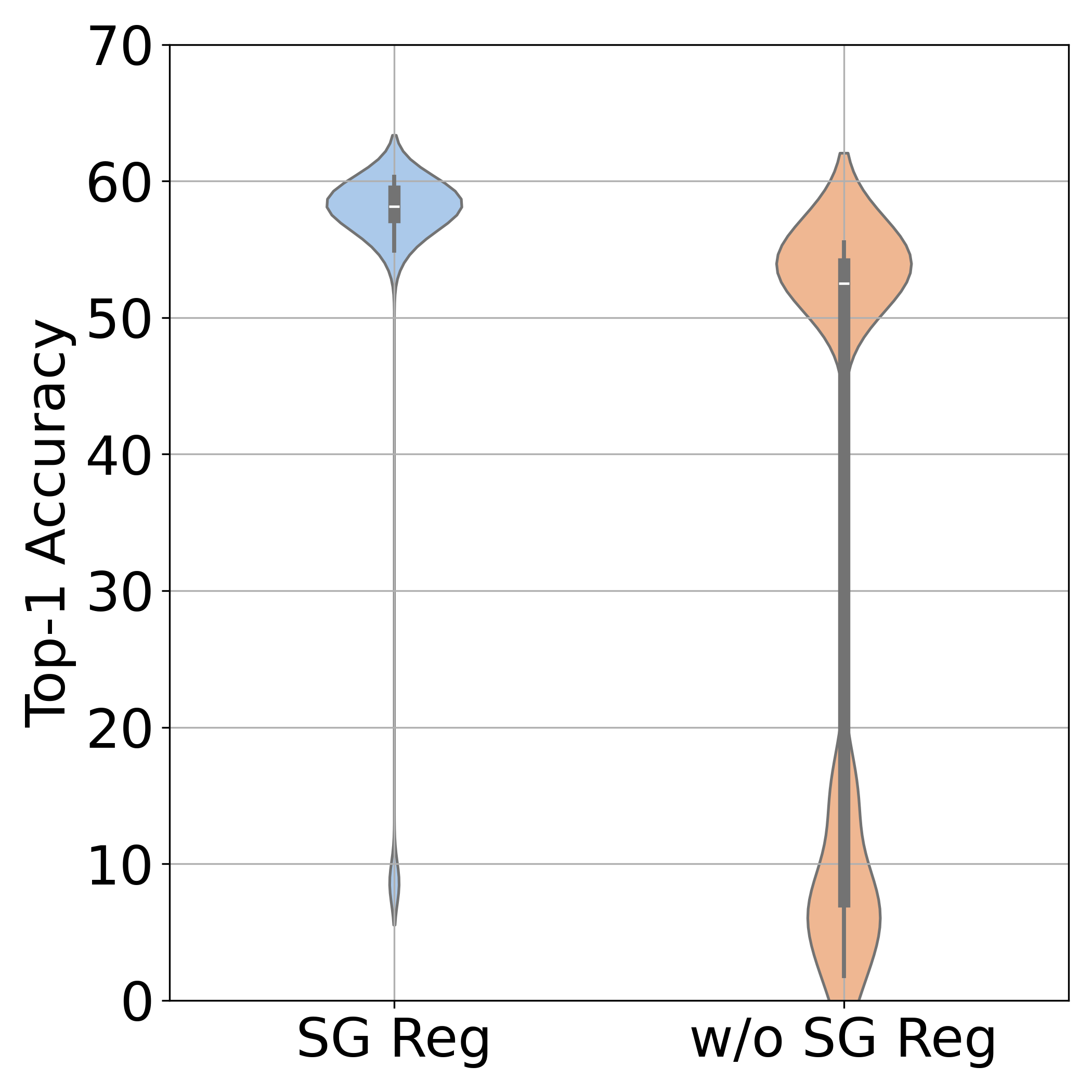}
 \caption{\textbf{Simulated annealing of $\epsilon$ and spectral geometric regularization are effective in stabilizing the solver and improving the accuracy of the assignment.} Left: simulated annealing schedule used. Middle: distribution of the alignment error (measured as FOSCTTM) over $20$ runs with and without $\epsilon$-annealing. Right: distribution of the top-1 assignment accuracy with and without the spectral geometric regularization of the transport cost.}
 \label{fig:sim_annealing}
\end{figure}

\paragraph{Simulated annealing of $\epsilon$.} Evaluating the solver on the scSNARE-seq data~\citep{chen2019high}, where the goal is to align transcriptomic readouts against those of chromatin accessibility (ground-truth available via co-assaying~\citep{chen2019high}), reveals sensitivity to initialization of the neural networks $f$ and $g$. As a result of symmetries in the metric spaces, the assignment sometimes consistently mismaps the cell line GM12878 to H1, and vice-versa.
The middle panel of \cref{fig:sim_annealing} depicts the distribution of alignment errors (lower is better) obtained by solving \cref{eq:final_objective_ours} with multiple random initializations of the embedding parameters $\theta$ and $\phi$. The error is the fraction of samples closer than the true match (FOSCTTM): for each sample, once both domains are represented in a common space, the fraction of samples of the other domain that lie closer to it than its true match, averaged over both domains ($0$ for perfect alignment). In the right column of that panel (without annealing), the largest mode corresponds to accurate assignment, whereas the two other modes with larger errors represent symmetry-induced cell-line mismappings.

This problem is mitigated by performing \textit{simulated annealing} on $\epsilon$ of the entropic OT problem within the Sinkhorn layer. A schedule for $\epsilon$ that starts high and is gradually decayed (see \cref{fig:sim_annealing}, left) results in coarse-to-fine refinement of the learned cost (implicitly parametrized via $f$ and $g$) during training, similar in spirit to multi-scale kernel matching in shape correspondence problems~\citep{vestner2017efficient, melzi2019zoomout,holzschuh2020simulated}. When $\epsilon$ is high, the entropic regularization is strong, and the resulting assignment is softer. By scheduling $\epsilon$ from a large value to a small one, the learned cost matrix and the resulting assignment are refined during training. In practice, the proposed $\epsilon$-scheduling works well, practically reducing variance across seeds to zero and effectively breaking symmetries in the metric spaces that lead to bad local minima (\cref{fig:sim_annealing}, middle).

\looseness=-1\paragraph{Spectral representation on graphs.} Before introducing the spectral-geometric regularization, a brief background on graphs is provided. A familiar reader may skip to the following paragraph.
Let $\mathcal{G} = (\mathcal{V}, \mathcal{E}, \mW)$ be a weighted graph with the vertex set $\mathcal{V}$, edge set $\mathcal{E}$, and weighted adjacency matrix $\mW$. The \textit{combinatorial graph Laplacian} is defined as $\mL = \operatorname{diag}(\mW\1) - \mW$, where $\operatorname{diag}(\mW\1)$ is \textit{the degree matrix}. Given a scalar-valued function $\vh \in \R^{\lvert\mathcal{V}\rvert}$ on the graph $\mathcal{G}$, the \textit{Dirichlet energy} is defined to be $\vh^\top\mL\vh$, and it measures the \textit{smoothness} of $\vh$ on $\mathcal{G}$ \citep{spielman2012spectral}.
Given two graphs $\mathcal{G}_{\mathcal{X}} = (\mathcal{V}_{\mathcal{X}}, \mathcal{E}_{\mathcal{X}}, \mW_{\mathcal{X}})$ and $\mathcal{G}_{\mathcal{Y}} = (\mathcal{V}_{\mathcal{Y}}, \mathcal{E}_{\mathcal{Y}}, \mW_{\mathcal{Y}})$, the Cartesian product of $\mathcal{G}_{\mathcal{X}}$ and $\mathcal{G}_{\mathcal{Y}}$, denoted by $\mathcal{G}_{\mathcal{X}}\,\Box\,\mathcal{G}_{\mathcal{Y}}$, is defined as a graph with the vertex set $\mathcal{V}_{\mathcal{X}} \times \mathcal{V}_{\mathcal{Y}}$, on which two nodes $(u, v), (u', v')$ are adjacent if either $u=u'$ and $(v, v') \in \mathcal{E}_{\mathcal{Y}}$ or $v = v'$ and $(u, u') \in \mathcal{E}_{\mathcal{X}}$.
The Laplacian of $\mathcal{G}_{\mathcal{X}}\,\Box\,\mathcal{G}_{\mathcal{Y}}$ is defined as a tensor sum of the Laplacians $\mL_{\mathcal{X}}$ and $\mL_{\mathcal{Y}}$, i.e., $\mL_{\mathcal{G}_{\mathcal{X}} \Box \mathcal{G}_{\mathcal{Y}}} = \mL_{\mathcal{X}} \oplus \mL_{\mathcal{Y}} = \mL_{\mathcal{X}} \otimes \mI + \mI \otimes \mL_{\mathcal{Y}}$.
Denote the spectral decompositions of the Laplacians by $\mL_{\mathcal{X}} = \mPhi_{\mathcal{X}} \mLambda_{\mathcal{X}} \mPhi_{\mathcal{X}}^\top$ and $\mL_{\mathcal{Y}} = \mPhi_{\mathcal{Y}} \mLambda_{\mathcal{Y}} \mPhi_{\mathcal{Y}}^\top$. A signal $\mH$ on the product graph $\mathcal{G}_{\mathcal{X}}\,\Box\,\mathcal{G}_{\mathcal{Y}}$ can be represented using the bases of the individual Laplacians as $\mH = \mPhi_{\mathcal{X}} \hat{\mF} \mPhi_{\mathcal{Y}}^\top$, with the coefficients $\hat{\mF}$.

\paragraph{Spectral-geometric regularization of the OT cost.} A spectral-geometric regularization on the learned OT cost demands that similar items in $\mathcal{X}$ incur similar cost with respect to all items in $\mathcal{Y}$, and vice-versa. To formalize this notion, let $\mathcal{G}_{\mathcal{X}} = (\mathcal{V}_{\mathcal{X}}, \mathcal{E}_{\mathcal{X}}, \mW_{\mathcal{X}})$ and $\mathcal{G}_{\mathcal{Y}} = (\mathcal{V}_{\mathcal{Y}}, \mathcal{E}_{\mathcal{Y}}, \mW_{\mathcal{Y}})$ be two graphs inferred on the samples of $\mathcal{X}$ and $\mathcal{Y}$, respectively, and let $\mL_{\mathcal{X}}$ and $\mL_{\mathcal{Y}}$ be their corresponding graph Laplacians. The learned OT cost $\mC=c(f_\theta(\mX), g_\phi(\mY))$ from \cref{eq:final_objective_ours} is interpreted as a signal on the product graph $\mathcal{G}_{\mathcal{X}}\,\Box\,\mathcal{G}_{\mathcal{Y}}$, with the requirement that $\mC$ is \textit{smooth} on $\mathcal{G}_{\mathcal{X}} \, \Box \, \mathcal{G}_{\mathcal{Y}}$. This smoothness can be expressed as the Dirichlet energy of $\mC$ measured on $\mathcal{G}_{\mathcal{X}} \, \Box \, \mathcal{G}_{\mathcal{Y}}$,
\begin{equation}
 \label{eq:dir_energy}
 \mathcal{R}_{\mathrm{sm}} = \operatorname{vec}(\mC)^\top\left(
 \mL_{\mathcal{X}} \oplus \mL_{\mathcal{Y}}
 \right)\operatorname{vec}(\mC) =
 \Tr\left( \mC^\top \mL_{\mathcal{X}} \mC +
 \mC \mL_{\mathcal{Y}} \mC^\top \right),
\end{equation}
with $\operatorname{vec}$ the row-major vectorization, and added to \cref{eq:final_objective_ours} as an additional regularization. \cref{fig:sim_annealing} (right) demonstrates the effectiveness of the spectral regularization on the task of aligning embeddings from neural latent spaces.

From the spectral perspective, interpreting the OT cost $\mC$ as a signal on $\mathcal{G}_{\mathcal{X}}\,\Box\,\mathcal{G}_{\mathcal{Y}}$, learning $\mC$ given the pointwise features from domains $\mathcal{X}$ and $\mathcal{Y}$ is equivalent to directly learning the spectral coefficients $\mPhi_{\mathcal{X}}^\top \mC \mPhi_{\mathcal{Y}}$ of $\mC$ (a functional-map-like object; \cref{ch:latent_functional_maps} uses $\mC$ for the functional map itself). This connection relates the approach to methods that learn functional maps~\citep{litany2017deep, halimi2019unsupervised, vestner2017efficient, boyarski2022spectral, kalofolias2014matrix} from shape correspondence and geometric matrix completion literature. This connection to functional maps is explored further in \cref{ch:latent_functional_maps}, which presents spectral methods for correspondence-coupled operators.

\paragraph{Matching ranks instead of distances.} The choice of comparison criterion for pairwise distances strongly influences the usability of the GW problem for real applications. Consider two point clouds that differ only by a scale factor; since distances are not scale-invariant, solving \cref{eq:final_objective_ours} to match distances would produce meaningless results. As a remedy, the \textit{ranks} of pairwise distances are matched instead of their absolute values. Ranks preserve order and are insensitive to scale or, more generally, monotone transformations. This departs from the standard framework of Gromov-Hausdorff and GW, which align metric spaces, generalizing to a problem of performing unpaired alignment by matching non-metric quantities. To differentiate the objective with respect to ranks---an inherently non-differentiable function---the differentiable soft ranking operators introduced by \citet{blondel2020fast} are employed.
The following modified objective is optimized:
\begin{equation}
\begin{aligned}
 \label{eq:final_objective_ours_rank}
 (\theta^*, \phi^*) =& \argmin_{\theta, \phi} \,
 \left\|
 \operatorname{srank}_\delta\left(\mD_{\mathcal{X}}\right) - \operatorname{srank}_\delta\left(\mP(\theta, \phi) \mD_{\mathcal{Y}} \mP^\top(\theta, \phi) \right)
 \right\|_F^2 \\
 & \,\, \text{ s.t. } \mP(\theta, \phi) = \argmin_{\mP \in U(\mu, \nu)} \langle \mP, {c}(f_\theta(\mX), g_\phi(\mY))\rangle,
\end{aligned}
\end{equation}
where $\operatorname{srank}_\delta$ is a soft-ranking operator applied separately to each row of the matrix, and $\delta$ controls the level of softness of the rank; as before, the alignment is $\mP^* = \mP(\theta^*, \phi^*)$. Because ranking is a nonlinear operation, this results in a problem that is no longer quadratic in $\mP$. It is unclear how standard GW solvers can be adapted to such settings, highlighting the benefit of a gradient-descent-based solver.
Applying ranking to other groups of distances effectively results in a different GW-like distance. Systematic exploration of this new family of distances is deferred to future work.

\textbf{Further extensions.} The framework admits several natural extensions, none of which is tested here: (i) extension to a fused GW~\citep{fusedgw} setting by adding a linear objective to \cref{eq:final_objective_ours,eq:final_objective_ours_rank}; (ii) controlling the rank of the OT cost by modifying the dimension of the embeddings' output by $f$ and $g$; and (iii) incorporating partial supervision on the assignment (semi-supervised alignment) into the loss as a data term.

\section{Experimental validation}
\label{sec:learning_experiments}
The experimental evaluation is organized into three parts.
First, the solver's performance in the inductive setting and its scalability to large sample sizes are demonstrated. Second, experiments show the effects of (i) simulated annealing of $\epsilon$, (ii) spectral geometric regularization, and (iii) the ranking-based formulation. Third, the solver's performance in the transductive setting is compared against the entropic GW solver on two single-cell multiomics benchmarks. Both real and synthetic data are employed as appropriate.

\textbf{Inductivity and scale.} To evaluate the method's inductive capability and benchmark it against the entropic GW solver, two experiments are considered: (i) when $\mathcal{X}$ and $\mathcal{Y}$ are isometric, and (ii) when $\mathcal{X}$ and $\mathcal{Y}$ are not exactly isometric.

\begin{figure}[t]
 \centering
 \includegraphics[width=0.4\textwidth]{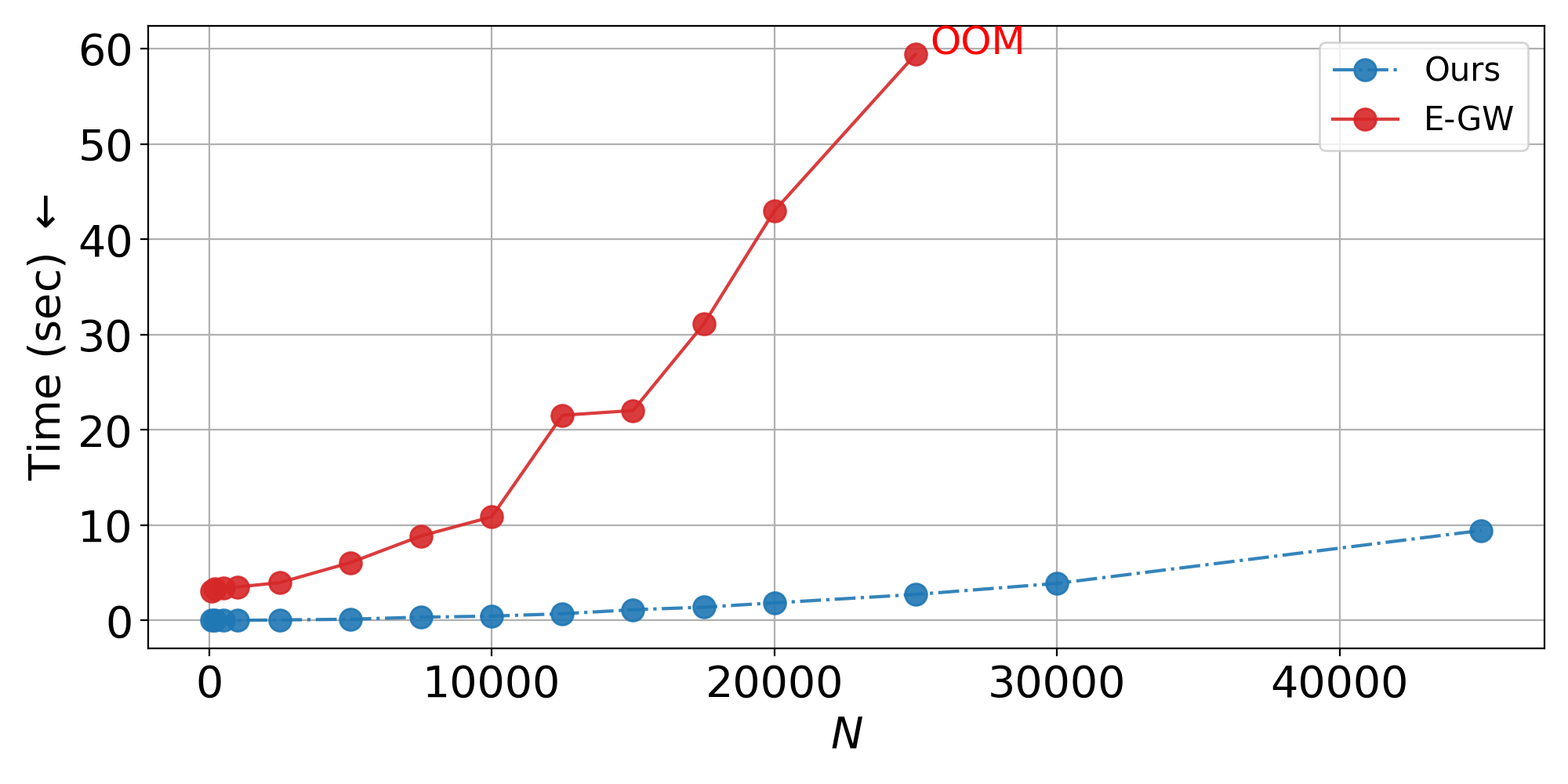}
 \includegraphics[width=0.4\textwidth]{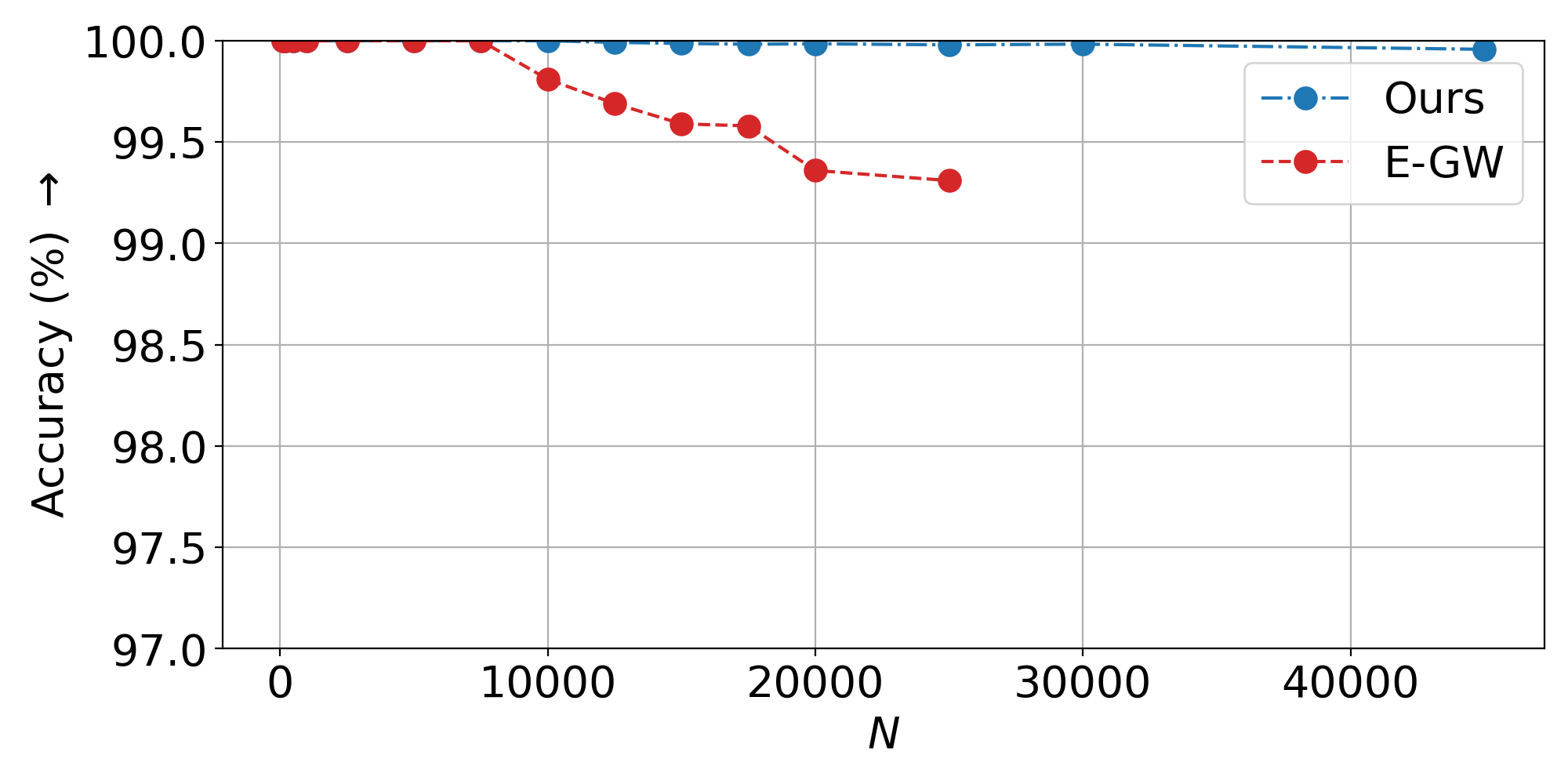}\\
 \includegraphics[width=0.4\textwidth]{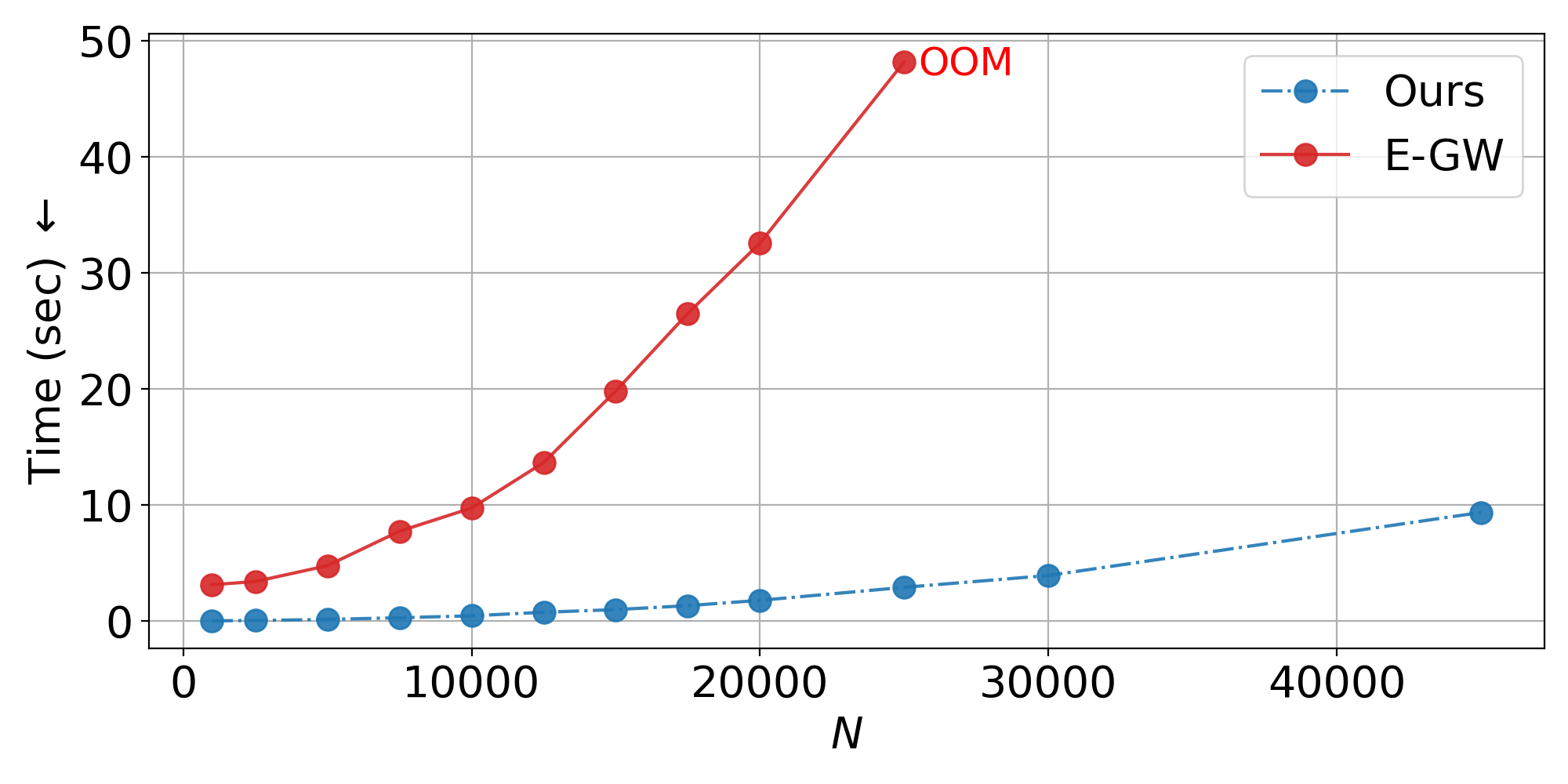}
 \includegraphics[width=0.4\textwidth]{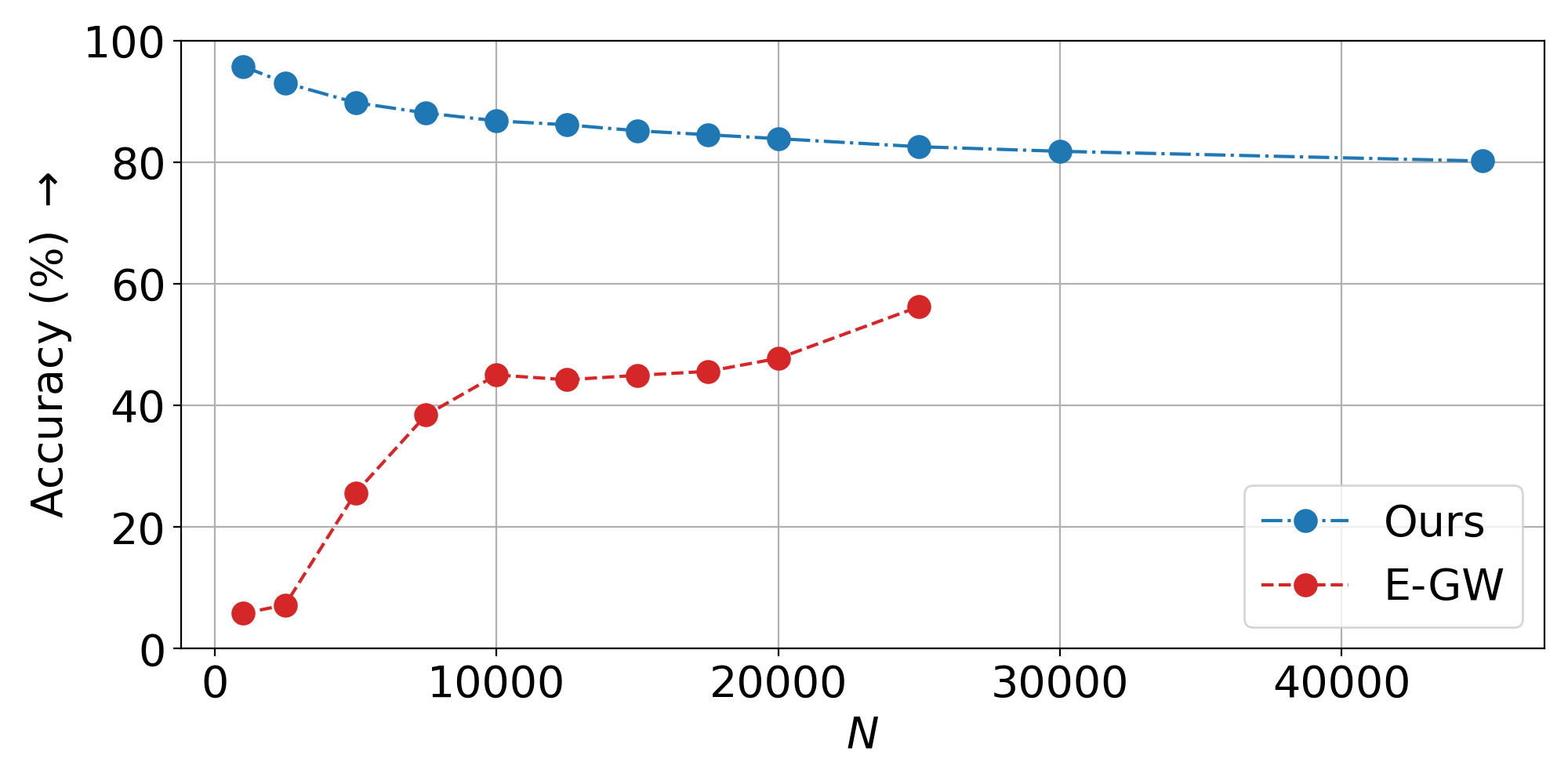}
 \caption{\textbf{The proposed solver generalizes to unseen samples and scales to large-sample sizes post-training.} In both top and bottom experiments, $\mathcal{X}$ and $\mathcal{Y}$ are ViT embeddings. The entropic GW solver can only operate in the transductive regime and runs out of memory for $n>25000$ ($N$ on the horizontal axis is $n$; E-GW is the entropic GW solver; OOM marks where it runs out of memory).}
 \label{fig:scale_inductive}
\end{figure}

For the \textbf{first experiment}, $\mathcal{X}$ comprises CIFAR-100 encodings obtained from a vision transformer~\citep{dosovitskiy2021an}. An orthogonal transformation is applied to each element of $\mathcal{X}$ to generate $\mathcal{Y}$. The encoders $f$ and $g$ are parametrized as 3-layer multi-layer perceptrons (MLPs), and \cref{eq:final_objective_ours} is optimized with respect to their parameters on $200$ unaligned samples for $500$ iterations ($12$ seconds). The method is then evaluated in an inductive setting with an increasing number of unaligned samples available at inference up to $n=45000$, benchmarked against the GPU-accelerated entropic GW solver from \texttt{ott-jax}~\citep{cuturi2022optimal}. The results are presented in the top panels of \cref{fig:scale_inductive}. Accuracy is measured as whether the predicted correspondence is correct in terms of class label. Both solvers recover the orthogonal transformation almost exactly (above 99\% accuracy). An inductive solver, because it solves only an $\epsilon$-OT problem at inference, is much faster and more memory efficient. The entropic GW solver, in contrast, runs out of memory for $n>25000$. The reported times do not include the time required to compute a geodesic distance matrix for both $\mathcal{X}$ and $\mathcal{Y}$, which is time-consuming at large sample sizes (>10 mins for $n=20000$). In contrast, the learned solver does not require computing $\mD_{\mathcal{X}}$ and $\mD_{\mathcal{Y}}$ at inference. For the \textbf{second experiment}, data from \citet{maiorca2023semantic} is used: $\mathcal{X}$ comprises ViT embeddings as in the previous experiment, while $\mathcal{Y}$ comprises ViT embeddings generated from rescaled images. The functions $f$ and $g$ are trained for $1000$ iterations ($\sim 2$ minutes), using $1000$ unpaired samples. The results are presented in the bottom two panels of \cref{fig:scale_inductive}. The results demonstrate that the solver generalizes well and scales gracefully with sample sizes, whereas the entropic GW solver produced inferior results in this setting. These experiments validate that the solver attains high-quality solutions and scales well in the inductive regime.

\textbf{Spectral geometric regularization.} For this experiment, the setting where $\mathcal{X}$ and $\mathcal{Y}$ are two unaligned sets of embeddings obtained from a pretrained vision transformer~\citep{dosovitskiy2021an} is considered. The functions $f$ and $g$ are set to be 3-layer MLPs, and \cref{eq:final_objective_ours} is solved with and without $\mathcal{R}_{\mathrm{sm}}$ regularization (\cref{eq:dir_energy}).
The problem is solved on $20$ unaligned datasets drawn from $\mathcal{X}$ and $\mathcal{Y}$, each of size $n=1000$. \cref{fig:sim_annealing} (right panel) presents the accuracy of the assignment by measuring if the predicted corresponding point belongs to the same class as the groundtruth correspondence. Geometric regularization improves the accuracy of the assignment (+20\% in terms of mean accuracy over trials). Moreover, it also reduces the variance, thereby inducing meaningful inductive bias into the solver.

\textbf{Simulated annealing of $\epsilon$.} As discussed in \cref{sec:extensions}, the scSNARE-seq data~\citep{chen2019high} is considered---a co-assay of transcriptome and chromatin accessibility measurements performed on $n=1047$ cells. Experiments are run with and without the simulated annealing of $\epsilon$ for 20 random initializations of $f$ and $g$; the results are presented in \cref{fig:sim_annealing}. The annealing strongly stabilizes the training process. This is used as a default choice across all real data experiments.

\begin{figure}[t]
 \centering
 \includegraphics[width=0.75\textwidth]{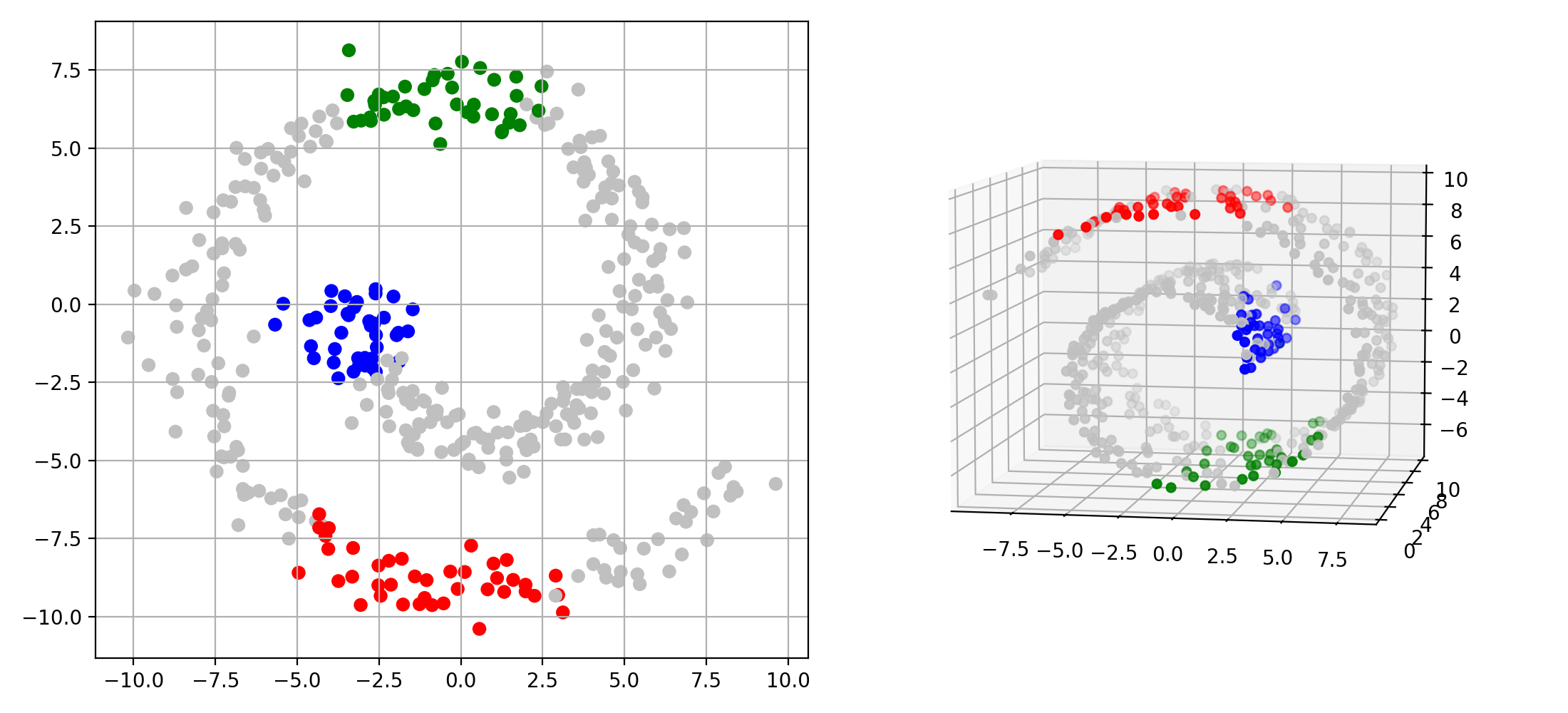}
 \caption{\textbf{Qualitative evaluation of the proposed GW solver in inductive setting.} The plot depicts the assignment produced by the distance-based GW solver (\cref{eq:final_objective_ours}) on a new set of samples.}
 \label{fig:swiss_roll}
\end{figure}

\begin{figure}[htbp]
 \centering
 \includegraphics[width=0.3\textwidth]{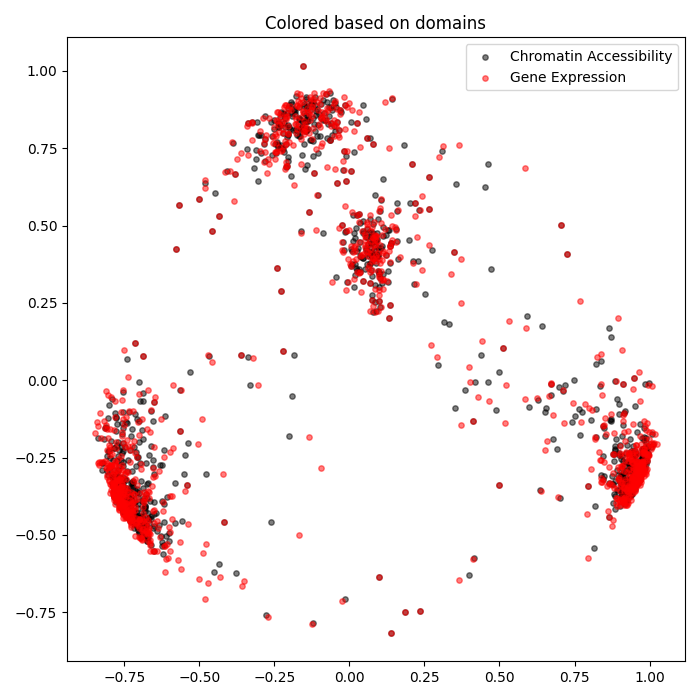}
 \includegraphics[width=0.3\textwidth]{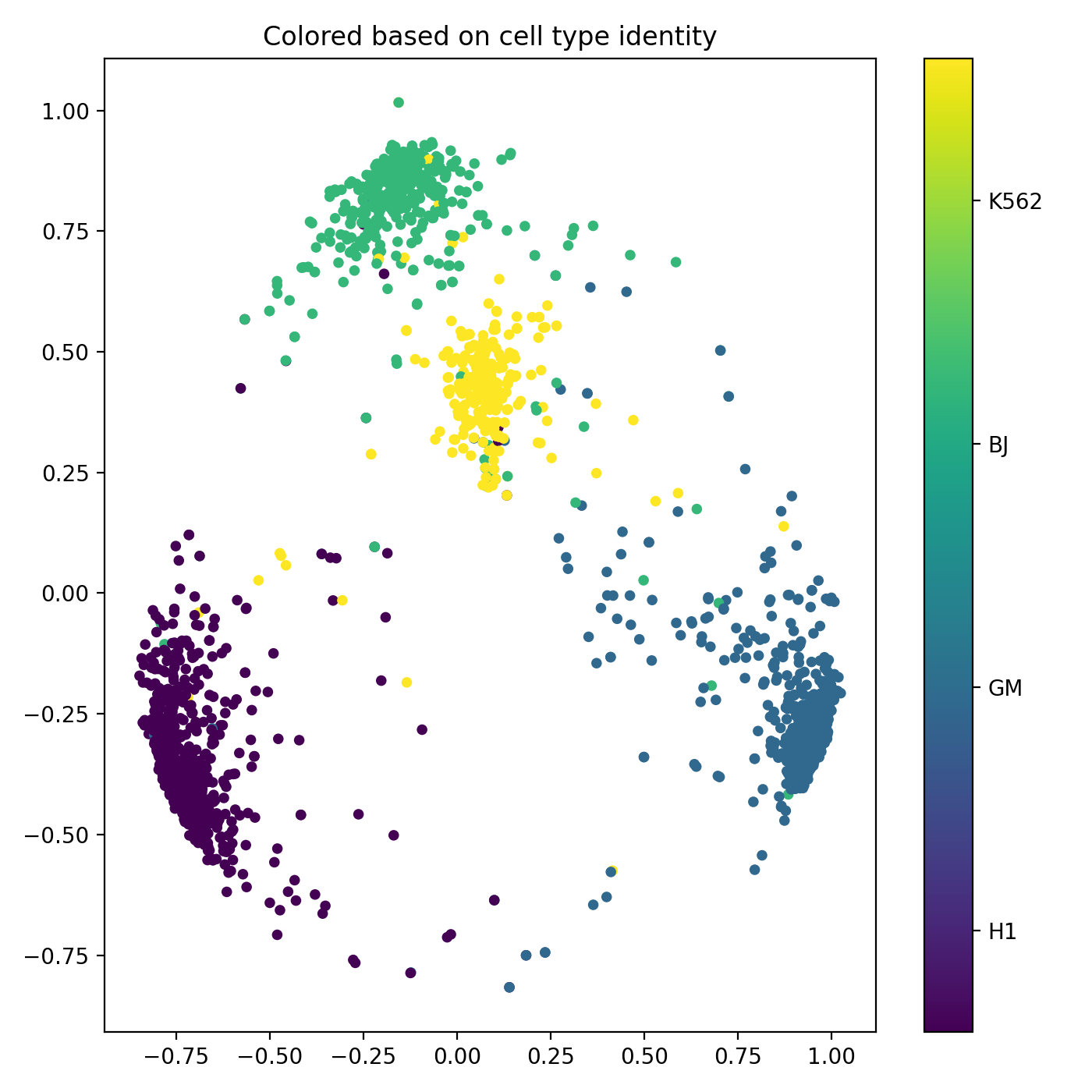}
 \includegraphics[width=0.3\textwidth]{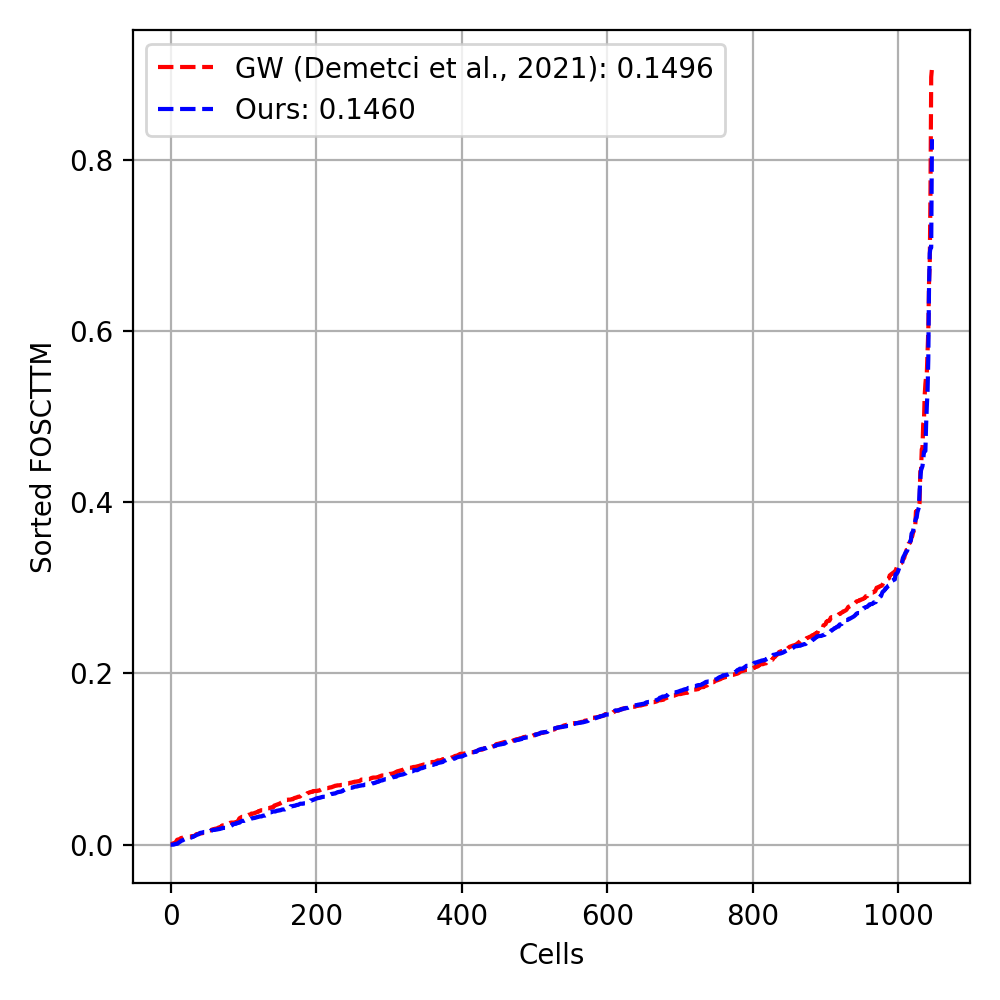}
 \caption{\textbf{Qualitative and quantitative results on the scSNARE-seq dataset.} Left and middle: Aligned samples from ATAC and RNA, colored by the domains (ATAC: black, RNA: red) and cell types, respectively. Right: the sorted FOSCTTM plot, a quantitative metric measuring the quality of the assignment (GW in the legend is the entropic GW baseline of \citet{demetci2022scot}).}
 \label{fig:scot}
\end{figure}

\begin{figure}[t]
 \centering
 \includegraphics[width=0.8\textwidth]{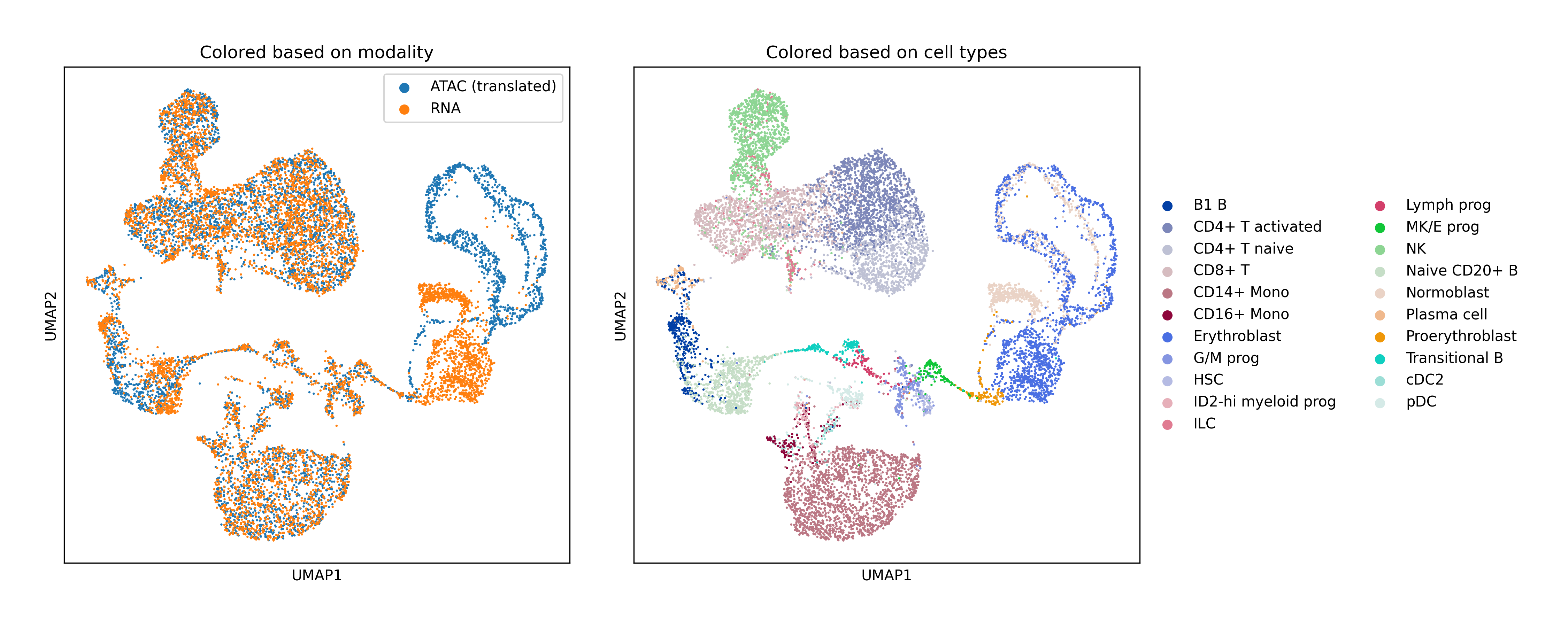}\\
 \includegraphics[width=0.4\textwidth]{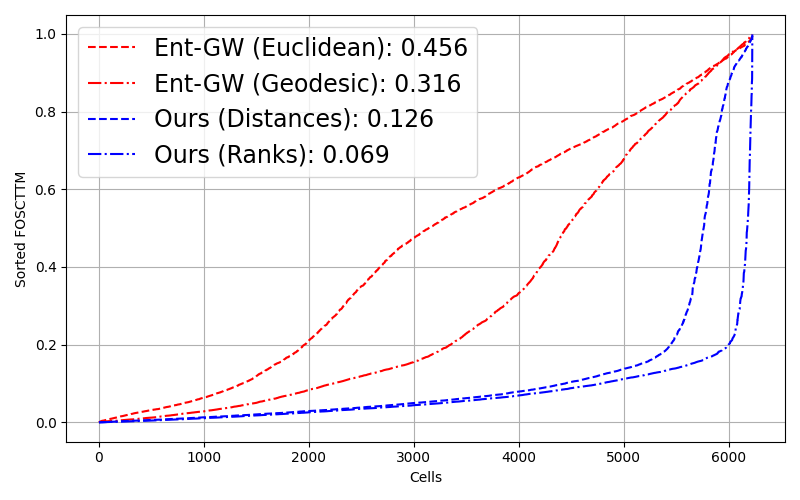}
 \includegraphics[width=0.4\textwidth]{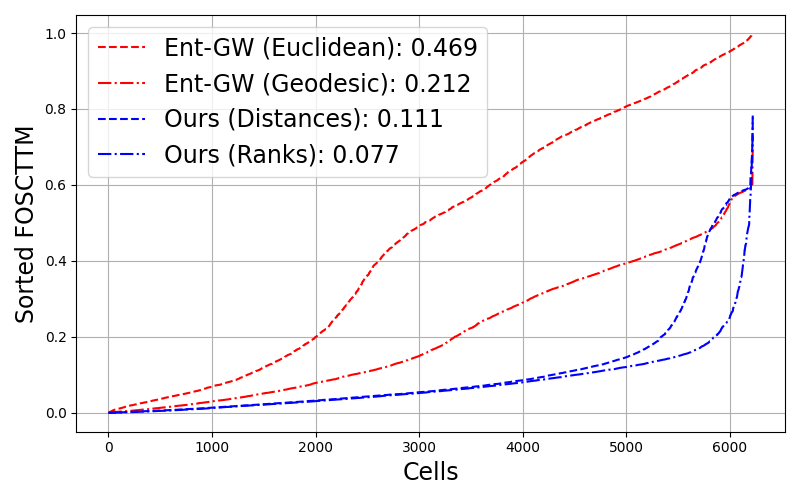}
 \caption{\textbf{Qualitative and quantitative results on the human bone marrow single-cell dataset.} Top plots depict the UMAP of the translated cells colored by domain (left) and by the cell type (right). Bottom plots report the FOSCTTM metrics for $\mathcal{Y}$ projected onto $\mathcal{X}$ (left) and $\mathcal{X}$ projected onto $\mathcal{Y}$ (right).}
 \label{fig:moscot}
\end{figure}

\Cref{fig:swiss_roll} shows, for reference, the assignment produced by the distance-based GW solver in an inductive setting. In single-cell multiomic alignment, the ranking-based variant outperforms this distance-based counterpart. Consequently, the results presented in \cref{fig:moscot,fig:scot} use ranking-based loss and outperform both the entropic GW solver and, on bone marrow (the only dataset where it is reported, \cref{fig:moscot}), the distance-based variant of the solver.

\textbf{Single-cell multiomic alignment.} Two real-world datasets are considered: (i) \textbf{scSNARE-seq} data which contains gene expression (RNA) and chromatin accessibility (ATAC) profiles from 1047 individual cells from four cell lines: H1, BJ, K562, and GM12878, with known groundtruth via a co-assaying technique~\citep{chen2019high}. The processed data of RNA and ATAC features is obtained from \citet{demetci2022scot}, whose method uses entropic GW to align these two modalities and is the baseline.
(ii) \textbf{human bone marrow} single-cell dataset that contains paired measurements of single-cell RNA-seq and ATAC-seq measurements released by \citet{moscotdata}. The processed data is obtained from \texttt{moscot}~\citep{klein2023moscot}. In the RNA space, a PCA embedding of 50 dimensions is used; in the ATAC space, an LSI (latent semantic indexing) embedding is used, followed by $L_2$ normalization.

In the scSNARE-seq experiment, the entropic GW solver with the same hyperparameters used by \citet{demetci2022scot} is employed as the baseline. This was shown by \citet{demetci2022scot} to outperform other baselines for unpaired alignment on this data.
In the bone marrow single-cell experiment, comparison is made to the entropic GW solver with Euclidean metric and geodesic distance metric. To establish a fair baseline, following the methodology of \citet{demetci2022scot}, a grid search is performed on the $\epsilon$ used in Sinkhorn iterations of the solver, and $k$ corresponding to the $k$-NN graph constructed for geodesic computation (for the latter setting), picking the hyperparameters with the least GW loss. In both bone marrow data and scSNARE-seq data, the ranking-based solver produces the best FOSCTTM score. In bone marrow data especially, the solver produces a large margin over the entropic GW solvers (FOSCTTM $0.069$ against $0.316$ for the best entropic GW baseline in the left panel of \cref{fig:moscot}, $0.077$ against $0.212$ in the right panel). In scSNARE-seq, the margin of improvement is lower, which could be attributed to limited diversity in cell-lines and small sample-size in scSNARE-seq compared to the bone-marrow data (\cref{fig:scot}).

\chapter{Domain Applications of Correspondence-Parallel Operators}
\label{ch:domain-applications}
\label{ch:08}
Correspondence-parallel operators (\gls{OperatorParallel}, \cref{sec:operator-families}) extend beyond standard vision and language benchmarks to three scientific domains: computational neuroscience (\cref{sec:cosyne}), uncertainty quantification (\cref{sec:uncertainty}), and reinforcement learning (RL, \cref{sec:r3l}). Each domain presents distinct challenges that demand domain-specific innovations within the correspondence framework.

We instantiate $\OperatorParallel_{\gls{Corr}}^{\mathcal{X}}$ using \emph{relative representations} (\gls{relrep}) across all three domains: anchoring to shared reference points provides a simple, interpretable transformation that factors out coordinate variations while preserving semantic structure.

However, each domain imposes unique constraints that require method-specific adaptations. In neuroscience, noisy fMRI signals from individually variable neural architectures require \emph{task-informed encoding}---subject-specific supervised encoders trained before applying \gls{relrep} anchoring. In uncertainty quantification, we introduce an \emph{alignment metric} inspired by Fisher's Linear Discriminant that quantifies how well \gls{relrep} removes reparameterization noise from ensemble variance. In reinforcement learning, continuous encoder updates during online training create coordinate drift in the anchor embeddings themselves; \emph{exponential moving average (EMA) stabilization} adapts \gls{relrep} to moving targets by slowing anchor drift while enabling gradual adaptation.

The commonality lies in the correspondence-parallel framework instantiated via \gls{relrep}; the diversity lies in domain constraints that drive innovation beyond the base method.

\section{Neuroscience: Multi-subject fMRI Alignment}
\label{sec:cosyne}

\sectionpublication{maiorca2024multi}{\textbf{Cosyne 2024}}{A task-informed bottleneck before projecting to semantic coordinates allows reconciling fMRI scans across subjects experiencing the same stimuli.}

Aligning brain activity across subjects confronts individual variations in neural architecture, vascular properties, and anatomical geometry. The decoding process is generally specific to each subject and faces limitations in generalizability across different individuals, primarily due to the significant variability observed among subjects.

We apply correspondence-parallel operators to map neural representations of distinct subjects into a common representational space agnostic of subject-specific attributes for downstream decoding. The framework builds on the assumption of a \emph{second-order isomorphism} between the neural activity of different individuals subject to the same stimuli, similarly to established neural analysis techniques like Representational Similarity Analysis (\gls{rsa}).

We instantiate $\OperatorParallel_{\Corr}^{\mathcal{X}_s}$ ($\Corr$ pairs the recordings of the same stimulus across subjects) using \gls{relrep} with shared anchor stimuli $\anchors = \{a_1, \dots, a_\nanch\}$ (300 images) and cosine similarity. The anchor-based transformation handles high-dimensional, noisy fMRI data by encoding subjects in a common coordinate frame.

Our approach relies on \emph{task-informed encoding}: training subject-specific supervised encoders (\glspl{mlp}) on classification tasks \emph{before} applying \gls{relrep} anchoring. Unlike task-agnostic methods like \gls{pca}, this dual objective---mapping fMRI to latent space while classifying visual stimuli---aims to preserve task-relevant semantic structure. The supervised classification objective is meant to force encoders to learn discriminative features, so that the dimensionality reduction follows the task, creating latent spaces where \gls{relrep} can capture meaningful cross-subject correspondences despite individual neural architectures.

\paragraph{Method.}

For each subject $s$, we train an \gls{mlp} encoder $\Enc_s: \Obs_s \to \mathcal{X}_s$ that simultaneously (1) maps fMRI recordings to a latent space $\mathcal{X}_s$ and (2) performs classification of visual stimuli according to image labels. After training, we project latent representations using relative coordinates. Given a latent encoding $\vx = \Enc_s(\obs)$ where $\obs \in \Obs_s$ is an fMRI recording, and $\anchors_s = \{a_{s,1}, \dots, a_{s,\nanch}\} \subset \Obs_s$, where $a_{s,j}$ is subject $s$'s recording of the anchor image $a_j$, the relative representation is:
\begin{equation}
 \relproj_{\anchors_s}(\vx) = [\cos(\vx, \Enc_s(a_{s,1})), \dots, \cos(\vx, \Enc_s(a_{s,\nanch}))]
\end{equation}
where the $\nanch = 300$ anchor images are randomly sampled from the dataset.

The critical difference from \gls{rsa} is that we first extract task-relevant features via supervised encoders; \gls{rsa} computes second-order similarities directly on raw fMRI signals. The encoders use nonlinear artificial neural networks to discover discriminative patterns in sparse, high-dimensional fMRI data.

\paragraph{Experiments.}
We evaluate on the Natural Scenes Dataset from the Algonauts challenge \citep{gifford2023algonauts}, chosen for: (i) suitability for deep learning pipelines; (ii) modality variation (fMRI and images); (iii) subject variation (8 participants total).

For each subject and each hemisphere, we train an \gls{mlp} that takes fMRI recordings as input and performs classification according to the corresponding image labels. After training, we project each subject's latent space using 300 anchor images randomly sampled from the dataset. Evaluation uses paired data: image samples with corresponding fMRI recordings from all subjects.

We measure cross-subject alignment via \glslink{mrr}{Mean Reciprocal Rank} (MRR):
\[
 \operatorname{MRR}(\mX, \mY) = \frac{1}{n} \sum_{i=1}^{n} \frac{1}{\operatorname{rank}(\vx_i, \mY)}
\]
where $\mX$, $\mY$ stack the representations of two subjects for the same $n$ stimuli in the condition under test (absolute, \gls{pca}-reduced to 300 components, or relative, $\vr_i = \relproj_{\anchors_s}(\vx_i)$), and $\operatorname{rank}(\vx_i, \mY)$ is the rank of $\vy_i$ among the rows of $\mY$ sorted by similarity to $\vx_i$. This quantifies how well corresponding stimuli align across subject-specific representational spaces. We compare three conditions: (i) \textbf{Absolute}---original latent spaces without projection; (ii) \textbf{\gls{pca}}---dimensionality reduction to 300 components applied independently per subject; (iii) \textbf{Relative}---correspondence-parallel projection using shared anchors. Results aggregate all subject-hemisphere pairs across three random seeds for anchor sampling.

\begin{figure}[t]
 \centering
 \includegraphics[width=0.36\linewidth]{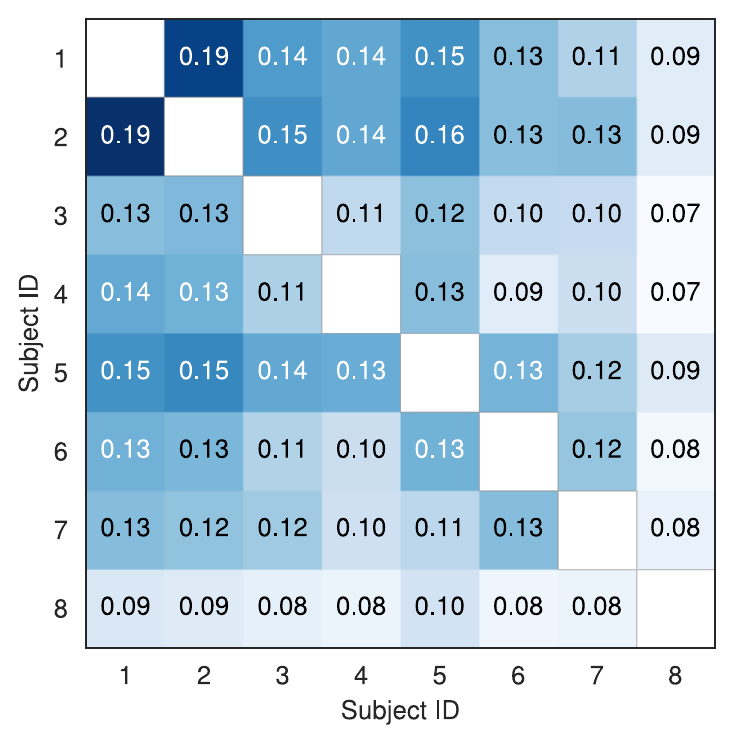}
 \hfill
 \includegraphics[width=0.51\linewidth]{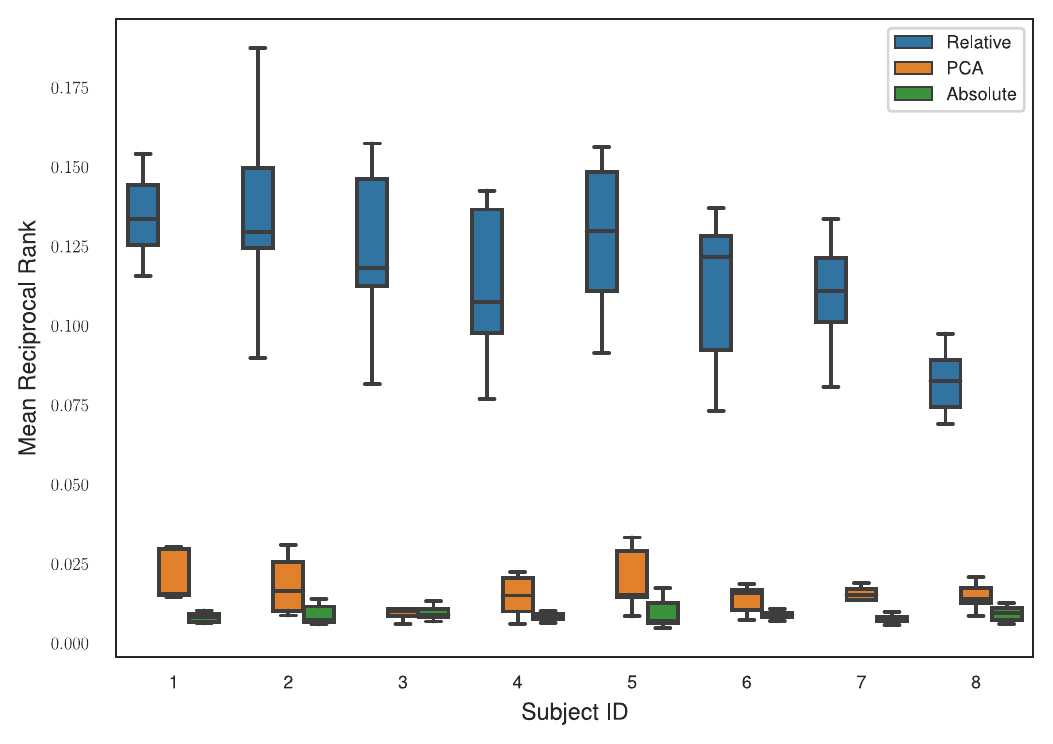}
 \caption{\textbf{Cross-subject fMRI alignment via task-informed encoding.} \textit{(Left)} Confusion matrix of \glslink{mrr}{MRR} similarity between 8 subjects' relative representation spaces; the diagonal is left empty. \textit{(Right)} Comparison with baselines: Relative coordinates (blue) achieve substantially higher similarity than \gls{pca} dimensionality reduction (orange) or absolute coordinates (green). Boxes aggregate the scores of all subject-hemisphere pairs over three anchor sampling seeds, grouped by subject.}
 \label{fig:cosyne-mrr}
\end{figure}

\paragraph{Results.}
Relative representations achieve substantially higher cross-subject similarity compared to both baselines (\cref{fig:cosyne-mrr}). The confusion matrix (left) shows \glslink{mrr}{MRR} similarity across all 8 subjects: the diagonal is left empty, while off-diagonal entries (0.07--0.19) reveal cross-subject alignment enabled by relative coordinates. The box plot (right) confirms that relative spaces (blue) reach markedly higher \glslink{mrr}{MRR} scores than \gls{pca} (orange) or absolute (green) baselines, demonstrating that correspondence-parallel projection successfully captures the common structure in brain activation patterns across different subjects.

The success of relative projection on top of task-informed encoders---using supervised classification before relative projection---suggests that correspondence-parallel operators may extend beyond artificial neural networks to natural neural systems, with semantic correspondence providing a common principle for cross-system alignment.\footnote{This work was presented at Cosyne 2024. While no public code repository was released, the method builds directly on the relative representations framework.}

\section{Uncertainty Quantification in Latent Spaces}
\label{sec:uncertainty}

\sectionpublication[https://github.com/fmager/it-s-all-relative]{mager2024s}{\textbf{NeurIPS UniReps Workshop 2024}}{\gls{relrep} can factor out part of the coordinate noise from genuine uncertainty---an alignment metric (between- versus within-observation variance) measures how much.}

The neuroscience application confronted natural variability across subjects---different brains processing the same stimuli. Here, we shift to artificial variability within a single model class: ensembles of networks trained on identical data that nonetheless produce incompatible latent coordinates. The challenge moves from cross-subject alignment to quantifying genuine uncertainty when parameterization noise artificially inflates variance estimates.

Ensemble uncertainty quantification in latent spaces confronts the reparameterization problem: different parameterizations of functionally equivalent models create artificial coordinate variations that inflate uncertainty estimates. When measuring latent uncertainty across ensemble members, observed variance $\sigma_O^2$ conflates two sources: genuine model uncertainty $\sigma_M^2$ (epistemic variability in what function was learned) and reparameterization noise $\sigma_R^2$ (arbitrary coordinate variations from how that function is parameterized).

We instantiate $\OperatorParallel_{\Corr}^{\mathcal{X}}$ using \gls{relrep} with $\nanch = d$ anchors (latent dimension: $d=512$ for VGG-16, $d=256$ for PreResNet-110) and, as the main choice, cosine similarity (Euclidean distance and basis norm are also evaluated below). Anchoring all ensemble members to fixed reference points transforms model-specific coordinates into a shared frame that, with the cosine similarity, is invariant to rotations and rescalings.

Our contribution is an \emph{alignment metric} $\rho$ inspired by Fisher's Linear Discriminant that quantifies how effectively \gls{relrep} removes reparameterization noise. This metric measures the share of between-observation variance (meaningful semantic distances) in the total variance, the remainder being within-observation variance (reparameterization noise from ensemble sampling): $\rho = \sigma_B^2/(\sigma_B^2 + \sigma_W^2)$. When $\rho \to 1$, \gls{relrep} has successfully factored out coordinate noise; observations cluster tightly within-observation while maintaining semantic separation between-observation.

We sample ensemble models along mode-connecting curves in weight space---paths where loss remains constant. This experimental design reveals where uncertainty changes most: in most panels, at curve midpoints.

\paragraph{Method.}
Neural networks trained from different initializations on the same data learn functionally similar solutions in incompatible coordinate systems. \emph{Mode connectivity theory} reveals that these independently trained networks can be connected by simple curves in weight space along which loss remains approximately constant \citep{garipov2018loss}. This geometric property of loss landscapes enables principled ensemble sampling: we can sample models along loss-constant curves to study how reparameterization affects uncertainty. The models do not need to be trained independently.

Given two converged models with parameters $\theta_1, \theta_2$, we exploit mode connectivity by fitting a Bézier curve $\phi_{\vc}(t)$ parameterized by $t \in [0,1]$, with endpoints $\theta_1, \theta_2$ and control points $\vc$ in weight space, by minimizing:
\[
\mathcal{L}_{\mathrm{curve}}(\vc) = \E_{t \sim U(0,1)}[\mathcal{L}(\phi_{\vc}(t))]
\]
where $\mathcal{L}$ is the original training loss. Once fitted, sampling $\phi_{\vc}(t)$ at discrete steps yields an ensemble of models living on a loss-constant manifold, allowing us to isolate reparameterization effects from genuine functional differences.

The reparameterization problem manifests as artificial variance. Under the assumption that reparameterization effects and model uncertainty are independent sources of variance, observed latent uncertainty decomposes additively as:
\begin{equation}
 \sigma_O^2(\obs_i) = \sigma_R^2(\obs_i) + \sigma_M^2(\obs_i)
\end{equation}
where each term is a variance across ensemble members of the representations of observation $\obs_i$, $\sigma_R^2$ is reparameterization noise (coordinate transformations between equivalent representations) and $\sigma_M^2$ is genuine model uncertainty. With cosine similarity, relative representations eliminate rotation and scaling components of $\sigma_R^2$ by anchoring coordinates to semantic content, allowing us to isolate $\sigma_M^2$.

We define an alignment metric that quantifies this noise removal. Given $J$ ensemble realizations of latent vector $\vx_i$ (one per sampled model), the metric measures compactness (within-observation variance) relative to separation (between-observation variance):
\begin{align}
\sigma_W^2 &= \frac{1}{n}\sum_{i=1}^n\sum_{j=1}^{J} \lVert \vx_i^{(j)} - \bar{\vx}_{i} \rVert^2 \\
\sigma_B^2 &= \frac{1}{n} \sum_{i=1}^n \lVert \bar{\vx}_{i} - \bar{\vx} \rVert^2\\
\rho &= \frac{\sigma_B^2}{\sigma_B^2 + \sigma_W^2}
\end{align}
where $\bar{\vx}_i = \frac{1}{J}\sum_j \vx_i^{(j)}$ is the ensemble mean for observation $i$, $\bar{\vx} = \frac{1}{n}\sum_i \bar{\vx}_i$ the global mean, and $n$ is the number of test observations. This formulation applies Fisher's Linear Discriminant (FLD) objective at the observation level: FLD measures the ratio of between-class to within-class variance to assess separability; here, we measure the share of between-observation variance in the total. When $\rho \to 1$, observations cluster tightly within-observation (low $\sigma_W^2$) while separating between-observation (high $\sigma_B^2$), indicating reparameterization noise has been factored out.

For transformation to relative space, we use anchor sets matching the latent dimension ($\nanch = d$), ensuring the relative transformation preserves dimensionality while changing the coordinate frame. We evaluate three similarity functions: (i) cosine similarity to each anchor; (ii) Euclidean distance to each anchor; (iii) basis norm, the projection coefficient onto each anchor, $\langle\vx,\va_j\rangle/\lVert\va_j\rVert^2$. For cosine similarity and basis norm transformations, spaces are centered using training set mean; for Euclidean distance, spaces are additionally scaled to unit variance.

\paragraph{Experiments.}
We evaluate on \gls{cifarh} using VGG-16 and PreResNet-110 architectures. For each architecture, we train three models to convergence from different random seeds ($\theta_1, \theta_2, \theta_3$). For each of the three pairwise combinations ($\theta_1$--$\theta_2$, $\theta_2$--$\theta_3$, $\theta_1$--$\theta_3$), we fit a Bézier curve connecting the modes.
Once fitted, we sample $J=21$ models uniformly along each curve ($t \in \{0, \frac{1}{20}, \frac{2}{20}, \dots, 1\}$).

For each sampled model, we extract latent representations (penultimate layer before classification) on a held-out test set. We then compare alignment in two coordinate systems: (i) \textbf{Absolute}---original latent coordinates; (ii) \textbf{Relative}---transformed via correspondence-parallel projection using randomly chosen anchors. We measure cumulative alignment as ensemble size grows (models added in order of increasing $t$).

As baseline, we compute alignment across eleven independently trained networks (no curve constraint), establishing the expected alignment level when models are sampled without enforcing mode connectivity.

\begin{figure}[t]
 \centering
 \includegraphics[width=\linewidth, trim={0 0 0 2.5cm}, clip]{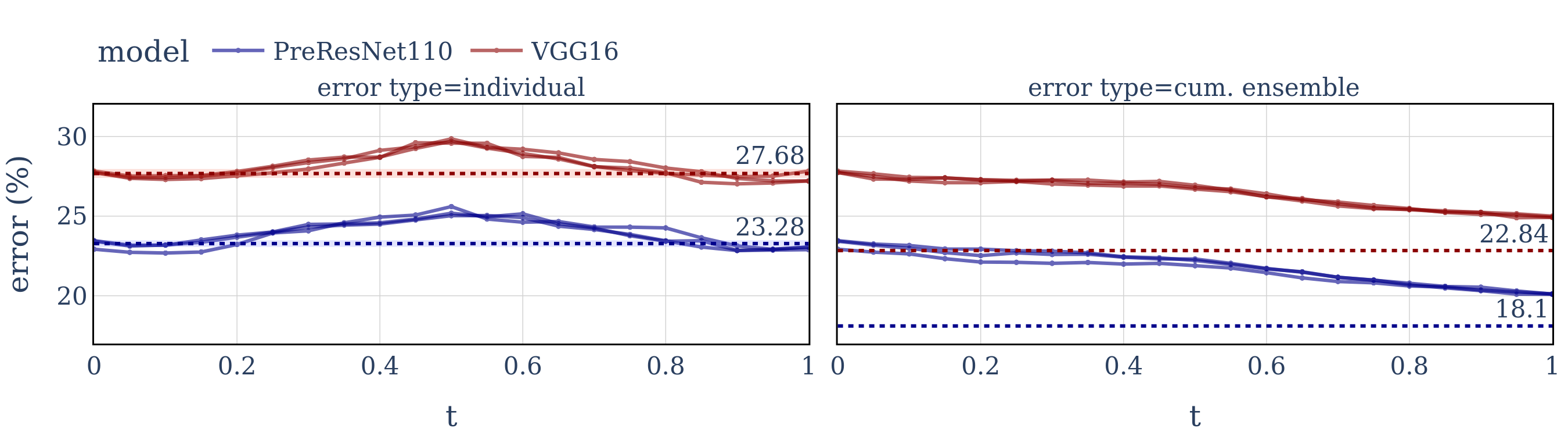}

 \vspace{0.3cm}

 \includegraphics[width=\linewidth, trim={0 0.4cm 0 0.2cm}, clip]{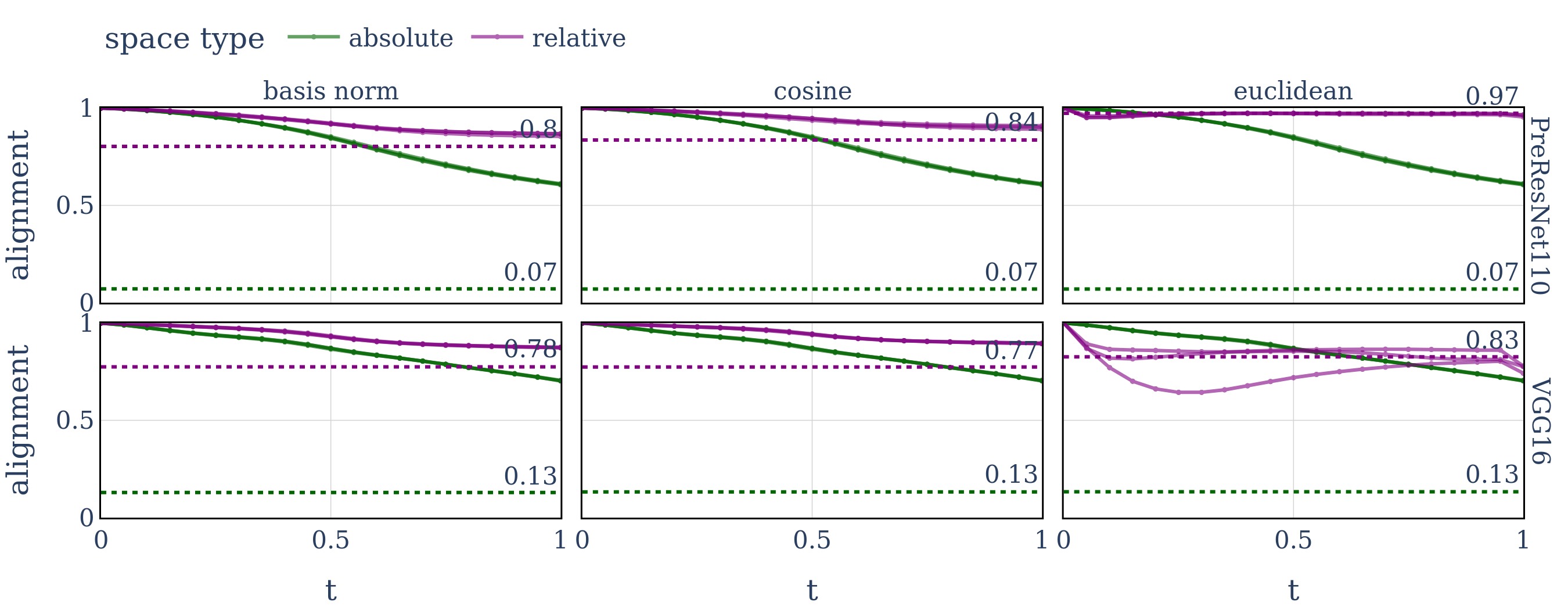}
 \caption{\textbf{Alignment quantifies reparameterization removal in ensemble latent spaces.} \textit{(Top)} Curve fitting results for VGG-16 (red) and PreResNet-110 (blue), pairwise connecting three modes. Individual model error rate (left) and cumulative ensemble error rate (right). \textit{(Bottom)} Alignment of embedding spaces along curves for absolute (green) and relative (purple) coordinates. Dashed lines show error rates (top) and alignment (bottom) across eleven independently trained networks (baseline). In most panels, relative spaces achieve higher alignment, indicating reduced reparameterization noise, and the steepest slope occurs at curve midpoint ($t=0.5$).}
 \label{fig:uncertainty-alignment}
\end{figure}

\paragraph{Results.}
Relative representations achieve substantially higher alignment than absolute coordinates in all panels but one, confirming that reparameterization noise inflates observed uncertainty (\cref{fig:uncertainty-alignment}). The top panel shows ensemble error along curves: individual model error (left) rises near curve midpoints before recovering, while cumulative ensemble error (right) decreases as ensemble size increases. Pairwise curves show consistent behavior across all three combinations.

The bottom panel reveals the core finding: alignment measured in relative space (purple) substantially exceeds absolute space (green) across all curve positions, except for VGG-16 with Euclidean distance, where relative alignment falls below absolute alignment for $t$ below about 0.5 (about 0.75 for one of the three curves). For VGG-16, cosine similarity achieves the strongest alignment; for PreResNet-110, Euclidean distance performs best. Alignment in absolute space decreases nearly linearly along curves, whereas relative space alignment stabilizes at levels closer to the independently-trained baseline (dashed lines) than absolute space. This indicates relative representations successfully factor out curve-induced reparameterization.

The alignment profile along curves reveals where alignment changes most: in most panels, cumulative alignment decreases most rapidly (steepest negative slope) near $t=0.5$ (curve midpoint), while endpoints ($t \to 0$ or $t \to 1$) show minimal variation. This suggests that ensemble sampling for uncertainty quantification may benefit from prioritizing midpoint regions, where model disagreement---and thus epistemic uncertainty---is highest.

\begin{figure}[t]
 \centering
 \includegraphics[width=0.4\linewidth]{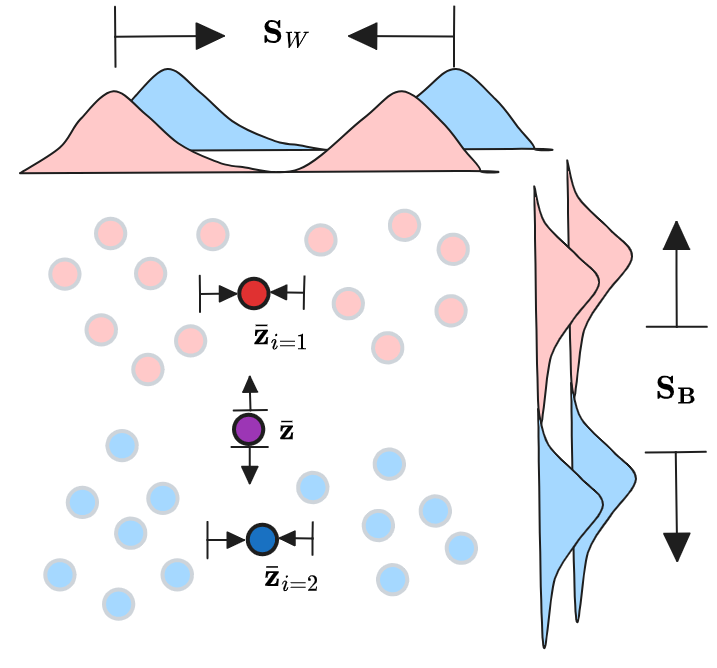}
 \hfill
 \includegraphics[width=0.55\linewidth]{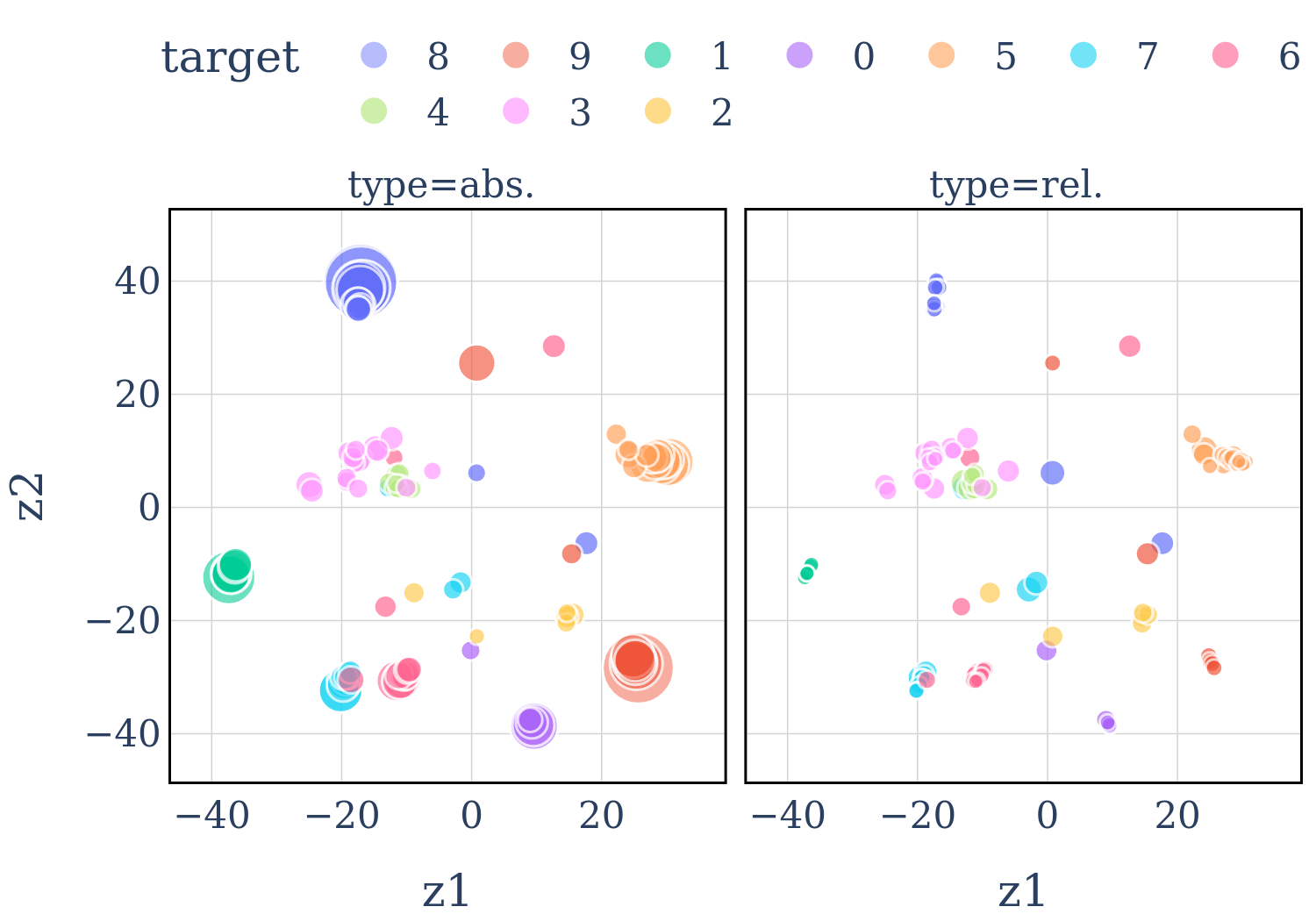}
 \caption{\textbf{Conceptual visualization and empirical measurement of alignment metric.} \textit{(Left)} Schematic: given observations $\obs_{i=1}$ and $\obs_{i=2}$ with $J$ ensemble realizations each (the drawing writes their ensemble means $\bar{\vx}_i$ as $\bar{\vz}_i$ and the global mean as $\bar{\vz}$, and labels the two variances $\sigma_W^2$, $\sigma_B^2$ as $\mS_W$, $\mS_B$), high alignment requires compact within-observation clustering (small $\sigma_W^2$) and large between-observation separation (large $\sigma_B^2$). \textit{(Right)} t-SNE projection of VGG-16 embeddings for 10 \gls{cifarh} classes (same t-SNE layout; relative space via cosine similarity). Dot size represents uncertainty. Absolute space (left projection) shows larger uncertainty; relative space (right projection) shows smaller uncertainty.}
 \label{fig:uncertainty-concept}
\end{figure}

\Cref{fig:uncertainty-concept} visualizes the alignment concept. The left schematic illustrates the Fisher-inspired objective: given $J$ ensemble samples per observation, alignment maximizes between-observation variance relative to within-observation variance. The right panels show t-SNE projections of latent representations (10 classes, VGG-16), where dot size encodes uncertainty magnitude. In absolute coordinates, many observations, often inside class clusters, show large uncertainty; in relative coordinates, the same observations have much smaller uncertainty.

\subsection{Discussion}
\label{sec:uncertainty:discussion}

The question behind these experiments is how much of the uncertainty measured in the latent space of an ensemble reflects the input, and how much of it is an artifact of the coordinates that each member happens to use. The comparison between absolute and relative coordinates answers it in part. Along almost every curve, the embeddings of the relative representations remain more aligned than those of the absolute representations (the exception is VGG-16 with the Euclidean distance for small $t$). A good part of the disagreement between ensemble members in absolute coordinates is therefore the reparameterization that separates two functionally equivalent solutions; it is not disagreement about the input at all. Once the ensemble is projected into a space of relative proximity, much of that disagreement is removed and the measured uncertainty drops accordingly. How much is removed depends on the projection, and the best projection depends in turn on the model: cosine similarity performs best for VGG-16 and the Euclidean distance for PreResNet-110. In both cases, however, the relative alignment ends closer to its own baseline, measured across eleven independently trained networks, than the absolute alignment does. In this respect, once the coordinates are factored out, an ensemble sampled along a curve comes to resemble a conventional ensemble of independent models.

The shape of the alignment curves carries a practical lesson as well. In most panels, alignment decreases most steeply around the midpoint of the curve and varies little in the neighborhood of either endpoint, so the members that bring new information to the ensemble are those sampled away from the two solutions that define it. If the purpose of the ensemble is to quantify uncertainty, this suggests drawing its members far from the endpoints of the connecting curve.

What the experiments leave open is which projection to use in general. Since the best choice differed between the two architectures, no single projection can be recommended, and the choice is not innocuous: a projection can bias the very estimate it is meant to clean, as the unit-norm scaling of the cosine similarity does by inflating the uncertainty of points close to the zero vector. The anchors, moreover, were a random set, so whether a more deliberate choice, such as archetypes or centroids, would improve the projection remains to be investigated.

This work establishes correspondence-parallel operators as tools for principled uncertainty quantification. By anchoring representations to semantic content, the framework factors extrinsic parameterization noise from intrinsic epistemic uncertainty, enabling honest assessment of what models know versus how they encode that knowledge.

Within the framework of this thesis, this section uses the correspondence-parallel operator of \cref{ch:relrep} in a different role from the chapters around it. Here the relative projection does not transport information from one model to another. It is a measurement device: it separates the effect of the coordinates from the effect of the models, and in doing so makes the latent coordinates of the ensemble members comparable, so that their spread can be read as uncertainty. \cref{sec:r3l} returns to transfer, in a setting where the anchor embeddings themselves move during training.

\section{Reinforcement Learning: Zero-Shot Module Stitching}
\label{sec:r3l}

\sectionpublication[https://github.com/antoniopioricciardi/rl_relrepr_gymnasium]{ricciardi2023r3l}{\textbf{EWRL 2023}}{EMA-stabilized anchors let \gls{relrep} work in online reinforcement learning, so that separately trained encoders and controllers can be stitched with a large reduction in training time.}

Where neuroscience addressed static cross-subject variability and uncertainty addressed ensemble parameterization noise, reinforcement learning introduces temporal instability: continuous encoder updates during online training create coordinate drift. The correspondence framework must now stabilize a moving target---anchors whose embeddings shift at every gradient step---while enabling zero-shot composition of visual encoders and task-specific controllers across environment variations.

Visual Reinforcement Learning agents typically require retraining for each new environment variation, even when changes should not affect world semantics (e.g., background color changes) or involve tasks that should be achievable with already-learned exploration patterns (e.g., respecting a velocity limit). This necessitates new training for each variation, consuming significant time and computational resources.

We instantiate $\OperatorParallel_{\Corr}^{\mathcal{X}}$ using \gls{relrep} with anchor observations and cosine similarity, creating modular representations where encoders from one visual variation compose with controllers from one task variation without retraining. This reduces training from $O(n_v \times n_t)$ visual-task combinations to $O(n_v + n_t)$ components, for $n_v$ visual variations and $n_t$ tasks.

\begin{sloppypar}
We formally model an RL problem as a Markov decision process $\mathcal{M} = (\mathcal{S}, \mathcal{U}, \Obs, R, P, \gamma)$ where $\mathcal{S}$ defines the set of states, $\mathcal{U}$ the set of actions, $\obs \in \Obs$ the observation produced as input for the agents, $P : \mathcal{S} \times \mathcal{U} \to \Delta(\mathcal{S})$ the probability distribution $P(s' \mid s, u)$ of transitioning to state $s'$ upon executing action $u$ in state $s$, $R : \mathcal{S} \times \mathcal{U} \to \R$ the reward function, and $\gamma$ the discount factor. The policy $\pi : \Obs \to \mathcal{U}$ is trained to maximize discounted long-term returns $\E [\sum_{t=0}^{\infty} \gamma^t R(s_t, u_t)]$.
\end{sloppypar}

We formalize environment variations as $\mathcal{M}_v^{(i)}$, with visual variations indexed by $v, w$ and tasks by $i, i'$: $\mathcal{M}_v^{(i)}$ is the pairing of an observation space $\Obs_v$ (capturing visual variations like background color or camera angle) with task $i$ (affecting transition dynamics, action spaces, or reward functions through $\mathcal{S}^{(i)}, \mathcal{U}^{(i)}, R^{(i)}, P^{(i)}$). Observation spaces $\Obs_v$ differ when they exhibit significant changes (e.g., background color, camera perspective). Task variations affect agent behavior through these task-specific components. Since task $i$ information is not explicitly provided, agents must infer it from reward signals.

We train agents separately on different visual-task variations. During testing, we evaluate on unseen visual-task combinations $\mathcal{M}_v^{(i')}$, meaning no agent has been trained end-to-end on that specific $(v,i')$ pair. Standard practice trains policies end-to-end as neural networks that decompose into encoder $\Enc_v^{(i)} : \Obs_v \to \R^d$ (visual feature extraction) and controller $\Dec_v^{(i)} : \R^d \to \mathcal{U}^{(i)}$ (action selection): $\pi_v^{(i)}(\obs_v) = \Dec_v^{(i)}(\Enc_v^{(i)}(\obs_v))$. Different training contexts yield incompatible latent spaces: $\Enc_v^{(i)}(\obs_v) \neq \Enc_w^{(i')}(\obs_w)$ even for corresponding observations.

Our contribution is \emph{exponential moving average (EMA) stabilization} that adapts \gls{relrep} to online RL's moving target problem. Standard \gls{relrep} assumes fixed anchor embeddings, but continuous encoder updates create coordinate drift---the anchors themselves shift at every gradient step. EMA stabilizes \gls{relrep} by preserving most of the previous anchor embeddings ($\alpha=0.999$) at each step, damping coordinate drift while enabling gradual adaptation. This makes \gls{relrep}'s modular zero-shot composition tractable despite continuous training.

\paragraph{Method.}
We instantiate correspondence-parallel operators ($\Corr$ pairs observations of the same state under two variations, $\{(\obs_v, \obs_w)\}$) using \gls{relrep} (\cref{ch:relrep}), a method referred to as R3L (Relative Representations for Reinforcement Learning), encoding latent states via cosine similarity to anchors $\anchors_v = \{a_{v,1}, \dots, a_{v,\nanch}\}$:
\begin{equation}
 \vr_{v}^{(i)} = \relproj(\vx_{v}^{(i)}) = [\cos(\vx_{v}^{(i)}, \Enc_v^{(i)}(a_{v,j}))]_{j=1}^{\nanch}
\end{equation}
where $\vx_{v}^{(i)} = \Enc_v^{(i)}(\obs_v)$. This factors out absolute coordinate positions while preserving semantic structure: $\vr_{v}^{(i)} \approx \vr_{w}^{(i')}$ for corresponding observations across environment variations. Training controllers $\Dec_w^{(i')}: \R^{\nanch} \to \mathcal{U}^{(i')}$ directly on \gls{relrep} spaces (in place of the $\R^d$ controllers above) yields universal controllers composable across encoders: given policies $\pi_v^{(i)}, \pi_w^{(i')}$, zero-shot stitching combines $\Enc_v^{(i)}$ with $\Dec_w^{(i')}$ to create $\hat{\pi}_v^{(i')}(\obs_v) = \Dec_w^{(i')}(\vr_v^{(i)})$ acting on unseen combination $\mathcal{M}_v^{(i')}$.

\paragraph{EMA Stabilization.}
Online RL training creates coordinate drift: as encoder $\Enc$ updates via gradient descent, anchor embeddings $\Enc(\anchors)$ shift continuously, forcing controller $\Dec$ to track a moving target. Naive relative representations amplify this instability---the correspondence-parallel transformation itself drifts as anchors update.

We stabilize training via exponential moving average (EMA):
\begin{equation}
 \bar{\anchmat}_k = \alpha\,\bar{\anchmat}_{k-1} + (1-\alpha)\,\anchmat_k
\end{equation}
where $\anchmat_k \in \R^{\nanch\times d}$ stacks the anchor embeddings $\Enc_v^{(i)}(a_{v,j})$ computed by the encoder at training step $k$, and $\bar{\anchmat}_k$ is their running average, updated at every training step; during training, the relative representation is computed against the rows of $\bar{\anchmat}_k$. With $\alpha=0.999$, each update keeps weight $\alpha$ on the previous running average and $1-\alpha$ on the new embeddings, damping coordinate drift while enabling gradual adaptation to encoder updates. The time constant $-1/\ln\alpha \approx 1000$ steps characterizes the memory horizon: the weight of a change decays by a factor $e$ over about 1000 gradient updates.

\paragraph{Data Collection.}
Prior work on latent space alignment relies on supervision from parallel anchors---training data samples with partial correspondence across domains. However, online RL lacks such predefined training data. We overcome this through two methods, the first of which requires a \emph{critical assumption: deterministic environments}.

\begin{figure}[!t]
 \centering
 \includegraphics[width=0.4\textwidth]{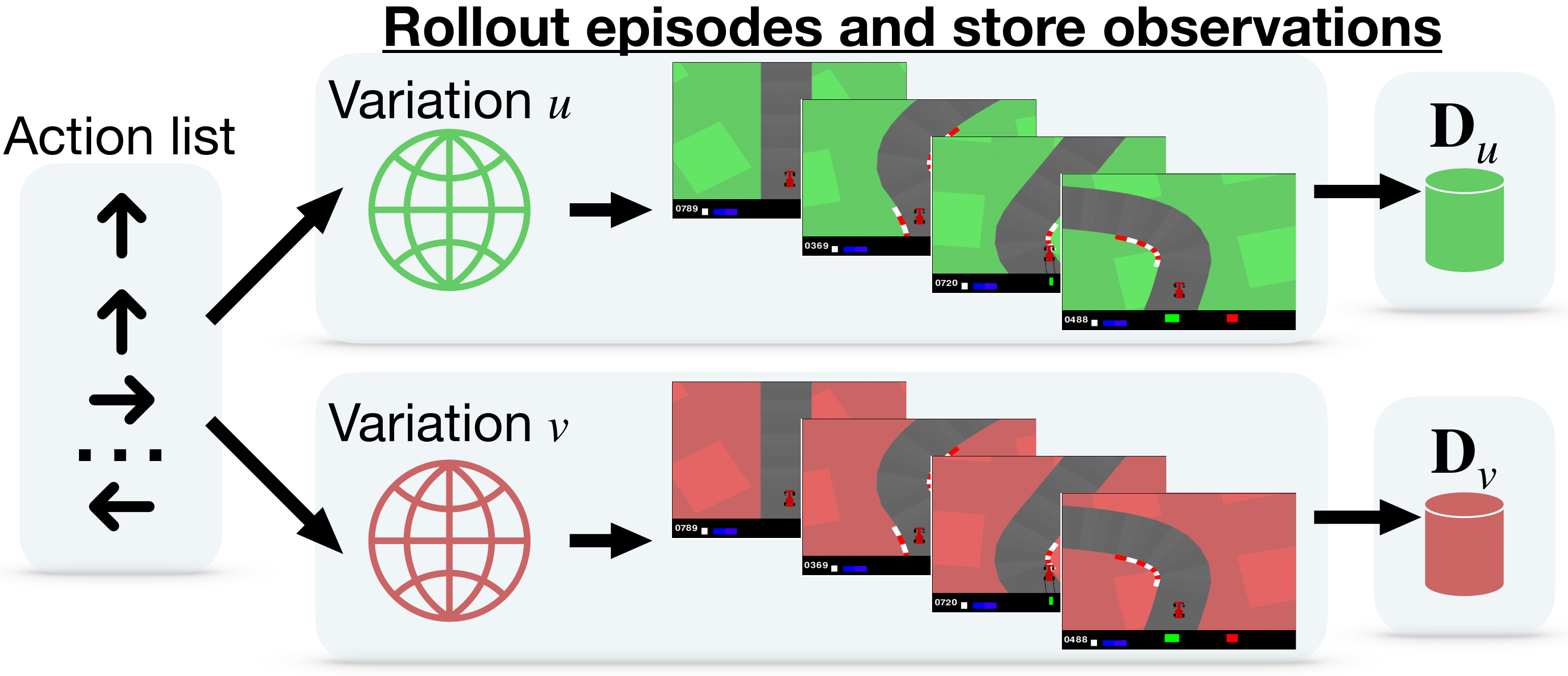}
 \caption{\textbf{Data collection for parallel anchors.} Correspondence across visual variations established via deterministic replay (same action sequence with fixed seed) or pixel-space transformations ($u$, $v$ and $\mD_u$, $\mD_v$ in the figure are the variations $v$, $w$ and their observation sets). Both methods assume deterministic environments or well-defined transformations to establish correspondence.}
 \label{fig:data-collection}
\end{figure}

We obtain parallel observations $\{(\obs_v, \obs_w)\}$ via (\cref{fig:data-collection}): (1) \textbf{deterministic replay}---executing identical action sequences with fixed random seeds across environment variations, ensuring that each observation $\obs_v$ has a corresponding $\obs_w$; or (2) \textbf{pixel-space transformations}---directly manipulating observations when visual variations are well-defined (e.g., programmatically changing grass color). The first method requires environments to be deterministic; the second requires that visual changes preserve semantic correspondence. Exploring other methods for approximating observation translation across environments remains future work.

We divide architectures at the flatten layer: encoders (convolutional layers) extract visual features; controllers (linear layers) select actions from latent states. This split enables independent embedding of single-frame anchors before stacking for controller input.

\section{Reinforcement Learning Experiments}\label{sec:domain:experiments}
We evaluate on CarRacing \citep{klimov2016carracing} with visual variations (grass color: green/\allowbreak red/\allowbreak blue; camera zoom) and task variations (standard/slow dynamics; scrambled actions; no-idle action space). Training uses PPO \citep{schulman2017proximal} via CleanRL \citep{huang2022cleanrl}.

\paragraph{Notation.}
\textbf{E. Abs/R3L}: end-to-end policies with absolute/relative representations. \textbf{S. Abs/R3L}: zero-shot stitched policies (encoder from one visual variation + controller from one task variation). \textbf{Multicolor}: models trained with background colors randomized among green, red, and yellow after each episode reset.

We report: (1) latent similarity (\cref{sec:exp-analysis}); (2) end-to-end performance (\cref{sec:exp-end-to-end-performance}); (3) zero-shot stitching---the main result (\cref{sec:zero-shot-stitching}).

\begin{figure}[t!]
 \centering
 \begin{subfigure}[b]{0.6\linewidth}
  \centering
  \includegraphics[width=\linewidth]{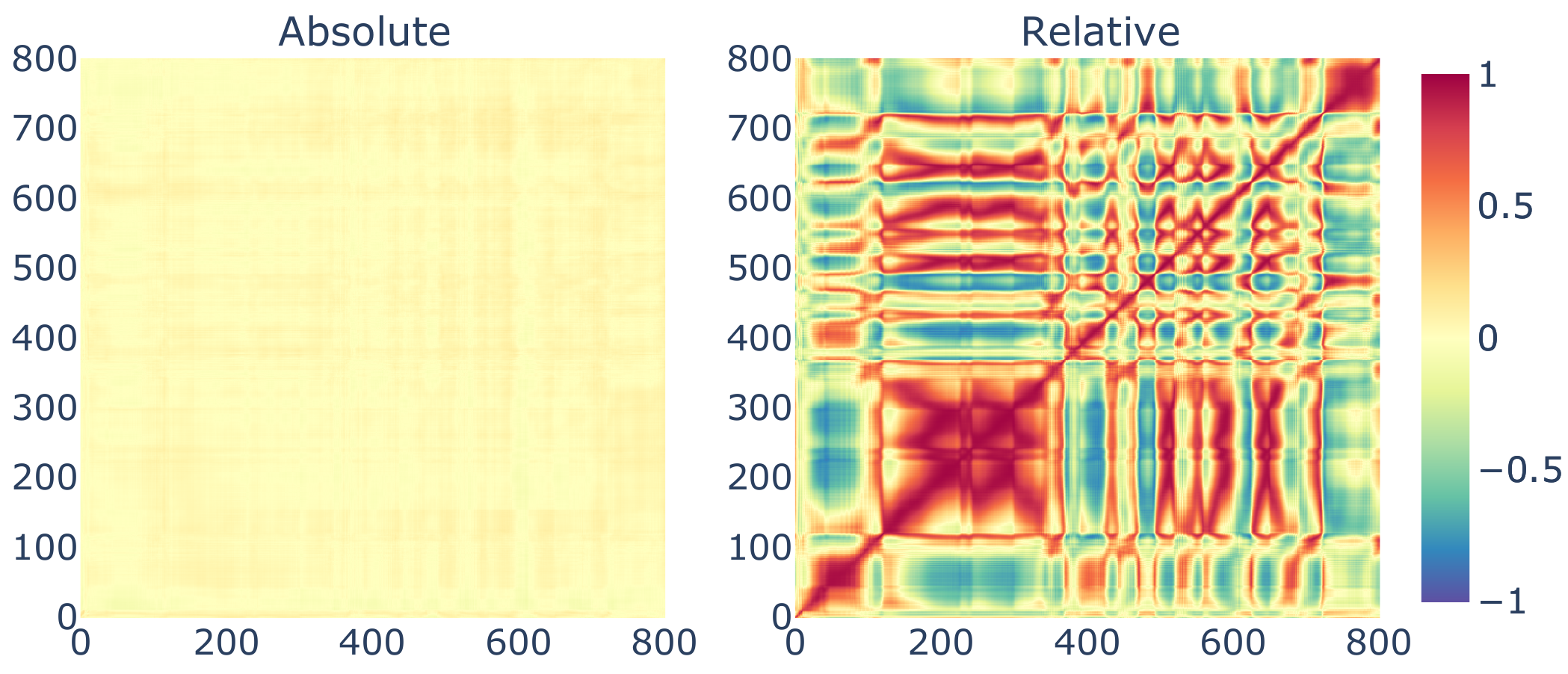}
  \caption{}
  \label{fig:matrix-sim}
 \end{subfigure}%
 \begin{subfigure}[b]{0.4\linewidth}
  \centering
  \includegraphics[width=\linewidth]{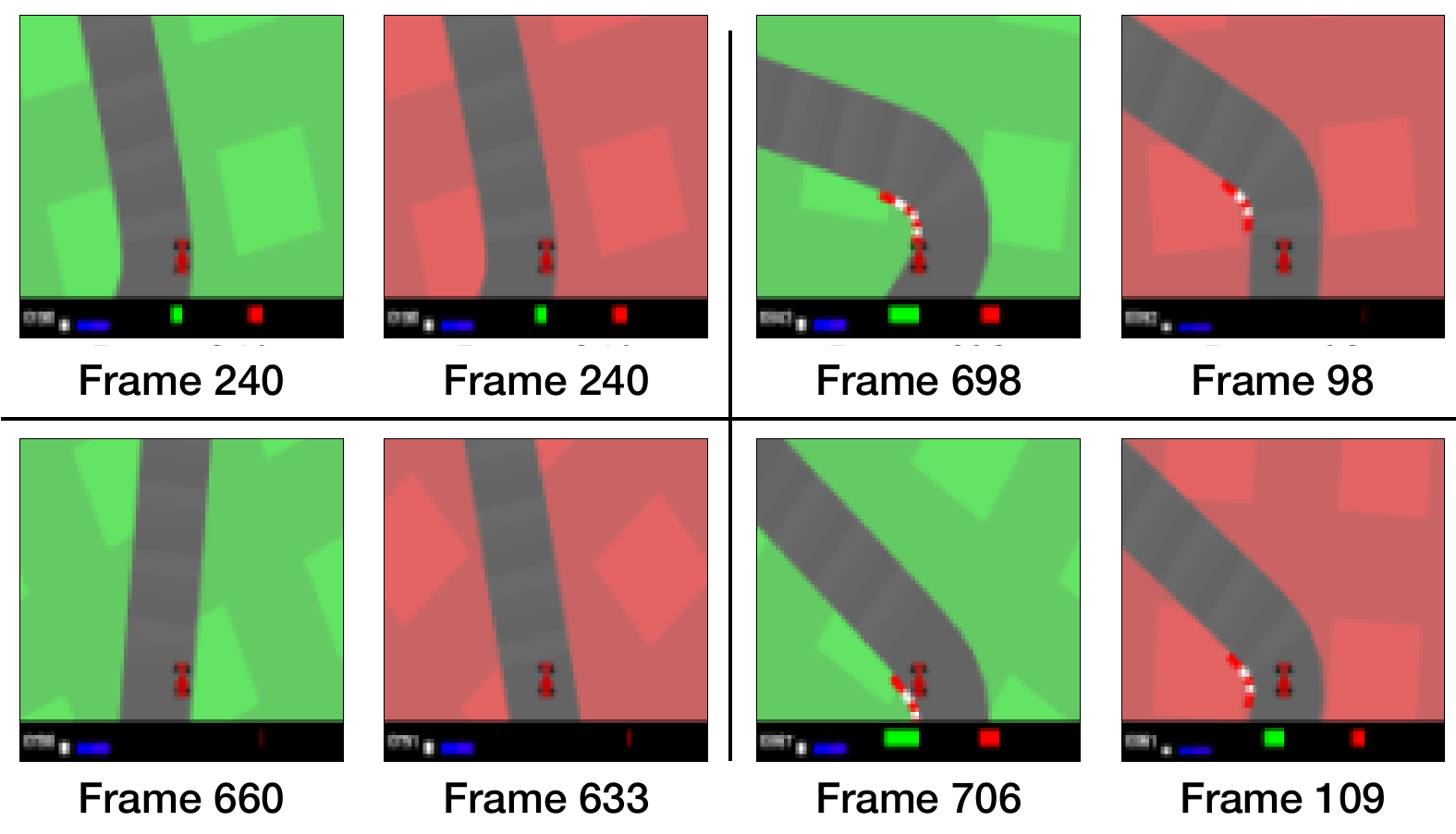}
  \caption{}
  \label{fig:frames-sim}
 \end{subfigure}
 \caption{\textbf{Correspondence-parallel projection reveals semantic structure.} (a) Cosine similarity matrices for absolute (left) vs.\ relative (right) representations from aligned frames (green/red grass). Absolute space shows uniformly low similarity, even on the diagonal; relative space reveals semantically meaningful off-diagonal matches---functionally similar track positions (e.g., a turn or straight segment of similar shape) despite temporal misalignment. (b) Example frame pairs from high-similarity regions of (a); three of the four are off-diagonal.}
 \label{fig:matrix-frames-sim}
\end{figure}

\subsection{Latent space analysis}\label{sec:exp-analysis}
\Cref{fig:matrix-sim} compares cosine similarities (green vs.\ red grass, $\sim$800 aligned frames). Absolute space (left) shows low similarity even on-diagonal; relative space (right) achieves strong diagonal similarity plus meaningful off-diagonal matches. In \cref{fig:frames-sim}, most of the pairs shown are off-diagonal and still correspond to semantically similar track positions despite temporal misalignment.

\subsection{End-to-end performance}\label{sec:exp-end-to-end-performance}
R3L achieves comparable training convergence and stability to absolute baselines on CarRacing environments (scores 797--832 across visual variations), except on the slow task (624~$\pm$~125 for R3L vs.\ 996~$\pm$~6 for absolute). Naive data augmentation fails: training with randomized backgrounds achieves 839~$\pm$~127 (green) and 900~$\pm$~32 (red) but catastrophically fails on unseen blue (158~$\pm$~106 vs.\ 800+ for direct training)---a $5\times$ drop demonstrating that exposure to variation does not enable zero-shot compositional generalization.

\subsection{Zero-shot stitching}\label{sec:zero-shot-stitching}
R3L's advantage emerges when composing encoders and controllers zero-shot to operate on unseen environment variations. \cref{tab:carracing-stitching_performance} shows stitching performance across seed, visual, and task variations (160 aggregated scores per cell: 10 track seeds $\times$ 4 encoders $\times$ 4 controllers). Diagonal cells (matching variations, e.g., green-green) stitch components trained with different seeds. Visual and task variations are analyzed independently.

R3L outperforms absolute baselines across all combinations. The multicolor baseline fails catastrophically at zero-shot stitching (all three evaluated cells near zero), confirming that naive data augmentation does not enable modular composition. The slow task variation causes performance decline for R3L.

\begin{table*}[t!]
 \caption{Comparing R3L and absolute in zero-shot stitching performance.
 The original domains for the encoders and the controllers are listed in the rows and columns, respectively.
 R3L outperforms the absolute model in all the tests.}
 \label{tab:carracing-stitching_performance}
 \centering
 \resizebox{0.8\textwidth}{!}{
 \begin{tabular}{ccccccccccc}
 \toprule
 & & & \multicolumn{6}{c}{\textbf{Controller}} & \\
 \cmidrule{4-9}
 & & & \multicolumn{3}{c}{\textbf{Visual Variations} (standard task)} & \multicolumn{3}{c}{\textbf{Task Variations} (green)} \\
 \cmidrule(r){4-6} \cmidrule(l){7-9}
 & & & \texttt{green} & \texttt{red} & \texttt{blue} & \texttt{slow} & \texttt{scrambled} & \texttt{no idle} \\
 \cmidrule(r){1-6} \cmidrule(l){7-9}
 \multirow{8}{*}{\rotatebox{90}{\textbf{Encoder}}}
 & \multirow{2}{*}{\rotatebox{90}{\texttt{green}}}
 & \emph{S. Abs} & 175 $\pm$ 304 & 167 $\pm$ 226 & $-4$ $\pm$ 79 & 148 $\pm$ 328 & 106 $\pm$ 217 & 213 $\pm$ 201 \\
 & & \emph{\textbf{S. R3L}} & \textbf{781} $\pm$ 108 & \textbf{787} $\pm$ 62 & \textbf{794} $\pm$ 61 & \textbf{268} $\pm$ 14 & \textbf{781} $\pm$ 126 & \textbf{824} $\pm$ 82 \\[1.5ex]
 & \multirow{2}{*}{\rotatebox{90}{\texttt{red}}}
 & \emph{S. Abs} & 157 $\pm$ 248 & 43 $\pm$ 205 & 22 $\pm$ 112 & 83 $\pm$ 191 & 138 $\pm$ 244 & 252 $\pm$ 228 \\
 & & \emph{\textbf{S. R3L}} & \textbf{810} $\pm$ 52 & \textbf{776} $\pm$ 92 & \textbf{803} $\pm$ 58 & \textbf{476} $\pm$ 430 & \textbf{790} $\pm$ 72 & \textbf{817} $\pm$ 69 \\[1.5ex]
 & \multirow{2}{*}{\rotatebox{90}{\texttt{blue}}}
 & \emph{S. Abs} & 137 $\pm$ 225 & 130 $\pm$ 274 & 11 $\pm$ 122 & 95 $\pm$ 128 & 138 $\pm$ 224 & 144 $\pm$ 206 \\
 & & \emph{\textbf{S. R3L}} & \textbf{791} $\pm$ 64 & \textbf{793} $\pm$ 40 & \textbf{792} $\pm$ 48 & \textbf{564} $\pm$ 440 & \textbf{804} $\pm$ 41 & \textbf{828} $\pm$ 50 \\[1.5ex]
 & \multirow{2}{*}{\rotatebox{90}{\texttt{far}}}
 & \emph{S. Abs} & 152 $\pm$ 204 & 65 $\pm$ 180 & 2 $\pm$ 152 & $-49$ $\pm$ 9 & 351 $\pm$ 97 & 349 $\pm$ 66 \\
 & & \emph{\textbf{S. R3L}} & \textbf{527} $\pm$ 142 & \textbf{605} $\pm$ 118 & \textbf{592} $\pm$ 86 & \textbf{303} $\pm$ 100 & \textbf{594} $\pm$ 39 & \textbf{673} $\pm$ 91 \\[1.5ex]
 \hline
 & & \emph{S. multicolor} & $-67$ $\pm$ 41 & $-59$ $\pm$ 54 & 13 $\pm$ 161 & \multicolumn{1}{c}{--} & \multicolumn{1}{c}{--} & \multicolumn{1}{c}{--} \\
 \bottomrule
 \end{tabular}
 }
\end{table*}

\paragraph{EMA Ablation.}
An ablation on CarRacing over $\alpha \in \{0, 0.1, 0.5, 0.9, 0.999\}$ (5M frames) selects $\alpha=0.999$, achieving training stability matching absolute baselines while enabling zero-shot stitching.

\paragraph{Computational advantage.}
Zero-shot stitching yields substantial computational savings through component reuse. Consider the CarRacing experiments with $n_v=4$ visual variations and $n_t=4$ task variations. Naive training requires training all $n_v \times n_t = 16$ visual-task combinations (52 hours total, \cref{tab:carracing-time-saved}). R3L requires training only the four highlighted combinations---which provide one encoder per visual variation and one controller per task variation (13 hours total, blue cells). Zero-shot stitching then assembles the remaining 12 combinations without additional training.

This is a 75\% time reduction (39 hours saved), but the advantage scales quadratically: with $n_v$ visual variations and $n_t$ task variations, naive training requires $O(n_v \times n_t)$ runs, while R3L requires only $O(n_v + n_t)$ runs. As environment complexity grows, correspondence-parallel operators transform prohibitive combinatorial training costs into linear scaling.

\begin{table}[h]
\caption{Table of training times. We only need to train combinations for the cells highlighted in blue and then perform zero-shot stitching to assemble all the other agents (\emph{totaling 13 hrs}). Normally, it would be required to train agents for each domain-task combination (\emph{totaling 52 hrs}). Visual variations \emph{V1}: green, \emph{V2}: red, \emph{V3}: blue, \emph{V4}: far (green) and task variations \emph{T1}: standard, \emph{T2}: slow, \emph{T3}: no idle, \emph{T4}: scrambled.}
\label{tab:carracing-time-saved}
\centering
\begin{tabular}{ccccc}
\toprule
& V1 & V2 & V3 & V4 \\
\midrule
T1 & \cellcolor{cellblue}3h & 3h & 3h & 3h \\
T2 & 4h & \cellcolor{cellblue}4h & 4h & 4h \\
T3 & 3h & 3h & \cellcolor{cellblue}3h & 3h \\
T4 & 3h & 3h & 3h & \cellcolor{cellblue}3h \\
\bottomrule
\end{tabular}
\end{table}

EMA stabilization mitigates the coordinate drift problem in online RL, enabling zero-shot policy stitching across visual and task variations. Zero-shot stitched policies achieve strong performance on unseen combinations while reducing training time by 75\%, with savings scaling quadratically as variations increase. This demonstrates that correspondence-parallel operators extend to sequential decision-making domains where continuous updates would otherwise destabilize alignment.

\makeatletter
\let\old@endpart\@endpart
\def\@endpart{\vspace{3cm}}
\makeatother

\part{Direct Translation: Coupled Operators}

\begin{center}
\includegraphics[width=0.5\textwidth]{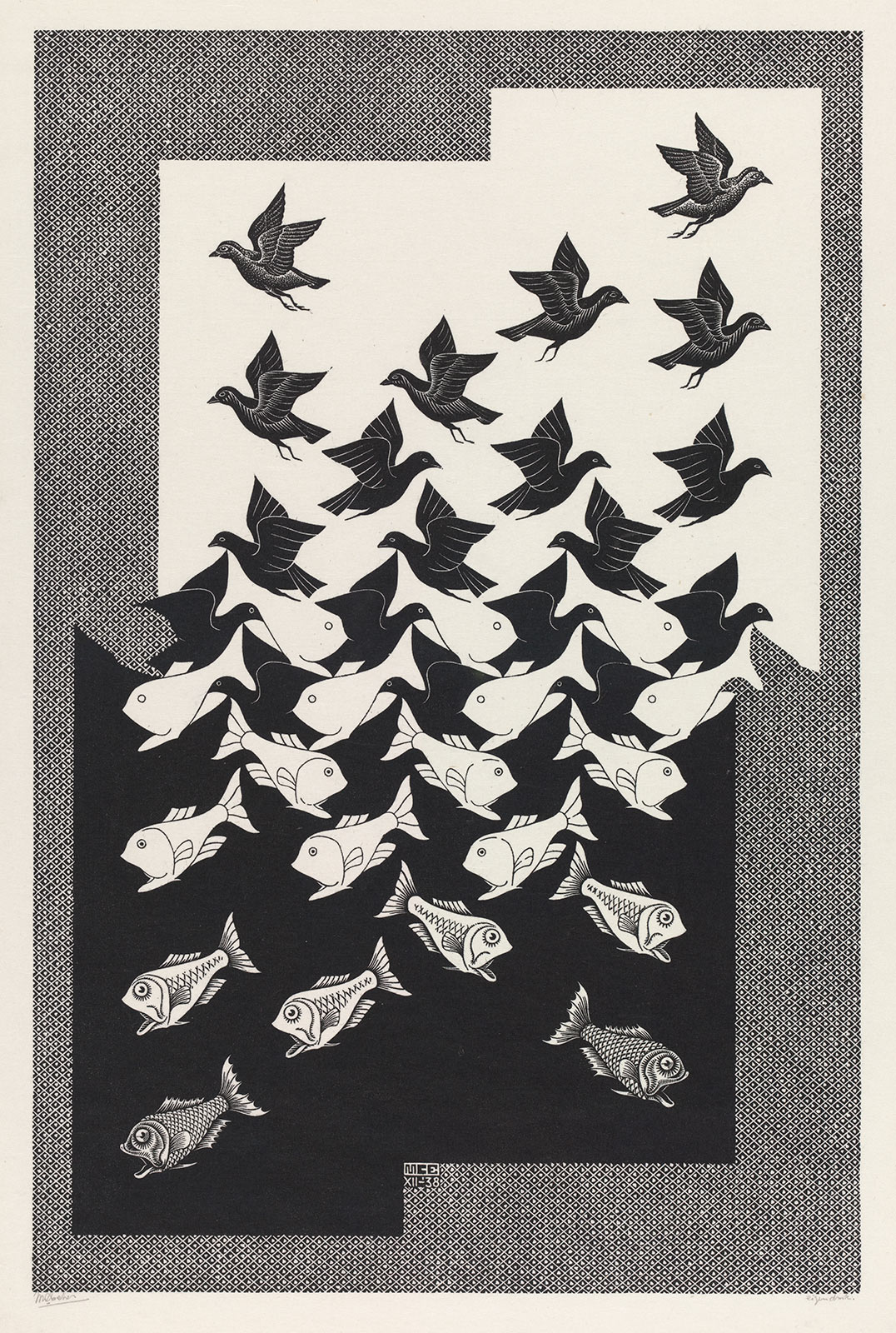}
\captionof*{figure}{\textbf{Sky and Water II} by M.C. Escher (1938). Birds and fish transform gradually into one another, illustrating the coupling principle: both sides of the correspondence work together to enable direct transformation between fundamentally different representations.}
\end{center}
\vfill
\begin{center}\footnotesize\textcopyright{} The M.C. Escher Company, The Netherlands. All rights reserved. Reproduced for academic, non-commercial use.\end{center}

\makeatletter
\let\@endpart\old@endpart
\makeatother

\clearpage

Describing every space relative to shared elements makes models comparable, but it leaves behind what each model already knows how to do with its own coordinates: a classifier, a decoder or a generator trained on one model's native representation cannot read the shared description. The operators of this part keep that machinery in use. Given the same correspondence, they learn a translation from one space directly into the native coordinates of another, so that the target's own components work unchanged, whether the two spaces belong to different models, to different layers of one model, or to one model before and after an intervention on its activations.

Part II established correspondence-parallel operators that transform spaces independently to semantic coordinates---intrinsic representations where models meet in auxiliary space. This part develops correspondence-coupled operators that sacrifice modularity to gain decoder compatibility. By using both sides of correspondence together, we construct direct transformations whose outputs live in target extrinsic coordinates---the native coordinate system of the target space.

The geometric principles remain unchanged: correspondence anchors arbitrary coordinate systems to shared semantic structure, revealing which geometric relationships persist across optimization trajectories. What differs is the operational mode. Coupled operators construct direct paths from source to target extrinsic coordinates, with no auxiliary semantic space in between. This design enables immediate reuse of pretrained decoders without modification, since outputs already inhabit the coordinate system decoders expect.

The same correspondence methodology applies whether the target is a different model, a different layer within the same model, a different temporal state, or a different distributional region. \cref{ch:inverse-relative-projection} bridges the two operator families: it composes a correspondence-parallel projection with an inverse reconstruction, so that outputs return to the native coordinates of the target space. \cref{ch:semantic_alignment,ch:lfm} present cross-model translation: fitting orthogonal transformations between models and constructing spectral operators that preserve functional relationships. \cref{ch:residual,ch:headpursuit,ch:lineas} apply correspondence principles within single models: decomposing residual streams spectrally, selecting attention heads for concept control, and steering representations through distributional alignment. In each case, correspondence links source and target---whether across models or within components---enabling transformations that respect semantic structure while operating in native extrinsic coordinates.

\chapter{Inverse Relative Projection}\label{ch:inverse-relative-projection}\label{ch:10}

\chapterpublication{maiorca2024inverse}{\textbf{arXiv 2024}}{Composing a relative projection with a pseudo-inverse reconstruction gives outputs in the target's native coordinates, so frozen decoders can be reused.}

\section{Introduction: The Bridge Between Operator Families}

The correspondence framework (\cref{ch:methodology}) establishes two operator families distinguished by fundamental trade-offs. \Glspl{correspondence-parallel} ($\gls{OperatorParallel}_{\gls{Corr}}$) transform spaces independently using their respective sides of \gls{correspondence-concept}, producing outputs in auxiliary \glspl{semantic-coordinate} that enable modular one-to-many communication. \Glspl{correspondence-coupled} ($\gls{OperatorCoupled}_{\Corr}$) use both sides together to construct direct transformations, producing outputs in target \glslink{extrinsic-factor}{extrinsic} coordinates compatible with pretrained \glspl{decoder}. The former provides modularity; the latter enables decoder reuse. Can we achieve both simultaneously?

This chapter presents a method that bridges these two operator families through operator composition. By chaining a forward \gls{correspondence-parallel} projection with an inverse reconstruction operator, effective translation $\mathcal{X} \to \mathcal{Y}$ is obtained that maintains the modularity of parallel processing while producing outputs in target-native coordinates suitable for frozen pretrained decoders. The key technical contribution is formalizing and stabilizing the inverse transformation: given \glspl{semantic-coordinate} produced by \gls{relrep} (\cref{ch:relative-representations}), target absolute coordinates are reconstructed through pseudo-inverse operations, enabling decoder compatibility without sacrificing the independence property that makes \gls{correspondence-parallel} operators attractive for model ecosystems.

\section{Method}

\subsection{Relative Representations}
Recall from \cref{ch:relative-representations} that \gls{relrep} factors out \glspl{extrinsic-factor} by anchoring to the anchors $\anchors[X] \subseteq \CorrSide{X}$ taken from the source side of the \gls{correspondence-concept} $\Corr$, producing \glspl{semantic-coordinate} through the relative projection $\mX_{\mathrm{rel}} = \relproj_{\anchors[X]}(\mX) = \mX \anchmat[X]^\top$, where $\mX \in \R^{n \times \dimof{X}}$ holds $n$ absolute encodings of dimension $\dimof{X}$, $\anchmat[X] \in \R^{\nanch \times \dimof{X}}$ holds the embeddings of the $\nanch$ anchors in $\anchors[X]$, $\mX_{\mathrm{rel}} \in \R^{n \times \nanch}$, and representations and anchors are normalized to unit norm (see \cref{ch:product-invariances} for extended discussion of angular invariances). This chapter uses ``relative space'' to denote the \glspl{semantic-coordinate} produced by \gls{relrep}, and ``absolute space'' to denote the native \glslink{extrinsic-factor}{extrinsic} coordinates of a model. The key observation enabling the inverse method is that semantically similar spaces differ primarily by isometry transformations (translation, rotation, reflection) and local rescaling, allowing separate estimation of these components and decoding of a single relative space into multiple target absolute spaces.

\subsection{Latent Space Translation}
The key benefit of the relative projection is its non-injective nature, as it maps different absolute spaces into a single relative space. The method formalizes the inverse process by exploiting the information contained in the anchor points. At a high level, the relative space is an intermediate space through which an encoding is translated from an absolute space $\mathcal{X}$ to any other semantically similar absolute space $\mathcal{Y}$. This can be formalized as:
\begin{equation} \label{eq:rel2abs}
\hat{\mY} = \mX_{\mathrm{rel}} \cdot (\anchmat[Y]^\top)^{+}\,,
\end{equation}
under the assumption that $\mX_{\mathrm{rel}} \approx \mY_{\mathrm{rel}}$, where $\mY \in \R^{n \times \dimof{Y}}$ holds the target encodings of the same $n$ samples, $\anchmat[Y] \in \R^{\nanch \times \dimof{Y}}$ the target embeddings $\va'_j$ of the anchors paired by $\Corr$, $\mY_{\mathrm{rel}} = \mY \anchmat[Y]^\top$, $\hat{\mY}$ is the reconstruction of $\mY$, and $(\cdot)^{+}$ denotes the Moore--Penrose pseudo-inverse.
This transformation of relative to absolute coordinates allows transferring any encoding between absolute spaces using only a subset of points from both spaces (the anchors). The anchors of the source space, the \textit{encoding space} $\mathcal{X}$, project into the shared relative space, while the anchors of the target one, the \textit{decoding space} $\mathcal{Y}$, project back to the specified absolute space.

As with other correspondence-based methods (see \cref{ch:relative-representations,ch:semantic_alignment,ch:latent_functional_maps}), the anchor points must be in \gls{correspondence-concept} to effectively translate between spaces. They represent the only bridge, a partial correspondence between them, much like a ``Rosetta stone'' (see \cref{ch:asif} and \citealp{norelli2023asif}).

\subsection{Stability improvements}
This section presents techniques for enhancing the stability of the process for inverting the anchors and evaluates their impact.
Consider the anchor matrix $\anchmat \in \R^{\nanch\times d}$ of either side, with $d$ the latent dimension of that side ($\dimof{X}$ or $\dimof{Y}$). $(\anchmat^\top)^{+}$ inverts the relative projection exactly when $\anchmat$ has rank $d$; with a pruned subset of fewer than $d$ rows it yields the least-squares reconstruction within the span of the selected rows. In a synthetic setting, where the vectors in $\anchmat$ can span the whole ambient space $\R^d$, the curse of dimensionality \citep{curse} makes a rank-deficient $\anchmat$ highly unlikely. However, in practical scenarios, $\anchmat$ is composed of only a reduced set of parallel anchors, and it is difficult to arbitrarily expand it using the entire training set (see \cref{ch:bootstrapping} and \citealp{cannistraci2023bootstrapping,vedula2024scalable}).
Additionally, the manifold hypothesis restricts the subspace where the encodings live, which is not the whole ambient space. This makes it more likely for two random points to be similar enough to negatively impact the inverse process's feasibility.
This motivates the following stability improvements.

\paragraph{Anchor Pruning}
Let the anchor set $\anchors[X]$ be drawn at random from the training data, with $\nanch = \dimof{X}$ anchors; the goal is to refine this set to enhance the stability of the inverse process. As a subset of the data manifold, the anchors may exhibit high correlation, leading to many near-zero singular values during the inverse computation. To mitigate this, a technique called ``anchor pruning'' is introduced.

The greedy farthest point sampling (FPS) algorithm~\citep{fps} selects a subset $\anchmat[X]^{(i)}$ of the rows of $\anchmat[X]$ with maximum orthogonality. The dissimilarity used by FPS is a customized cosine distance, calculated as:
\begin{equation} \label{eq:dcos}
  \operatorname{dcos} (\anchmat) = \1\1^\top - \big|\anchmat \anchmat^\top\big|\,,
\end{equation}
where the rows of $\anchmat$ have unit norm, so that $\anchmat \anchmat^\top$ holds the pairwise cosine similarities, and $| \cdot |$ denotes the entry-wise absolute value to avoid quasi-colinear points with opposite directions. A stopping condition ($\delta$) is applied based on the minimum acceptable distance between points. The subset of anchors selected by FPS is defined as:
\begin{equation} \label{eq:fps}
 \anchmat[X]^{(i)} = \operatorname{FPS}(\anchmat[X], \operatorname{dcos}, \delta)\,,
\end{equation}
for $i = 1, \dots, \omega$, one run per random seed with $\omega$ the number of runs, while $\anchmat[Y]^{(i)}$ keeps the corresponding rows of $\anchmat[Y]$.
The remaining vectors in $\anchmat[X]^{(i)}$ span a subspace of the row space of $\anchmat[X]$.
This process has the effect of selecting a highly informative set of anchors while reducing the correlation between them, which improves the stability of the inverse process.

\paragraph{Anchor Subspaces}
To balance the reduction in the cardinality of the anchor set given by the anchor pruning, the pruning procedure is applied multiple times with different random seeds, controlled by the parameter $\omega$. For each anchor subset $\anchmat[X]^{(i)}$ (spanning a subspace), the absolute space is independently reconstructed. The reconstructed spaces are then average-pooled in a parameter-free ensemble fashion:
\begin{equation} \label{eq:subpool}
 \hat{\mY} =
 \frac{1}{\omega}\times
 \sum_{i=1}^{\omega}{\mX_{\mathrm{rel}}^{(i)} \cdot (\anchmat[Y]^{(i)\,\top})^{+}} \,,
\end{equation}
where $\mX_{\mathrm{rel}}^{(i)}$ is the relative encoding of $\mX$ with respect to $\anchmat[X]^{(i)}$, the anchors of $\mathcal{X}$ in correspondence with $\anchmat[Y]^{(i)}$.
This approach maximizes the coverage of the original set of anchors while maintaining stability in the inverse process. It can be seen as an ensembling step, where the absolute encodings are reconstructed as the mean of the predictions of different estimators.

Overall, this process increases the method's robustness to the stochastic factors introduced by the anchor pruning while also mitigating information loss.

\paragraph{Anchor Completion}
The objective of this step is to recover $\dimof{X}$ points as anchors, such that the dimensionality of the points is not altered when transforming $\mX$ into $\mX_{\mathrm{rel}}$. To achieve this, the identity matrix is first encoded using any anchor subset $\anchmat[X]^{(i)}$ and decoded using the corresponding subset $\anchmat[Y]^{(i)}$. This process reveals how the canonical basis of $\mathcal{X}$ maps to relative space and back to space $\mathcal{Y}$. The canonical basis in $\mathcal{X}$ becomes the new set of anchors for $\mathcal{X}$ (denoted as $\widetilde{\anchmat}_{\mathcal{X}} = \mI$; these are anchors with no observation behind them), with the corresponding points as the new set of anchors for $\mathcal{Y}$ (denoted as $\widetilde{\anchmat}_{\mathcal{Y}}$). Intuitively, this process: i) stabilizes the computation without altering the information encoded by the anchors, redistributing it across different dimensions; ii) amounts to estimating a linear map from $\mathcal{X}$ to $\mathcal{Y}$ (an isometry when the two spaces differ by one), since the relative encoding of $\mX$ with respect to the new $\widetilde{\anchmat}_{\mathcal{X}}$ is $\mX$ itself.

This yields a number of anchor points equal to $\dimof{X}$, ensuring that projecting $\mX$ into the relative space does not change the dimensionality of the points.

\subsection{Scale invariance and temperature}\label{sec:scale-invariance-temperature}\label{sec:scale-invariance}

 \begin{figure}
 \centering
 \begin{overpic}[trim=1.25cm 2cm 1 3cm,clip,width=0.85\linewidth]{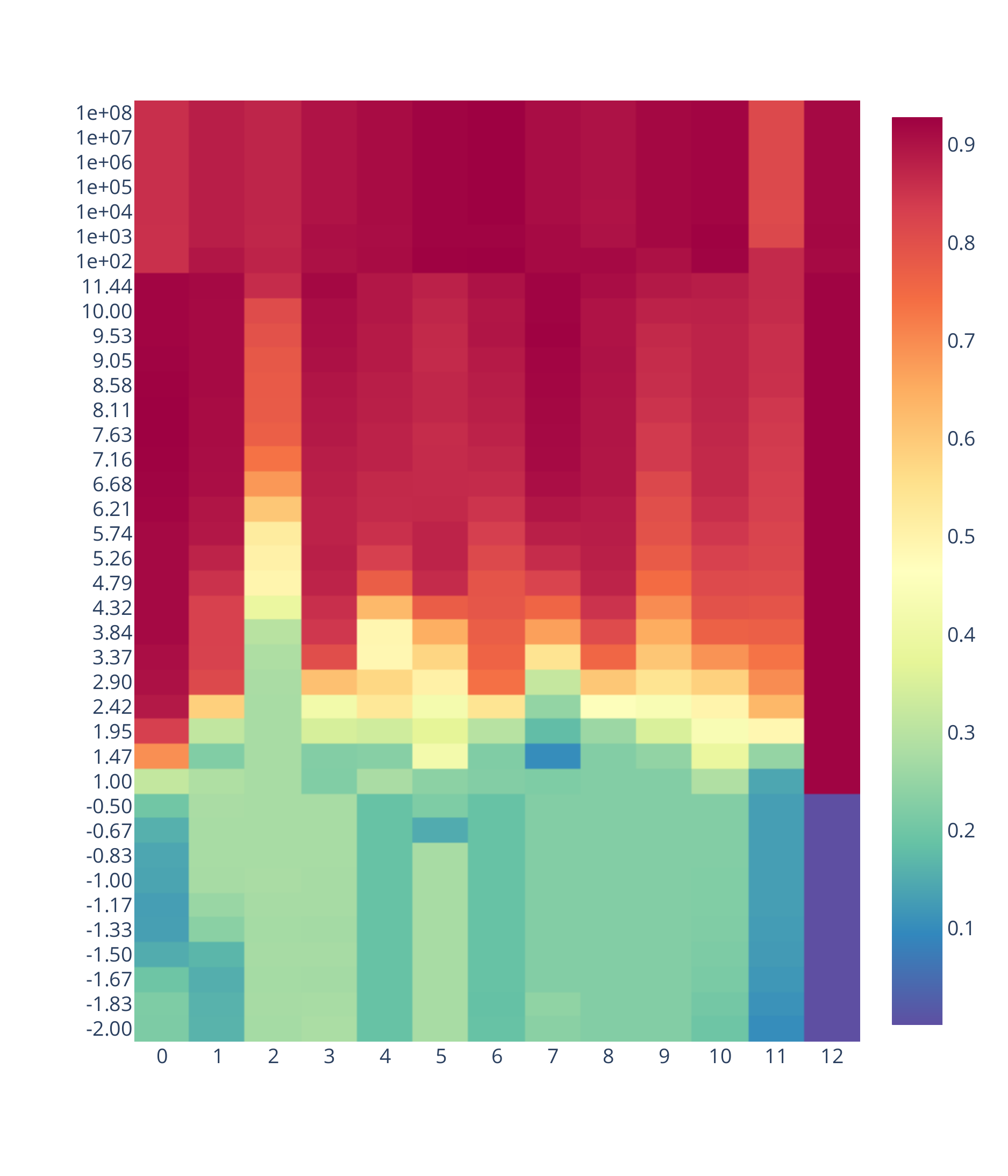}
  \put(33, 0){Rescaled layer}
  \put(0, 35){\rotatebox{90}{Rescaling factor $\alpha$}}
  \put(93, 59){\rotatebox{-90}{Accuracy}}
 \end{overpic}
 \caption{Scale invariance of RoBERTa according to the performance of a downstream classifier trained on the encodings of the last attention layer. At each layer (with 0 being the embedding layer and 12 the output one), for each run, the encodings are normalized to unit norm and multiplied by $\alpha$, and the effect on final accuracy is measured. The performance without any rescaling is $0.92$. }
 \label{fig:irp:rescaled-layer-acc}
\end{figure}

This section investigates the scale-invariance properties of neural classifiers that utilize the softmax activation function. The focus is on the effect of rescaling operations on the latent input encodings, arguing that, by construction, certain classifiers exhibit scale-invariance properties without the need for additional priors.

The softmax function, commonly used in neural classifiers, is known to be a temperature-controlled variant of the argmax function:
\begin{equation} \label{eq:irp:softmax}
\softmax(\vs)_i=\frac{e^{s_i/\beta}}{\sum_{j=1}^C e^{s_j/\beta}}\,,
\end{equation}
where $\vs \in \R^C$ are the logits over $C$ classes and $\beta$ is the temperature; as $\beta \to 0$ the output approaches the one-hot argmax.

This means that the softmax temperature can be used to control the level of confidence of the classifier's predictions. This section shows that a similar effect can also be achieved by rescaling the latent encodings given as input to a trained (and frozen) classifier.

To show this, note that the rescaling factor, $\alpha$, can be factored out of the matrix multiplication in the Linear layers of the classifier. This can be represented mathematically as:
$\vs = \alpha \mW\vx + \vb$, where $\vx$ is the input latent encoding, $\mW$ is the weight matrix, $\vb$ is the bias vector, $\alpha$ is the rescaling factor, and $\vs$ is the output of the final linear layer, i.e.\ the logits of \cref{eq:irp:softmax}. Strictly, for $\alpha > 0$ and a bias-free layer ($\vb = \0$), this implies that the rescaling operation can be ``pushed through'' the classifier without affecting its final prediction as it becomes equivalent to some temperature value applied at the softmax level.

Regarding the effect of rescaling when non-linear activation functions are involved, it is conjectured that as long as the function has a monotonic interval, rescaling all dimensions by an amount similar to the mean scale of the encodings on which the classifier was trained keeps the encodings in the monotonic interval, preserving the scale-invariance property. \cref{sec:irp:scale-distribution} links the monotonic interval of activation functions to the scale range of the data encodings.

In summary, this analysis provides an argument that neural classifiers that utilize the softmax activation function can, in practice, maintain their scale-invariance properties when the input latent encodings are rescaled (\cref{fig:irp:rescaled-layer-acc}). This property is essential to the method, as it allows ignoring the exact scale when decoding back to the absolute space.

\section{Relative Inversion}\label{sec:relative-inversion}
This section evaluates the capabilities and effectiveness of the proposed method. The results first demonstrate inverting a relative space by decoding it back into the original absolute space on which it was computed, referred to as \textit{intra-space inversion}. This provides a valuable illustration of the method's effectiveness and components. Building upon this, the property that relative representations of semantically similar spaces are relatively consistent enables generalization to \textit{inter-space inversion}, which decodes into an absolute space different from the one used for relative encoding, effectively enabling latent translation.

\paragraph{Experimental setting}
All experiments follow a relative inversion pattern:
given a dataset $\Data$ and a set of encoders $\{\Enc_i\}$, where $i = 1, 2, \dots, m$, the method translates a sample $\vx$ from the latent space produced by $\Enc_i$ to the latent space produced by $\Enc_j$. The translation consists of the following steps: (1) centering and L2 normalization; (2) relative encoding with the anchors embedded by $\Enc_i$; (3) relative decoding with the anchors embedded by $\Enc_j$; (4) de-normalization and de-centering according to the anchor statistics. The dimensionality of $\Enc_i$ and $\Enc_j$ is arbitrary and may differ, enabling zero-shot translation between latent spaces of different dimensionalities.

\subsection{Space Inversion}
The performance of the method is evaluated by analyzing its sensitivity to two hyperparameters: $\omega$ (number of subspaces) and $\delta$ (pruning threshold). The analysis examines how these parameter choices affect overall performance and identifies optimal values for each.

\paragraph{Vision}
The experiments use \gls{cifarh}, an image classification dataset, embedding each image with 5 different encoders: 3 vision transformers \citep{dosovitskiy2021an}, RexNet \citep{rexnet}, and the visual encoder of CLIP \citep{clip}. For each sample and encoder, the relative encoding followed by relative inversion is applied with different values for $\omega$ and $\delta$. The analysis measures: i) \textit{reconstruction similarity} in terms of cosine similarity (since scale can be ignored) between the original encoding and the reconstructed one; ii) whether $\omega$ and $\delta$ affect performance in the same way for both intra-space and inter-space settings. The experiments use 768 anchors and group all results by $\omega$ and $\delta$.

\begin{figure}
 \centering
 \begin{overpic}[trim=0 -1.5 0 -2.5ex,clip,width=0.9\linewidth]{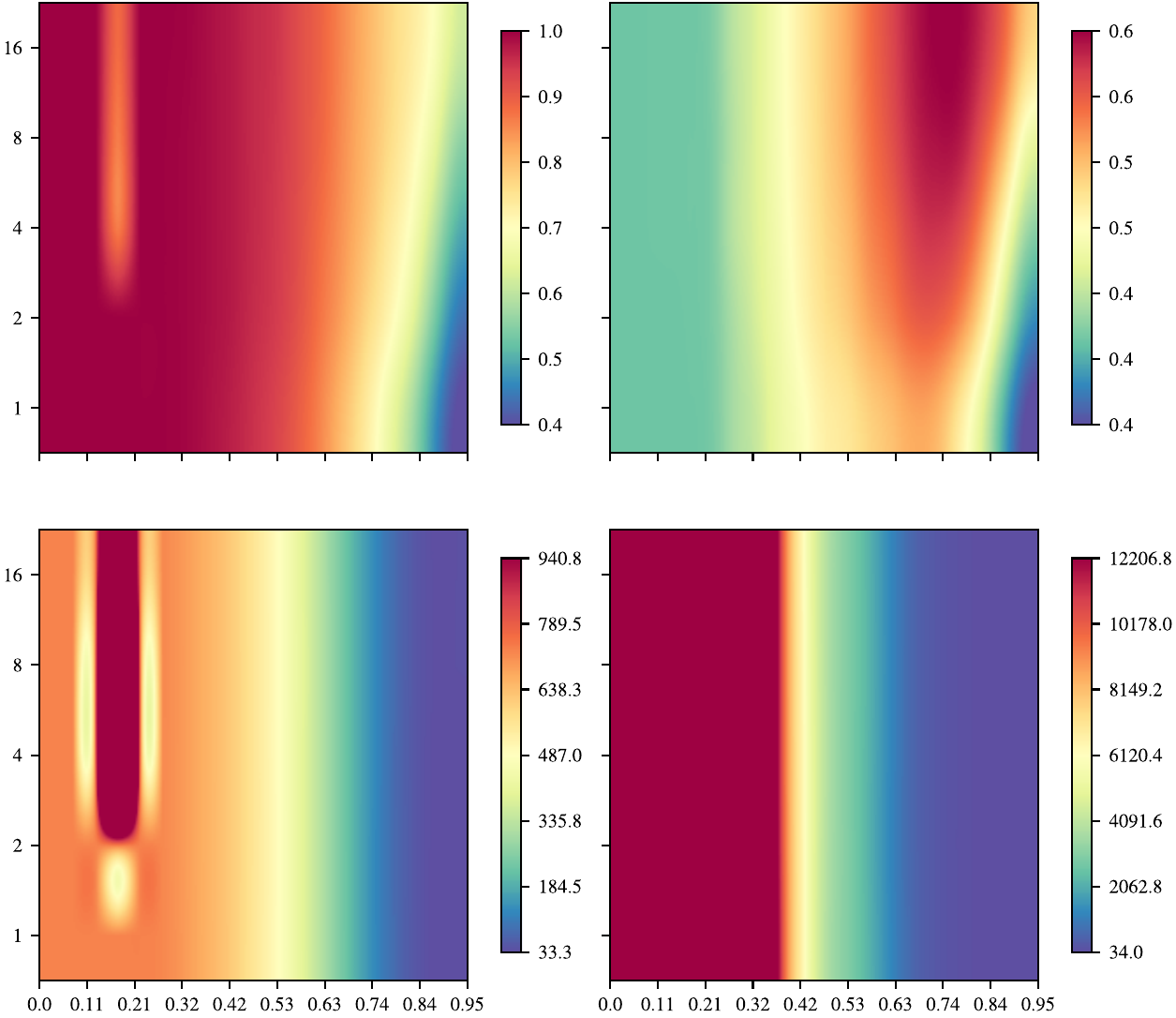}
  \put(62, 89){\small Inter space}
  \put(13, 89){\small Intra space}

  \put(-3, 58){\rotatebox{90}{\scriptsize \# Subspaces ($\omega$)}}
  \put(-3, 12){\rotatebox{90}{\scriptsize \# Subspaces ($\omega$)}}

  \put(100.5, 84){\rotatebox{-90}{\scriptsize Reconstruction Similarity}}
  \put(100.5, 35){\rotatebox{-90}{\scriptsize Condition Number}}

  \put(8, -3){\scriptsize Pruning Threshold ($\delta$)}
  \put(57, -3){\scriptsize Pruning Threshold ($\delta$)}
 \end{overpic}
 \vspace{2ex}
 \caption{On the top row, reconstruction similarity sensitivity at different number of subspaces and pruning threshold ($\delta$) of intra-space inversion (\textit{left}) and inter-space inversion between different encoders (\textit{right}) on the coarse-grained \gls{cifarh}. On the bottom row, the corresponding condition number average over the subspaces. Higher pruning thresholds lower the condition number, stabilizing the matrix inverse; in the inter-space setting this increases the reconstruction similarity up to $\delta \approx 0.8$.}
 \label{fig:inner-space-inversion}
\end{figure}

The results in \cref{fig:inner-space-inversion} demonstrate that: i) the inverse works effectively, achieving medium to high similarity between the original encoding and the reconstructed one; ii) $\omega$ has little to no impact in the intra-space setting when the number of anchors pruned is low (low $\delta$), but effectively compensates for aggressive pruning (high $\delta$); iii) the inter-space setting reveals synergy between the two parameters, with optimal balance found at medium-high pruning combined with high resampling (high $\omega$).

\paragraph{Language}
The experiments use WikiMatrix \citep{wikimatrix} as a source for parallel multi-lingual data, consisting of translations of the same sentence in different languages automatically extracted from Wikipedia. The selection includes $\sim$4000 samples with high probability of being different translations of the same sentence in 4 languages. For each language, a language-specific RoBERTa-based encoder is selected: English \citep{roberta}, Spanish \citep{es_roberta}, French, and Japanese. Each sentence in a sample is encoded using its corresponding language-specific encoder and a language-agnostic one, specifically XLM-R \citep{xlmr}. The task translates a mono-lingual encoding in a source language to its corresponding mono-lingual encoding in a target language. Cosine similarity between the two is the score metric used to compare: \textit{i)} the decoded representation of the method, Inverse Relative Projection (IRP); \textit{ii)} the multi-lingual representations of XLM-R; \textit{iii)} the original mono-lingual encodings of the two languages.

The results in \cref{tab:cross-lang-inversion} demonstrate that the latent translation method works effectively across languages.
For reference, the mean \texttt{en} similarities between all possible pairs of different embeddings in the absolute and XLM-R spaces are respectively $0.93 \pm 0.05$ and $0.98 \pm 0.03$.

These experiments demonstrate that the reconstruction is effective across various domains (including different languages) and with different encoders.

\begin{table}
\centering
\caption{Reconstruction similarities between parallel encodings across languages. \textit{Absolute} refers to the original encodings, \textit{XLM-R} to the encodings of a multilingual LLM, \textit{IRP} to Inverse Relative Projection, comparing the translated source embeddings with the target ones.
}
\label{tab:cross-lang-inversion}
\begin{tabular}{llccc}
\toprule
\texttt{Enc.} & \texttt{Dec.} & \texttt{Absolute} & \texttt{XLM-R} & \texttt{IRP} \\
\midrule
\texttt{en} & \texttt{en} &    1.00 &    1.00 &   0.97 \\
\texttt{es} & \texttt{en} &    0.00 &    0.99 &   0.97 \\
\texttt{fr} & \texttt{en} &   -0.02 &    0.99 &   0.97 \\
\texttt{ja} & \texttt{en} &    0.03 &    0.99 &   0.97 \\
\bottomrule
\end{tabular}
\end{table}

\subsection{Empirical Validation of Scale Invariance}

This section examines scale invariance in neural networks and its implications for model compositionality. By analyzing the behavior of networks when subjected to a specific type of input manipulation, \textit{rescaling injection}, the robustness and versatility of neural networks in handling different scales of input data is demonstrated, a property that enhances the applicability of the inverse relative projection method.

\paragraph{Rescale Injection}

\begin{figure}
 \centering
 \begin{overpic}[trim=0 0 0 0, clip,width=.9\linewidth]{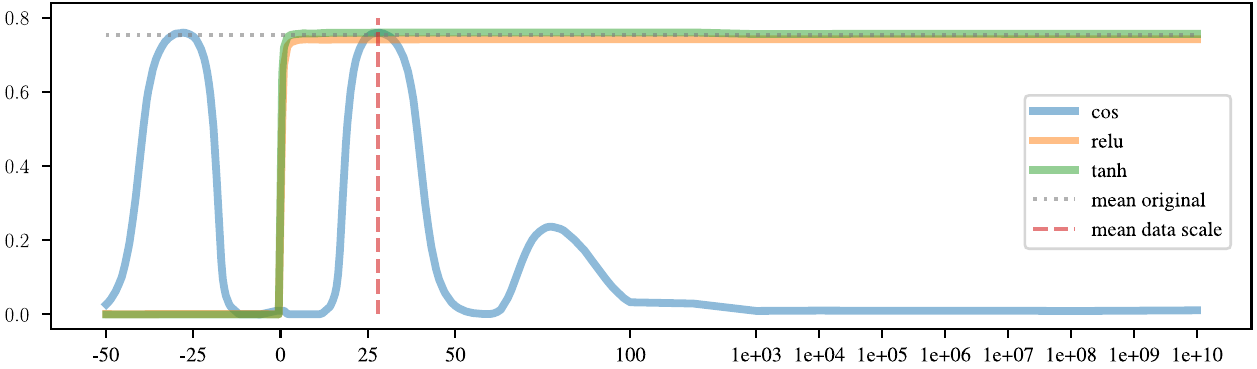}
  \put(-3.5, 7){\rotatebox{90}{\scriptsize Accuracy}}
  \put(40, -3){\scriptsize Rescale Factor}
 \end{overpic}
 \vspace{2ex}

 \caption{Performance comparison of three Multilayer Perceptrons (MLPs) with different activation functions, namely cosine (blue), ReLU (orange), and tanh (green) at different rescaling factors $\alpha$. The ReLU and tanh MLPs exhibit scale invariance for $\alpha > 0$, while the cosine activation function is only invariant on the mean data scale and its periodic cycles.
 }
 \label{fig:irp:scale-inv-mono}
\end{figure}

The \textit{rescaling injection} is defined as the operation of artificially altering the scale of the features produced at a specific layer of the network. This is achieved by normalizing the embeddings to unit norm and then rescaling them by a factor of $\alpha$; this per-sample rescaling sets the norm of each embedding to $\alpha$, unlike the global factor multiplying the raw encoding in $\vs = \alpha \mW\vx + \vb$ above. By varying the value of $\alpha$, the network's performance at different scales is observed. This empirical analysis provides insight into the scale invariance properties of neural networks and their potential for use in model compositionality.

\Cref{fig:irp:scale-inv-mono} presents experimental results regarding this investigation. Simple multi-layer perceptrons (MLPs) composed of two hidden layers, with no normalization layers, were trained using encodings produced by the CLIP vision transformer (\texttt{clip-vit-base-patch32}) on the \gls{cifarh} (fine) dataset. The MLPs were evaluated using different activation functions: cosine (blue), ReLU (orange), and tanh (green). The rescaling injection technique was applied directly to the input embeddings, scaling the unit-normalized embeddings by $\alpha$.

For $\alpha > 0$, the scale of the embeddings has little impact on the MLPs' performance when using monotone activation functions that do not flip signs. This is a non-trivial result, as the nonlinearity of the activation function, the presence of bias terms $\vb$, and the absence of normalization layers make it difficult to predict the performance impact of rescale injection.
The cosine activation function shows an oscillatory performance comparable to the original embeddings when rescaled by the mean embeddings scale (vertical red line) or its opposite since it is symmetric.

Even the internal layers of large deep learning models exhibit \textit{positive scale invariance}, as illustrated in \cref{fig:irp:rescaled-layer-acc}. The underlying mechanism for this behavior is not straightforward, but may result from the interplay between various factors, such as the choice of activation function, the use of normalization layers, the optimization objective and regularization techniques employed during training. While further research is needed to fully understand this phenomenon, it can be exploited for inverse projection purposes since it allows ignoring the encoding scale.

\begin{figure}
 \centering
 \begin{overpic}[clip,width=0.95\linewidth]{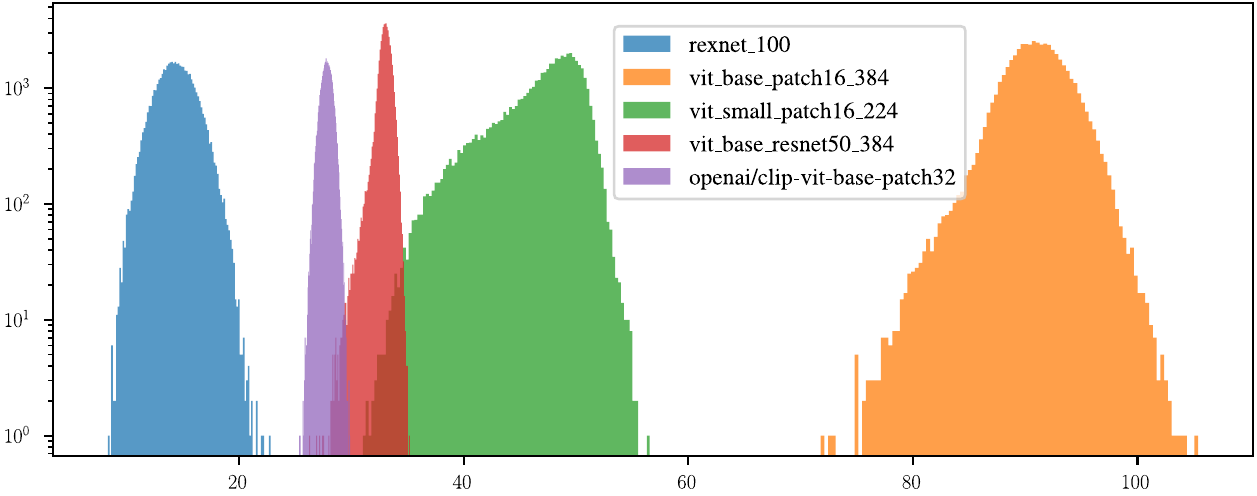}
  \put(-3, 4){\rotatebox{90}{\scriptsize Number of samples ($\log$)}}
  \put(45, -3){\scriptsize Scale}
 \end{overpic}
 \vspace{2ex}
 \caption{Distribution of the embedding scales in different pretrained encoders on the \gls{cifarh} dataset. Well-behaved roughly Gaussian distributions with a single mode and well-defined mean are crucial for the method, as they support the ability to rescale the embeddings to a mean scale.}
 \label{fig:irp:norm-ranges}
\end{figure}

\paragraph{Scale distributions}\label{sec:irp:scale-distribution}
\Cref{fig:irp:norm-ranges} presents the scale distribution of the embeddings produced by several encoders on the \gls{cifarh} (fine) dataset.
This empirical analysis shows a consistent pattern among encoders, in that the scale distribution of their embeddings follows a roughly Gaussian one with a single mode and a well-defined mean, with minimal presence of outliers. This consistent behavior across encoders is likely attributed to their architectural choices, such as the normalization techniques, regularizations and the optimization problems they are designed to solve.

Overall, the scale invariance property and the encoders' well-behaved scale distribution enable the zero-shot translation between arbitrary latent spaces via the IRP method.

\section{Zero-Shot Model Stitching via Relative Space}
\begin{table*}
\centering
\caption{\textbf{Zero-shot stitching for classification}. Comparison between stitching and non-stitching accuracies. Stitching includes the translation method (\textit{zero-shot}) with its hyperparameters and the stitching of \textit{absolute} spaces. The \textit{achievable} rows show the number of possible combinations the stitching methods can tackle among the \textit{available} ones. The method works between encoders of arbitrary embedding dimensionality, enabling stitching of all available pairs. The results are averaged over 5 different random seeds. The \textit{non-stitch} column reports, for reference, each classification head applied to the encoder it was trained on.}
\label{tab:stitching}
\setlength{\tabcolsep}{4pt}
\begin{tabular}{llcccccc}
\toprule
 & & & & \multicolumn{3}{c}{Accuracy} & \\
 \cmidrule(lr){5-7}
 & & $\omega$ & $\delta$ & zero-shot & absolute & non-stitch & \textit{available} \\
\midrule
\multirow{6}{*}{\rotatebox{90}{Vision}} & \gls{mnist} & 8 & 0.65 & $0.88 \pm 0.03$ & $0.10 \pm 0.03$ & $0.97 \pm 0.01$ & 20 \\
& \gls{fmnist} & 8 & 0.65 & $0.82 \pm 0.03$ & $0.10 \pm 0.02$ & $0.91 \pm 0.01$ & 20 \\
& \gls{cifart} & 8 & 0.75 & $0.92 \pm 0.05$ & $0.11 \pm 0.04$ & $0.95 \pm 0.03$ & 20 \\
& \gls{cifarh} \textit{(coarse)} & 8 & 0.70 & $0.72 \pm 0.10$ & $0.05 \pm 0.01$ & $0.88 \pm 0.05$ & 20 \\
& \gls{cifarh} \textit{(fine)} & 8 & 0.75 & $0.62 \pm 0.09$ & $0.01 \pm 0.00$ & $0.82 \pm 0.07$ & 20
\\[1.15ex]
& \textit{achievable} & & & 20 & 6 & & \\
\midrule
\multirow{3}{*}{\rotatebox{90}{NLP}} & \gls{trec} & 8 & 0.70 & $0.73 \pm 0.09$ & $0.22 \pm 0.06$ & $0.87 \pm 0.04$ & 20 \\

& \gls{n24news} & 8 & 0.65 & $0.55 \pm 0.10$ & $0.05 \pm 0.01$ & $0.70 \pm 0.07$ & 20
\\[1.15ex]
& \textit{achievable} & & & 20 & 20 & & \\
\bottomrule
\end{tabular}
\end{table*}

This section combines the inter-space translation capabilities (\cref{sec:relative-inversion}) of the method with the scale-invariance of neural classifiers (\cref{sec:scale-invariance-temperature,sec:irp:scale-distribution}) to obtain zero-shot stitching of independently pretrained models and classification heads. This method is distinct from previous approaches that rely on relative representations, as it does {\em not} assume that the decoders have been independently trained on relative representations.
Additionally, since the relative space itself is the bridge between absolute ones, the transformation remains independent and not conditioned on the target space as in \citet{maiorca2023semantic} (see \cref{ch:semantic_alignment}). In fact, each absolute space gets \textit{independently encoded} to the relative one, and each absolute space can be \textit{independently decoded} just by selecting appropriate anchors. This makes the approach more versatile and applicable to a wider range of scenarios, especially when the target space is unknown beforehand.

\paragraph{Experimental setting} The experiments use a variety of Computer Vision (\gls{mnist} \citep{Lecun1998-we}, \fmnist{} \citep{Xiao2017-er}, \gls{cifart}, \gls{cifarh} \citep{Krizhevsky2009-hv}) and Natural Language Processing (\gls{trec} \citep{hovy-etal-2001-toward} and \news{} \citep{wang-EtAl:2022:LREC3}) datasets.
For the text domain, five language models are the encoders, and for the image domain, five encoders, all pretrained and frozen.
For each dataset and encoder, a classification head is trained on top of their specific encodings. The mean performance over all combinations of (encoder, classification head) for each test set is then measured in different settings: i) \textit{zero-shot}, the result of applying inter-space translation, for which $\omega$ and $\delta$ are also reported; ii) \textit{absolute}, the result of using the encodings without any transformation, considered as a probe for any pre-existing compatibility of encodings; iii) \textit{non-stitch}, the performance of the classification head applied to the original space it was trained on, considered as an upper bound.

Lastly, one dataset, \gls{trec}, is selected to examine the sensitivity to the hyperparameters $\omega$ and $\delta$, considering their impact on: \textit{i)} the reconstruction similarity; \textit{ii)} the accuracy on the downstream task; \textit{iii)} the mean condition number across the subspaces; \textit{iv)} the mean number of anchors after pruning across subspaces.

\paragraph{Result analysis.}
The complete stitching results are in \cref{tab:stitching}. As expected, the \textit{absolute} encodings achieve a score comparable to random guessing while also considering fewer encoder combinations in the vision setting, due to the dimensionality discrepancy between some of them.

\begin{figure}[h]
 \centering
 \begin{overpic}[trim=0 0 0 -4ex, width=0.9\linewidth]{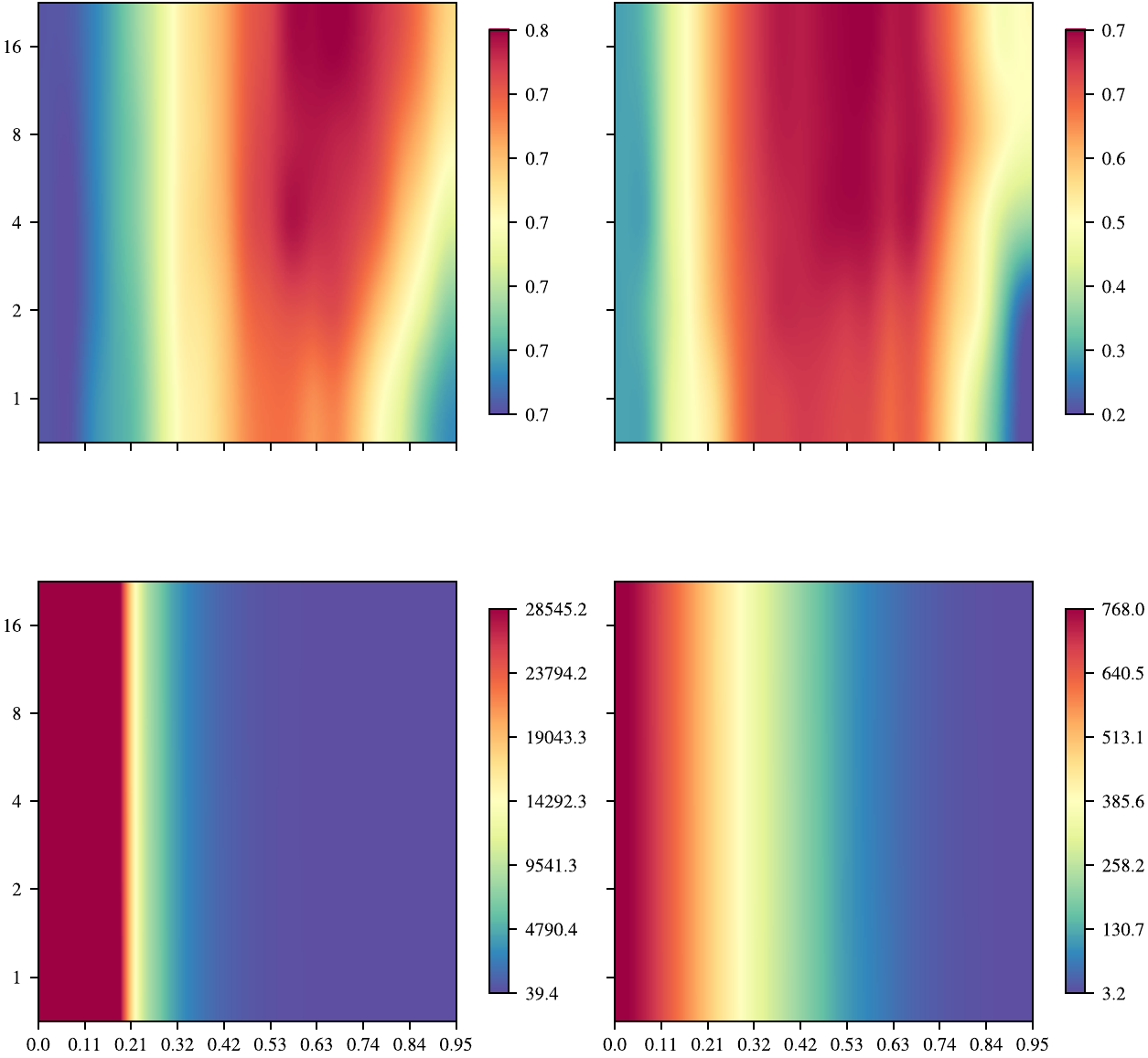}
  \put(1.5, 94){\small Reconstruction Similarity}
  \put(65, 94){\small Accuracy}

  \put(6, 44){\small Condition Number}
  \put(56, 44){\small Number of anchors}

  \put(-3, 62){\rotatebox{90}{\scriptsize \# Subspaces ($\omega$)}}
  \put(-3, 12){\rotatebox{90}{\scriptsize \# Subspaces ($\omega$)}}

  \put(8, -3){\scriptsize Pruning Threshold ($\delta$)}
  \put(57, -3){\scriptsize Pruning Threshold ($\delta$)}
 \end{overpic}
 \vspace{2ex}
 \caption{On the top row, the reconstruction similarity (\textit{left}) and accuracy (\textit{right}) sensitivity analysis on the number of subspaces ($\omega$) and pruning threshold ($\delta$) on the \gls{trec} dataset. On the bottom row, the corresponding condition number (\textit{left}) and the number of anchors after pruning (\textit{right}) averaged across subspaces.
 The metrics are averaged over stitching between all possible pairs of language encoders across 3 seeds. }
 \label{fig:stitching-trec}
\end{figure}

Additionally, as expected, \cref{fig:stitching-trec} shows a high correlation between the reconstruction similarity and the classification accuracy: their Pearson correlation is $0.7$.
Performance-wise, having a stable inverse (i.e., a low condition number) is more important than many anchors, which intuitively should better represent the data. Indeed, the best performance is obtained with few anchors and a low condition number. Moreover, the anchor pruning technique successfully stabilizes the inverse, as demonstrated by the high Pearson correlation between the condition number and the number of anchors ($0.93$).

This result shows the effectiveness and stability of the inversion method in zero-shot classification scenarios, as well as its ability to maximize the information extracted from the starting anchor set.

\subsection{Cross-domain stitching}
As a final experiment, the capabilities of the inversion method to translate representations between distinct domains in a zero-shot fashion are tested.

\begin{figure*}
 \centering
 \begin{minipage}{.7\columnwidth}
  \begin{overpic}[trim=-1cm -1cm 0 0,width=1\linewidth]{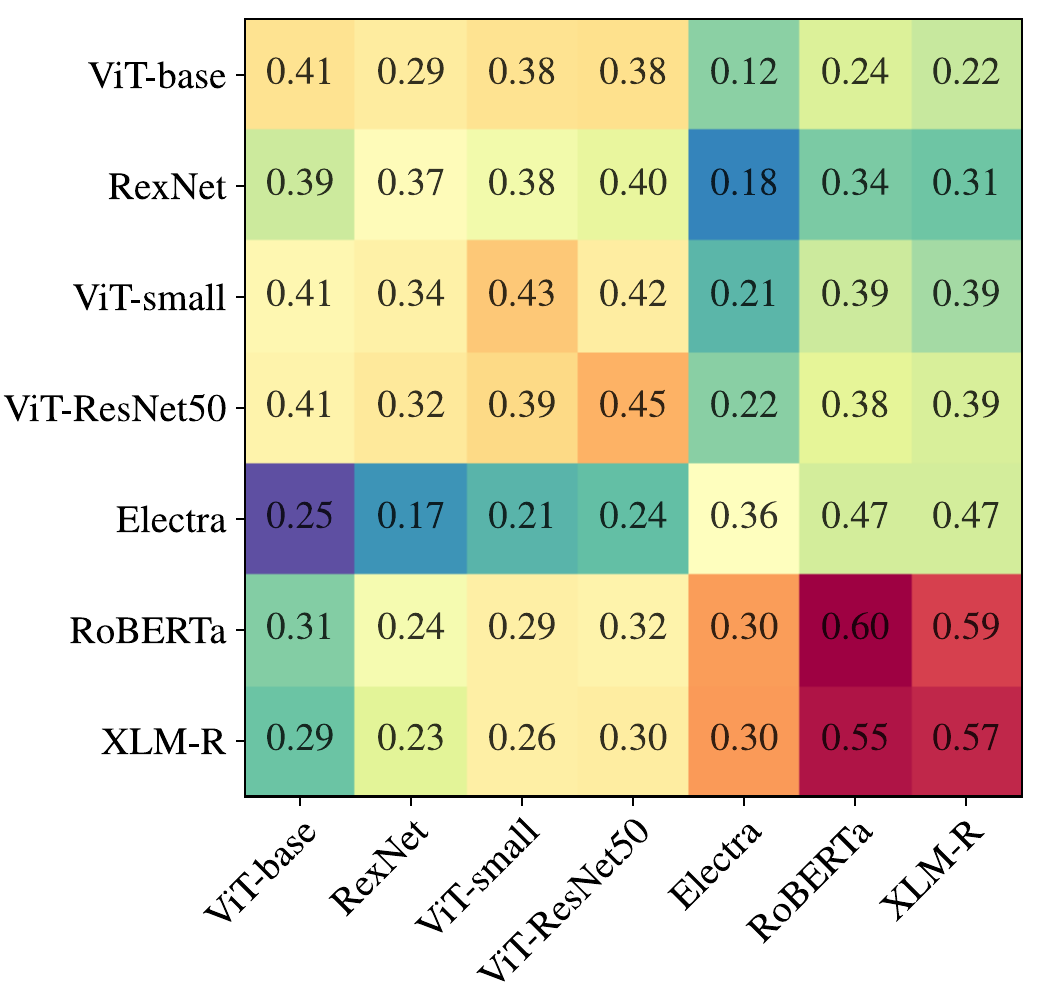}
   \put(-0.25, 50){\rotatebox{90}{\scriptsize Decoder}}
   \put(52, 0){\scriptsize Encoder}
  \end{overpic}
 \end{minipage}
 \begin{minipage}{.29\columnwidth}
  \scriptsize
   \setlength{\tabcolsep}{0pt}
  \begin{tabular}{p{3ex}p{14ex}p{6ex}}
  \toprule
    & Model & Acc \\
  \midrule
  \multirow{4}{*}{\rotatebox{90}{Vision}} & ViT-base & 0.50 \\
   & RexNet & 0.42 \\
   & ViT-small & 0.48 \\
   & ViT-ResNet50 & 0.50 \\[1.5ex]
  \multirow{3}{*}{\rotatebox{90}{NLP}} & Electra & 0.60 \\
   & RoBERTa & 0.76 \\
   & XLM-R & 0.71 \\
  \bottomrule
  \end{tabular}

 \end{minipage}

 \caption{Performance comparison between different encoders and data modalities on the \news{} multimodal dataset. On the right, each classification head applied to the encoder it was trained on. On the left the stitching performance between pairs of encoders and decoders. The stitching achieves satisfactory results even when translating across modalities. Results obtained with $1000$ anchors, $\omega=16$ and $\delta=0.8$.}
 \label{fig:cross-domain-stitching}
\end{figure*}

\paragraph{Experimental setting}
The experiments select \news{} as in \citet{maiorca2023semantic}, since it is a multimodal news classification dataset that contains both text and associated pictures. Different pretrained uni-modal encoders apply the standard encoding procedure to these two features separately. A classification head is then trained on top of each one. Finally, each encoder is stitched with a classification head different from its corresponding one, measuring classification accuracy.

\paragraph{Result analysis.}
The results are in \cref{fig:cross-domain-stitching}. The discrepancy in the mean accuracy across the rows and the columns of the heatmap is a signal that can be used to identify spaces more suited to be \textit{decoded into} and the ones that are stronger in \textit{encoding from}. In fact, the RoBERTa and XLM-R language models as source space for the translation exhibit stronger performance than the vision encoders, as observed by \citet{maiorca2023semantic}. This behavior relates to the higher generality of the text domain data used during pre-training with respect to the image domain one \citep{lit}.

Overall, these results show that the method can effectively zero-shot translate across different domains.

\chapter[Latent Space Translation]{Latent Space Translation via Semantic Alignment}
\label{ch:semantic_alignment}

\chapterpublication[https://github.com/Flegyas/latent-translation]{maiorca2023semantic}{\textbf{NeurIPS 2023}}{Semantic correspondence can be used to efficiently estimate cross-space transformations, mostly in closed form, with no extra training.}

\section{Introduction}

This chapter presents the first \gls{correspondence-coupled} operator (\cref{ch:methodology}) in the thesis: \textbf{affine and orthogonal alignment}, the latter via Procrustes analysis. Given a correspondence $\gls{Corr}$ establishing semantic relationships between observation spaces through paired anchors, a direct transformation $\gls{OperatorCoupled}_\Corr: \mathcal{X} \to \mathcal{Y}$ is constructed that exploits both sides of the correspondence jointly. Unlike $\gls{OperatorParallel}_\Corr$ methods (\cref{ch:relrep,ch:asif}) that transform spaces independently into \glspl{semantic-coordinate}, $\OperatorCoupled_\Corr$ produces outputs in the target's \glslink{extrinsic-factor}{extrinsic} coordinate system, enabling immediate decoder compatibility without auxiliary decoder training.

The main contribution is \textbf{direct translation between representation spaces}, using mostly closed-form algebraic procedures. While relative representations (\cref{ch:relrep}) showed that spaces sharing semantic structure differ primarily by angle-preserving transformations, they require decoders trained on \glspl{semantic-coordinate}. Complementary to inverse relative projection (\cref{ch:inverse-relative-projection})---which enables projecting semantic coordinates back to \glslink{extrinsic-factor}{extrinsic} spaces---this chapter directly estimates the transformation $\OperatorCoupled_\Corr$ between \glslink{extrinsic-factor}{extrinsic} representations, enabling zero-shot stitching of arbitrary pretrained encoders and decoders across different trainings, architectures, and even modalities.

\begin{figure}
 \centering
 \begin{overpic}[trim=0 0cm 0 0cm,clip,width=0.7\linewidth,tics=2]{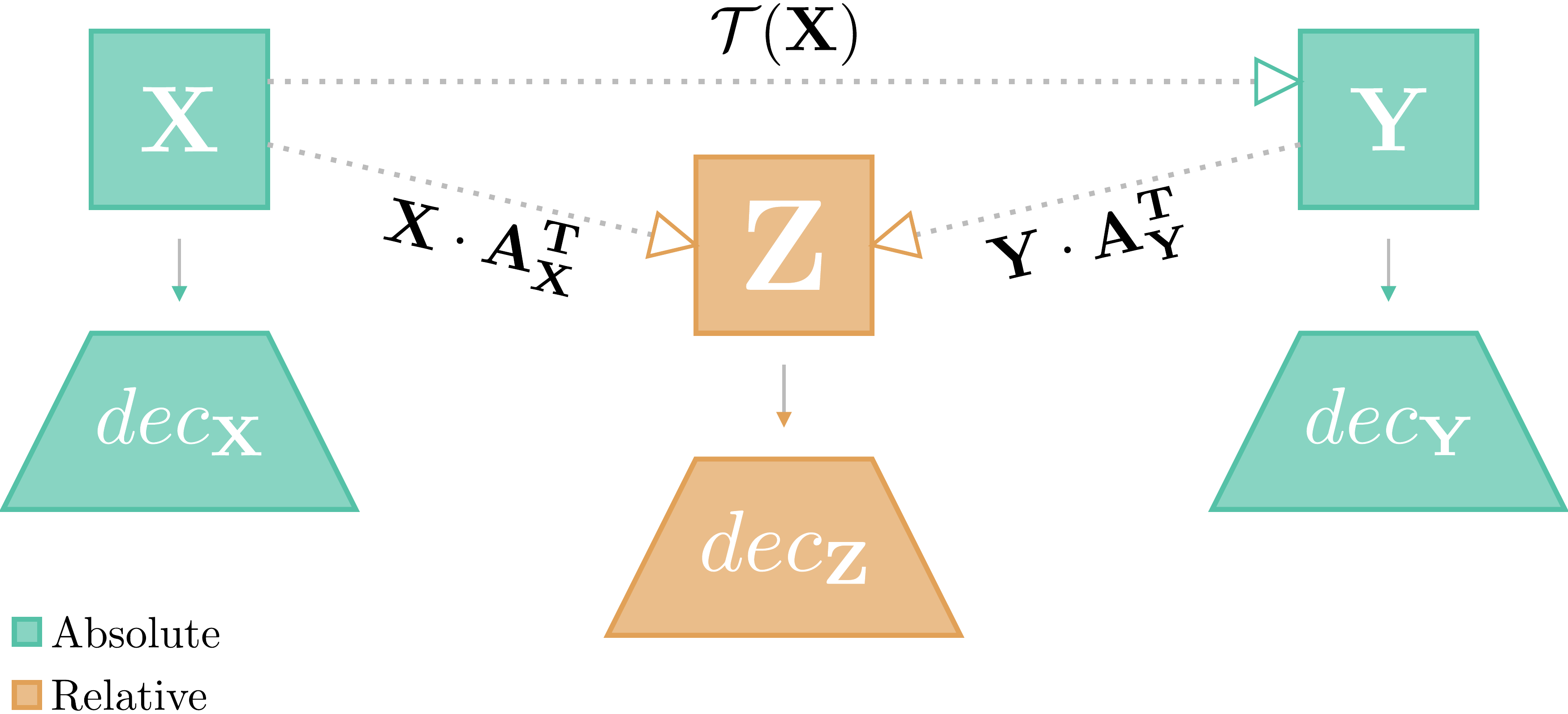}
 \end{overpic}
 \caption{
  Zero-shot stitching of $\mathcal{X}$ and $\mathcal{Y}$ \glslink{extrinsic-factor}{extrinsic} spaces (labelled X and Y in the figure, where Z is the relative space and A\textsubscript{X}, A\textsubscript{Y} are the anchor matrices $\anchmat[X]$, $\anchmat[Y]$; in the edge labels, X and Y are the stacked representations $\mX$, $\mY$, and the superscript T is the transpose ${}^\top$) utilizing relative representations and semantic alignment (the estimation of $\OperatorCoupled_\Corr$, $\mathcal{T}$ in the figure).
  This approach does not require a decoder specifically trained on relative representations ($\Dec$ on the relative space $\R^{\nanch}$, the central decoder in the figure). Instead, it directly translates representation spaces, enabling the use of arbitrarily pretrained decoders (dec\textsubscript{X}, dec\textsubscript{Y} in the figure, i.e.\ the decoders $\Dec[X]$, $\Dec[Y]$) originally trained on \glslink{extrinsic-factor}{extrinsic} spaces.
 }
 \label{fig:semantic:teaser}
\end{figure}

Building on the intuition that representation spaces are related by simple transformations, this chapter demonstrates the effectiveness and applications of \textit{directly translating between different representation spaces}---provided that a partial (and possibly sparse) correspondence between data points is given.
The process of combining different neural networks, pretrained on diverse datasets, modalities, architectures, domains and downstream tasks, proves straightforward.
For instance, this capability enables effective integration of any pretrained text encoder with any image classification head, and vice versa, without requiring additional re-training or assumptions about decoder training (\citealt{moschella2023relative} assume decoders are trained on \glspl{semantic-coordinate}).
The method difference is emphasized in \cref{fig:semantic:teaser}. While zero-shot stitching with correspondence-parallel methods assumes the use of a single decoder specifically trained on \glspl{semantic-coordinate}, the $\OperatorCoupled_\Corr$ approach permits the reuse of decoders originally trained on their native \glslink{extrinsic-factor}{extrinsic} coordinate systems.

Semantic alignment demonstrates that $\OperatorCoupled_\Corr$ operators bridge the gap between modularity and decoder compatibility: by estimating mostly closed-form transformations directly from correspondence, zero-shot stitching across architectures and modalities is achieved without auxiliary decoder training. \cref{ch:latent_functional_maps} extends this approach using spectral methods to exploit functional structure.

\section{Method}

\begin{figure}
 \centering
 \begin{overpic}[trim=0 0cm 0 -1cm,clip,width=1\linewidth,tics=4,]{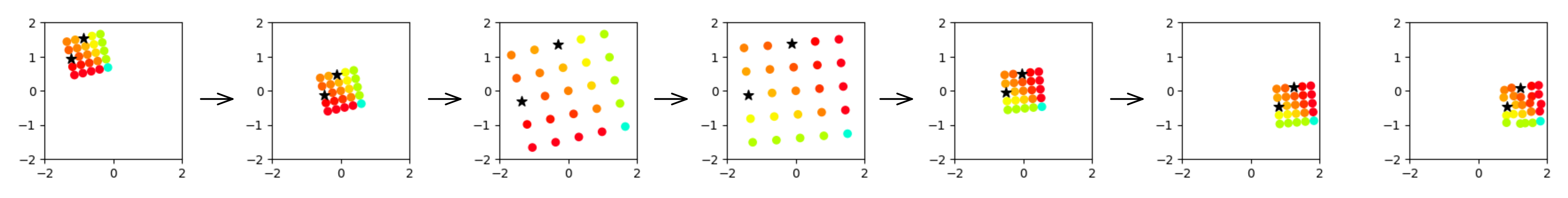}
  \put(4,-1.25){\tiny{ Source $\mathcal{X}$}}
  \put(91,-1.25){\tiny{ Target $\mathcal{Y}$}}
  \put(16,14){\rotatebox{180}{\makebox(25,3){\upbracefill}}}
  \put(23.5,14){\tiny Normalization}

  \put(60,14){\rotatebox{180}{\makebox(25,3){\upbracefill}}}
  \put(67,14){\tiny De-normalization}

  \put(19,-1.25){\tiny Centering}
  \put(34.1,-1.25){\tiny Scaling}
  \put(50,-1.25){\tiny $\OperatorCoupled_\Corr$}
  \put(61.8,-1.25){\tiny De-Scaling}
  \put(75.5,-1.25){\tiny De-Centering}

  \put(85.5,6){$\approx$}
 \end{overpic}
 \caption{Method illustration on a synthetic example. Given a source space $\mathcal{X}$, the steps to translate it to a target $\mathcal{Y}$ are sequentially applied as described in \Cref{sec:translation}. Note that the translation is not perfect due to an arbitrary distortion of the data.}
 \label{fig:method}
\end{figure}

\subsection{From Correspondence-Parallel to Correspondence-Coupled}

Recall from \cref{ch:relrep} that correspondence-parallel methods transform spaces independently using their respective sides of the correspondence, producing outputs in \glspl{semantic-coordinate}. While this enables zero-shot communication between models, it requires decoders trained specifically on \glspl{semantic-coordinate}---a practical limitation.

Semantic alignment eliminates this constraint through a \gls{correspondence-coupled} approach. Given a correspondence $\Corr \subseteq \anchors[X] \times \anchors[Y]$ between paired anchor sets, transformation $\OperatorCoupled_\Corr$ is directly estimated that maps from source to target \glslink{extrinsic-factor}{extrinsic} coordinates. This joint exploitation of both correspondence sides enables immediate decoder reuse without auxiliary training, while relaxing the strict angle-preserving assumption of \gls{relrep} to broader transformation classes obtained via mostly closed-form algorithms.

\subsection{Latent Space Translation}
\label{sec:translation}
Consider two representation spaces $\mathcal{X}$ and $\mathcal{Y}$, with $n$ paired samples stacked as $\mX \in \R^{n \times \dimof{X}}$ and $\mY \in \R^{n \times \dimof{Y}}$. The objective is to estimate the transformation $\OperatorCoupled_\Corr: \mathcal{X} \to \mathcal{Y}$ that translates $\mathcal{X}$ into $\mathcal{Y}$, exploiting the \textbf{semantic alignment} between the two spaces: $\OperatorCoupled_\Corr$ is estimated from the anchor pairs, $\anchmat[Y] \approx \anchmat[X]\mT^\top + \1\vb^\top$, and applied row-wise, $\mY \approx \mX\mT^\top + \1\vb^\top$, with $\mT$ and $\vb$ the parameters of the affine map introduced below. Here $\anchmat[X] \in \R^{\nanch \times \dimof{X}}$ and $\anchmat[Y] \in \R^{\nanch \times \dimof{Y}}$ stack the encodings $\va_j = \Enc[X](a_j)$ and $\va'_j = \Enc[Y](a'_j)$ of the $\nanch$ pairs $(a_j, a'_j) \in \Corr$ in the same row order (shapes before zero-padding); in the fit, $\anchmat[X]$, $\anchmat[Y]$, $\mX$ and $\mY$ denote the pre-processed (zero-padded and standardized) matrices, and the full translation applies $\OperatorCoupled_\Corr$ followed by de-normalization.
The translation process involves two main steps: pre-processing the spaces and estimating the transformation $\OperatorCoupled_\Corr$, as outlined in \cref{fig:method}.

\paragraph{Pre-processing.}
Generally, the two spaces may have different dimensionalities---in those cases, the method zero-pads the smaller one to match the dimension of the other without changing its underlying structure \citep{Williams_etal_2021_shapes}. Moreover, each feature in the encoding is standardized to have zero mean and unit variance (standard scaling) if not otherwise specified, whose statistics are computed only on the anchor sets for both source and target space, to perform the necessary de-normalization.

\paragraph{Estimating $\OperatorCoupled_\Corr$.}
\label{sec:estimatingt}
\citet{moschella2023relative} empirically show that spaces mostly differ by an angle-preserving transformation. Nevertheless, the investigation broadens by considering different ways of obtaining $\OperatorCoupled_\Corr$ to evaluate the robustness of that assumption and the versatility of the approach. The method primarily operates under the assumption that $\OperatorCoupled_\Corr$ can be constrained to encode, at most, an affine transformation: $\OperatorCoupled_\Corr(\vx) = \mT \vx + \vb$.

This general formulation, without additional constraints, corresponds to the \texttt{affine} method in the experiments, and it is optimized via gradient descent. The other transformations are trivially obtained by progressively adding constraints on this one:
\vspace{-0.3pt}
\begin{itemize}
 \item \texttt{linear}. To model a linear transformation, the bias is set to zero, $\vb = \0$, and $\mT$ is fitted by least squares. Here the class of transformations is simplified and the optimization switches from gradient descent to a closed-form procedure.
 \item \texttt{l-ortho}. Additionally, $\mT$ is constrained to an orthogonal $\mR$ to encode an isometry. In this case, $\mR$ is obtained by applying Singular Value Decomposition (SVD) on the corresponding $\mT$ obtained by the \texttt{linear} solution. Through this, the implications of enforcing orthogonality on a transformation that was originally not constrained to be so are examined, in a setting similar to \citet{xing-etal-2015-normalized}.
 \item \texttt{ortho}. Here too $\mT$ is constrained to an orthogonal $\mR$: to obtain the optimal one, Procrustes analysis is applied \citep{schonemann1966generalized,gower1975generalized}.
\end{itemize}

This methodology facilitates efficient and precise zero-shot translation between disparate representation spaces. The transformation $\OperatorCoupled_\Corr$, derived solely from the subset of corresponding points, provides a robust and versatile foundation for model reuse and interoperability in diverse machine learning contexts.

\begin{figure}
 \centering
 \begin{overpic}[trim=-0cm -1cm 0 -0cm, clip,width=0.95\linewidth]{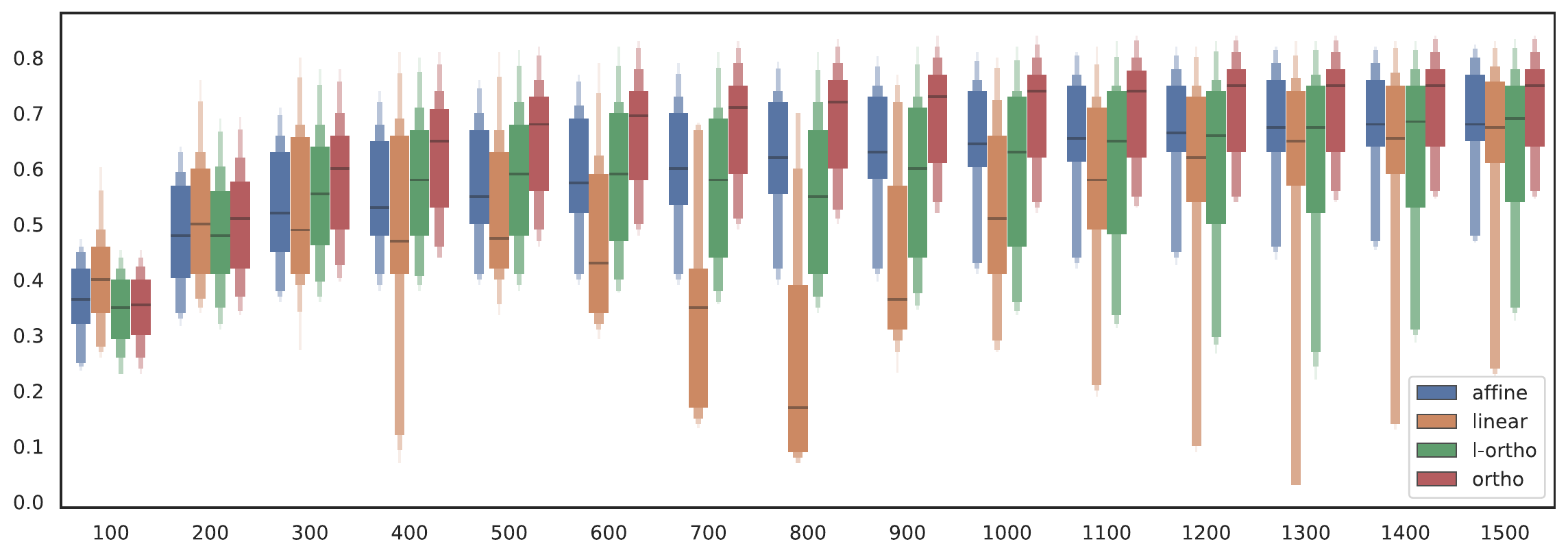}
  \put(-2.5, 15){\rotatebox{90}{Accuracy}}
  \put(42, -.5){Number of anchors}
 \end{overpic}
 \caption{Performance comparison of \texttt{affine}, \texttt{linear}, \texttt{l-ortho}, and \texttt{ortho} at varying number of anchors on classification accuracy. Results on \gls{cifarh} fine-grained.}
 \label{fig:anchors-num}
\end{figure}

\section{Latent Communication via Translation}\label{sec:latent-translation}

This section evaluates the capabilities and effectiveness of the translation method through various scenarios, highlighting its applicability in diverse contexts. Empirical results are presented in three different settings: i) cross-architecture; ii) cross-modality; iii) autoencoding.
In each case, the translation performance of each method for obtaining the transformation $\OperatorCoupled_\Corr$ is evaluated against two baselines: the naive absolute baseline and the relative representation baseline.

\paragraph{Stitching Procedure.}
\label{sec:exp_baseline}
Following the \textit{zero-shot stitching} concept from \citet{moschella2023relative}, independent encoders and decoders (e.g., classifiers, generators) are combined without subsequent training or fine-tuning. This approach does not necessitate a decoder trained on relative representations; instead, the original decoders trained on \glslink{extrinsic-factor}{extrinsic} spaces are directly employed.
Each benchmark follows the same procedure unless otherwise specified: the mean performance is measured over all the possible combinations of (encoder, decoder) for each test set in different settings:
\begin{itemize}
 \item \texttt{no-stitch}. The end-to-end performance of the decoder applied to the original space it was trained on. This is useful to establish an upper-bound in performances;
 \item \texttt{absolute}. The result of using the encodings without any transformation, considered as a probe for any pre-existing compatibility among encodings and, therefore, a lower-bound;
 \item \texttt{relative}. Stitching through relative representations~\citep{moschella2023relative};
 \item \textit{translation}. The results of applying the latent translation method, with the estimation of $\OperatorCoupled_\Corr$ via \texttt{affine}, \texttt{linear}, \texttt{l-ortho} and \texttt{ortho}.
\end{itemize}

In each instance, the same paired anchors (establishing correspondence $\Corr$) are used, uniformly chosen, in a quantity comparable with the dimensionality of the \glslink{extrinsic-factor}{extrinsic} representation.

\subsection{Cross-Architecture}
\label{sec:stitch-cross-architecture}
Firstly, the method is tested in a cross-architecture setting, zero-shot stitching together encodings coming from a variety of pretrained networks and their associated absolute decoders (classifiers). This scenario provides an extensive testing ground and demonstrates robustness across different architectures.

\begin{table}
 \tiny
 \centering
 \caption{Cross-architecture stitching with various methods for estimating $\OperatorCoupled_\Corr$ and applying standard scaling (mean $\pm$ standard deviation over seeds). The stitched decoders are SVMs with a linear kernel. 5 runs for each encoder-decoder pair. (C) and (F) next to \gls{cifarh} indicate, respectively, coarse-grained and fine-grained.}
 \label{tab:stitching:svm}
 \begin{tabular}{lllllllll}
  \toprule
               & \multicolumn{1}{l}{\texttt{Dataset}} & \multicolumn{1}{c}{\texttt{no-stitch}} & \multicolumn{1}{c}{\texttt{absolute}} & \multicolumn{1}{c}{\texttt{relative}} & \multicolumn{1}{c}{\texttt{affine}} & \multicolumn{1}{c}{\texttt{linear}} & \multicolumn{1}{c}{\texttt{l-ortho}} & \multicolumn{1}{c}{\texttt{ortho}} \\
  \midrule
  \multirow{5}{*}{\rotatebox{90}{\tiny \textbf{Vision}}} & \gls{cifart}       & $0.95 \pm 0.03$       & $0.16 \pm 0.22$      & $0.80 \pm 0.22$      & $0.92 \pm 0.05$      & $0.88 \pm 0.11$      & $0.90 \pm 0.09$      & $0.93 \pm 0.04$     \\
               & \gls{cifarh}-\texttt{C}     & $0.85 \pm 0.07$       & $0.11 \pm 0.21$      & $0.54 \pm 0.25$      & $0.78 \pm 0.09$      & $0.73 \pm 0.16$      & $0.77 \pm 0.11$      & $0.81 \pm 0.07$     \\
               & \gls{cifarh}-\texttt{F}     & $0.76 \pm 0.09$       & $0.07 \pm 0.21$      & $0.30 \pm 0.24$      & $0.68 \pm 0.11$      & $0.62 \pm 0.19$      & $0.64 \pm 0.16$      & $0.71 \pm 0.09$     \\
               & \gls{fmnist}      & $0.88 \pm 0.01$       & $0.15 \pm 0.20$      & $0.63 \pm 0.23$      & $0.86 \pm 0.01$      & $0.83 \pm 0.06$      & $0.82 \pm 0.05$      & $0.85 \pm 0.02$     \\
               & \gls{mnist}        & $0.96 \pm 0.01$       & $0.15 \pm 0.21$      & $0.50 \pm 0.22$      & $0.94 \pm 0.01$      & $0.89 \pm 0.08$      & $0.81 \pm 0.11$      & $0.91 \pm 0.02$     \\
  \midrule \multirow{4}{*}{\rotatebox{90}{\textbf{Text}}} & \gls{trec}        & $0.87 \pm 0.12$       & $0.20 \pm 0.06$      & $0.36 \pm 0.13$      & $0.82 \pm 0.12$      & $0.74 \pm 0.25$      & $0.57 \pm 0.25$      & $0.79 \pm 0.11$     \\
               & \ag{}        & $0.73 \pm 0.09$       & $0.25 \pm 0.02$      & $0.39 \pm 0.13$      & $0.65 \pm 0.08$      & $0.62 \pm 0.08$      & $0.61 \pm 0.10$      & $0.66 \pm 0.10$     \\
               & \gls{dbpedia}       & $0.78 \pm 0.23$       & $0.07 \pm 0.01$      & $0.16 \pm 0.10$      & $0.66 \pm 0.24$      & $0.62 \pm 0.23$      & $0.57 \pm 0.23$      & $0.66 \pm 0.22$     \\
               & \imdb{}        & $0.61 \pm 0.04$       & $0.50 \pm 0.01$      & $0.51 \pm 0.02$      & $0.59 \pm 0.04$      & $0.57 \pm 0.04$      & $0.56 \pm 0.03$      & $0.59 \pm 0.04$     \\
  \bottomrule
 \end{tabular}
\end{table}

\begin{table}
 \tiny
 \centering
 \caption{Cross-architecture stitching with various methods for estimating $\OperatorCoupled_\Corr$ and applying $L_2$ normalization (mean $\pm$ standard deviation over seeds). The stitched decoders are SVMs with linear kernel. 5 runs for each encoder-decoder pair. (C) and (F) next to \gls{cifarh} indicate, respectively, coarse-grained and fine-grained. }
 \label{tab:stitching:svm:l2}
 \begin{tabular}{lllllllll}
  \toprule
               & \multicolumn{1}{l}{\texttt{Dataset}} & \multicolumn{1}{c}{\texttt{no-stitch}} & \multicolumn{1}{c}{\texttt{absolute}} & \multicolumn{1}{c}{\texttt{relative}} & \multicolumn{1}{c}{\texttt{affine}} & \multicolumn{1}{c}{\texttt{linear}} & \multicolumn{1}{c}{\texttt{l-ortho}} & \multicolumn{1}{c}{\texttt{ortho}} \\
  \midrule
  \multirow{5}{*}{\rotatebox{90}{\tiny \textbf{Vision}}} & \gls{cifart}       & $0.95 \pm 0.03$       & $0.16 \pm 0.22$      & $0.80 \pm 0.22$      & $0.93 \pm 0.04$      & $0.78 \pm 0.27$      & $0.88 \pm 0.12$      & $0.91 \pm 0.09$     \\
               & \gls{cifarh}-\texttt{C}     & $0.85 \pm 0.07$       & $0.11 \pm 0.21$      & $0.54 \pm 0.25$      & $0.79 \pm 0.07$      & $0.65 \pm 0.25$      & $0.73 \pm 0.17$      & $0.79 \pm 0.10$     \\
               & \gls{cifarh}-\texttt{F}     & $0.76 \pm 0.09$       & $0.07 \pm 0.21$      & $0.30 \pm 0.24$      & $0.69 \pm 0.10$      & $0.52 \pm 0.25$      & $0.62 \pm 0.19$      & $0.68 \pm 0.13$     \\
               & \gls{fmnist}      & $0.88 \pm 0.01$       & $0.15 \pm 0.20$      & $0.63 \pm 0.23$      & $0.86 \pm 0.01$      & $0.65 \pm 0.23$      & $0.83 \pm 0.06$      & $0.84 \pm 0.05$     \\
               & \gls{mnist}        & $0.96 \pm 0.01$       & $0.15 \pm 0.21$      & $0.50 \pm 0.22$      & $0.94 \pm 0.01$      & $0.61 \pm 0.23$      & $0.90 \pm 0.08$      & $0.90 \pm 0.04$     \\
  \midrule \multirow{4}{*}{\rotatebox{90}{\textbf{Text}}} & \gls{trec}        & $0.87 \pm 0.12$       & $0.20 \pm 0.06$      & $0.36 \pm 0.13$      & $0.82 \pm 0.12$      & $0.44 \pm 0.20$      & $0.74 \pm 0.23$      & $0.77 \pm 0.12$     \\
               & \ag{}        & $0.73 \pm 0.09$       & $0.25 \pm 0.02$      & $0.39 \pm 0.13$      & $0.66 \pm 0.08$      & $0.56 \pm 0.10$      & $0.62 \pm 0.08$      & $0.64 \pm 0.10$     \\
               & \gls{dbpedia}       & $0.78 \pm 0.23$       & $0.07 \pm 0.01$      & $0.16 \pm 0.10$      & $0.66 \pm 0.24$      & $0.44 \pm 0.20$      & $0.62 \pm 0.23$      & $0.60 \pm 0.22$     \\
               & \imdb{}        & $0.61 \pm 0.04$       & $0.50 \pm 0.01$      & $0.51 \pm 0.02$      & $0.59 \pm 0.04$      & $0.55 \pm 0.03$      & $0.58 \pm 0.04$      & $0.59 \pm 0.04$     \\
  \bottomrule
 \end{tabular}
\end{table}

\paragraph{Experimental setting.} The evaluation considers a variety of Computer Vision (\gls{mnist} \citep{Lecun1998-we}, \gls{fmnist} \citep{Xiao2017-er}, \gls{cifart}, \gls{cifarh} \citep{Krizhevsky2009-hv}) and Natural Language Processing (\gls{trec} \citep{hovy-etal-2001-toward,li-roth-2002-learning}, \gls{dbpedia} \citep{Auer2007-fb}, \ag{} \citep{zhang2015AGNews}, \imdb{} \citep{maas-EtAl:2011:ACL-HLT2011}) datasets. Across the experiments, 7 language models are used as encoders (uncased and cased BERT \citep{bert}, Electra \citep{electra}, RoBERTa base \citep{roberta}, XLM-RoBERTa base \citep{xlmr}, ALBERT \citep{albert}, and the text encoder of CLIP \citep{clip}), and 7 image encoders (RexNet100 \citep{rexnet}, CSPDarkNet53 \citep{wang2020cspnet}, 4 variations of ViT \citep{dosovitskiy2021an}, and the image encoder of CLIP \citep{clip}), all pretrained and frozen.
For each dataset and for each encoder, an SVM classification head (decoder) is trained on top of their specific encodings.
The standard stitching procedure outlined in \cref{sec:exp_baseline} is then applied and the results are collected.

\paragraph{Result analysis.}
The stitching results are in \cref{tab:stitching:svm}. As expected, the \texttt{absolute} encodings obtain a score comparable to random guessing while also considering fewer encoder combinations out of the possible ones due to the dimensionality mismatch between some of them.
These results show how close the transformation relating to these pretrained encoders is to orthogonal:
i) \texttt{ortho} and \texttt{affine}, the narrowest and the broadest transformation classes considered, are the better-performing translation methods. But while the former is obtained via a simple and efficient closed-form algorithm, the latter is optimized by gradient descent (\cref{sec:estimatingt}).
ii) The \texttt{l-ortho} version improves or has small drops in performances over the \texttt{linear} transformation it is obtained from, suggesting that on most datasets the least-squares solution $\mT$ is almost orthogonal; the exceptions are \gls{trec} and \gls{mnist}, where it drops by 0.17 and 0.08.
Note that these results demonstrate the feasibility of combining pretrained models without the need for retraining or fine-tuning, with drops of at most 0.12 in performances across the board and without any additional assumption on the decoders. The drop is measured between \texttt{no-stitch} and the best translation method on each dataset; the largest is on \gls{dbpedia}, from 0.78 to 0.66.

\paragraph{Sensitivity to Anchor Quantity.}
The number of anchors is an essential parameter in the approach.
\Cref{fig:anchors-num} evaluates how the quantity of these anchors impacts the classification accuracy for this experimental setting. This analysis offers insights into the number of anchors necessary for efficient representation space translation: from about 300 anchors on, \texttt{ortho} obtains the highest median accuracy, while the accuracy of \texttt{linear} drops sharply around 700--800 anchors before recovering.

\paragraph{Role of Scaling.}
The approach is designed to accommodate generic (re)scaling methods as pre-processing steps. Standard scaling shows reliable performance in the experiments, indicating that the scale of the data points is useful in estimating the transformation $\OperatorCoupled_\Corr$.
However, for completeness, $L_2$ normalization is also considered, which is the standard normalization in relative representations. This normalization method generalizes the class of transformations handled by the method and introduces an element of complete scale invariance (see \cref{sec:scale-invariance} for an analysis of scale invariance in neural classifiers). When this level of generalization is introduced, a scale-invariant decoder is required since the norm information is effectively removed. In the relative representation work, this is implicitly accomplished by training a decoder on relative representations. In this setting, since the decoder is not trained, scale invariance is assumed. This investigation exemplifies the flexibility of the approach, capable of adapting to different normalization and pre-processing strategies based on the specific requirements of the task at hand.

The results presented in \cref{tab:stitching:svm:l2}, when compared with \cref{tab:stitching:svm}, indicate a stronger reliance of the \texttt{linear} translation on the information encoded in the norm: it drops by 0.30 on \gls{trec}, 0.28 on \gls{mnist} and 0.18 on both \gls{fmnist} and \gls{dbpedia}, while \texttt{affine} and \texttt{ortho} change by at most 0.06. For the text datasets, this is aligned with existing literature in the \gls{nlp} domain \citep{oyama2023norm}, which suggests that the scale of the encodings contains information (e.g., it is correlated with the token frequency).

These results in diverse scenarios show the flexibility and adaptability of the method, especially its robustness in translating between representation spaces of different dimensionality and domains.

\subsection{Cross-Modality}

This scenario illustrates the applicability of the method in cross-modality settings, translating between representation spaces of different modalities: text and image.

\begin{figure}
 \centering
 \begin{minipage}{.65\columnwidth}
  \begin{overpic}[trim=-1cm -1cm 0 0,width=1\linewidth]{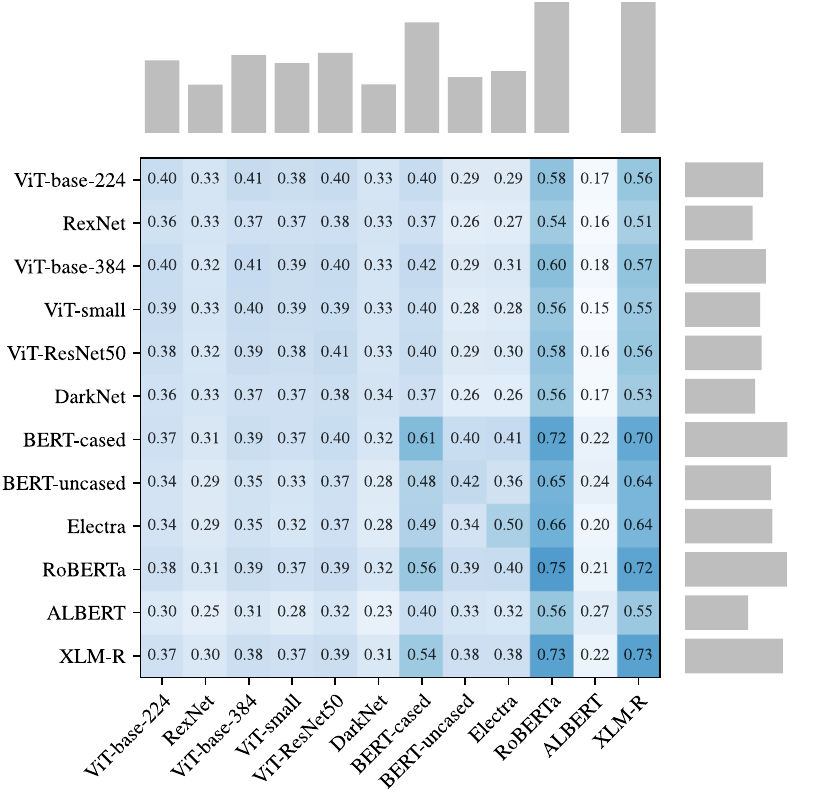}
   \put(0.5, 44){\rotatebox{90}{ Decoder}}
   \put(45, 2){ Encoder}
  \end{overpic}
 \end{minipage}
 \begin{minipage}{.29\columnwidth}
  \scriptsize
  \setlength{\tabcolsep}{2pt}
  \begin{tabular}{lllr}
   \toprule
   {}            & \textbf{Encoder} & \textbf{Score} & \textbf{Scale} \\
   \midrule
   \multirow{6}{*}{\rotatebox{90}{\textbf{Vision}}} & ViT-base-224  & 0.40   & 90.45   \\
                & RexNet   & 0.33   & 13.46   \\
                & ViT-base-384  & 0.41   & 89.66   \\
                & ViT-small  & 0.39   & 50.17   \\
                & ViT-ResNet50  & 0.41   & 32.10   \\
                & DarkNet   & 0.34   & 11.62   \\
   \midrule
   \multirow{6}{*}{\rotatebox{90}{\textbf{Text}}} & BERT-cased  & 0.61   & 15.43   \\
                & BERT-uncased  & 0.42   & 14.54   \\
                & Electra   & 0.50   & 11.94   \\
                & RoBERTa   & 0.75   & 11.06   \\
                & ALBERT   & 0.27   & 32.27   \\
                & XLM-R   & 0.73   & 18.75   \\
   \bottomrule
  \end{tabular}

 \end{minipage}

 \caption{Performance comparison between different encoders and data modalities on the \gls{n24news} multimodal dataset. On the right the accuracy of models trained end-to-end on a single data modality (Score) and their average norm (Scale). On the left the stitching performance between pairs of encoders and decoder. This shows the importance of translating from good encoders, that can even improve unimodal decoder performances. Results obtained with $2000$ anchors and \texttt{ortho}, with an SVM as classification head.}
 \label{fig:cross-modality-stitching}
\end{figure}

\paragraph{Experimental setting.}
The evaluation adopts \gls{n24news} \citep{wang-EtAl:2022:LREC3}, a multimodal news classification dataset that contains both text and associated pictures. The standard encoding procedure is applied to these two features separately, using different pretrained uni-modal encoders. Then, a classification head (an SVM) is trained on top of each one. Lastly, each encoder is zero-shot stitched with a classification head different from its corresponding one, measuring its classification accuracy, without further training or fine-tuning.

\paragraph{Result analysis.}
The discrepancy in the mean accuracy represented by the marginal bar plots in \cref{fig:cross-modality-stitching} is a signal that can be used to identify spaces more suited to be \textit{decoded into} and the ones that are stronger in \textit{encoding from}. In fact, three of the language models (BERT-cased, RoBERTa and XLM-R), used as source space for the translation, exhibit stronger performance than the vision encoders. This behavior may relate to the higher generality of the text domain data used during pre-training with respect to the image domain one \citep{lit}. A further finding in this setting is the improvement in classification performance when a modality-specific classifier trained on images is fed zero-shot with corresponding text encodings translated to the image domain via the method. This result underlines the significance of a good encoder and demonstrates the broad applicability of the technique. In practice, this means image classifiers can be applied on textual data, and vice versa.

These results show that the method: i) obtains effective zero-shot translation over different modalities; ii) improves unimodal decoders when translating from a better encoder than the one it was trained on.

\subsection{Autoencoding}

\begin{figure}

 \hfill \begin{overpic}[trim=-0.27cm 0cm 0cm 0cm, clip,width=.96\linewidth,]{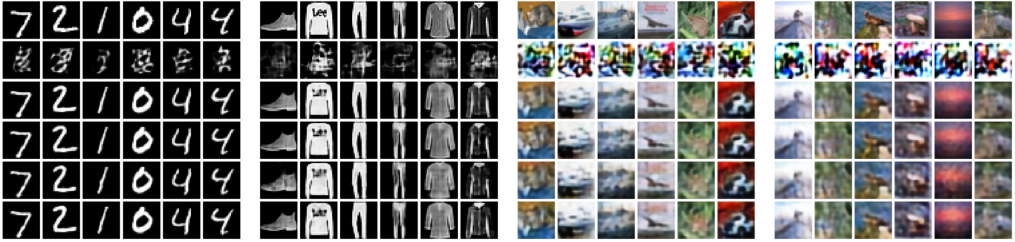}
  \put(.1, 20.7){\rotatebox{0}{\tiny \texttt{S}}}
  \put(-1.5, 16.7){\rotatebox{0}{\tiny \texttt{Abs.}}}
  \put(-3.4, 13){\rotatebox{0}{\tiny \texttt{affine}}}
  \put(-3.4, 9){\rotatebox{0}{\tiny \texttt{linear}}}
  \put(-4.25, 5.2){\rotatebox{0}{\tiny \texttt{l-ortho}}}
  \put(-2.75, 1.5){\rotatebox{0}{\tiny \texttt{ortho}}}
 \end{overpic}

 \caption{Reconstruction examples grouped by dataset (from left to right: \gls{mnist}, \gls{fmnist}, \gls{cifart}, \gls{cifarh}). Each column is a different image, from top to bottom: original image, \texttt{absolute} stitching, \texttt{affine} stitching, \texttt{linear} stitching, \texttt{l-ortho} stitching, and \texttt{ortho} stitching. No additional normalization applied on the decoder part.}
 \label{fig:aes-decoders-without-l2-norm}
\end{figure}

In this setting, the method is applied to align representation spaces of different trainings of the same autoencoder. The novelty of this scenario lies in the generation setting itself, as most prior works primarily focus on classification tasks.
One key observation of \citet{cannistraci2024bricks}, presented in \cref{ch:product-invariances}, is that the \textit{task} at hand (e.g., classification, generation) defines a certain \textit{class of transformations} (e.g. rotations) which act among the representation spaces. Restricting the search for the transformation to the right class is fundamental to guarantee optimal performance and efficiency.

\paragraph{Experimental setting.}
Four datasets are utilized for these experiments: \gls{mnist} \citep{Lecun1998-we}, \gls{fmnist} \citep{Xiao2017-er}, and \gls{cifart} and \gls{cifarh} \citep{Krizhevsky2009-hv}. For each dataset, two standard \gls{cnn}-based autoencoders are trained, with convolutions in the encoder and deconvolutions in the decoder; please refer to the appendix of \citet{maiorca2023semantic} for further implementation details. The two autoencoders are identical in structure, differing only in the random seed used for weight initialization and data shuffling. To perform zero-shot stitching, each data point is first translated from the representation space of the first encoder to the representation space of the second using $1000$ anchors. The second decoder is then applied to the translated data, without any further training or fine-tuning.

\paragraph{Result analysis.}

\begin{table}
 \small
 \centering
 \caption{Zero-shot stitching for generation with various methods for estimating $\OperatorCoupled_\Corr$. Standard scaling used as normalization and the stitched decoders do not have any additional normalization. Reported metrics: latent cosine similarity (\textit{lcos}) and MSE (\textit{lmse}) between the target encoding and the translated one, and reconstruction MSE (\textit{rmse}) between the input and the output. 1000 anchors used on 500-dimensional spaces.}
 \label{table:aes-decoder-without-norm-1000-anchors}
 \setlength{\tabcolsep}{4pt}
 \begin{tabular}{lrrrrrrrrrrrr}
  \toprule
      & \multicolumn{3}{c}{ \gls{mnist} } & \multicolumn{3}{c}{ \gls{fmnist} } & \multicolumn{3}{c}{ \gls{cifart} } & \multicolumn{3}{c}{ \gls{cifarh}}                             \\
  \cmidrule(lr){2-4} \cmidrule(lr){5-7} \cmidrule(lr){8-10} \cmidrule(lr){11-13}
      & \emph{lcos}      & \emph{lmse}      & \emph{rmse}      & \emph{lcos}      & \emph{lmse} & \emph{rmse} & \emph{lcos} & \emph{lmse} & \emph{rmse} & \emph{lcos} & \emph{lmse} & \emph{rmse} \\
  \midrule
  \texttt{absolute} & 0.09        & 0.27        & 0.14        & 0.17        & 0.23  & 0.23  & 0.30  & 0.29  & 0.34  & 0.34  & 0.53  & 0.40  \\
  \texttt{affine} & 0.94        & 0.08        & 0.02        & 0.94        & 0.06  & 0.03  & 0.96  & 0.03  & 0.05  & 0.96  & 0.04  & 0.05  \\
  \texttt{linear} & 0.92        & 0.09        & 0.02        & 0.93        & 0.07  & 0.04  & 0.94  & 0.03  & 0.05  & 0.94  & 0.04  & 0.06  \\
  \texttt{l-ortho} & 0.79        & 0.14        & 0.02        & 0.78        & 0.12  & 0.05  & 0.85  & 0.05  & 0.06  & 0.84  & 0.07  & 0.07  \\
  \texttt{ortho} & 0.90        & 0.10        & 0.02        & 0.90        & 0.08  & 0.04  & 0.94  & 0.03  & 0.06  & 0.93  & 0.04  & 0.06  \\
  \bottomrule
 \end{tabular}
\end{table}

This experiment analyzes the alignment of representation spaces in different training regimens of the same autoencoder.
The performance evaluation, as shown in \cref{table:aes-decoder-without-norm-1000-anchors}, demonstrates that all methods \texttt{affine}, \texttt{linear}, \texttt{l-ortho}, and \texttt{ortho} yield satisfactory results. Moreover, qualitative results depicted in \cref{fig:aes-decoders-without-l2-norm} reveal minimal visual differences in the stitching outcomes across various datasets using different methods.
Since \texttt{affine} reaches a higher latent cosine similarity than \texttt{ortho} on every dataset (e.g., 0.94 against 0.90 on \gls{mnist}), these results suggest that the representation spaces of image autoencoders are not exclusively correlated by orthogonal transformations.
Therefore, further research is warranted to explore and model the specific class of transformations that govern the correlation between neural networks during image autoencoding to constrain and enhance their approximation.

Overall, these results, combined with \citet{cannistraci2024bricks} and \cref{sec:stitch-cross-architecture}, confirm that representation spaces in image autoencoders trained end-to-end are related by a class of transformations larger than orthogonal transformations.

\chapter{Latent Functional Maps}
\label{ch:latent_functional_maps}
\label{ch:lfm}

\chapterpublication[https://github.com/marc0git/lfmap]{fumero2024latent}{\textbf{NeurIPS 2024}}{Representing correspondence as a functional operator on Laplacian eigenbases yields a compact spectral object that serves to compare spaces, expand correspondences, and transfer information.}

\section{Introduction}

This chapter introduces latent functional maps (LFM), a \gls{correspondence-coupled} operator ($\gls{OperatorCoupled}_{\gls{Corr}}$) that elevates the problem of aligning representations from manipulating individual points to transforming entire function spaces. As established in \cref{ch:methodology}, correspondence-coupled operators use both sides of a correspondence jointly to construct transformations producing outputs in target-native coordinates compatible with pretrained decoders. Where \gls{correspondence-parallel} methods like \gls{relrep} (\cref{ch:relrep}) transform spaces independently using angular structure, and correspondence-coupled methods like Semantic Alignment (\cref{ch:semantic_alignment}) or Inverse Projection (\cref{ch:inverse-relative-projection}) use paired samples directly, LFM exploits the spectral geometry of the representation manifolds themselves. The approach treats \gls{correspondence-concept} as a continuous linear operator $\mC$ that maps between function spaces defined over each manifold.

The technical mechanism builds on principles from geometric correspondence \citep{ovsjanikov2012fmap}: approximate each representation manifold through a k-nearest-neighbor (k-NN) graph, compute its Laplacian eigenbasis, then represent correspondence as a compact matrix $\mC$ operating on spectral coefficients. This functional lifting achieves three objectives simultaneously. First, it compresses correspondence into low-dimensional spectral space ($k \ll n$ eigenvectors). Second, eigendecomposition acts as a natural noise filter: high-frequency components, which tend to carry noise, occupy higher eigenvectors that can be truncated. Third, the structure of $\mC$ itself becomes interpretable: orthogonality reveals isometric relationships between spaces, while the off-diagonal structure of $\mC^\top\mC$ measures departure from isometry and the first eigenvector of $\mC^\top\mC$ localizes regions of geometric distortion.

\begin{figure}
  \centering
  \begin{overpic}[width=\linewidth,trim = {0 2cm 0 2cm}, clip
]{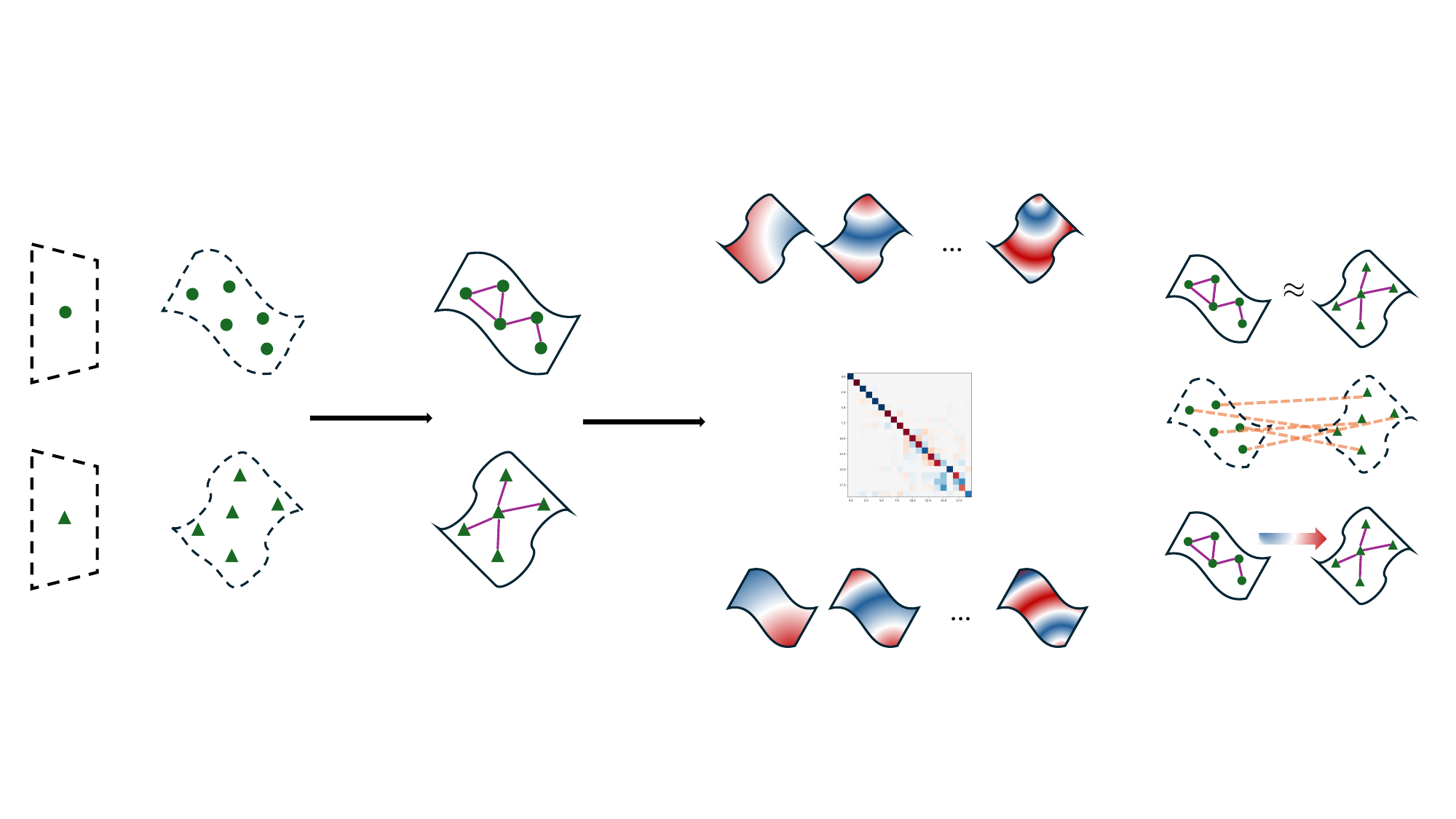}%
   \put(17,32){$\mathcal{M}$}
  \put(12,8){$\mathcal{N}$}
  \put(26 ,35){\footnotesize{$\mathcal{G}_{\mathcal{X}} = \mathrm{kNN}(\mathcal{V}_{\mathcal{X}},\operatorname{dist})$}}
  \put(26,7){\footnotesize{$\mathcal{G}_{\mathcal{Y}} = \mathrm{kNN}(\mathcal{V}_{\mathcal{Y}},\operatorname{dist})$}}
  \put(52,38.5) {\small $\phi_1^{\mathcal{X}}$}
  \put(59,38.5) {\small $\phi_2^{\mathcal{X}}$}
  \put(71,38.5) {\small $\phi_k^{\mathcal{X}}$}
  \put(52,3.) {\small $\phi_1^{\mathcal{Y}}$}
  \put(59,3.) {\small $\phi_2^{\mathcal{Y}}$}
  \put(71,3.) {\small $\phi_k^{\mathcal{Y}}$}%
 \put(52,26){\small $\mC \in \R^{k\times k}$}
  \put(98,28){\emph{(i)}}
  \put(98,21){\emph{(ii)}}
  \put(98,13){\emph{(iii)}}
 \end{overpic}\caption{\label{fig:lfm_teaser}\textbf{Framework overview:} Given two spaces $\mathcal{X}$, $\mathcal{Y}$, their samples lie on manifolds $\mathcal{M}$, $\mathcal{N}$, which are approximated with k-NN graphs $\mathcal{G}_{\mathcal{X}}$, $\mathcal{G}_{\mathcal{Y}}$. A \emph{latent functional map} $\mC$ is optimized between the eigenbases of operators defined on the graphs. This map is a mapping between functions defined on the two manifolds and can be used for (i) comparing representational spaces, (ii) solving correspondence problems, and (iii) transferring information between the spaces.}
\end{figure}

This functional perspective yields a versatile framework. The same spectral operator $\mC$ is a similarity measure between spaces (through analysis of its orthogonality), enables correspondence discovery from few or zero seed anchors (through constrained optimization), and provides efficient information transfer (through spectral coefficient mapping). \cref{fig:lfm_teaser} illustrates how correspondence between manifolds $\mathcal{M}$ and $\mathcal{N}$ lifts to a structured operator in function space.

These three capabilities make LFM versatile: unlike methods requiring explicit anchors (\gls{relrep}) or shared observations (Semantic Alignment), LFM operates with few or zero seed correspondences while simultaneously providing similarity measurement, correspondence discovery, and information transfer in a unified framework, validated across stitching, retrieval, and similarity measurement on vision and language modalities.

\paragraph{Relationship to Other Methods.} While \gls{relrep} (\cref{ch:relrep}) also produces structured representations---angular coordinates relative to anchors---LFM operates on the spectral decomposition of the manifold Laplacian, revealing frequency content and smooth eigenfunctions. This spectral perspective complements the scale invariance principle explored in \cref{ch:inverse-relative-projection}, since the k-NN graph (and, with a scale-invariant distance, its weights) is unchanged by a global rescaling. The method's ability to discover correspondences with minimal supervision distinguishes it from Semantic Alignment (\cref{ch:semantic_alignment}), which assumes correspondence is given, and from Inverse Projection (\cref{ch:inverse-relative-projection}), which bridges the parallel and coupled operator families through inversion but requires known anchor sets.

\section{Related Work}

\paragraph{Functional Maps and Spectral Methods.} The approach derives from the functional maps framework for smooth manifolds \citep{ovsjanikov2012fmap}, extended with regularization \citep{nogneng2017informative} and refinement techniques \citep{melzi2019zoomout} for 3D shape correspondence. Extensions to graph domains \citep{wang2019functional,GRASP,pegoraro2023spectral} demonstrated that functional maps retain advantageous properties when Laplace bases are computed on graph structures. This chapter extends these principles to neural latent spaces. For broader context on representation alignment and spectral methods in deep learning, refer to \cref{ch:background}.

\section{\label{sec:lfm_background} Method}

This section presents the technical details of latent functional maps as a \glslink{correspondence-coupled}{corre\-spon\-dence-coupled} operator in the framework established in \cref{ch:methodology}. The method uses both sides of correspondence jointly---paired observations and their relationships---to construct a spectral transformation between function spaces.

\subsection{Background}
This section provides the basic notions to understand the framework of functional maps applied to manifolds. For a comprehensive overview, refer to \citet{ovsjanikov2016computing}.

Consider two manifolds $\mathcal{M}$ and $\mathcal{N}$ equipped with a basis $\mPhi_{\mathcal{M}}$ (respectively $\mPhi_{\mathcal{N}}$). Any squared integrable function $f: \mathcal{M} \to \R$ can be represented as a linear combination of basis functions $\mPhi_{\mathcal{M}}$: $f = \sum_i \hat{f}_i \phi_i^{\mathcal{M}} = \mPhi_{\mathcal{M}} \hat{f}$. Given a bijective correspondence $\Corr : \mathcal{M} \to \mathcal{N}$ between points on these manifolds (the correspondence of \cref{ch:methodology}, carried to the representations by the encoders and assumed one-to-one, written here as a map), for any real-valued function $f: \mathcal{M} \to \R$, one can construct a corresponding function $g: \mathcal{N} \to \R$ such that $g = f \circ \Corr^{-1}$. In other words, the correspondence $\Corr$ defines a mapping between two function spaces, the push-forward $\Corr_\sharp : \mathcal{F}(\mathcal{M}) \to \mathcal{F}(\mathcal{N})$, $\Corr_\sharp(f) = f \circ \Corr^{-1}$. Such a mapping is \textit{linear} \citep{ovsjanikov2012fmap} and can be represented as a matrix $\mC$ such that for any function $f$ represented as a vector of coefficients $\hat{f}$, the coefficient vector of $\Corr_\sharp(f)$ is $\mC\hat{f}$, i.e.\ $\Corr_\sharp(\mPhi_{\mathcal{M}}\hat{f}) = \mPhi_{\mathcal{N}}\mC\hat{f}$.

The functional representation is particularly well-suited for map inference (i.e., constrained optimization problems). When the underlying map $\Corr$ (and by extension the matrix $\mC$) is unknown, many natural constraints on the map become linear constraints in its functional representation. In practice, the simplest method for recovering an unknown functional map is to solve the following optimization problem:
\begin{equation}
 \argmin_{\mC} \| \mC \hat{\mF}_{\mathcal{M}} - \hat{\mF}_{\mathcal{N}} \|_F^2 + \mathcal{R}(\mC)
 \label{eq:lfm_fmap_opt}
\end{equation}
where $\hat{\mF}_{\mathcal{M}}$ and $\hat{\mF}_{\mathcal{N}}$ are sets of corresponding functions, denoted as \emph{descriptors}, expressed in the bases on $\mathcal{M}$ and $\mathcal{N}$, respectively, and $\mathcal{R}(\mC)$\glsadd{Reg} represents additional constraints deriving from the properties of the matrix $\mC$ \citep{ovsjanikov2016computing}. When the manifolds $\mathcal{M}$ and $\mathcal{N}$ are approximately isometric and the descriptors are well-preserved by the (unknown) map, this procedure provides a good approximation of the underlying map.
In cases where the correspondence $\Corr$ is encoded as a permutation matrix $\mP$, the functional map can be retrieved as $\mC = \mPhi_{\mathcal{N}}^{+} \mP \mPhi_{\mathcal{M}}$ where $\mPhi_{\mathcal{M}}$ and $\mPhi_{\mathcal{N}}$ are the bases of the functional spaces $\mathcal{F}(\mathcal{M})$ and $\mathcal{F}(\mathcal{N})$, respectively, and $(\cdot)^{+}$ denotes the Moore-Penrose pseudo-inverse.

\subsection{Latent Functional Maps}

\subsubsection{\label{sec:lfm_setting}Setting}
Consider deep neural networks $F := f_L\circ\dots\circ f_1$ where each layer $f_\ell$ is associated to a representational space $\mathcal{X}_\ell$ corresponding to the image of $f_\ell$. In the following, the subscript $\ell$ is dropped when the layer is clear from context. Elements $\vx \in \mathcal{X}$ are sampled from a latent manifold $\mathcal{M}$. For pairs of spaces $(\mathcal{X},\mathcal{Y})$ and corresponding manifolds $\mathcal{M} \subset \mathcal{X}$\glsadd{Manifold}, $\mathcal{N} \subset \mathcal{Y}$, the objective is to characterize their relation by mapping the space of functions $\mathcal{F}(\mathcal{M})$ to $\mathcal{F}(\mathcal{N})$. The framework comprises three steps: (i) representing $\mathcal{M}$ from a sample $\mathcal{V}_{\mathcal{X}} = \{\vx_1, \dots, \vx_n\} \subset \mathcal{X}$ (and likewise $\mathcal{V}_{\mathcal{Y}} = \{\vy_1, \dots, \vy_n\} \subset \mathcal{Y}$) by building a graph in the latent space $\mathcal{X}$, (ii) encoding known preserved quantities between the two spaces by defining descriptor functions for each domain, (iii) optimizing for the latent functional map between $\mathcal{M}$ and $\mathcal{N}$. \cref{fig:lfm_teaser} provides an illustrative description.
\paragraph{Building the graph.}
To exploit the geometry of the underlying manifold, we model the latent space of a neural network building a symmetric k-nearest neighbor (k-NN) graph \citep{MAIER20091749}. Given the samples $\mathcal{V}_{\mathcal{X}}$, we construct an undirected weighted graph $\mathcal{G}_{\mathcal{X}} = (\mathcal{V}_{\mathcal{X}}, \mathcal{E}_{\mathcal{X}}, \mW_{\mathcal{X}})$ with nodes $\mathcal{V}_{\mathcal{X}}$, edges $\mathcal{E}_{\mathcal{X}}$, and weight matrix $\mW_{\mathcal{X}}$. The weight matrix is determined by a distance function $\operatorname{dist}: \mathcal{X} \times \mathcal{X} \to \R$, the neighbourhood size $k_{\mathrm{nn}}$ and a weighting function $\omega$ (below).
Suitable choices for $\operatorname{dist}$ include the $L_2$ metric or the angular distance.
Edges $\mathcal{E}_{\mathcal{X}}$ are defined as
\[
 \mathcal{E}_{\mathcal{X}} = \{(\vx_i, \vx_j) \in \mathcal{V}_{\mathcal{X}} \times \mathcal{V}_{\mathcal{X}} \mid \vx_i \sim_{k_{\mathrm{nn}}} \vx_j \text{ or } \vx_j \sim_{k_{\mathrm{nn}}} \vx_i\},
\]
where $\vx_i \sim_{k_{\mathrm{nn}}} \vx_j$ indicates that $\vx_j$ is among the $k_{\mathrm{nn}}$ nearest neighbors of $\vx_i$ (the subscript $k_{\mathrm{nn}}$ distinguishes the number of neighbors from the number $k$ of eigenvectors used below). The weight matrix $\mW_{\mathcal{X}} \in \R^{n \times n}_{\geq 0}$ assigns a weight $\omega(\vx_i, \vx_j)$, a monotonic function of $\operatorname{dist}(\vx_i,\vx_j)$, to each edge $(\vx_i, \vx_j) \in \mathcal{E}_{\mathcal{X}}$, and $(\mW_{\mathcal{X}})_{i, j} = 0$ otherwise.
Next, we define the associated weighted graph Laplacian operator $\mL_{\mathcal{X}} = \mI - \mD_{\mathcal{X}}^{-1/2} \mW_{\mathcal{X}} \mD_{\mathcal{X}}^{-1/2}$, where $\mD_{\mathcal{X}}$ is the diagonal degree matrix with entries $(\mD_{\mathcal{X}})_{i, i} = \sum_{j=1}^{n} (\mW_{\mathcal{X}})_{i, j}$. $\mL_{\mathcal{X}}$ is a positive semi-definite, self-adjoint operator \citep{von2007tutorial}, therefore, it admits an eigendecomposition $\mL_{\mathcal{X}} = \mPhi_{\mathcal{X}} \mLambda_{\mathcal{X}} \mPhi_{\mathcal{X}}^\top$, where $\mLambda_{\mathcal{X}}$ is a diagonal matrix containing the eigenvalues, and $\mPhi_{\mathcal{X}}$ is a matrix whose columns are the corresponding eigenvectors. The eigenvectors form an orthonormal basis for the space of functions defined on the graph nodes (i.e., $\mPhi_{\mathcal{X}}^\top \mPhi_{\mathcal{X}} = \mI$).

\paragraph{Choice of descriptors.}
To define the alignment problem between a pair of spaces $\mathcal{X}, \mathcal{Y}$ we will introduce the notion of \emph{descriptor function}. We define as descriptor function any real-valued function $f: \mathcal{V}_{\mathcal{X}} \to \R$ defined on the nodes of the graph (functions on the $n$ nodes are written in italic although they are vectors in $\R^n$; so are their spectral coefficient vectors, e.g.\ $\hat{f}$). Informally descriptors should encode the information which is \emph{shared} between $\mathcal{X}$ and $\mathcal{Y}$, either explicitly or implicitly. We categorize descriptor functions into supervised, weakly supervised and unsupervised descriptors. For supervised descriptors, we assume access to a partial correspondence between the anchor sets $\anchors[X] \subset \Obs[X]$ and $\anchors[Y] \subset \Obs[Y]$, a bijection $\Corr^{\mathrm{sup}}: \anchors[X] \to \anchors[Y]$ with embeddings $\va_j = \Enc[X](a_j) \in \mathcal{V}_{\mathcal{X}}$ and $\va'_j = \Enc[Y](\Corr^{\mathrm{sup}}(a_j)) \in \mathcal{V}_{\mathcal{Y}}$. Then descriptors take the form of distance functions $\delta_{\mathcal{X}}(\vx, \va_j) \in \R_{\geq 0}$, $\delta_{\mathcal{Y}}(\vy, \va'_j) \in \R_{\geq 0}$. For example, the geodesic distance (shortest path distance on the k-NN graph) from each node of the graph to the nodes in the anchor set is a supervised descriptor. In the case of weakly supervised descriptor, we assume to have access to a many-to-many relation $\Corr^{\mathrm{weak}} \subseteq \Obs[X] \times \Obs[Y]$ (for example, shared class labels), written as an indicator on pairs of samples. Unsupervised descriptors are quantities that depend on the topology and the geometry of the graph itself and they don't rely on any pre-given correspondence or many-to-many relation. An example of this is the Heat Kernel Signature \citep{sun2009hks} node descriptor, based on heat diffusion over the manifold.

\paragraph{Computing LFMs.}
The optimization problem to compute a latent functional map between $\mathcal{X}$ and $\mathcal{Y}$ proceeds as follows. Each space is modeled using a subset of training samples $\mathcal{V}_{\mathcal{X}} = \{\vx_1, \dots, \vx_n\}$ and $\mathcal{V}_{\mathcal{Y}} = \{\vy_1, \dots, \vy_n\}$, from which k-NN graphs $\mathcal{G}_{\mathcal{X}}$ and $\mathcal{G}_{\mathcal{Y}}$ are constructed. For each graph, the graph Laplacian $\mL_{\mathcal{X}}$ (respectively $\mL_{\mathcal{Y}}$) is computed, yielding the first $k$ eigenvalues $\mLambda_{\mathcal{X}}$ and eigenvectors $\mPhi_{\mathcal{X}} = [\phi_1^{\mathcal{X}}, \dots, \phi_k^{\mathcal{X}}]$ (respectively $\mLambda_{\mathcal{Y}}$, $\mPhi_{\mathcal{Y}}$), which form the basis of the function space defined on the latent spaces.

Given $n_{\mathrm{f}}$ descriptors per space, $\mF_{\mathcal{X}} = [f^{\mathcal{X}}_1, \dots, f^{\mathcal{X}}_{n_{\mathrm{f}}}]$ and $\mF_{\mathcal{Y}} = [f^{\mathcal{Y}}_1, \dots, f^{\mathcal{Y}}_{n_{\mathrm{f}}}]$, the optimization problem defined in \cref{eq:lfm_fmap_opt} incorporates two regularizers which enforce commutativity for the Laplacian and the descriptors, respectively, with the map $\mC$:
\begin{equation}
 \argmin_{\mC} \| \mC \hat{\mF}_{\mathcal{X}}- \hat{\mF}_{\mathcal{Y}} \|_F^2 +
 \alpha \mathcal{R}_{\mL}(\mC) +
 \beta \mathcal{R}_{f}(\mC)
 \label{eq:lfm_fmap_opt_ls}
\end{equation}
where $\hat{\mF}_{\mathcal{X}} = \mPhi_{\mathcal{X}}^\top \mF_{\mathcal{X}}$ (respectively $\hat{\mF}_{\mathcal{Y}} = \mPhi_{\mathcal{Y}}^\top \mF_{\mathcal{Y}}$) are the spectral coefficients of the functions $\mF_{\mathcal{X}}$ (respectively $\mF_{\mathcal{Y}}$),
$\mathcal{R}_{\mL}$ and $\mathcal{R}_{f}$ are the Laplacian and descriptor operator commutativity regularizers respectively, weighted by $\alpha,\beta\ge 0$, akin to \citet{nogneng2017informative}.
Once the optimization problem defined in \cref{eq:lfm_fmap_opt_ls} is solved, the resulting functional map $\mC$ is refined using the spectral upsampling algorithm proposed by \citet{melzi2019zoomout}.

\subsection{LFMs as a similarity measure} \label{sec:lfm_similarity}
Once computed, the functional map $\mC$ can be used as a measure of similarity between spaces. The reason is that for isometric transformations between manifolds, the functional map is volume preserving (see Thm 5.1 in \citealp{ovsjanikov2012fmap}), and this is manifested in orthogonal $\mC$. By defining the inner product between functions $h_1,h_2 \in \mathcal{F}(\mathcal{M})$ as $\langle h_1,h_2 \rangle_{\mathcal{M}} =\int_\mathcal{M} h_1(\vx) h_2(\vx) \, \mathrm{d}\mu(\vx)$, with $\mu$ the measure on $\mathcal{M}$, the map preserves inner products, $\langle \Corr_\sharp h_1, \Corr_\sharp h_2\rangle_{\mathcal{N}} = \langle h_1, h_2\rangle_{\mathcal{M}}$, i.e.\ $\langle \mC\hat{h}_1, \mC\hat{h}_2\rangle = \langle \hat{h}_1,\hat{h}_2\rangle$, exactly when it preserves local area, i.e.\ when $\mC$ is orthogonal, where $\hat{h}$ denotes the functional representation of $h$. In other words, when the transformation between the two manifolds is an isometry, the matrix $\mC^\top\mC$ will be diagonal. By measuring the ratio between the norm of the off-diagonal elements of $\mC^\top\mC$ and the norm of the full matrix, we can define a measure of similarity $\simf_{\mathrm{LFM}}(\mathcal{X},\mathcal{Y})= 1-\frac{\|\operatorname{off}(\mC^\top\mC)\|_F^2}{\|\mC^\top\mC\|_F^2}$.
Furthermore, this quantity is interpretable; the first eigenvector of $\mC^\top\mC$ can act as a signal to localize the region where the map has higher distortion \citep{ovsjanikov2013analysis}, as we show in \cref{fig:lfm_cka_distorsion}.
\subsection{Transferring information with LFM}
The map $\mC$ computed between two latent spaces can be utilized in various ways to transfer information from one space to the other. In this chapter, we focus on two methods: (i) Expressing arbitrary points in the latent space as distance function on the graph and transferring them through the functional domain; (ii) Obtaining a point-to-point correspondence between the representational spaces from the LFM, starting from none to few known pairs, and using off-the-shelf methods to learn a transformation between the spaces. Additional strategies could be explored in future work.

\paragraph{Space of Functional Coefficients.}
The space of functional (or spectral) coefficients offers an alternative representation for points in the latent space $\mathcal{X}$. Using the equation $\hat{f}_{\mathcal{X}} = \mPhi_{\mathcal{X}}^\top f_{\mathcal{X}}$, any function $f_{\mathcal{X}} \in \mathcal{F}(\mathcal{G}_{\mathcal{X}})$ is represented, up to truncation to $k$ eigenvectors, by its functional coefficients $\hat{f}_{\mathcal{X}}$. We use this property to represent any point $\vx \in \mathcal{X}$ as a distance function $f_{\vx} \in \mathcal{F}(\mathcal{G}_{\mathcal{X}})$ from the set of points $\mathcal{V}_{\mathcal{X}}$, which correspond to the nodes of the graph $\mathcal{G}_{\mathcal{X}}$.
The functional map $\mC$ between two latent spaces $\mathcal{X}$ and $\mathcal{Y}$ aligns their functional representations, enabling the transfer of any function from the first space to the second. This functional alignment can be used similarly to the method proposed by \citet{moschella2023relative} to establish a shared space where the representational spaces $\mathcal{X}$ and $\mathcal{Y}$ are aligned.

\paragraph{Finding correspondences.}
As explained in \cref{sec:lfm_background}, the functional map $\mC$ represents the bijection $\Corr$ in a functional form. \citet{ovsjanikov2012fmap} demonstrated that this bijection can be retrieved as a point-to-point map by finding the nearest neighbor for each row of $\mPhi_{\mathcal{Y}} \mC$ in $\mPhi_{\mathcal{X}}$. This process can be efficiently implemented and scaled up using kd-tree structures or approximation strategies \citep{faiss, indyk1998approximate}.
Starting from an empty set or few known correspondences (anchors) between the two spaces $\mathcal{X}$ and $\mathcal{Y}$, the correspondence can be extended to the entire set of nodes $\mathcal{V}_{\mathcal{X}}$ and $\mathcal{V}_{\mathcal{Y}}$. This extended set of anchors can then be used to fit an arbitrary transformation between the latent spaces, for example an orthogonal mapping \citep{maiorca2023semantic}. The operator $\OperatorCoupled_\Corr$ of this chapter is obtained in this way, by fitting an orthogonal map to the point correspondence recovered from $\mC$.
The experiments demonstrate that using a very small number of anchors suffices to retrieve transformations that facilitate improved stitching and near-perfect retrieval tasks.

\section{\label{sec:lfm_experiments}Experiments}
This section presents experiments evaluating the effectiveness of the LFM framework across various tasks, including similarity evaluation, stitching, retrieval performance, and robustness to perturbations in latent space. Comparisons with existing methods demonstrate the versatility and performance of LFM in aligning and analyzing neural network representations.

\subsection{Analysis}
This subsection demonstrates the benefits of using latent functional maps for comparing distinct representational spaces, using the similarity metric
$\simf_{\mathrm{LFM}}(\mathcal{X},\mathcal{Y})$ defined in \cref{sec:lfm_similarity}.

\begin{figure}
 \centering
 \begin{subfigure}{0.28\textwidth}
  \centering
  \begin{overpic}[trim=0cm 0cm 0cm -0.6cm,clip,width=\linewidth, grid=false]{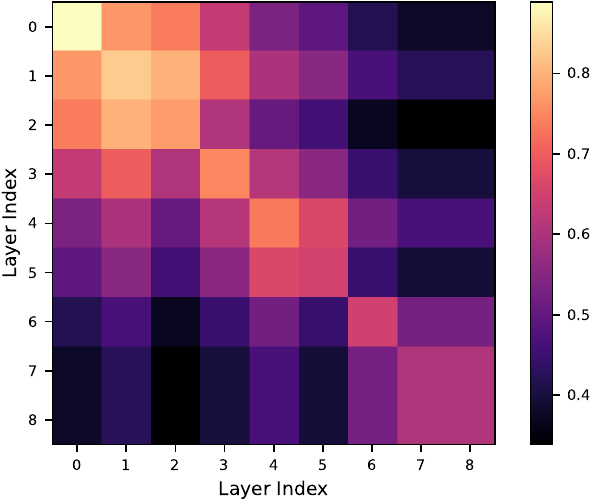}
  \put(25,86){CCA (41.4 \%)}
  \end{overpic}
 \end{subfigure}
 \hfill
 \begin{subfigure}{0.28\textwidth}
  \centering
  \begin{overpic}[trim=0cm 0cm 0cm -0.6cm,clip,width=\linewidth]{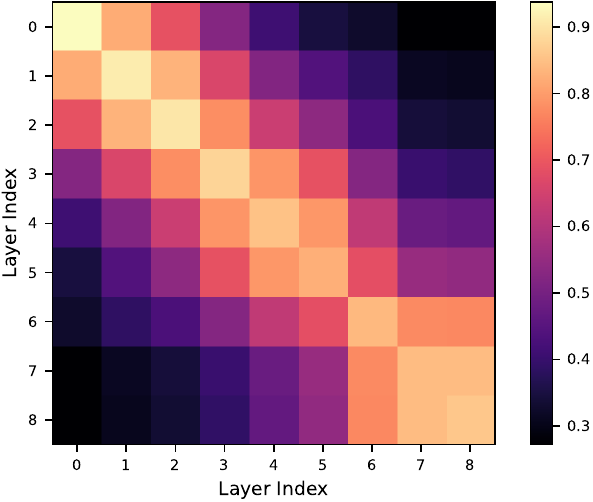}
  \put(25,86){CKA (99.6 \%)}
  \end{overpic}
 \end{subfigure}
 \hfill
 \begin{subfigure}{0.28\textwidth}
  \centering
  \begin{overpic}[trim=0cm 0cm 0cm -0.6cm,clip,width=\linewidth]{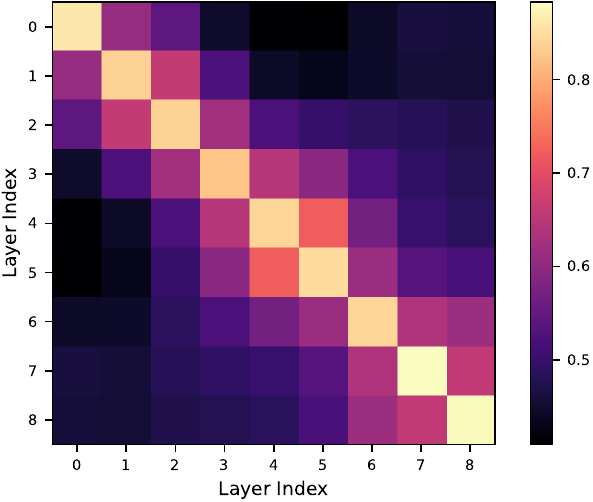}
  \put(25,86){LFM (\textbf{99.8 \%})}
  \end{overpic}
 \end{subfigure}
 \caption{\textbf{Similarity across layers.} Similarity matrices between internal layer representations of \gls{cifart} comparing the LFM-based similarity with the CCA and CKA baselines, averaged across 10 models. For each method, the accuracy scores for matching the corresponding layer by maximal similarity are reported.}
 \label{fig:lfm_side_by_side}
\end{figure}

\paragraph{Experimental setting.}
To validate experimentally whether LFMs are a good measure of similarity between distinct representational spaces, the same sanity check as \citet{Kornblith2019-sz} is performed. Ten CNN models are trained on the \gls{cifart} dataset \citep{Krizhevsky2009-hv} with different initialization seeds. Their inner representations at each layer (excluding logits) are compared and plotted as a similarity matrix, alongside Centered Kernel Alignment (CKA) \citep{Kornblith2019-sz} and Canonical Correlation Analysis (CCA) \citep{hotelling1992relations,raghu2017svcca} baselines. The accuracy of identifying corresponding layers across models is measured and reported. For CCA, average pooling is applied on the spatial dimensions to the embeddings of internal layers, making it more stable numerically and boosting the results for this baseline compared to what was observed in \citet{Kornblith2019-sz}.

\paragraph{Result analysis.}
\Cref{fig:lfm_side_by_side} shows that the LFM-based similarity measure behaves correctly as CKA does. Furthermore, the similarities are less spread around the diagonal, favoring a slightly higher accuracy score in identifying the corresponding layers across models.

While CKA (Centered Kernel Alignment) is a widely used similarity metric in deep learning, recent research by \citet{davari2022reliability} has shown that it can produce unexpected or counter-intuitive results in certain situations. Specifically, CKA is sensitive to transformations that preserve the linear separability of two spaces, such as local translations. The LFM similarity measure is robust to these changes and demonstrates greater stability compared to CKA.

\begin{figure}
 \centering
 \begin{subfigure}[b]{0.5\linewidth}
  \centering
  \begin{overpic}[trim=0cm 0cm 0cm 0cm,clip,width=\linewidth,grid=false]{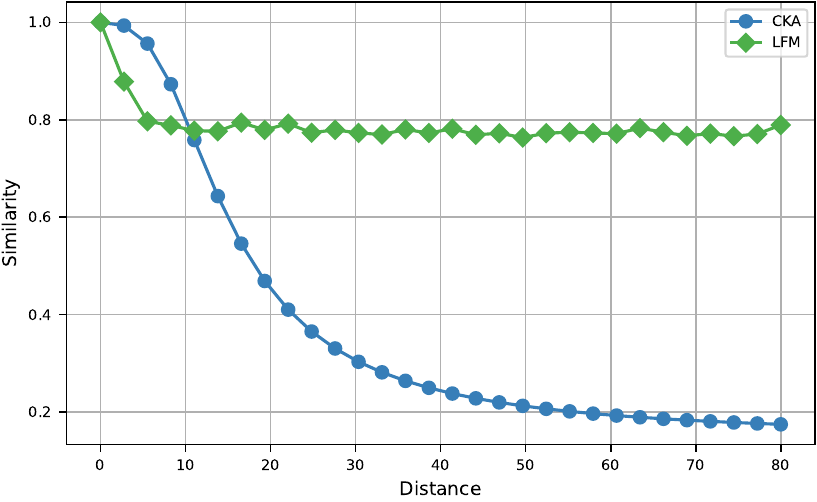}
  \end{overpic}
  \caption{CKA vs LFM similarity}
  \label{fig:lfm_ckaVSlfm}
 \end{subfigure}
 \hfill
 \begin{subfigure}[b]{0.48\linewidth}
  \centering
  \begin{overpic}[trim=0cm 0.1cm 0cm 0cm,clip,width=\linewidth,grid=false]{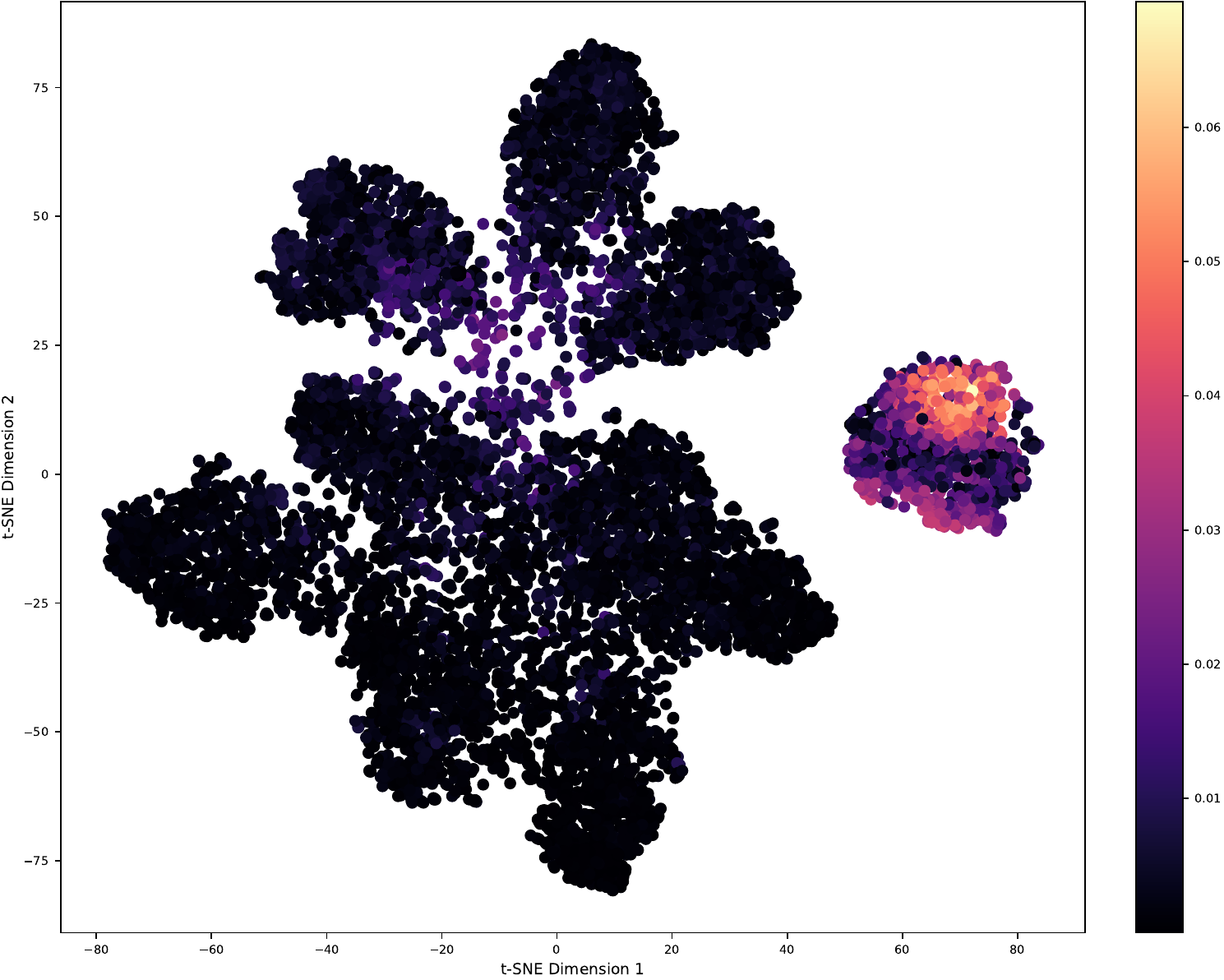}
   \put(65,6){\begin{overpic}[trim=0cm 0.1cm 0cm 0cm,clip,width=0.28\linewidth,grid=false]{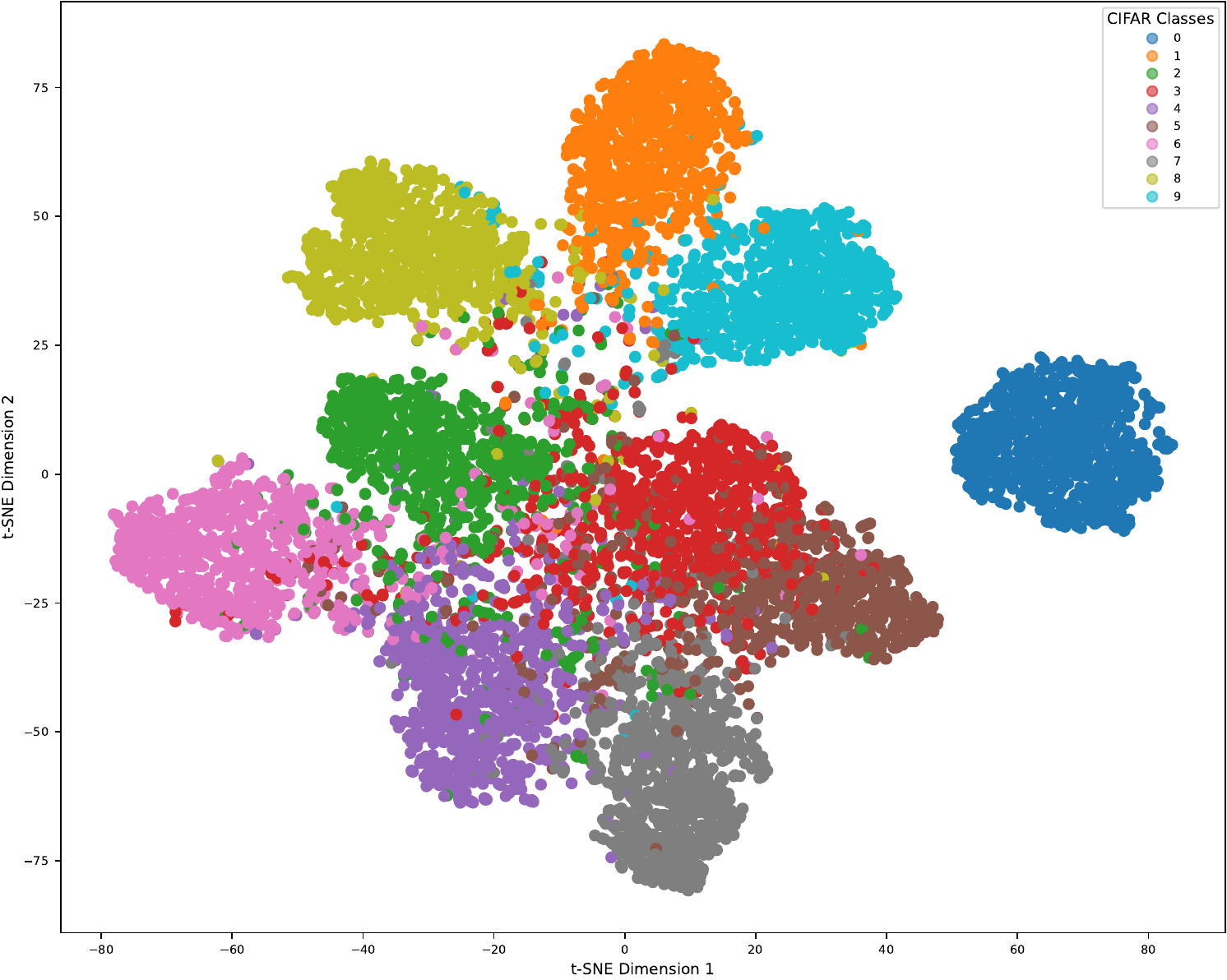}
    \end{overpic}}
  \end{overpic}
  \caption{\label{fig:lfm_cka_distorsion} Localized area distortion}
 \end{subfigure}
 \caption{\textbf{Robustness of LFM similarity.} \emph{Left:} Similarity scores as a function of perturbation strength (``Distance'' on the horizontal axis is the norm of the perturbation vector): while the CKA baseline degrades, LFM similarity scores are robust to perturbations that preserve linear separability of the space. \emph{Right:} Visualization of area distortion of the map by projecting the first singular component of the LFM in the perturbed space: the distortion localizes on the samples of the perturbed class, making LFM similarity interpretable.}
 \label{fig:lfm_CKAvs}
\end{figure}

\paragraph{Experimental setting.}
Latent representations are computed from the pooling layer just before the classification head for the \gls{cifart} train and test sets. Following the setup in \citet{davari2022reliability}, a Support Vector Machine (SVM) classifier is trained on the latent representations of the training samples to find the optimal separating hyperplane between samples of one class and others. The samples are then perturbed by translating them in a direction orthogonal to the hyperplane, ensuring the space remains linearly separable. The CKA and LFM similarities are measured as functions of the perturbation vector norm, as shown in \cref{fig:lfm_ckaVSlfm}.

\paragraph{Result analysis.}
When the latent space is perturbed in a way that still preserves its linear separability, it should be considered identical from a classification perspective, as this does not semantically affect the classification task.
\Cref{fig:lfm_ckaVSlfm} shows that while CKA degrades as a function of perturbation intensity, the LFM similarity, after an initial drop, stays at about 0.78.
To understand this difference, the area distortion is visualized as a function of the samples by projecting the first singular component of $\mC$ onto the perturbed space. In \cref{fig:lfm_cka_distorsion}, t-SNE \citep{van2008visualizing} projects the perturbed samples and the distortion function into 2D. The visualization reveals that distortion is localized to the samples corresponding to the perturbed class.

\subsection{\label{sec:lfm_stitching} Zero-shot stitching}
In latent communication tasks, a common challenge is combining an encoder that embeds data with a decoder specialized in a downstream task, such as classification or reconstruction---this process is often referred to as \textit{stitching}. Previous works, like \citet{Lenc2014-gy} and \citet{Bansal2021-oj}, have used trainable linear projections, known as stitching layers, to enable the swapping of different network components. However, \citet{moschella2023relative} introduced the concept of zero-shot stitching, where neural components can be merged without any additional training procedure. While \citet{moschella2023relative} still required training a decoder module once for processing relative representations before stitching could be performed, LFM is fully zero-shot, with no need for additional training. In the following experiments, stitching is conducted without any training or fine-tuning of the encoder or decoder, adhering strictly to a \textit{zero-shot fashion}.

\begin{figure}[htbp]
\begin{subfigure}[b]{0.32\linewidth} \centering
 \begin{overpic}
  [trim=0cm 0cm 0cm 0cm,clip,width=\linewidth, grid=false]{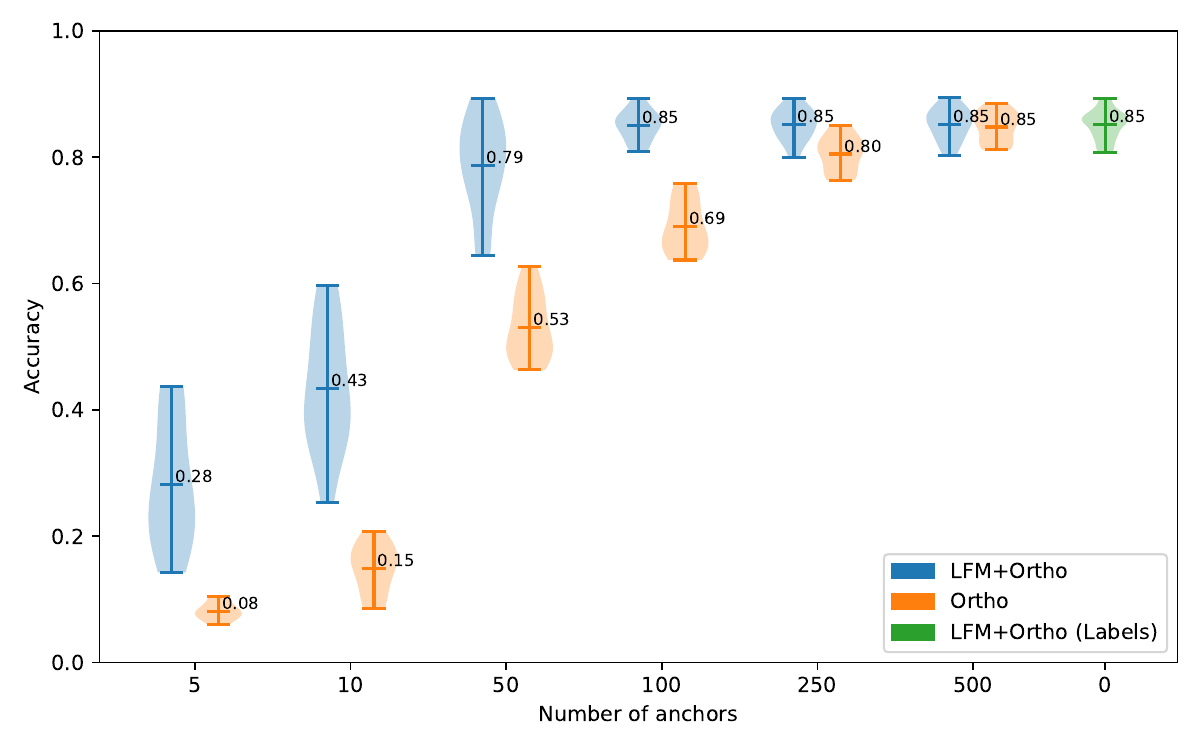}
  \end{overpic}
 \caption{\gls{cifarh} coarse}
 \end{subfigure}
 \hfill
  \begin{subfigure}[b]{0.32\linewidth} \centering
 \begin{overpic}
  [trim=0cm 0cm 0cm 0cm,clip,width=\linewidth, grid=false]{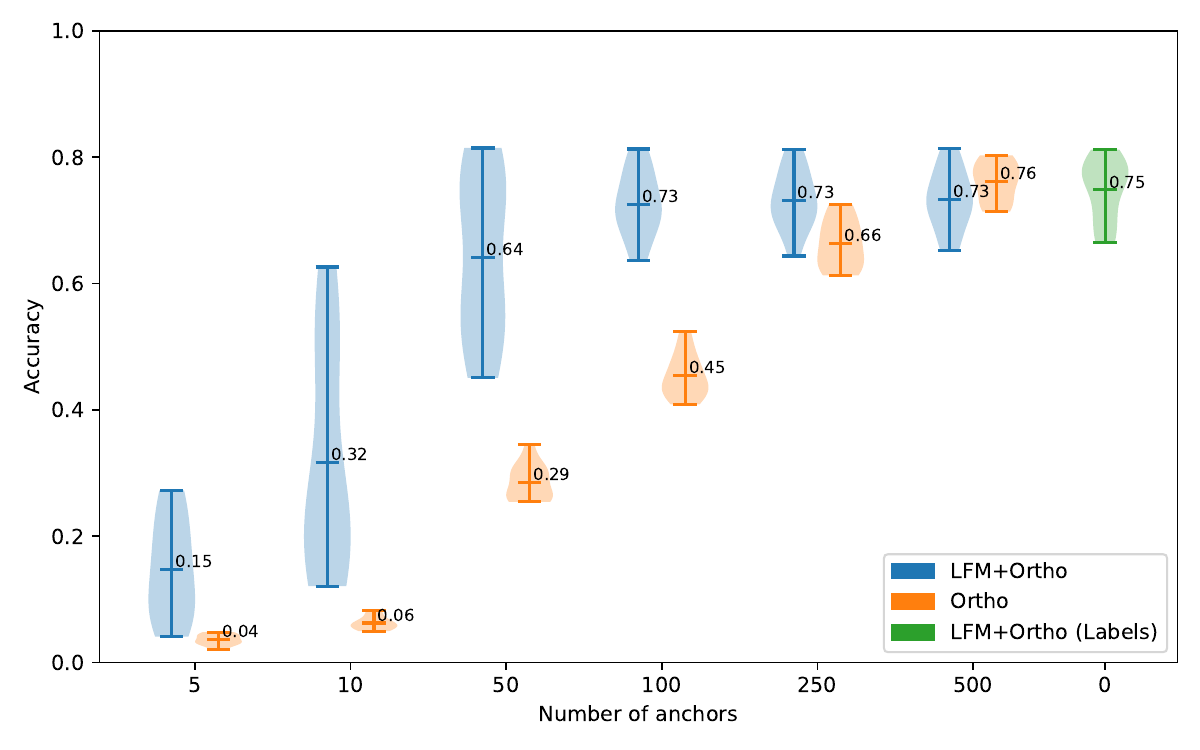}
  \end{overpic}
 \caption{\gls{cifarh} fine}
 \end{subfigure}
 \hfill
 \begin{subfigure}[b]{0.32\linewidth} \centering
 \begin{overpic}
  [trim=0.1cm 0.1cm 0.1cm 0.4cm,clip,width=\linewidth, grid=false]{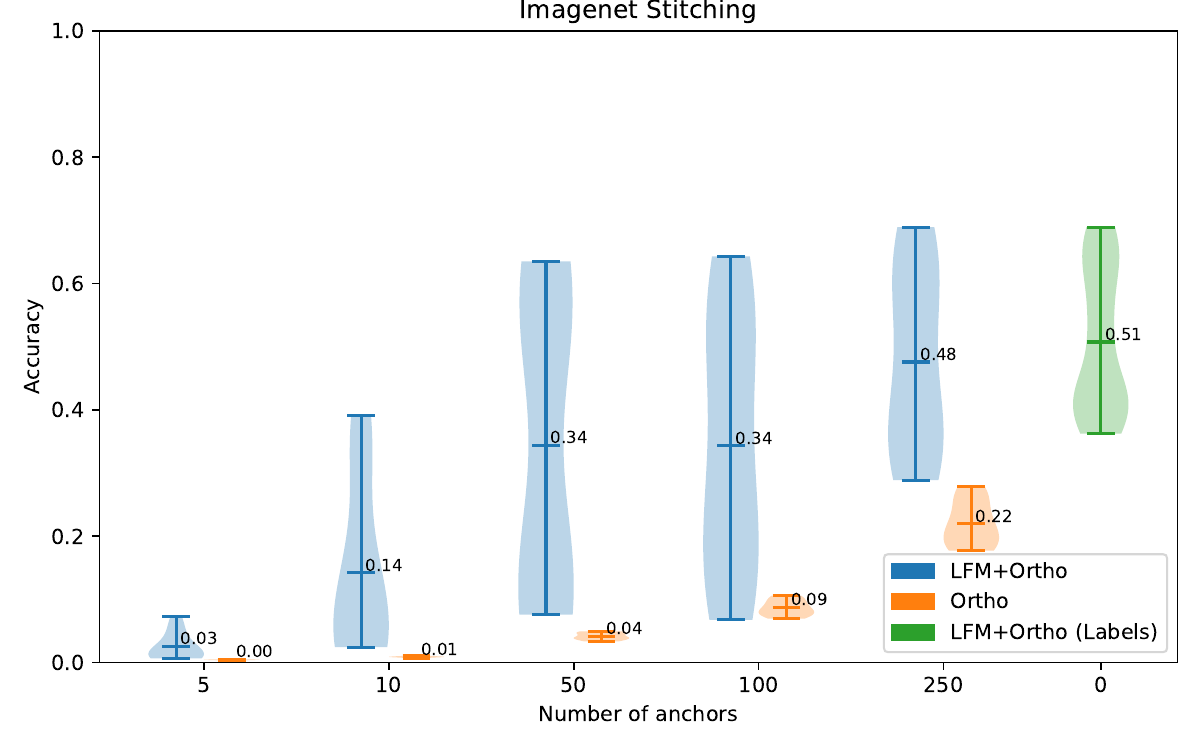}
  \end{overpic}
 \caption{ImageNet}
 \end{subfigure}
 \hfill
  \begin{subfigure}[b]{0.32\linewidth} \centering
 \begin{overpic}
  [trim=0cm 0.4cm 0cm 0.4cm,clip,width=\linewidth, grid=false]{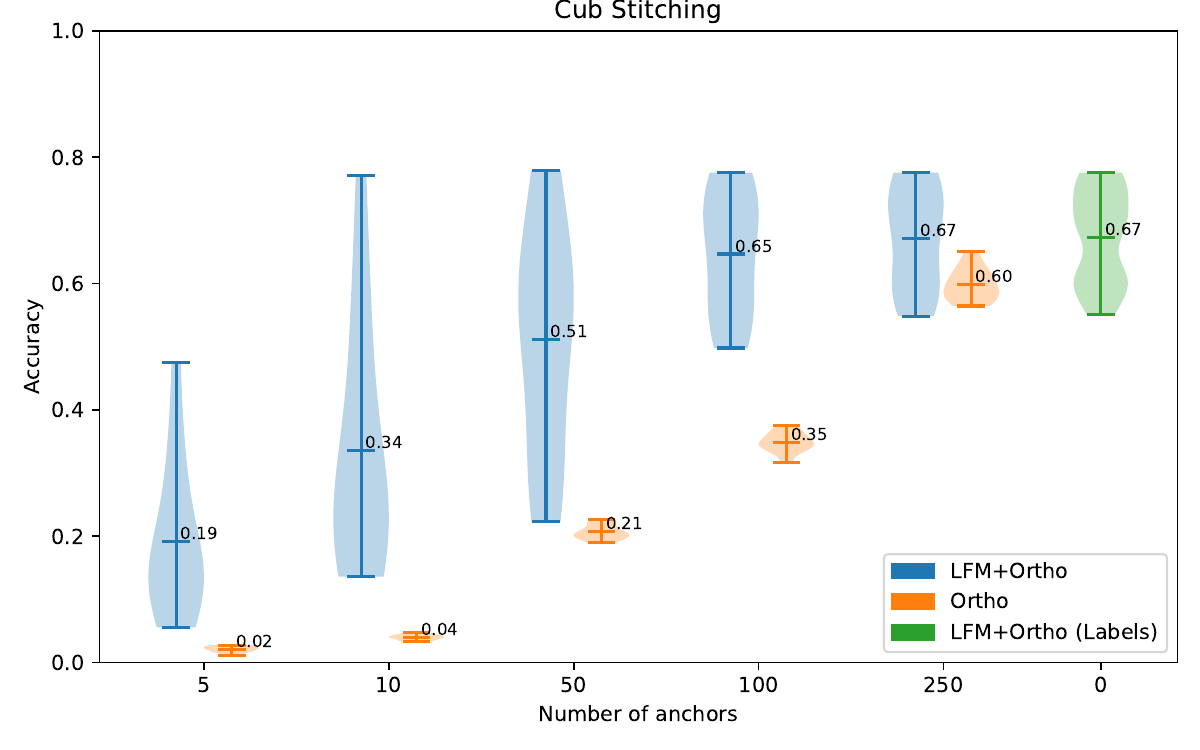}
  \end{overpic}
 \caption{CUB}
 \end{subfigure}
 \hfill
  \begin{subfigure}[b]{0.32\linewidth} \centering
 \begin{overpic}
  [trim=0cm 0.4cm 0cm 0.4cm,clip,width=\linewidth, grid=false]{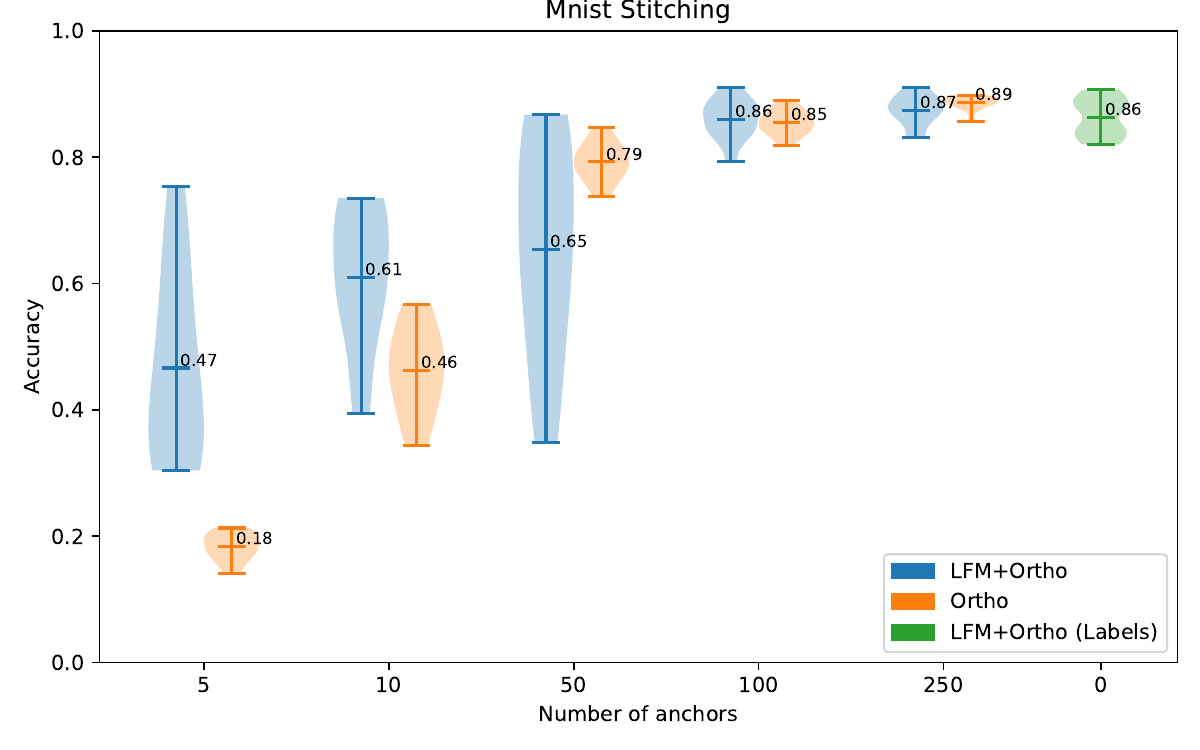}
  \end{overpic}
 \caption{\gls{mnist}}
 \end{subfigure}
 \hfill
  \begin{subfigure}[b]{0.32\linewidth} \centering
 \begin{overpic}
  [trim=0cm 0.4cm 0cm 0.4cm,clip,width=\linewidth, grid=false]{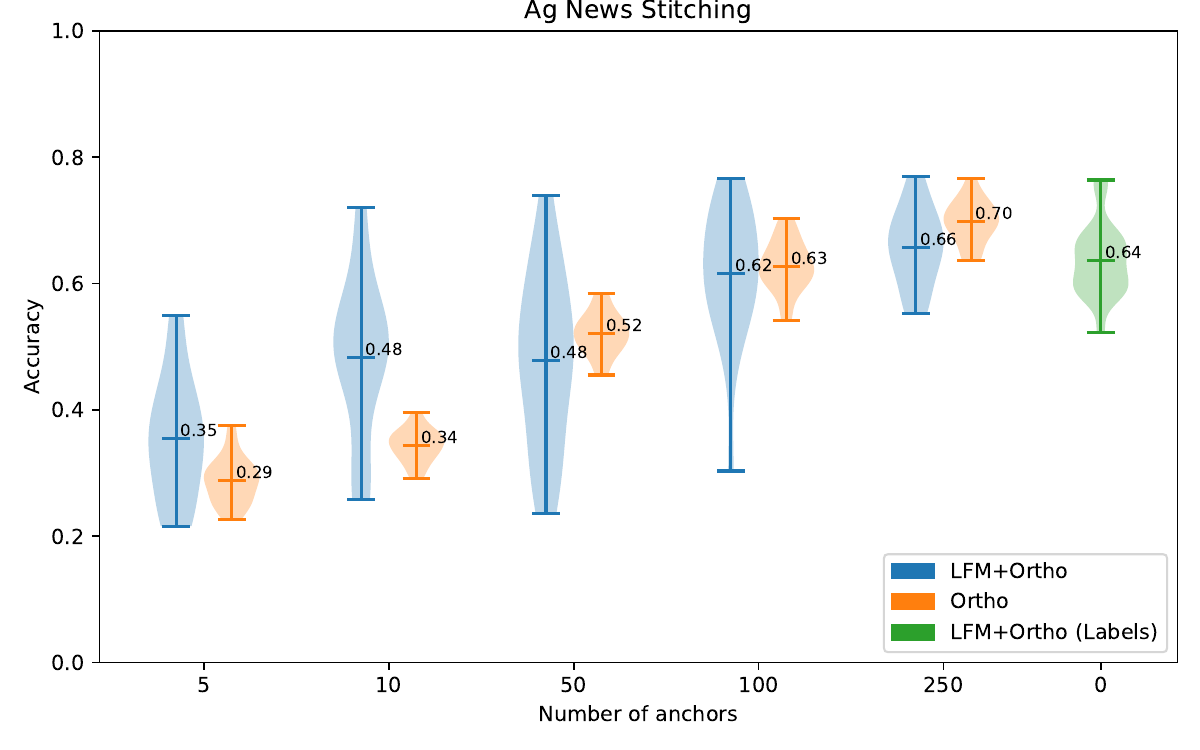}
  \end{overpic}
 \caption{AgNews}
 \end{subfigure}
 \caption{\textbf{Stitching results.} Accuracy performance in stitching between pretrained encoders on five datasets (\gls{cifarh} with two labelings), comparing orthogonal transformation (Ortho) and LFM+Ortho at varying anchor counts. Also shown is LFM+Ortho (Labels), which uses the dataset labels instead of anchors. Results are presented with mean accuracy values reported on each violin.}
 \label{fig:lfm_stitching}
\end{figure}

\begin{figure}
  \begin{subfigure}[b]{0.32\linewidth} \centering
 \begin{overpic}
  [trim=2.8cm 2.3cm 2.2cm 3cm,clip,width=\linewidth, grid=false]{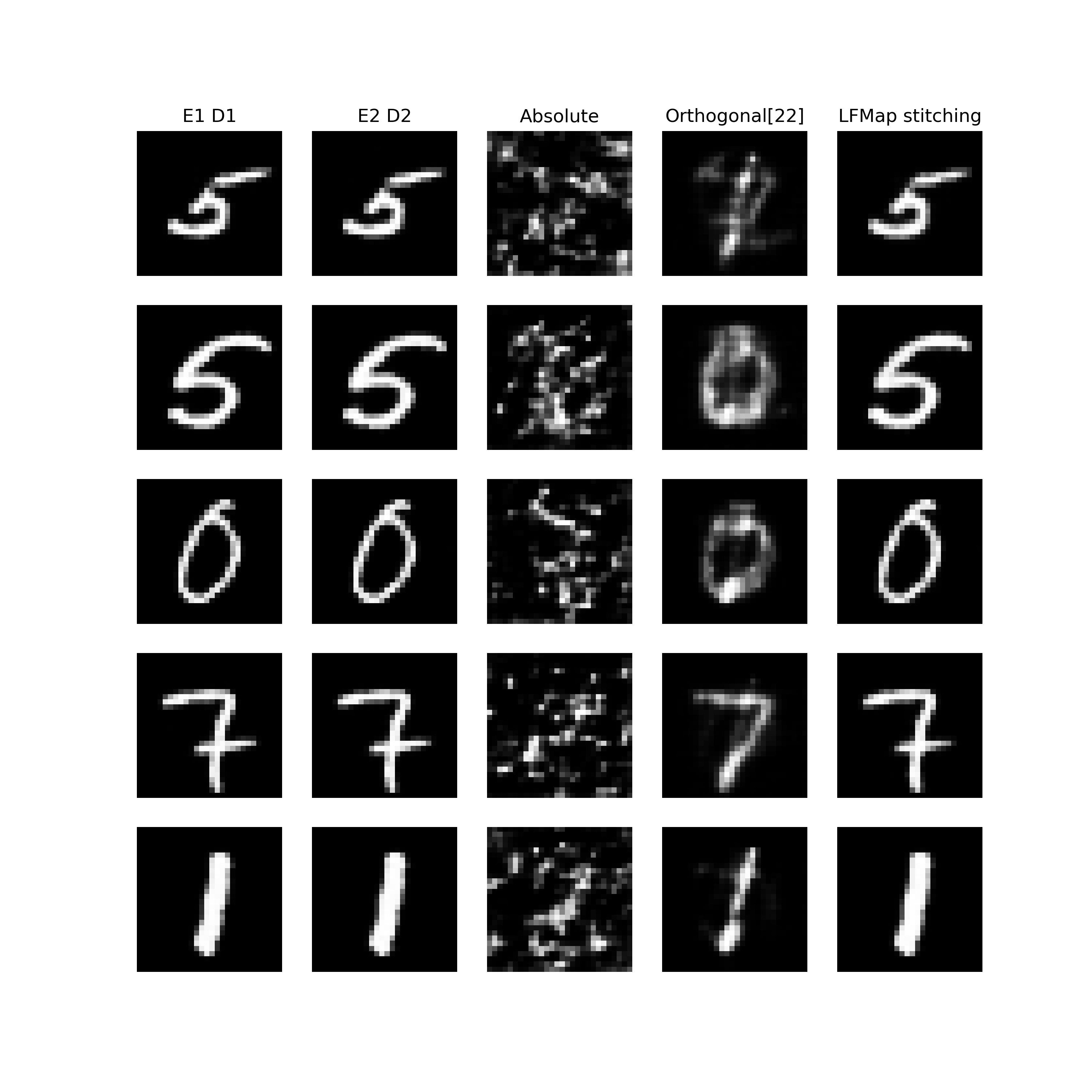}
  \put(4,99){\tiny E1 D1}
  \put(24,99){\tiny E2 D2}
  \put(44,99){\tiny Absol.}
  \put(64,99){\tiny Ortho}
  \put(85,99){\tiny LFM}
  \end{overpic}
 \caption{\gls{mnist}}
 \end{subfigure}
 \hfill
 \begin{subfigure}[b]{0.32\linewidth} \centering
 \begin{overpic}
  [trim=2.8cm 2.3cm 2.2cm 3cm,clip,width=\linewidth, grid=false]{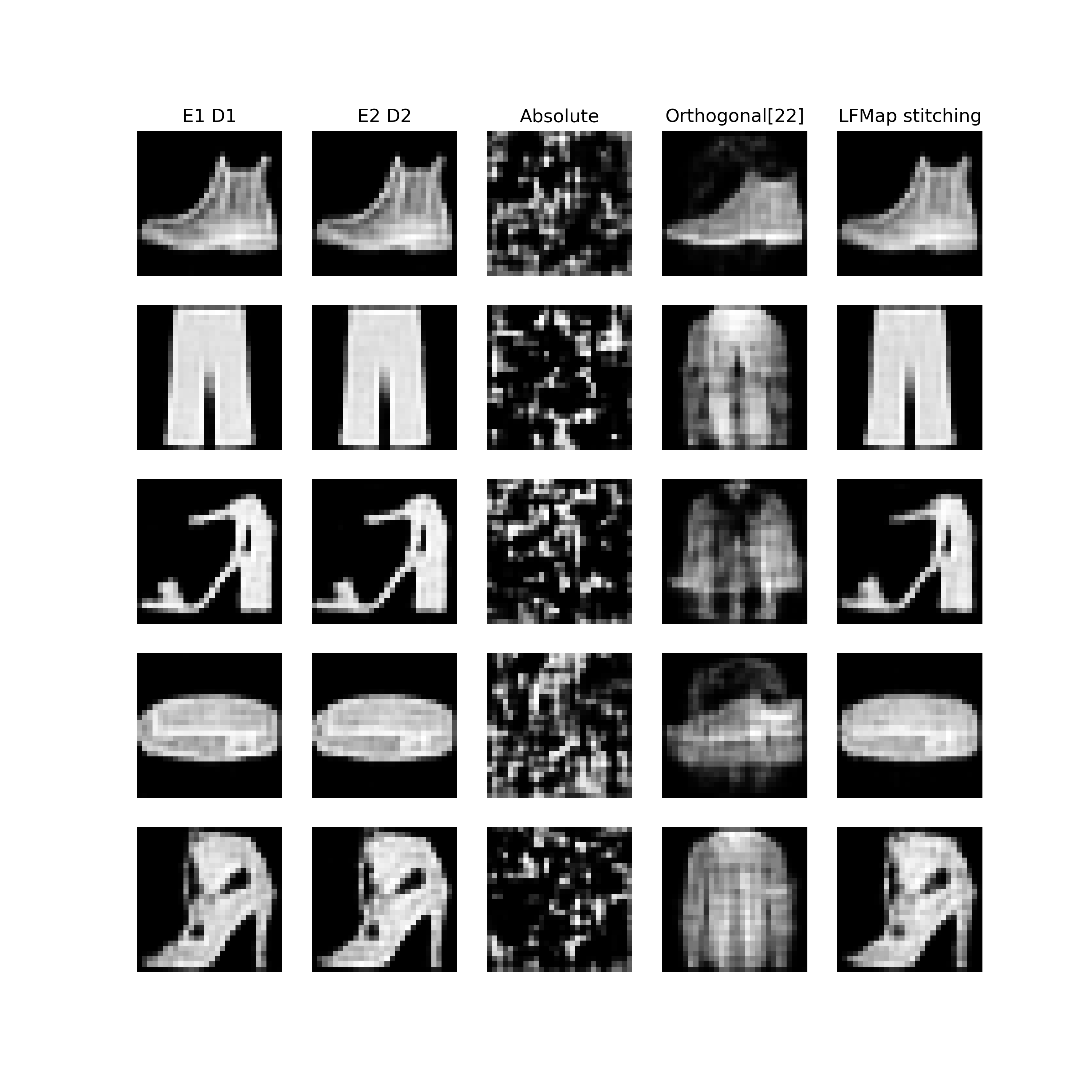}
  \put(4,99){\tiny E1 D1}
  \put(24,99){\tiny E2 D2}
  \put(44,99){\tiny Absol.}
  \put(64,99){\tiny Ortho}
  \put(85,99){\tiny LFM}
  \end{overpic}
 \caption{\gls{fmnist}}
 \end{subfigure}
 \hfill
  \begin{subfigure}[b]{0.32\linewidth} \centering
 \begin{overpic}
  [trim=2.8cm 2.3cm 2.2cm 3cm,clip,width=\linewidth, grid=false]{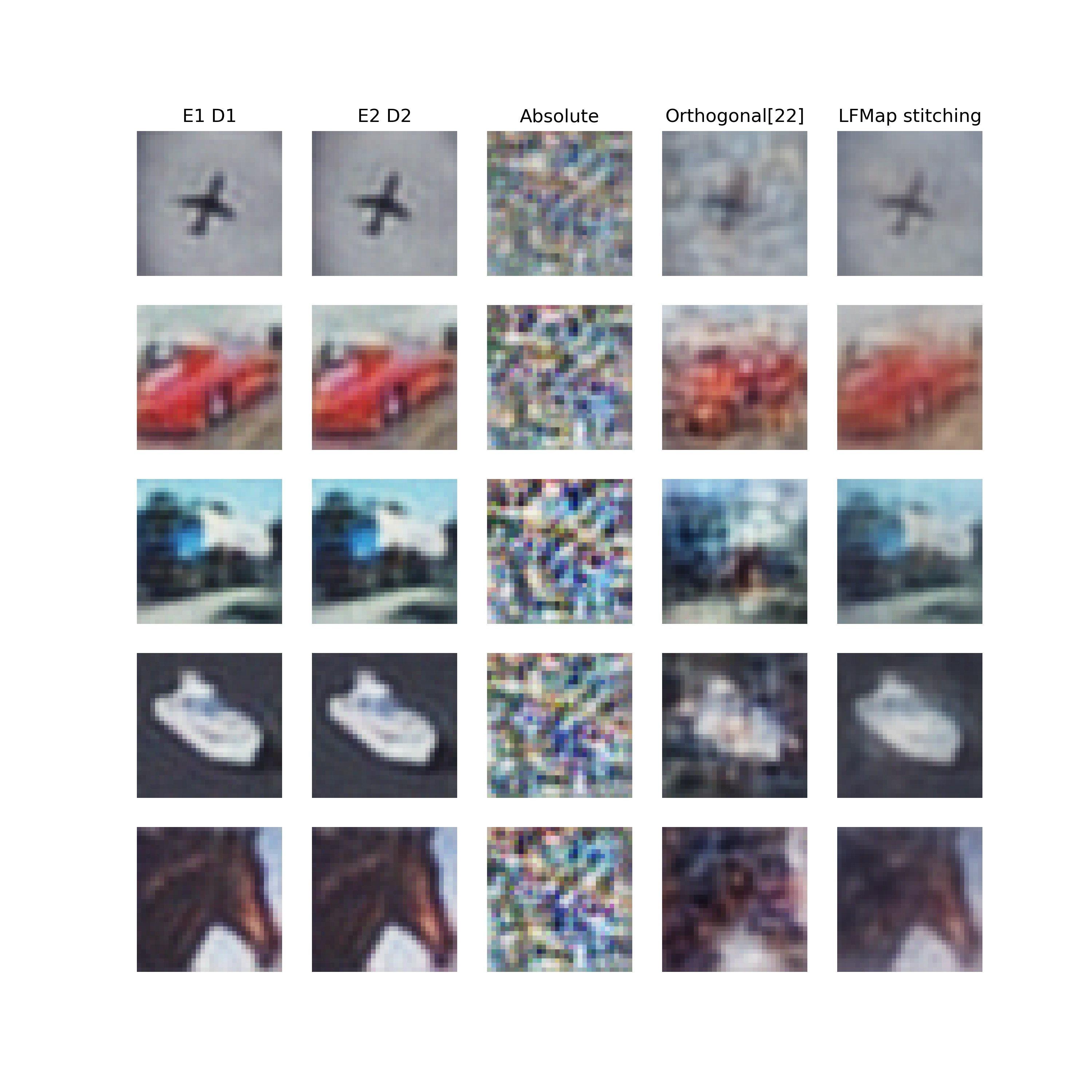}
  \put(4,99){\tiny E1 D1}
  \put(24,99){\tiny E2 D2}
  \put(44,99){\tiny Absol.}
  \put(64,99){\tiny Ortho}
  \put(85,99){\tiny LFM}
  \end{overpic}
 \caption{\gls{cifart}}
 \end{subfigure}
 \caption{\textbf{Qualitative results on stitching.} We combine the encoder decoder modules of two convolutional autoencoders (\textit{E1 D1}, \textit{E2 D2}; E1, D1 and E2, D2 are $\Enc[X], \Dec[X]$ and $\Enc[Y], \Dec[Y]$) using three different approaches: no transformation (\textit{Absol.}), the orthogonal transformation from \citet{maiorca2023semantic} (\textit{Ortho}), the orthogonal transformation augmented with LFM (\textit{LFM}). }
 \label{fig:lfm_stitching_qualitative}
\end{figure}

\paragraph{Experimental setting.}
Four pretrained encoders are considered, and their latent spaces are stitched to perform classification using a Support Vector Machine (SVM) on five different datasets: \gls{cifarh} \citep{Krizhevsky2009-hv} with coarse and fine-grained labeling, ImageNet \citep{deng2009imagenet}, Caltech-UCSD Birds-200-2011 (CUB) \citep{wah2011CUB}, \gls{mnist} \citep{deng2012mnist}, AgNews \citep{zhang2015AGNews}.
These datasets were chosen to encompass both large-scale, complex datasets like ImageNet and a variety of modalities, including text and images.
To evaluate the effectiveness of integrating the functional map, the correspondences are extended to determine an orthogonal transformation \citep{maiorca2023semantic} between the latent spaces. For each encoder, a graph of 3,000 points with 300 neighbors per node is computed. The problem in \cref{eq:lfm_fmap_opt_ls} is optimized using the first 50 eigenvectors of the graph Laplacian, and two different descriptors are considered: the distance functions defined from the anchors (LFM+Ortho) and the labels (LFM+Ortho (Labels)). For each dataset class, the latter provides an indicator function with 1 if the point belongs to the class and 0 otherwise. This descriptor does not require any anchor as input, representing an example of stitching requiring no additional information beyond the dataset labels.

\paragraph{Result analysis.}
\Cref{fig:lfm_stitching} shows the accuracy results for all possible combinations of encoder stitching. The latent functional map (LFM+Ortho) outperforms Ortho across all datasets with a low number of anchors (5 and 10), and up to 250 anchors on \gls{cifarh}, ImageNet and CUB. Even without anchors, using label descriptors (LFM+Ortho (Labels)) achieves strong performance across different labeling schemes. The orthogonal transformation computed directly from the anchors (Ortho) only matches or exceeds LFM's performance when 50 or more anchors are used on AgNews and \gls{mnist} (on \gls{mnist} it is within 0.01 at 100 anchors) and when 500 are used on \gls{cifarh}; at that point LFM's effectiveness is constrained by the number of eigenvectors.
Comparing the \gls{cifarh} coarse and fine-grained labeling, the fine-grained task is more complex due to the larger number of classes, making the estimation of the map more challenging. This complexity is especially evident when aligning self-supervised vision models with classification-based ones (e.g., DINOv2 vs.\ ViT), where the training strategies differ.

This experiment shows that the latent functional map is highly effective when few anchors are available ($\leq 10$). It enhances performance in zero-shot stitching tasks, outperforming direct orthogonal transformations at low anchor counts, and without anchors through label descriptors. This suggests that the latent functional map method provides a robust means of aligning latent spaces with minimal correspondence data, making it a valuable tool for tasks requiring the integration of independently trained models.

\subsubsection{Qualitative example}
\Cref{fig:lfm_stitching_qualitative} presents qualitative experiments on the \gls{mnist}, \gls{fmnist} \citep{Xiao2017-er}, and \gls{cifart} datasets, focusing on visualizing the reconstructions of stitched autoencoders. For these experiments, convolutional autoencoders are trained using two different seeds and their encoder and decoder modules are stitched together.
As a baseline, no transformation (Absol.) is used, where the latent representation from the first encoder is directly input into the second decoder. This baseline is compared with the orthogonal transformation from \citet{maiorca2023semantic} (Ortho) and its LFM-augmented version (LFM). For both LFM and \citet{maiorca2023semantic}, 10, 10 and 50 correspondences are used for the \gls{mnist}, \gls{fmnist}, and \gls{cifart} datasets, respectively.
LFM consistently produces superior reconstruction quality compared to both the baseline and the method from \citet{maiorca2023semantic}.

\subsection{\label{sec:lfm_Retrieval} Retrieval}
The analysis extends to the retrieval task, where the most similar embedding in the aligned latent space is sought.
\begin{figure}[t]
 \centering
 \begin{subfigure}[b]{0.5\linewidth} \centering
 \begin{overpic}
  [trim=0cm 0cm 0cm 0cm,clip,width=\linewidth, grid=false]{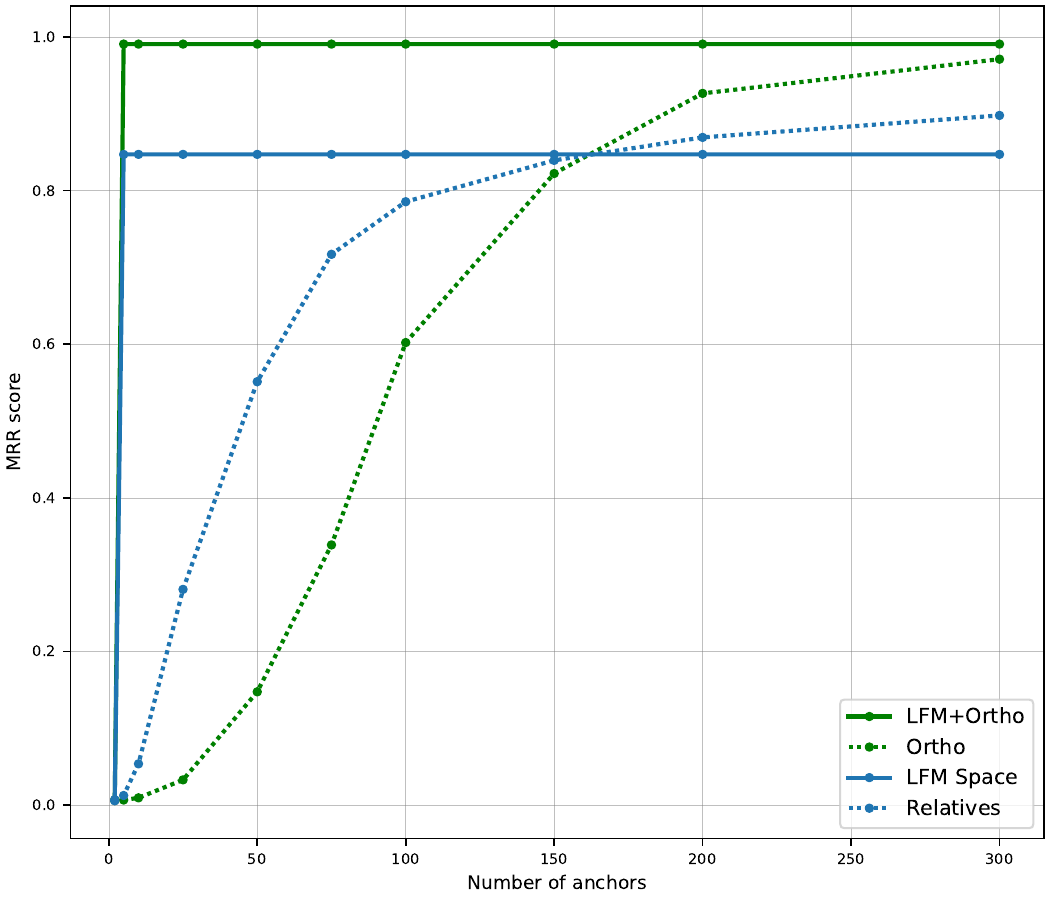}
  \end{overpic}
 \caption{MRR score at increasing number of anchors}
 \end{subfigure}
 \hfill
 \begin{subfigure}[b]{0.48\linewidth} \centering
 \begin{subfigure}[b]{\linewidth} \centering
 \vspace{0.2cm}
 \begin{overpic}
  [trim=0cm 0.1cm 0cm 0cm,clip,width=\linewidth, grid=false]{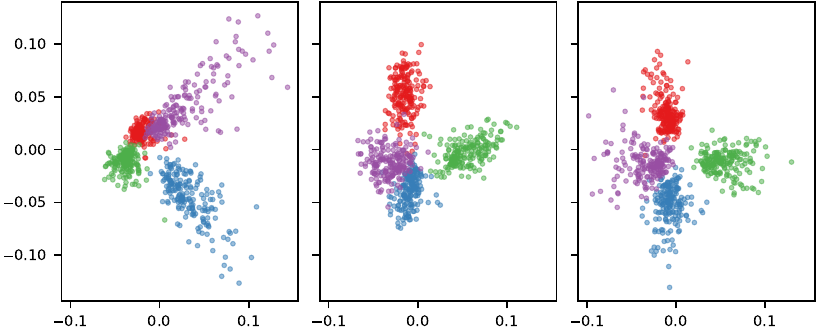}
  \put(16,41){\tiny FastText}
  \put(46,41){\tiny Word2Vec}
  \put(79,41){\tiny Aligned}
 \end{overpic}
 \caption{Latent space}
 \end{subfigure}

 \begin{subfigure}[b]{\linewidth} \centering
 \vspace{0.2cm}
 \begin{overpic}
  [trim=0cm 0.1cm 0cm 0cm,clip,width=\linewidth, grid=false]{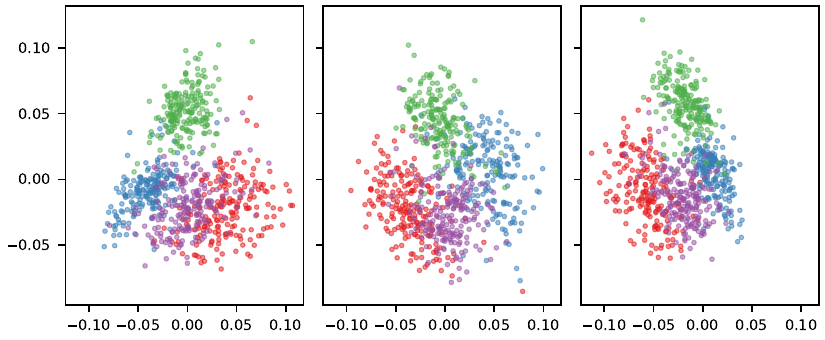}
  \put(16,41){\tiny FastText}
  \put(46,41){\tiny Word2Vec}
  \put(79,41){\tiny Aligned}
 \end{overpic}
 \caption{Functional space}
 \end{subfigure}
 \end{subfigure}
 \caption{\textbf{Retrieval of word embeddings.} We compare the retrieval performance of the functional map framework with baseline methods as the number of anchors increases. The left panel shows the Mean Reciprocal Rank (MRR) across different numbers of anchors. The right panels depict the first two components of PCA for a subsample of the latent space (b) and the functional space (c), both before and after alignment using the functional map.}
 \label{fig:lfm_retrieval_anchors_fast_w2v}
\end{figure}

\paragraph{Experimental setting.}
Two different English word embeddings are considered: FastText \citep{bojanowski2016enriching} and Word2Vec \citep{mikolov2013word2vec}. Following the approach of \citet{moschella2023relative}, embeddings of 20K words are extracted from their shared vocabulary using pretrained models. 2K random corresponding samples are used to construct the k-NN graphs and evaluate the retrieval performance on the remaining 18K word embeddings.
Two settings are tested:
\begin{enumerate}
 \item Aligning functional coefficients (LFM Space).
 \item Computing an orthogonal transformation using the correspondences obtained by the functional map (LFM+Ortho).
\end{enumerate}
For this experiment, k-NN graphs with a neighborhood size of 300 are constructed and the functional map is computed using the first 50 eigenvectors. The methods' performance is evaluated using the Mean Reciprocal Rank (MRR).
The functional map methods are compared with the method proposed by \citet{moschella2023relative} (Relatives) and the orthogonal transformation method proposed by \citet{maiorca2023semantic} (Ortho).
The same orthogonal transformation is fitted on top of the correspondence found by LFM to make the comparison the fairest possible, although in principle any off-the-shelf methods could be used to estimate the transformation once the new correspondence is found. This versatility is illustrated in Tables 2 and 3 (Appendix C.3) of \citet{fumero2024latent}, where LFM is combined with other transformations in the same task.

\paragraph{Result analysis.}
\Cref{fig:lfm_retrieval_anchors_fast_w2v} shows the performance of these methods as the number of anchors increases.
The functional map improves performance with just 5 anchors, achieving an MRR of over 0.8. As the number of anchors increases, the performance of competing methods improves but still falls short of LFM+Ortho at 300 anchors, which reaches an MRR of 0.99. The performance of the functional map methods does not improve beyond 5 anchors, suggesting that this number of anchors is sufficient to achieve an optimal functional map between the spaces.

These results confirm that the latent functional map is a valuable tool in settings with little knowledge about correspondences. It enhances retrieval performance with a minimal number of anchors, making it an efficient approach for aligning latent spaces. Moreover, its performance can be improved using a higher number of eigenvectors (Appendix C.3 of \citet{fumero2024latent}).
In particular, the framework comprises simultaneously (i) an interpretable similarity measure, (ii) a way to find correspondences, and (iii) an explicit mapping between different spaces in a single framework, unlike \citet{moschella2023relative} and \citet{maiorca2023semantic}, which attempt to solve just the latter implicitly.

\chapter[ResiDual]{ResiDual: Transformer Alignment with Spectral Decomposition}\label{ch:13}\label{ch:residual}

\chapterpublication[https://github.com/Flegyas/ResiDual]{basile2025residual}{\textbf{TMLR 2025}}{Spectral reweighting across layers allows transformer architectures to improve alignment by focusing on task-relevant geometry with no fine-tuning.}

\section{Introduction}
\begin{figure}
\centering
 \includegraphics[width=0.85\textwidth]{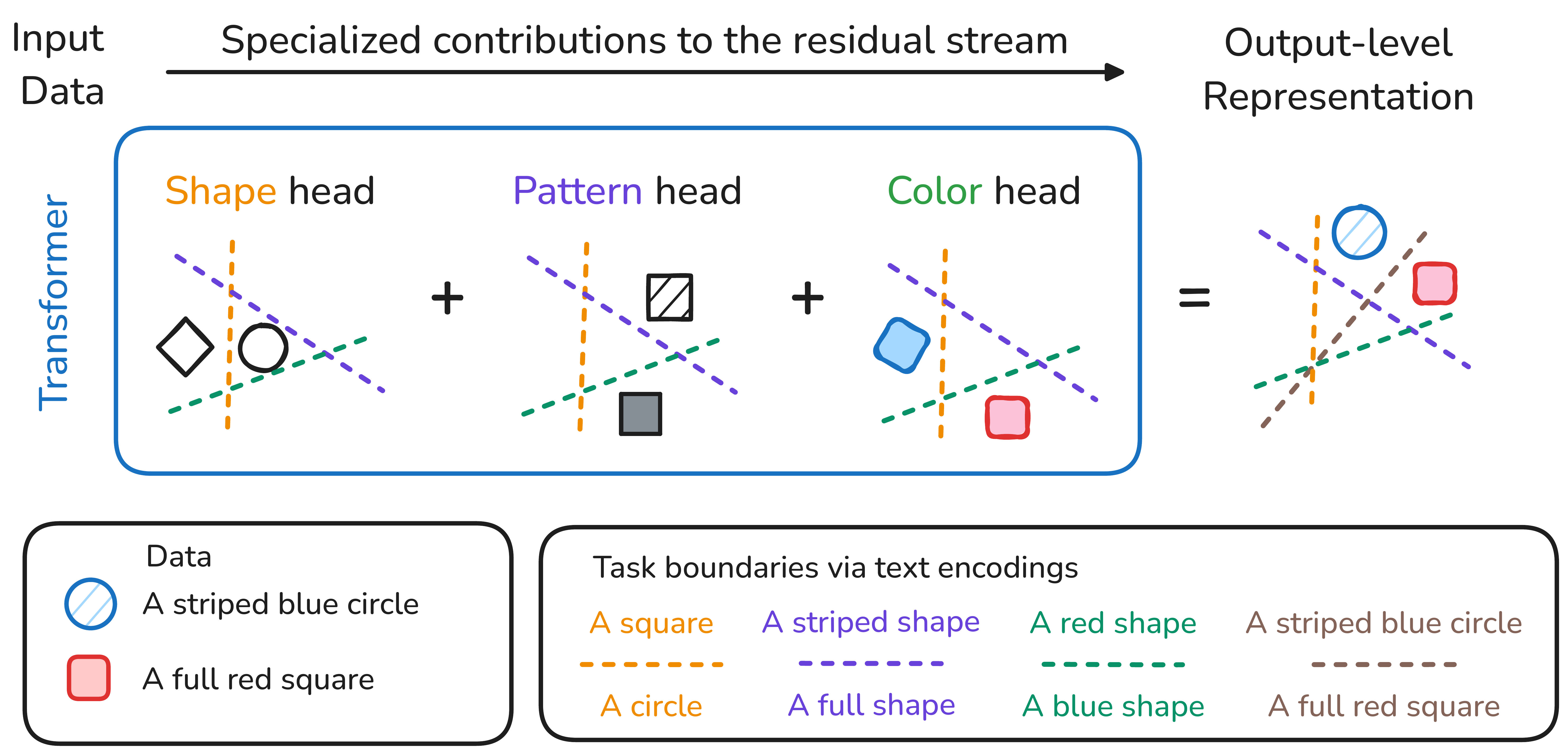}
\caption{Direct contributions from individual heads can be analyzed from the transformer's residual stream. In a multimodal, zero-shot classification setting (e.g., in CLIP), task boundaries are defined by text prompts that may vary in their conceptual granularity. When certain heads are specialized in particular features (e.g., \textcolor{rdorange}{shape}, \textcolor{rdpurple}{pattern}, \textcolor{rdgreen}{color}), they may more accurately apply these boundaries than the model's original output. In this example, only the fine-grained task (\textcolor{rdbrown}{brown}) effectively separates the samples at the output level.}
\label{fig:residual:teaser}
\end{figure}

Transformer architectures aggregate information through unweighted summation: $\mY = \sum_{i} \mU_i$ (\cref{eq:decomposition} below), where the input embedding, each attention head and each multi-layer perceptron (\gls{mlp}) contribute equally to the final output~\citep{elhage2021mathematical}. This additive structure underpins transformers' expressivity, but creates a critical challenge for alignment. When residual units encode conflicting or task-irrelevant information, specialized signals become obscured in the final aggregation.

As illustrated in \cref{fig:residual:teaser}, vision transformers develop attention heads specialized for particular features, such as detecting letters~\citep{gandelsman2024interpreting}, recognizing seasonal patterns, or processing grayscale textures (\cref{tab:descriptions_exp2}), yet only specialized subsets provide relevant signals for specific downstream tasks. The model's output sums all contributions equally, diluting precision through this unweighted aggregation.

ResiDual is a \gls{correspondence-coupled} operator ($\gls{OperatorSameSpace}_{\gls{Corr}}$, \cref{ch:methodology}) that recovers task-relevant alignment hidden within the residual stream through anisotropic spectral scaling. The method decomposes each residual unit into principal components and learns per-component scaling factors $\vlambda$, enabling fine-grained amplification of task-relevant semantic directions while suppressing noise from irrelevant units. This dual-level approach first analyzes which attention heads already align with task requirements by measuring correspondence with text encodings (\cref{ssec:coarse_selection}) in multimodal models like CLIP~\citep{clip}, then applies anisotropic scaling within each unit's principal component basis to refine the alignment at the spectral level. The result is high parameter efficiency: the restricted variant (heads-only, 90\% variance) achieves 0.91 accuracy on OpenCLIP-L across 10 datasets with only 13.2k parameters, matching linear aligners that require 590k parameters---a 45$\times$ reduction. On \gls{svhn}, the full method delivers substantial improvement: 0.86 accuracy versus 0.75 for output-level linear alignment, suggesting latent alignment that exists within specialized head components but washes out when all units combine at the output.

Unlike the cross-model operators in \cref{ch:relrep,ch:semantic_alignment,ch:inverse-relative-projection}, ResiDual demonstrates that correspondence-based alignment need not cross model boundaries. By establishing correspondences between internal components and task requirements within a single model, the alignment framework extends to model manipulation and steering. Like the spectral methods in \cref{ch:lfm}, which operate on functional maps, ResiDual exploits spectral structure, here applied to residual stream components. \cref{ch:headpursuit,ch:lineas} explore alternative geometric structures for within-model alignment: sparse recovery and distributional matching, respectively.

\section{The Geometry of Residual Units}
This section examines the relationship between head specialization and the low-dimensional nature of head manifolds. Initially, head representations exist in a relatively low-dimensional ambient space. Through a linear transformation, they are subsequently embedded into a higher-dimensional space within the residual stream \citep{elhage2021mathematical}, which shares the same dimensionality as the model's output. At this point, head representations are transcribed to the residual stream and contribute additively to the final output of the model. Throughout this chapter, the model output is assumed to be the summation of the encodings of all residual units (attention heads $\mH$, MLPs $\mM$ and input embeddings $\mX_0$):
\begin{equation}\label{eq:decomposition}
 \mY = \sum_{i=1}^{\lvert\mathcal{U}\rvert}\mU_i = \mX_0 + \sum_{h=1}^{\lvert\mathcal{U}_{\mathrm{attn}}\rvert}\mH_h + \sum_{j=1}^{\lvert\mathcal{U}_{\mathrm{mlp}}\rvert}\mM_j\,,
\end{equation}
with $\mY$ as the final output of the model, summing all the residual units $\mU \in \mathcal{U}$, where $\mathcal{U} = \{\mX_0\} \cup \mathcal{U}_{\mathrm{attn}} \cup \mathcal{U}_{\mathrm{mlp}}$ collects the input embedding, the attention heads and the MLPs.

This chapter focuses on the geometry of attention head representations, as prior works~\citep[e.g.,][]{voita2019analyzing,gandelsman2024interpreting} have demonstrated their specialization in distinct tasks. MLP layers mix outputs from multiple heads by design, leading to more entangled and less interpretable representations.

\subsection{Residual Dimensionality}\label{ssec:residual_dimensionality}

Despite being embedded into a higher-dimensional space, head representations exhibit an even lower \textit{intrinsic dimensionality} (ID) than that of the original ambient space. This indicates that irrespective of their high-dimensional embedding, the essential structure of head representations is highly compressed and governed by a compact, low-dimensional geometry.
In short, the intrinsic dimensionality of a dataset is the least number of variables required to satisfactorily describe the data points. Ideally, if data lie on a linear manifold (a hyperplane), the intrinsic dimensionality coincides with the number of principal components required to completely explain their variance. In a more realistic setting, data lie on curved, nonlinear manifolds, and linear estimators like PCA fail to capture their real intrinsic dimensionality. In such cases, nonlinear ID estimators can be employed. Among them, TwoNN \citep{facco2017estimating} is chosen because of its efficiency and stability on complex and non-uniform manifolds.

\paragraph{Experimental setting.}
The intrinsic dimensionality of head representations is evaluated across multiple transformer-based vision architectures pretrained with different objectives (supervised, unsupervised, self-supervised). Namely, OpenAI's CLIP \citep{clip}, OpenCLIP \citep{cherti2023reproducible}, BLIP \citep{li2022blip}, ViT \citep{dosovitskiy2021an} and DINOv2 \citep{oquab2024dinov} are employed, all in their version based on ViT-Large. These models are fed a subset of the training set of \gls{imagenet1k} \citep{deng2009imagenet} containing 80000 images stratified on the class labels, and representations are extracted for all attention heads. The intrinsic dimensionality of such representations is then computed using a linear estimator (PCA) and a nonlinear one (TwoNN). Linear ID is computed as the number of components needed by PCA to explain 99\% of head variance.
\begin{figure}[h]
\centering
 \includegraphics[width=\textwidth]{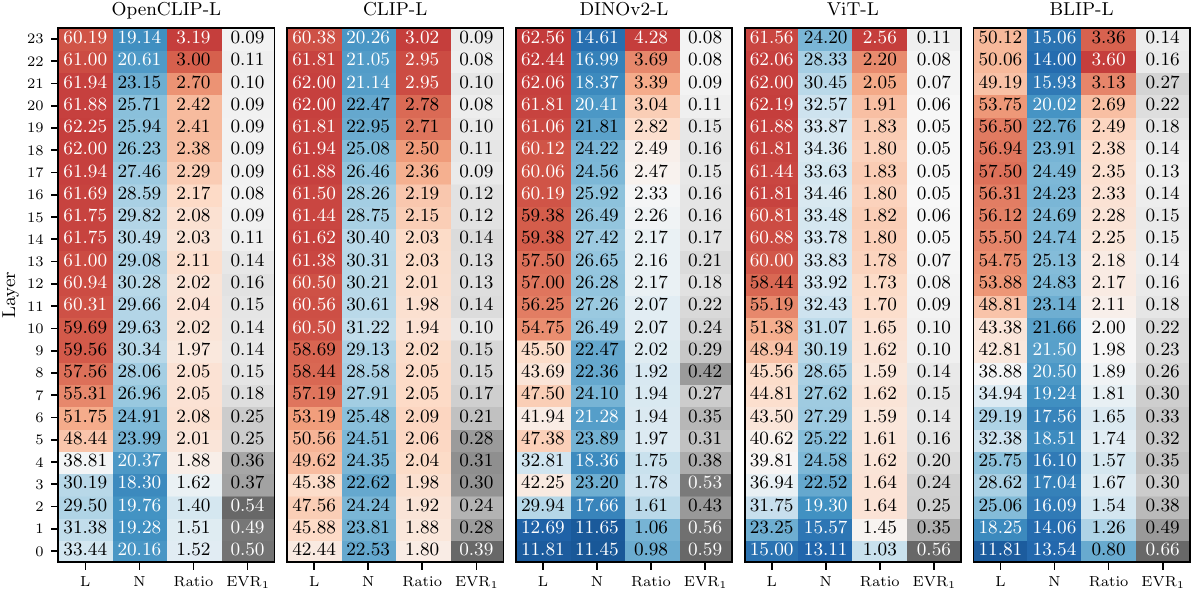}
\caption{Heads in early layers show low-dimensional, linear structures, as suggested by similar intrinsic dimension estimates from PCA and TwoNN (column labels L and N). Moving toward the output layer, the nonlinear dimensionality peaks and then decreases, while PCA's linear estimate continues to rise, indicating increasing nonlinearity in head manifolds ($\text{Ratio} = \frac{L}{N}$). The first principal component ({EVR$_1$}) explains around 50\% of the variance in early layers, dropping to around 10\% in later layers.
}
\label{fig:id_heads}
\end{figure}
\paragraph{Result analysis.} \cref{fig:id_heads} reports the results for head units on \gls{imagenet1k}. The \textit{true} head dimensionality (computed with a nonlinear estimator, TwoNN) tends to increase in the first half of the model and to decrease towards the last few layers, following a characteristic hunchback shape, similar to previous findings in other vision architectures \citep{ansuini2019intrinsic}. However, the number of dimensions returned by the linear estimator grows constantly through the model. This disparity, reflected in the growing ratio between the two estimates, suggests that the units in the early layers are close to linear, while those in the later layers lie on more curved manifolds. The last column shows the average explained variance ratio (EVR) of the first PCA component and highlights that heads in the first layers are largely explained by this direction, while it still accounts for a nontrivial 10\% of head variance in late layers.

These findings highlight that low head dimensionality and monotonically increasing nonlinearity arise along the residual streams of vision transformers, regardless of pre-training objective and data.

\subsection{Principal Components Encode Unit Semantics}\label{ssec:omp}
The low dimensionality of head encodings results in relevant consequences for their interpretability. Head representations can be easily approximated with sparse recovery algorithms in a way that is akin to performing PCA, but over a discrete set of vectors.
For CLIP models, this approach has been recently explored by \citet{gandelsman2024interpreting}. There, the authors introduce a sparse approximation algorithm, TextSpan (TS), and decompose head encodings using a set of textual descriptions coming from the text branch of CLIP as a dictionary. They observe strong specialization properties, witnessed by high coherence in the textual explanations of each head.

TextSpan is here linked to the more established family of Matching Pursuit (MP) \citep{mallat1993matching} algorithms, widely employed in signal processing. More specifically, TextSpan is analogous to Simultaneous Orthogonal Matching Pursuit (SOMP) \citep{tropp2006algorithms}, with light modifications; see the appendix of \citet{basile2025residual}.
TextSpan, like any MP algorithm, approximates the signal through linear combinations of basis functions.
Considering the high nonlinearity of later layers (\cref{ssec:residual_dimensionality}), and TS being a linear sparse approximation method, the investigation focuses on whether TS is, in reality, focusing on the first principal components of the signal (head-level representations).

\paragraph{Experimental setting.}
This experiment follows the setup of the original TS paper \citep{gandelsman2024interpreting}. The attention heads belonging to the last 4 layers of OpenCLIP-L are considered. Two sparse approximation algorithms are employed: the original TextSpan, which operates on the whole head representation, and Orthogonal Matching Pursuit (OMP) \citep{pati1993orthogonal}. Different from TextSpan, OMP computes sparse approximations of vectors, and head representations are matrices. Therefore, OMP is applied to the first principal component of each head. This method is denoted as OMP$_1$.
The dictionary contains the encodings produced by OpenCLIP-L for the set of image descriptions provided by \citet{gandelsman2024interpreting}. Both algorithms are applied to select 5 descriptions for each head and compute an agreement score between the two sets. The agreement is computed as the absolute Z-score of the cosine similarity ($\simf$) between them, compared with the average cosine similarity $\mu$ between the descriptions selected by TextSpan and the entire dictionary, with $\sigma$ its standard deviation: $Z=\frac{|{\simf(\text{TS}, \text{OMP}_1)}-\mu|}{\sigma}$.

\begin{figure}[h]
 \centering
 \begin{minipage}{0.23\textwidth}
  \includegraphics[width=\textwidth]{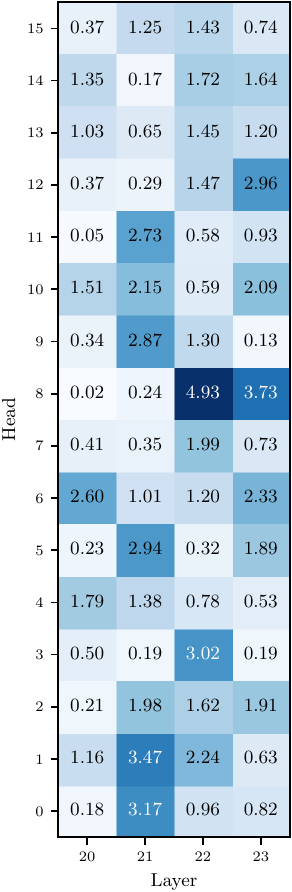} %
 \end{minipage}%
 \begin{minipage}{0.7\textwidth}
 \resizebox{\linewidth}{!}{
  \begin{tabular}{ll}
\toprule
\textbf{TextSpan} & \textbf{OMP$_1$} \\
\midrule
\textbf{L20.H8} (``Scenery'') & \textbf{L20.H8} ($Z=0.02$) \\
\midrule
Photo taken in Galápagos Islands & Unspoiled beauty \\
Image taken in Norway & Picture taken in Portugal \\
Evocative beauty & Crisp autumn leaves \\
Vibrant urban energy & Evocative candid gaze \\
A skirt & Picture taken in Cyprus \\
\midrule
\textbf{L21.H11} (``Location'') & \textbf{L21.H11} ($Z=2.73$) \\
\midrule
Picture taken in Cyprus & Picture taken in Cyprus \\
Picture taken in Ontario, Canada & Picture taken in the Canadian lakes \\
Photo taken in Rio de Janeiro, Brazil & Image taken in the Florida Everglades \\
Photo captured in the Arizona desert & Image taken in New England \\
Picture captured in the Scottish highlands & Warm and cozy indoor scene \\
\midrule
\textbf{L22.H8} (``Letters'') & \textbf{L22.H8} ($Z=4.93$) \\
\midrule
A photo with the letter F & A photo with the letter F \\
A photo with the letter V & A photo with the letter P \\
A photo with the letter D & A photo with the letter T \\
A photo with the letter T & A photo with the letter X \\
A photo with the letter X & A photo with the letter B \\
\midrule
\textbf{L23.H2} (``Animals'') & \textbf{L23.H2} ($Z=1.91$) \\
\midrule
Image showing prairie grouse & Picture of a feline \\
Image with a penguin & An image with dogs \\
A magnolia & Photo of a furry animal \\
An image with dogs & Photo taken in Grand Canyon \\
An image with cats & An image with cats \\
\bottomrule
\end{tabular}
}
\vspace{0.9cm}
 \end{minipage}
 \caption{Comparison between TextSpan and Orthogonal Matching Pursuit on the first principal component (OMP$_1$), applied to the heads of OpenCLIP-L. \textit{Left}: agreement score between the descriptions returned by the two methods. \textit{Right}: qualitative comparison of selected descriptions for 4 heads, one per layer, at different agreement levels.}
 \label{fig:ts_omp}
\end{figure}

\paragraph{Result analysis.} The left panel of \cref{fig:ts_omp} reports the agreement scores. The right panel reports a few examples (one per layer) of descriptions obtained using the two algorithms.
A high Z-score (e.g., head 8 of layer 22, which is almost $5\sigma$ away from $\mu$) is reflected in extremely similar descriptions from the two methods.
When the agreement is lower, as in the case of head 8 of layer 20, the two sets of descriptions substantially differ even though they share some high-level semantics.

Overall, this analysis indicates that, in some cases, the first principal component captures nearly all the essential information about the head's specialized semantics. In other cases, the head's role appears to be distributed across multiple components.

\subsection{Spectral Dataset Comparison}\label{ssec:cross_dataset}
The aim is to understand to what extent specialization generalizes across different input data distributions.
To do so, a spectral metric is introduced to compare the representations of residual units. Since units are low-dimensional (\cref{ssec:residual_dimensionality}) and their specialization is deeply impacted by a few principal components (\cref{ssec:omp}), a metric is defined to quantify the similarity between their PCA bases, inspired by principal angles \citep{principal_angles}.

Let \(\mathcal{S}_1\) and \(\mathcal{S}_2\) be two subspaces of dimensions \(p_1\) and \(p_2\) in a \(d\)-dimensional space. The principal angles \(\theta_r\) for \(r = 1, \dots, \min(p_1, p_2)\) are given by:
\begin{equation}
\cos \theta_r = \max_{\substack{\vu \in \mathcal{S}_1,\ \vu \perp \mathcal{P}_{r-1} \\ \vv \in \mathcal{S}_2,\ \vv \perp \mathcal{Q}_{r-1}}} \cos(\vu, \vv) \,,
\end{equation}
where \(\mathcal{P}_{r-1}\) and \(\mathcal{Q}_{r-1}\) are the spans of the earlier maximisers \(\vu_1^\star, \dots, \vu_{r-1}^\star\) and \(\vv_1^\star, \dots, \vv_{r-1}^\star\), respectively, so that \(\vu \in \mathcal{S}_1\) is restricted to \(\vu \perp \mathcal{P}_{r-1}\) and \(\vv \in \mathcal{S}_2\) to \(\vv \perp \mathcal{Q}_{r-1}\).

Let \(\mathcal{V}_1 = \{\vu_1, \dots, \vu_{p_1}\}\) and \(\mathcal{V}_2 = \{\vv_1, \dots, \vv_{p_2}\}\) represent sets of $\ell_2$-normalized discrete vectors (e.g., principal components) with (optional) associated weights \(w_{1,i}\) and \(w_{2,j}\) (e.g., singular values).

The \textbf{spectral cosine similarity} $s_r$ for \(r = 1, \dots, \min(p_1, p_2)\) is defined as:
\begin{equation}
(i_r, j_r) = \argmax_{\substack{i \notin \{i_1, \dots, i_{r-1}\} \\ j \notin \{j_1, \dots, j_{r-1}\}}} \vu_i^\top \vv_j \,, \qquad s_r = (\vu_{i_r}^\top \vv_{j_r})\, w_{1,i_r} w_{2,j_r} \,,
\end{equation}
where \(i_1, \dots, i_{r-1}\) and \(j_1, \dots, j_{r-1}\) are previously selected indices.

The original principal angles measure the alignment between \textbf{subspaces} by maximizing the cosine similarity of vectors in their \textbf{span}. In the discrete case, vectors are directly selected from sets \(\mathcal{V}_1\) and \(\mathcal{V}_2\) with optional weighting.
The final measure is the aggregation of the spectral cosine similarities along the $\min(p_1, p_2)$ entries, with normalization to bound the measure between 0 and 1 (weights sorted in decreasing order).
The \textbf{normalized spectral cosine similarity} between the two sets of principal components is defined as:
\begin{equation}
\label{eq:spectral_cosine}
\simf(\mathcal{V}_1, \mathcal{V}_2) =
\sqrt{\frac{\sum_{r=1}^{\min(p_1, p_2)} s_r^2}{\sum_{r=1}^{\min(p_1, p_2)} (w_{1,r} w_{2,r})^2}} \,.
\end{equation}
This measure is applied to compare residual units across different input datasets. The main advantage of this formulation, compared to standard approaches to representation similarity, is that the metric does not rely on the alignment between samples, as it operates in the dual spectral space. This property is crucial in the application to different datasets, which even vary in size.

\paragraph{Experimental setting.}
The same ViT-based encoders of \cref{ssec:residual_dimensionality} are considered, along with 14 different datasets:
\gls{imagenet1k} (the same split used in \cref{ssec:residual_dimensionality}), \gls{cifart}/\gls{cifarh} \citep{Krizhevsky2009-hv}, ImageNet-Sketch \citep{wang2019learning}, Cars \citep{krause20133d}, \gls{mnist} \citep{Lecun1998-we}, \gls{svhn} \citep{netzer2011reading}, EuroSAT \citep{helber2019eurosat}, RESISC45 \citep{cheng2017remote}, DTD \citep{cimpoi2014describing}, SUN397 \citep{xiao2016sun}, GTSRB \citep{stallkamp2011german}, PACS \citep{li2017deeper}, and random images (10000 samples with RGB values in $[-1,1]$). The original train/validation/test splits are used if available, otherwise the splits are produced through a stratified random sampling over the classes. For each encoder, the similarity measure is used to compare its unit representations produced on each training dataset with the ones obtained on the training split of \gls{imagenet1k}. \gls{imagenet1k} is taken as a reference under the assumption that being a general enough dataset, its head PCA bases are sufficiently comprehensive to approximate the primary features across other datasets. Additionally, a qualitative inspection of a few heads that stand out is performed by finding their textual decomposition. For this step, Simultaneous Orthogonal Matching Pursuit (SOMP) is used, having established its strong relationship with TextSpan (see the appendix of \citet{basile2025residual}).

\paragraph{Result analysis.}

\begin{figure}[h]
 \centering
 \begin{subfigure}{0.87\linewidth}
  \includegraphics[width=\linewidth]{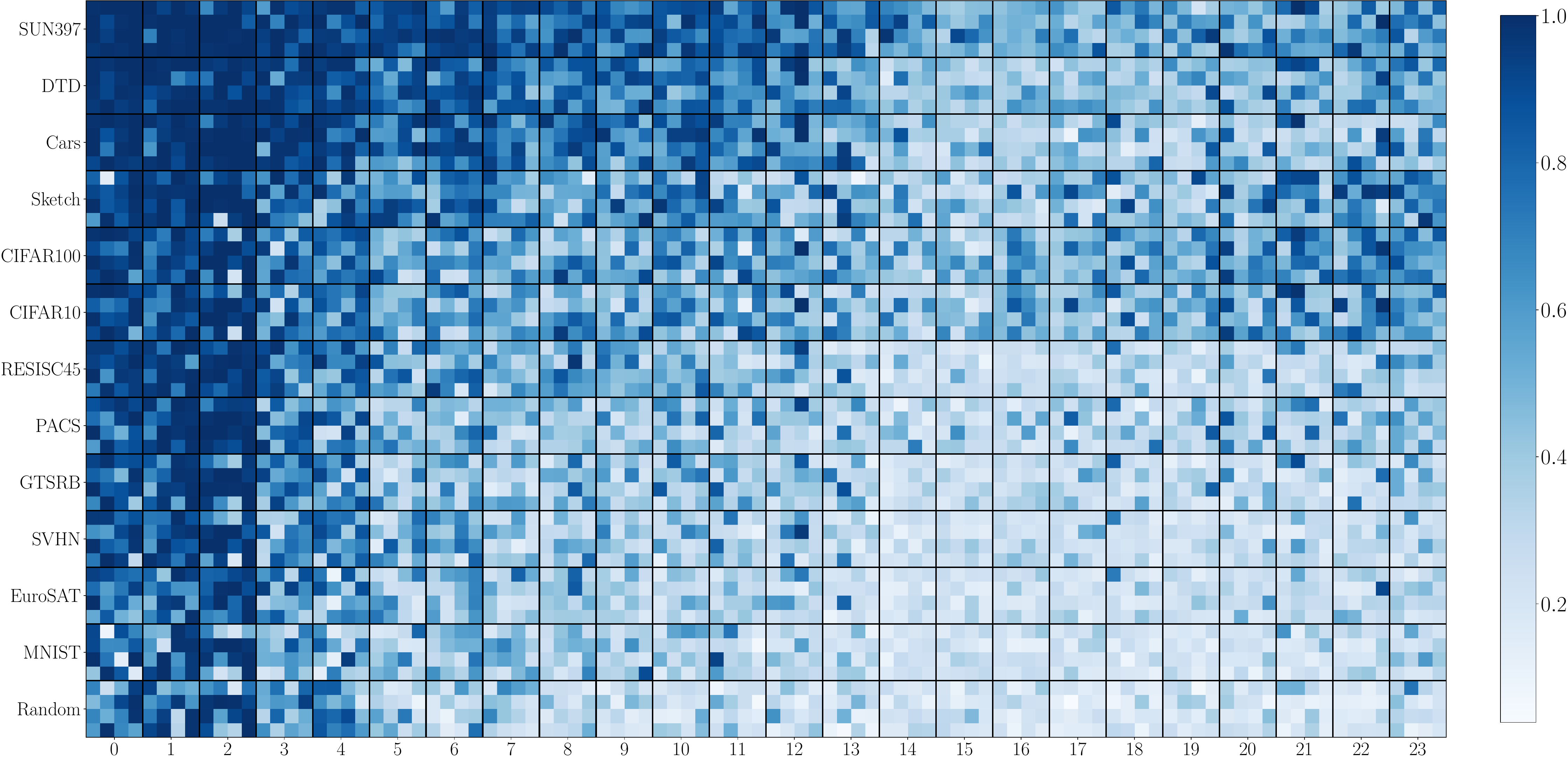}
  \caption{}
  \label{fig:head_comparison_a}
 \end{subfigure}

 \begin{subfigure}{0.9\linewidth}
  \centering
  \resizebox{\linewidth}{!}{
  \begin{tabular}{lll}
   \toprule
   \textbf{L22.H7} (``Seasons'') & \textbf{L22.H11} (``Grayscale'') & \textbf{L23.H10} (``Numbers'') \\
   \midrule
   Serene winter wonderland & A charcoal gray color & Image with six subjects \\
   A photo taken in the summer & Minimalist white backdrop & An image of three subjects \\
   A photo taken in the fall & Sepia-toned photograph & An image of the number 9 \\
   A photo taken in the spring & An amber color & The number fifteen \\
   Serene garden oasis & High-contrast black and white & Image with four people \\
   \bottomrule
  \end{tabular}
  }
  \caption{}
  \label{tab:descriptions_exp2}
 \end{subfigure}
 \caption{\textit{(a)} Attention head similarity across layers of OpenCLIP-L, computed between \gls{imagenet1k} head representations and those obtained on other datasets. \textit{(b)} Descriptions picked by SOMP for three specialized heads found in the analysis of panel \textit{(a)}.}
 \label{fig:head_comparison}
\end{figure}

\Cref{fig:head_comparison} reports the results of the spectral head-to-head comparison between \gls{imagenet1k} and all other datasets on OpenCLIP-L. Rows are ordered according to the mean overall similarity between the corresponding dataset and \gls{imagenet1k}. Dataset ordering is consistent across different encoders. The mean correlation coefficient between dataset similarities (averaged over all heads) across different models is 0.97.
On OpenCLIP-L, datasets that maximally align with \gls{imagenet1k} share classes with it (e.g., SUN397 and Sketch) and/or contain generic images (e.g., DTD and Cars). Moreover, datasets that share the same input image structure and concepts (\gls{cifart} and \gls{cifarh}) have an almost identical similarity distribution across heads.
Overall, a decreasing trend in head similarity scores as depth in the model increases is observed. The simple, linear heads of the first layers are responsible for the extraction of low-level patterns \citep{dosovitskiy2021an} and are almost always identical across different data distributions. On the last layers, just a few heads per dataset stand out: this is where specialization is found. Zooming in on a few of these heads, \cref{tab:descriptions_exp2} reports their textual descriptions obtained with SOMP. Head 7 of layer 22 (specialized on seasons) stands out because it is highly activated in many datasets that contain pictures of scenery (such as SUN397, GTSRB, and, more prominently, EuroSAT). Head 11 of layer 22 (specialized in shades of gray) is extremely different between \gls{imagenet1k} and Sketch, which contains grayscale drawings of \gls{imagenet1k} classes. Head 10 of layer 23 (specialized in numbers) is highly activated on both \gls{mnist} and \gls{svhn}. Other heads are shared between the two, such as head 1 of the same layer, but they are also activated by the random dataset, signaling that they are not as specific.

These findings show that the similarity measure yields intuitive scores, with \gls{imagenet1k}'s foundational attributes demonstrating generalizability across various data distributions.

\section{ResiDual Alignment}
With a refined understanding of head specialization, the objective is now to use this property to enhance alignment between visual unit representations (both heads and MLPs) and text encodings in CLIP-like models. From this point onwards, the experiments share a common objective: given a frozen zero-shot classifier (e.g., text encodings or prototypes for classes), manipulating the residual to improve its alignment with this target subspace. The focus is on multimodal architectures, where improved alignment directly benefits tasks such as zero-shot classification by strengthening the model's capacity to interpret visual data through text-based descriptors.

\subsection{Coarse Unit Alignment}\label{ssec:coarse_selection}

The exploration begins by investigating whether certain heads are already aligned with the text subspace relevant to the task.
Specifically, the goal is to optimize vision-text alignment by combining visual heads weighted using various scoring functions. These functions assign weights to entire head representations to modulate (in some cases completely remove) their contributions to the residual stream. Crucially, these methods operate exclusively at the level of whole-head reweighting, leaving the internal structure and intrinsic specialization of individual heads unchanged. As a result, only the output specialization is affected.

\paragraph{Experimental setting.}

Three different selection methods are employed: i) \textbf{Unsupervised (U):} The head-to-output correlation is used as a measure. Intuitively, the more one head correlates to the full output, the more information it carries. In practice, the Pearson correlation is computed between each sample at the head level and its corresponding output encoding, averaging across samples to obtain a scalar; ii) \textbf{Task-conditioned Unsupervised (U|T)}: Conditioning on the output alone does not necessarily imply that the selected heads will be suitable for a given task. Since the task is modeled by a text subspace, having one encoding for each class, the previous unsupervised measure can be conditioned to be applied only on the head and output subspaces spanned by the task encodings. This has the effect of ignoring features that might influence the correlation but are not related to the task at hand. This is a direct application of the CompAttribute metric introduced by \citet{balasubramanian2024decomposing}; iii) \textbf{Supervised (S):} When both the task encodings and labeled samples are assumed to be available, the head score can be directly estimated by looking at its performance on the downstream task (in the style of logit lens \citep{nostalgebraist2020interpreting}).

For each scoring function, each head is evaluated individually, ranked according to its score, and a greedy top-$k$ selection is applied. These selected heads $\mathcal{U}'_{\mathrm{attn}}\subseteq \mathcal{U}_{\mathrm{attn}}$ are then summed to create a partially recovered residual, which is subsequently evaluated on downstream task performance:
\begin{equation}
 \mY' = \sum_{\mH \in \mathcal{U}'_{\mathrm{attn}}} \mH \,.
\end{equation}
Three control measures are in place: i) \textbf{Heads (H):} The performance of the model when \textit{all} the attention heads, and only them, are used. This gives information about the heads' contribution to the residual and, symmetrically, how much the final performance depends on the MLP units; ii) \textbf{Random (R):} The average performance over 10 independent random samplings of $k$ \textit{head units}. This can be seen as a lower bound on the expected performance; iii) \textbf{Base (B):} The original performance of the model without any modification to its residual. Intuitively, this could represent a theoretical upper bound on the performance if there are no task-aligned units.

The greedy selection strategy scores heads independently, disregarding inter-head relationships. To make the selection aware of them, a scalar weight for each head is optimized simultaneously using gradient descent, providing an empirical upper bound for the selection performance. This procedure is referred to as \textbf{Optimized~(O)}.

These unit selection strategies are evaluated on two CLIP-like models (BLIP-L and OpenCLIP-L) and 10 of the datasets of \cref{ssec:cross_dataset}. The value of $k$ for the greedy and random selection is chosen so that 5\% of the total heads are considered.

\begin{table}
\footnotesize
\setlength{\tabcolsep}{3pt}
\centering
\begin{tabular}{lcccccccccccccc}
\toprule
 & \multicolumn{7}{c}{\textbf{BLIP-L}} & \multicolumn{7}{c}{\textbf{OpenCLIP-L}} \\
 \cmidrule(lr){1-1}\cmidrule(lr){2-8}\cmidrule(lr){9-15}

\textbf{Dataset}  &  U & U|T & S & R & H & B & O & U & U|T & S & R & H & B & O \\
\cmidrule(lr){1-1}\cmidrule(lr){2-5}\cmidrule(lr){6-8}\cmidrule(lr){9-12}\cmidrule(lr){13-15}
CIFAR-10 & 0.94 & 0.94 & 0.93 & 0.56 & 0.93 & 0.94 & 0.96 & 0.96 & 0.96 & 0.96 & 0.59 & 0.94 & 0.97 & 0.98 \\
CIFAR-100 & 0.70 & 0.71 & 0.72 & 0.33 & 0.69 & 0.71 & 0.75 & 0.80 & 0.78 & 0.79 & 0.38 & 0.76 & 0.82 & 0.84 \\
Cars & 0.67 & 0.68 & 0.71 & 0.27 & 0.66 & 0.72 & 0.77 & 0.92 & 0.92 & 0.93 & 0.39 & 0.89 & 0.93 & 0.93 \\
DTD & 0.52 & 0.54 & 0.54 & 0.27 & 0.50 & 0.55 & 0.61 & 0.59 & 0.59 & 0.59 & 0.29 & 0.54 & 0.63 & 0.69 \\
EuroSAT & 0.49 & 0.53 & 0.53 & 0.24 & 0.37 & 0.50 & 0.92 & 0.64 & 0.64 & 0.67 & 0.34 & 0.55 & 0.64 & 0.95 \\
GTSRB & 0.34 & 0.34 & 0.37 & 0.14 & 0.33 & 0.35 & 0.59 & 0.55 & 0.56 & 0.55 & 0.20 & 0.49 & 0.56 & 0.75 \\
MNIST & 0.62 & 0.64 & 0.65 & 0.27 & 0.41 & 0.52 & 0.94 & 0.74 & 0.84 & 0.85 & 0.25 & 0.41 & 0.54 & 0.97 \\
RESISC45 & 0.58 & 0.57 & 0.60 & 0.28 & 0.56 & 0.59 & 0.79 & 0.70 & 0.69 & 0.70 & 0.33 & 0.63 & 0.73 & 0.86 \\
SUN397 & 0.67 & 0.66 & 0.68 & 0.28 & 0.65 & 0.70 & 0.73 & 0.70 & 0.69 & 0.71 & 0.27 & 0.59 & 0.74 & 0.76 \\
SVHN & 0.25 & 0.36 & 0.42 & 0.16 & 0.21 & 0.33 & 0.56 & 0.48 & 0.57 & 0.53 & 0.24 & 0.41 & 0.41 & 0.70 \\
\midrule
\textbf{Average} & 0.58 & 0.60 & \textbf{0.61} & 0.28 & 0.53 & 0.59 & 0.76 & 0.71 & \textbf{0.73} & \textbf{0.73} & 0.33 & 0.62 & 0.70 & 0.84 \\
\bottomrule

\end{tabular}

\caption[Zero ablation of all units except the top 5\% of attention heads.]{Accuracy when doing \textbf{zero ablation of all units except top 5\%} of attention heads. Heads are assigned a binary weight using an Unsupervised (U), Task-conditioned Unsupervised (U|T), Supervised (S), and Random (R) strategy (a mean over 10 different seeds). H corresponds to using all the attention heads available, B is the original model performance, and O is the optimized continuous weighting case. Bold: best average among U, U\textbar T, S.}\label{tab:residual:ablation}
\end{table}

\paragraph{Result analysis.}
As reported in \cref{tab:residual:ablation}, the unsupervised scoring (\textbf{U}) performs unexpectedly well despite not explicitly considering the task. A likely reason is that core task information is often embedded in the first few principal components (PCs) of the output. By aligning with this information using only a few heads, a dual benefit is achieved: preserving essential task-relevant information while effectively filtering out noise. In fact, adding the conditioning on the task (\textbf{U|T}) is just slightly beneficial in terms of performance, with the exception of \gls{svhn} (and \gls{mnist} on OpenCLIP-L). The vast majority (on average around 90\%) of alignment between task and residual comes from the head contributions, as witnessed by the similarity between the columns \textbf{H} and \textbf{B}. The optimized selection strategy (\textbf{O}) is extremely powerful, having the best score among them, and sometimes almost doubling the original model performances (\textbf{B}), showing how task-relevant information is already present in the residual, just hidden.

These results illustrate that retaining only task-aligned units (effectively ``panning for the gold'' contained in the residual) is highly effective across the board.

\subsection{Spectral ResiDual Alignment}
\label{sec:residual_alignment}
Given that: i) unit specialization is essentially encoded in the principal components (\cref{ssec:omp}); ii) components of general enough data distributions (e.g., \gls{imagenet1k}) capture specialized behavior even on other datasets (\cref{ssec:cross_dataset}); iii) retaining only task-aligned units is beneficial for image-text alignment (\cref{ssec:coarse_selection}), a method to directly filter information along the residual at the spectral level is proposed: \textbf{ResiDual}.

Starting from the decomposition formula for the residual stream (\cref{eq:decomposition}), anisotropic scaling of each unit representation is allowed. Specifically, given a unit representation $\mU$, its corresponding principal component basis $\mPhi$, and associated mean $\vmu$, the ResiDual transformation of $\mU$ is defined as:
\begin{equation}\label{eq:residual}
 \operatorname{RD}_{\mPhi, \vmu}(\mU, \vlambda) = \mPhi^\top\operatorname{diag}(\vlambda)\mPhi(\mU-\vmu)^\top \,,
\end{equation}
where the learnable vector $\vlambda$ contains the weights associated with each principal component, and the rows of $\mPhi$ are the principal components of the unit, with $\mU \in \R^{n \times d}$ (rows are samples), $\vmu$ subtracted from every row, and the result transposed back to $n \times d$ before the sum. When all of them are kept (RD), $\mPhi$ is orthogonal and $\mPhi^\top = \mPhi^{-1}$; when only the leading ones are kept (the truncated variant RD$^*$ below), $\mPhi$ has orthonormal rows and $\mPhi^\top$ is its pseudo-inverse. \citet{basile2025residual} write $\mPhi^{-1}$ in both cases. Then, the transformation is applied to every residual unit independently, resulting in a transformed version $\widehat{\mY}$ of the output $\mY$:
\begin{equation}
 \widehat{\mY} = \sum_{i=1}^{\lvert\mathcal{U}\rvert} \operatorname{RD}_{\mPhi_i, \vmu_i}(\mU_i, \vlambda_i) \,.
\end{equation}
In summary, ResiDual models a simple spectral anisotropic scaling of residual units that results in more complex dynamics in the output space. Summed over units, the map $\operatorname{RD}$ realises $\OperatorSameSpace_\Corr$ on the representation space of the vision encoder, with $\Corr$ pairing the residual units with the text encodings that define the task.
The effectiveness of this method is extensively evaluated across a variety of configurations, models and datasets.
\paragraph{Experimental setting.} ResiDual is evaluated in 3 different configurations: i) \textbf{RD}, as presented in the original formulation; ii) \textbf{RD}$^*$, a version of ResiDual constrained on the number and type of units (restricted to heads) and the number of considered components (PCA bases truncated to explain 90\% of the variance). This is done because head units, and specifically their principal components, are strongly connected to specialization; iii) \textbf{RD$^\mY$}, where the ResiDual transformation is applied directly on the output encoding $\mY$, to assess whether output components can already be aligned with the task.
For all configurations, \gls{imagenet1k} is selected as a reference for the PCA bases $\mPhi$ and unit means $\vmu$ that appear in \cref{eq:residual}.

ResiDual is compared with 3 reference alignment methods: i) \textbf{Base}: the base zero-shot performance of the model, relying on the alignment coming from pre-training; ii) \textbf{Full Fine-tuning}: its score is considered the empirical upper bound for alignment. It is obtained by fine-tuning the whole vision transformer with frozen text encodings; iii) \textbf{Linear Aligner (Lin)}: the performance of a trained linear transformation of the output shows to what extent output alignment could be linearly recovered. This approach is well-supported by recent studies (\cref{ch:relrep,ch:semantic_alignment}), which indicate that even independently trained models with different architectures can often be aligned by a simple linear transformation \citep{moschella2023relative, maiorca2023semantic, norelli2023asif, lahner24a, balasubramanian2024decomposing}.

For this experiment, 3 CLIP-like models are employed: BLIP-L, CLIP-L, and OpenCLIP-L, tuned on the 10 datasets employed in \cref{ssec:coarse_selection}. All training runs use the Schedule-Free Adam optimizer \citep{defazio2024road} with the automatic learning rate finder by \citet{smith2017cyclical}, implemented in PyTorch Lightning \citep{Lightning}. The maximum number of epochs is 30, with an early-stopping policy on the validation set accuracy with patience of 5 epochs.
\begin{figure}
\centering
 \includegraphics[width=\textwidth]{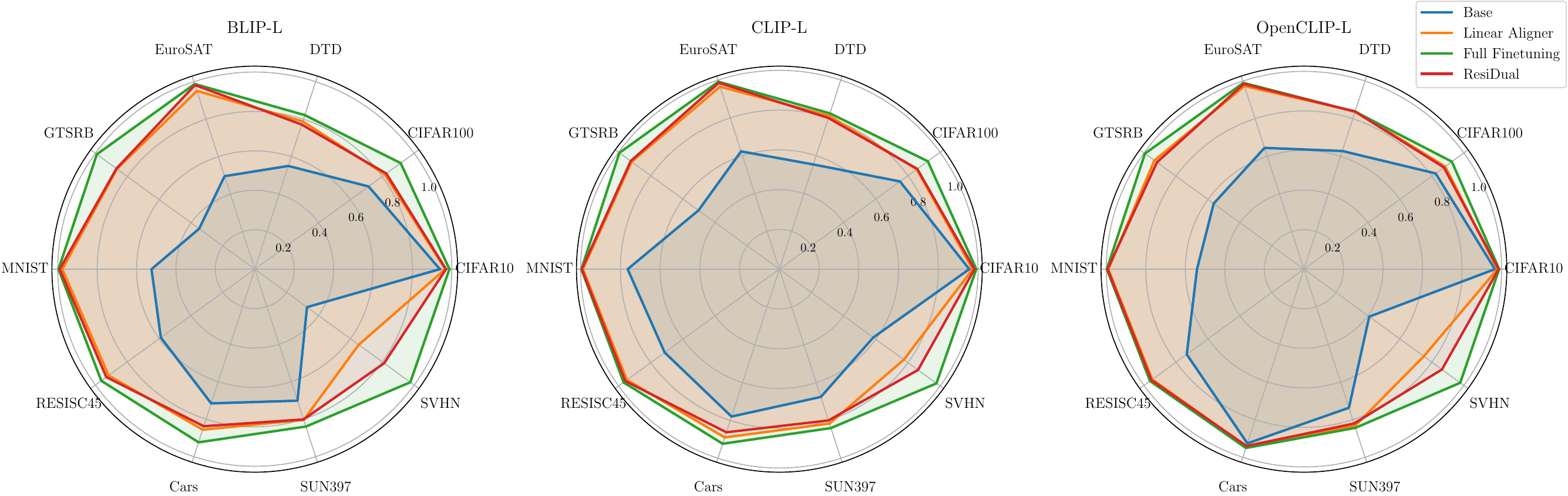}
\caption{Performance comparison between text-image alignment methods on zero-shot classification tasks.
}
\label{fig:residual}
\end{figure}

\paragraph{Result analysis.} \cref{fig:residual} reports a comparative analysis of ResiDual against reference alignment methods. A linear transformation of the output is sufficient to approximate full fine-tuning performance on most datasets. The spectral residual transformation modeled by ResiDual attains comparable (and, in some cases, better) results than the linear aligner on all datasets. These statements hold true across all models. Specifically, the case of \gls{svhn} stands out: on this input dataset, ResiDual has an advantage of 0.09--0.16 accuracy over the linear aligner on all models. This gap is hypothesized to be due to the absence of task-relevant features at the output level. This is confirmed by the results in \cref{tab:residual}: on \gls{svhn}, \textbf{RD$^\mY$} has one of the largest gaps from \textbf{RD}, meaning that output components are not well aligned with the task. While having approximately 30\% of the learnable parameters, \textbf{RD$^*$} achieves comparable results to \textbf{RD}, indicating that applying the ResiDual procedure to heads (and not MLPs) and their first principal components alone is sufficient. Moreover, in the cases where the performance of the original model is already satisfactory (e.g., \gls{cifart}), applying ResiDual does not compromise alignment.

Overall, these results show that, through spectral-level operations along the residual, ResiDual builds a concise yet expressive transformation that bridges the modality gap, closely approximating full fine-tuning-level performance on most datasets, even in its more parameter-efficient configuration.

\begin{table}[t]
\small
\setlength{\tabcolsep}{3pt}
\centering
\begin{tabular}{lcccccccccccc}
\toprule
 & \multicolumn{4}{c}{\textbf{BLIP-L}} & \multicolumn{4}{c}{\textbf{CLIP-L}} & \multicolumn{4}{c}{\textbf{OpenCLIP-L}} \\
\cmidrule(lr){2-5}\cmidrule(lr){6-9}\cmidrule(lr){10-13}
\textbf{Dataset} & \textbf{Lin} & \textbf{RD} & \textbf{RD$^*$} & \textbf{RD$^\mY$}& \textbf{Lin} & \textbf{RD} & \textbf{RD$^*$} & \textbf{RD$^\mY$} & \textbf{Lin} & \textbf{RD} & \textbf{RD$^*$} & \textbf{RD$^\mY$} \\
\midrule
 CIFAR-10 & 0.97 & 0.97 & 0.97 & 0.96 & 0.97 & 0.98 & 0.98 & 0.97 &  0.98 &  0.98 &  0.98 &  0.98 \\
CIFAR-100 & 0.82 & 0.83 & 0.81 & 0.74 & 0.85 & 0.86 & 0.85 & 0.80 &  0.88 &  0.88 &  0.87 &  0.85 \\
  DTD & 0.79 & 0.77 & 0.76 & 0.58 & 0.81 & 0.80 & 0.79 & 0.63 &  0.84 &  0.84 &  0.83 &  0.69 \\
 EuroSAT & 0.95 & 0.98 & 0.98 & 0.83 & 0.97 & 0.99 & 0.98 & 0.95 &  0.97 &  0.98 &  0.98 &  0.94 \\
 GTSRB & 0.87 & 0.87 & 0.84 & 0.59 & 0.92 & 0.92 & 0.92 & 0.77 &  0.94 &  0.92 &  0.92 &  0.78 \\
 MNIST & 0.98 & 0.99 & 0.99 & 0.93 & 0.99 & 0.99 & 0.99 & 0.97 &  0.99 &  0.99 &  0.99 &  0.97 \\
RESISC45 & 0.92 & 0.93 & 0.92 & 0.77 & 0.95 & 0.96 & 0.95 & 0.87 &  0.95 &  0.95 &  0.95 &  0.89 \\
 Cars & 0.86 & 0.84 & 0.82 & 0.75 & 0.89 & 0.86 & 0.85 & 0.81 &  0.94 &  0.94 &  0.94 &  0.93 \\
 SUN397 & 0.80 & 0.80 & 0.77 & 0.72 & 0.82 & 0.80 & 0.77 & 0.70 &  0.83 &  0.82 &  0.80 &  0.75 \\
 SVHN & 0.65 & 0.81 & 0.77 & 0.53 & 0.77 & 0.86 & 0.86 & 0.70 &  0.75 &  0.86 &  0.85 &  0.64 \\
\midrule
\textbf{Average} & 0.86 & \textbf{0.88} & 0.86 & 0.74 & 0.89 & \textbf{0.90} & \textbf{0.90} & 0.82 & 0.91 & \textbf{0.92} & 0.91 & 0.84 \\
\midrule
\textbf{\#params} & 65.8k & 30.7k & 8.3k & 256 & 590k & 43k & 14k &768 & 590k & 43k & 13.2k & 768 \\
\bottomrule
\end{tabular}
\caption{Accuracy produced by different configurations of ResiDual, compared with a linear aligner. (Lin): linear aligner at the output level; (RD): ResiDual in the original formulation (all principal components of all residual units); (RD$^*$): ResiDual limited to head units alone (no MLPs), and PCs truncated to 90\% of explained variance; (RD$^\mY$): ResiDual applied to the output encoding. Bold: best average per model.}
\label{tab:residual}
\end{table}

\chapter[Head Pursuit]{Head Pursuit: Probing Attention Specialization in Multimodal Transformers}
\label{ch:headpursuit}

\chapterpublication[https://github.com/lorenzobasile/HeadPursuit]{headpursuit}{\textbf{NeurIPS 2025} \textit{(\textcolor{blue}{Spotlight})}}{Sparse decomposition over the unembedding matrix reveals concept-specialized heads, enabling targeted interventions on LLMs and VLMs.}

\section{Introduction}
This chapter presents HeadPursuit, a \gls{correspondence-coupled} operator ($\gls{OperatorSameSpace}_{\gls{Corr}}$, \cref{ch:methodology}) that applies the Simultaneous Orthogonal Matching Pursuit (SOMP) framework, related to TextSpan in \cref{ch:residual}, to attention head analysis in generative transformers. The core \gls{correspondence-concept} $\Corr$ links attention head outputs to interpretable semantic directions from the unembedding matrix---a fixed dictionary where each row is a natural language token embedded in latent space. By decomposing residual stream contributions over this semantic dictionary, SOMP reveals which heads specialize in generating specific concepts (colors, numbers, sentiment), enabling targeted interventions to steer model outputs in both language and vision-language models.

The unembedding matrix provides semantic grounding: its rows align with interpretable tokens, establishing \glspl{correspondence-concept} between internal geometry and semantic content. SOMP then identifies which heads contribute to target concepts across multiple examples, generalizing the logit lens~\citep{nostalgebraist2020interpreting} from single-example, single-direction inspection to multi-example, multi-direction characterization.

Intervening on concept-specific heads identified through this \gls{correspondence-concept} enables precise control. Suppressing a few color-related heads almost removes color words from generated captions, while enhancing them increases their presence; sentiment and quantity words behave the same way. These capabilities demonstrate that sparse geometric structure within the residual stream supports targeted interventions across both language and vision-language models.

Unlike the spectral approach in \cref{ch:residual}, which operates through continuous spectral decomposition, HeadPursuit exploits sparse dictionary structure. Where ResiDual identifies relevant components through variance explained by principal directions, HeadPursuit uses sparse coding to select interpretable atoms from a fixed semantic dictionary. This sparsity-based approach complements the distributional methods explored in \cref{ch:lineas}, together demonstrating how different geometric structures within the residual stream enable distinct intervention strategies.

\begin{figure}[t]
 \centering
 \includegraphics[width=\textwidth]{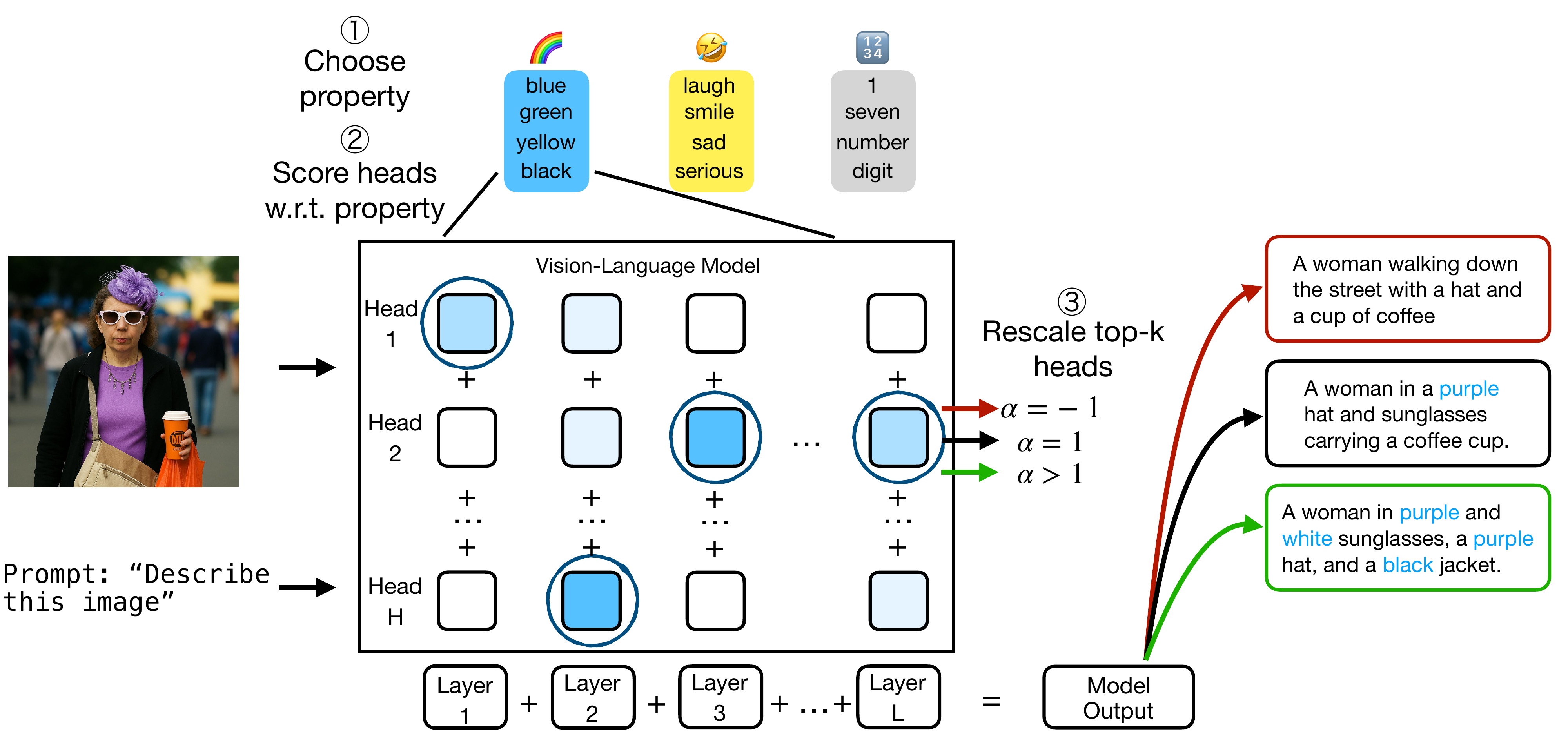}
 \caption{Overview of the method. Given a language or vision-language model and a target property defined by text (e.g., \textit{colors}, \textit{sentiments}, \textit{numbers}), HeadPursuit scores all attention heads according to how well they align with interpretable directions from a fixed dictionary, using a method based on Matching Pursuit~\citep{mallat1993matching} (1). The method then selects the top-$k$ heads (2) and intervenes by rescaling their contribution to the residual stream (3), either enhancing or suppressing the attribute in the model's output.}\label{fig:cartoon}
\end{figure}

\section{Pursuing specialized attention heads}\label{sec:pursuing}
The investigation begins by exploring whether individual attention heads of generative LLMs specialize in interpretable functions. To isolate the contribution of each head, a residual stream decomposition approach is employed. This allows assessment of how each attention head contributes to the residual stream at head-level granularity. Specifically, following \citet{elhage2021mathematical}, the output written by each head into the residual stream is modeled as a matrix $\mH_h^{(\ell)} \in \R^{n \times d}$ for head $h$ of layer $\ell$, where $n$ is the number of samples in the dataset and $d$ is the internal dimensionality of the transformer.

Motivated by the growing body of literature that uses latent decompositions for interpretability, especially in vision-language models~\citep{gandelsman2024interpreting, basile2025residual, parekh2024a,khayatan2025analyzing}, the goal is to identify sparse and interpretable directions for each attention head $\mH_h^{(\ell)}$ that maximally explain its variance on a given dataset. Concretely, the method seeks a sparse representation of $\mH_h^{(\ell)}$ using only directions drawn from a fixed dictionary of interpretable vectors, to ensure that the resulting components are meaningful and grounded in known semantic structures.

As a dictionary, the unembedding matrix of the language model $\mD\in \R^{v \times d}$, where $v$ is the vocabulary size, is adopted, as it contains directions aligned with semantically meaningful outputs, grounding latent structure in human-interpretable terms. Every row of this matrix is a $d$-dimensional vector that effectively represents in latent space a token that can be decoded into natural language.

An approximation of each head representation is then constructed using directions from this dictionary (the unembedding matrix) via a classical sparse coding algorithm: Simultaneous Orthogonal Matching Pursuit (SOMP)~\citep{tropp2006algorithms} (see \cref{sec:somp_algorithm}). SOMP is a multi-sample extension of Orthogonal Matching Pursuit~\citep{pati1993orthogonal}, itself a refinement of the original Matching Pursuit algorithm~\citep{mallat1993matching}. SOMP jointly considers all samples in a given dataset and selects the dictionary directions that are most informative across the representation.

Formally, given a head activation matrix $\mH \in \R^{n \times d}$ (the indices $h,\ell$ are dropped from here on) and a dictionary $\mD \in \R^{v \times d}$, SOMP aims to iteratively construct a column-sparse coefficient matrix $\mW^* \in \R^{n \times v}$ such that:
\begin{equation}
\mH \approx \mW^* \mD
\end{equation}
At each iteration $t$, the algorithm selects the dictionary atom (i.e., a row of $\mD$) that maximally correlates with the head residuals across all samples:
\begin{equation}
p^{(t)} = \argmax_j \| \mD[j] \mE^{(t)\top} \|_1
\end{equation}
Here, the head residual matrix $\mE^{(t)} \in \R^{n \times d}$ is defined as the difference between the original signal and its reconstruction at step $t$: $\mE^{(t)} = \mH - \mH_{\mathrm{r}}^{(t)}$. The selected index $p^{(t)}$ is added to the support set $\mathcal{J}^{(t+1)}$, and the coefficients are refit by solving a least-squares problem restricted to the current support:
\begin{equation}
\mW^{(t)} = \argmin_{\mW} \| \mH - \mW \mD[\mathcal{J}^{(t+1)}] \|_F
\end{equation}
Here $\mD[\mathcal{J}] \in \R^{\lvert\mathcal{J}\rvert \times d}$ keeps the rows indexed by $\mathcal{J}$, so $\mW^{(t)} \in \R^{n \times \lvert\mathcal{J}^{(t+1)}\rvert}$; after $n_{\mathrm{it}}$ iterations, $\mW^*$ is $\mW^{(n_{\mathrm{it}}-1)}$ padded with zero columns outside $\mathcal{J}^{(n_{\mathrm{it}})}$.
The reconstruction is updated as $\mH_{\mathrm{r}}^{(t+1)} = \mW^{(t)} \mD[\mathcal{J}^{(t+1)}]$, and the residuals are recomputed accordingly. This process runs for $n_{\mathrm{it}}$ iterations, so exactly $n_{\mathrm{it}}$ atoms are selected (the sparsity level; typically $n_{\mathrm{it}}=50$ in the experiments). The resulting decomposition expresses each head's output using a sparse set of semantically meaningful dictionary atoms, yielding an interpretable approximation of its behavior.

A conceptual connection exists between this reinterpretation of SOMP and the logit lens (LL)~\citep{nostalgebraist2020interpreting}, a tool widely used in mechanistic interpretability to probe internal representations of transformer models. Similarly to the method just described, LL works by projecting a single residual stream vector onto the unembedding directions to approximate the output logits of the model at intermediate layers. This is equivalent to performing a single step of Matching Pursuit on an individual example. The SOMP-based method generalizes this idea in two key ways: it operates on multiple examples simultaneously, and it selects multiple dictionary directions, each capturing distinct components of the signal. This leads to a more robust and semantically structured characterization of the attention head's functional role.

\Cref{tab:head_spec} presents examples of specialized attention heads, obtained by applying SOMP to Mistral-7B attention heads, prompted by questions from the TriviaQA dataset~\citep{joshi2017triviaqa}. Before applying SOMP, the tokens from the prompt were aggregated by averaging. LL applies to one sample at a time; to aggregate it over a dataset, the 5 tokens with the highest logits are stored for each sample, and the 5 most frequent tokens overall are reported.
\Cref{tab:head_spec_logitlens} shows that a direct application of LL in this setting results in noisier and highly redundant explanations.
\begin{table}[h]
\footnotesize
 \centering
 \caption{Top-5 tokens identified by SOMP on selected attention heads of Mistral-7B, evaluated on TriviaQA prompts. L$\ell$.H$h$ denotes head $h$ of layer $\ell$. Layers and heads are 0-indexed.}\label{tab:head_spec}
 \begin{tabular}{llll}
   \toprule
   \textbf{L18.H27} (``Politics'') & \textbf{L24.H20} (``Nationality'') & \textbf{L25.H14} (``Months'') &
   \textbf{L30.H28} (``Numbers'')\\
   \midrule
   COVID & British & December & 9\\
   Soviet & American & July & 1\\
   Obama & European & April & 3\\
   Biden & German & October & 7\\
   Clinton & English & February & five\\
   \bottomrule
  \end{tabular}
\end{table}

\begin{table}[h]
\footnotesize
 \centering
 \caption{Top-5 tokens identified by aggregated logit lens on selected attention heads of Mistral-7B, evaluated on TriviaQA data.}\label{tab:head_spec_logitlens}
 \begin{tabular}{llll}
   \toprule
   \textbf{L18.H27} (``Politics'') & \textbf{L24.H20} (``Nationality'') & \textbf{L25.H14} (``Months'') &
   \textbf{L30.H28} (``Numbers'')\\
   \midrule
   vaccine & American & Sunday & 8\\
   Covid & Americans & breakfast & u\\
   pandemic & California & Oct & u\\
   COVID & America & October & n\\
   Soviet & American & February & 9\\
   \bottomrule
  \end{tabular}
\end{table}

Besides returning lists of latent directions and associated natural language tokens that better characterize each head, SOMP produces a reconstruction of the head representation in the space spanned by those vectors. Building on this insight, the method automatically identifies the heads most relevant for a target attribute. Given a list of words related to the chosen semantic area, the unembedding matrix is restricted to the rows associated with these tokens, and SOMP is applied on this concept-specific dictionary. The fraction of head variance explained by SOMP in this setting is a measure of specialization, allowing heads to be ranked and selected by their relevance with respect to the target concept.

\subsection{Simultaneous Orthogonal Matching Pursuit Algorithm}\label{sec:somp_algorithm}
The pseudocode for the Simultaneous Orthogonal Matching Pursuit (SOMP) algorithm~\citep{tropp2006algorithms} is given in \cref{alg:somp}.

\begin{algorithm}[ht]
\caption{Simultaneous Orthogonal Matching Pursuit (SOMP)}
\label{alg:somp}
\begin{algorithmic}[1]
\Require Signal Matrix (head representation) $\mH \in \R^{n \times d}$, dictionary $\mD \in \R^{v \times d}$, number of iterations $n_{\mathrm{it}}$
\Ensure Reconstruction $\mH_{\mathrm{r}}^{(n_{\mathrm{it}})}$, support set $\mathcal{J}^{(n_{\mathrm{it}})}$
\State \textbf{Initialize:} Residual $\mE^{(0)} = \mH$, reconstruction $\mH_{\mathrm{r}}^{(0)}=\0$, support set $\mathcal{J}^{(0)} = \emptyset$
\For{$t \in \{0,\dots,n_{\mathrm{it}}-1\}$}
 \State $\mC \gets \mD \mE^{(t)\top}$
 \State $p^{(t)} \gets \argmax_{j\in\{1,\dots,v\}} (\|\mC[j]\|_1)$
 \State $\mathcal{J}^{(t+1)}\gets \mathcal{J}^{(t)}\cup \{p^{(t)}\}$
 \State $\mW^{(t)} \gets \argmin_{\mW} \|\mH-\mW\mD[\mathcal{J}^{(t+1)}]\|_F$
 \State $\mH_{\mathrm{r}}^{(t+1)} \gets \mW^{(t)}\mD[\mathcal{J}^{(t+1)}]$
 \State $\mE^{(t+1)} \gets \mH - \mH_{\mathrm{r}}^{(t+1)}$
\EndFor
\end{algorithmic}
\end{algorithm}

\section{Controlling language generation through specialized heads}
This section evaluates how the specialization of attention heads can be used to apply domain-specific targeted interventions to model behavior, effectively validating the selection. One approach disrupts the information flow from a selected subset of heads to the residual stream during the forward pass. Concretely, this intervention inverts the sign of the head representations. This intervention affects both the direct contribution of the head to the residual stream and the indirect contribution of its information content to subsequent layers.

The key preliminary step identifies relevant and specialized heads. To accomplish this, SOMP (see \cref{sec:pursuing}) is applied over a restricted unembedding dictionary, filtered to include only a set of tokens associated with the target property. These tokens can be selected using various strategies, such as user-defined word lists, class names, or keywords generated by an external LLM. Attention heads are then ranked by the proportion of variance in the data explained by SOMP (typically using $50$ iterations in the experiments) and intervention is applied to the top-$k$ ranked heads.

All experiments include a random control condition to verify the specificity of findings. This control involves intervening on a randomly selected set of attention heads that matches the original set in both size and layer distribution but is entirely disjoint from it. To ensure fair comparison, heads previously identified as specialized are explicitly avoided when constructing the control set. In all experiments, random control results are reported in terms of medians and interquartile ranges over $10$ independently sampled sets of heads.

\subsection{Question answering}
\paragraph{Experimental setting.} The generative LLM Mistral-7B~\citep{jiang2023mistral7b} is evaluated on textual prompts from the TriviaQA~\citep{joshi2017triviaqa} question answering dataset. The goal is to identify attention heads specialized in generating country names, a target attribute motivated by their relative abundance in the dataset: despite their specificity, country names account for over 6\% of the answers in the test split. For targeting the countries concept, the tokens in the dictionary (unembedding matrix) are restricted to those corresponding to names of countries, and the Matching Pursuit based method is applied to select specialized heads. As an additional baseline, heads are also selected with a simple adaptation of the logit lens: each head is scored by the mean logit that LL assigns to country-related tokens, and the top-$k$ heads are selected as in the SOMP-based method. Both sets of heads are computed using representations obtained from the training data, which is strictly disjoint from the data used in evaluation. Model performance on TriviaQA is evaluated in terms of the standard F1 score, which accounts for partial overlaps between the predicted and ground-truth answers.

\begin{figure}[h]
 \centering
 \includegraphics[width=0.6\textwidth]{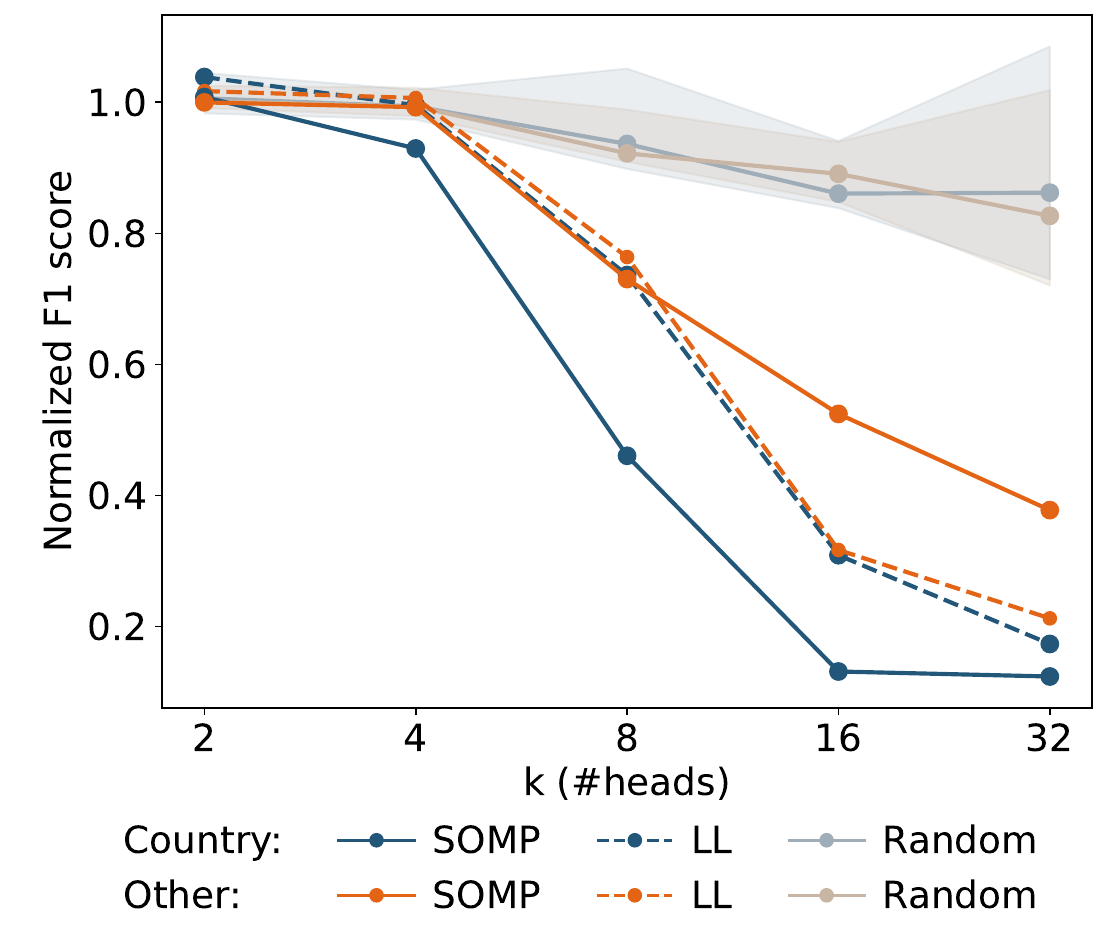}
 \caption{\looseness=-1Question answering performance of Mistral-7B on TriviaQA. F1 scores are shown for samples associated (country) or not (other) with the target attribute, normalized to the F1 score of the base model without intervention. Random baselines are displayed as medians and interquartile ranges.}
 \label{fig:trivia}
\end{figure}

\paragraph{Result analysis.}
The results of intervention on subsets of specialized heads are reported in \cref{fig:trivia}, varying the number of selected heads. Performance on the target attribute (solid blue line) noticeably degrades when the signs of 8 heads ($0.8\%$ of the total) or more are inverted. Performance on the remaining examples (solid orange line) declines more gradually, suggesting that these specialized heads have a targeted effect. Since the semantic domains of the questions answered by a country name and the remaining part of TriviaQA are not disentangled, intervening on the selected heads is expected to also have a (lower) impact on the remaining examples (orange line). Control experiments using random heads (faded lines) show a limited impact on performance, suggesting the specificity of the identified heads. Instead, heads selected by the logit lens (dashed lines) are relevant to the question answering task but not specific to the target concept, as they degrade performance equally within and outside the target domain.

Overall, the analysis suggests that HeadPursuit correctly identifies heads relevant to country-related examples, as intervention disproportionately impacts target performance, while control heads induce a limited effect and the logit lens selects relevant but non-specific heads.

\subsection{Mitigation of toxic content}
\paragraph{Experimental setting.}
This experiment evaluates HeadPursuit in a more realistic and less controlled scenario, where a complete list of target keywords is not available. Instead, only a limited and incomplete list of words represents the topic or concept.

The focus is on toxicity mitigation: specifically, reducing the occurrence of offensive content in text generated by Mistral. To accomplish this, a subset of toxic heads within the model is identified and intervened upon. Two datasets are considered: RealToxicityPrompts (RTP)~\citep{gehman2020realtoxicityprompts}, which contains real-world Web prompts, and Thoroughly Engineered Toxicity (TET)~\citep{luong2024realistic}, a benchmark with carefully constructed test cases, both of which are designed to elicit harmful responses from LLMs.

For both datasets, toxic words are extracted from Mistral's responses using Llama-3.1-8B~\citep{grattafiori2024llama} (the prompt is reported in the appendix of \citet{headpursuit}), and those words are used to identify and invert toxic heads.
Two complementary metrics quantify the toxicity of Mistral's responses: one semantic and one lexical. The semantic metric uses a RoBERTa-based toxicity classifier from \citet{logacheva-etal-2022-paradetox}, trained to detect toxic content in text. The lexical metric measures the frequency of a held-out subset of toxic words that were not used for head selection.

\begin{table}[h]
\centering
 \caption{Normalized count of toxic generations after intervention, as judged by the toxicity classifier. Lower values indicate better mitigation. Targeted heads reduce toxicity, while heads selected using LL have a weaker impact and random heads maintain or increase toxicity. For random baselines, only the median is shown; the interquartile ranges are reported in the appendix of \citet{headpursuit}.}
 \label{tab:toxicity}
 \centering
 \begin{tabular}{lccccccccc}
 \toprule
 & \multicolumn{3}{c}{\textbf{8 heads}} & \multicolumn{3}{c}{\textbf{16 heads}} & \multicolumn{3}{c}{\textbf{32 heads}} \\
 \cmidrule(lr){2-4}\cmidrule(lr){5-7}\cmidrule(lr){8-10}
 \textbf{Dataset} & \textbf{SOMP} & \textbf{LL} & \textbf{Rand.} & \textbf{SOMP} & \textbf{LL} & \textbf{Rand.} & \textbf{SOMP} & \textbf{LL} & \textbf{Rand.} \\
 \midrule
 RTP & \textbf{0.83} & $0.91$ & $1.02$ & \textbf{0.67} & $0.79$ & $1.00$ & \textbf{0.66} & $0.71$ & $1.13$ \\
 TET & $0.83$ & \textbf{0.81} & $0.97$ & \textbf{0.68} & $0.73$ & $0.95$ & \textbf{0.49} & $0.68$ & $0.95$ \\
 \bottomrule
 \end{tabular}
\end{table}

\paragraph{Result analysis.}
The results obtained by inverting the sign of toxic head activations are displayed in \cref{tab:toxicity}, for $8$, $16$ and $32$ heads. In both RTP and TET, intervening on such heads noticeably reduces the number of generations deemed toxic by the classifier, while intervening on heads identified by the logit lens has a generally weaker impact on toxicity. Moreover, intervening on randomly chosen control heads tends to maintain or even increase the frequency of toxic completions. A possible explanation is that random selection picks and disrupts heads that are deemed non-toxic (imposed to be distinct from the targeted ones), thus implicitly reinforcing toxic behavior. Analogous results are reported in the appendix of \citet{headpursuit} for the lexical metric, showing that the intervention reduces the frequency of toxic words, even those \textit{not used} for the head selection.

These results demonstrate that intervening on a small subset of heads can make generated text less toxic. Moreover, the approach impacts broader semantic areas, beyond the keywords used for head selection.

\section{Targeted control of visual attributes}
This section evaluates the extent and implications of head specialization in the LLM backbones within generative Vision-Language models (VLMs). These models are usually built by fine-tuning a pretrained LLM on multimodal tasks, such as visual question answering or image captioning, using visual tokens coming from a pretrained vision encoder as contextual information~\citep{liu2023visual}. In line with recent works~\citep{neo2025towards} that have applied the logit lens to visual tokens of LLaVA, a prominent example of VLM, the analysis applies SOMP-based head selection to the head representations of image patches, averaged over tokens. Two task scenarios are considered: image classification and image captioning.

\subsection{Image classification}
\label{sec:classification}

\paragraph{Experimental setting.}
For this experiment, LLaVA-NeXT-7B~\citep{liu2024llavanext} (referred to as LLaVA hereafter) is benchmarked on a range of image classification datasets, including: \gls{mnist}, \gls{svhn}~\citep{netzer2011reading}, GTSRB~\citep{stallkamp2011german}, EuroSAT~\citep{helber2019eurosat}, RESISC45~\citep{cheng2017remote} and DTD~\citep{cimpoi2014describing}. For each dataset, the set of $k$ most relevant heads is selected. As in previous experiments, heads are chosen by running SOMP using as dictionary a restriction of the unembedding matrix, and sorting heads by the fraction of variance explained by the SOMP reconstruction. Two settings are considered: one task-conditioned, in which the unembedding matrix is restricted to the tokens corresponding to class names; and one completely task-agnostic, in which a set of keywords extracted by an external VLM (Mistral-Small-3.1-24B) is considered. Results for the latter are reported in \cref{fig:keywords_classification}, and the prompt is reported in the appendix of \citet{headpursuit}. In this experiment, the model is prompted to classify the image, and the generated output is evaluated in terms of exact match with the ground truth class label.

\begin{figure}[h]
 \centering
 \includegraphics[width=\textwidth]{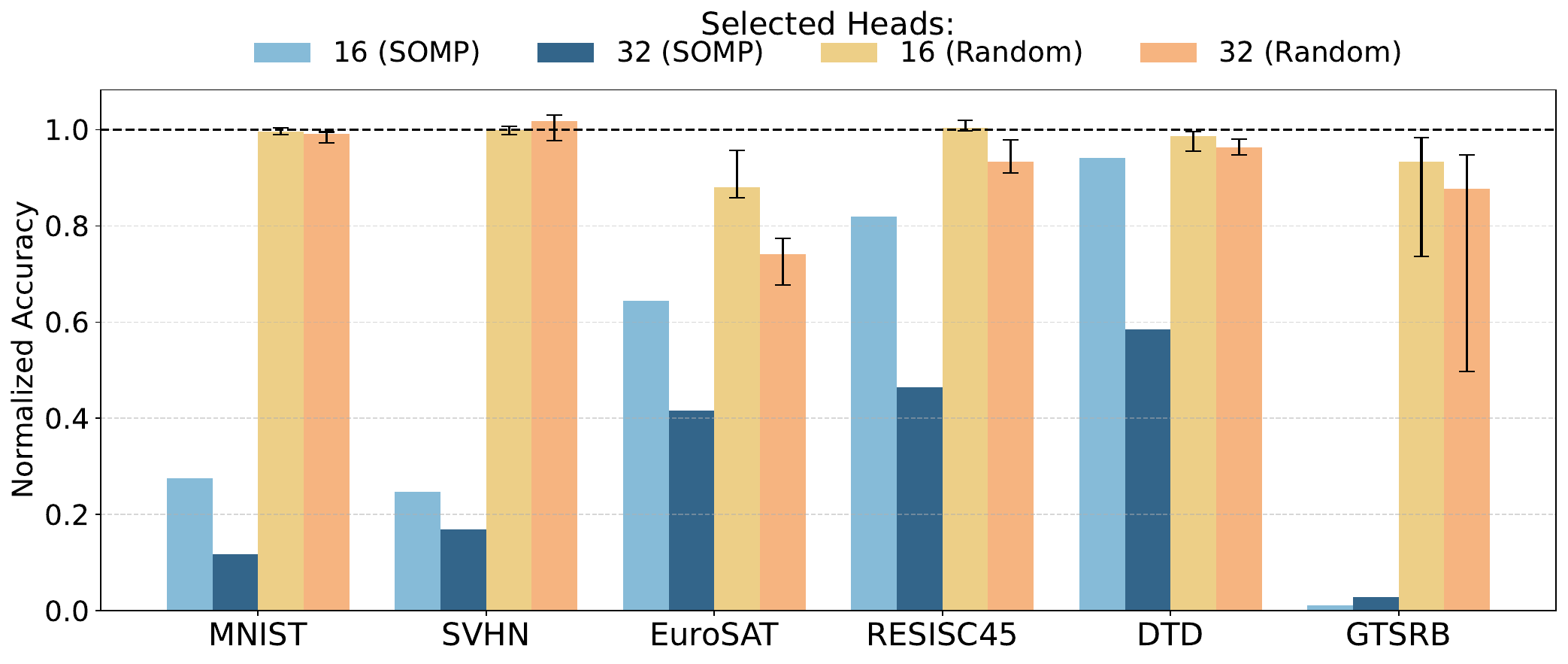}
 \caption{Classification results under different head selection strategies: (light blue) 16 heads with highest variance ratio explained by SOMP; (dark blue) 32 heads with highest explained variance ratio; (yellow) 16 random heads, with the same layer-wise counts of top 16; (orange) 32 random heads, with the same layer-wise counts of top 32.}\label{fig:classification}
\end{figure}

\begin{figure}[h]
 \centering
 \begin{minipage}[t]{0.49\textwidth}
 \centering
 \includegraphics[width=\textwidth]{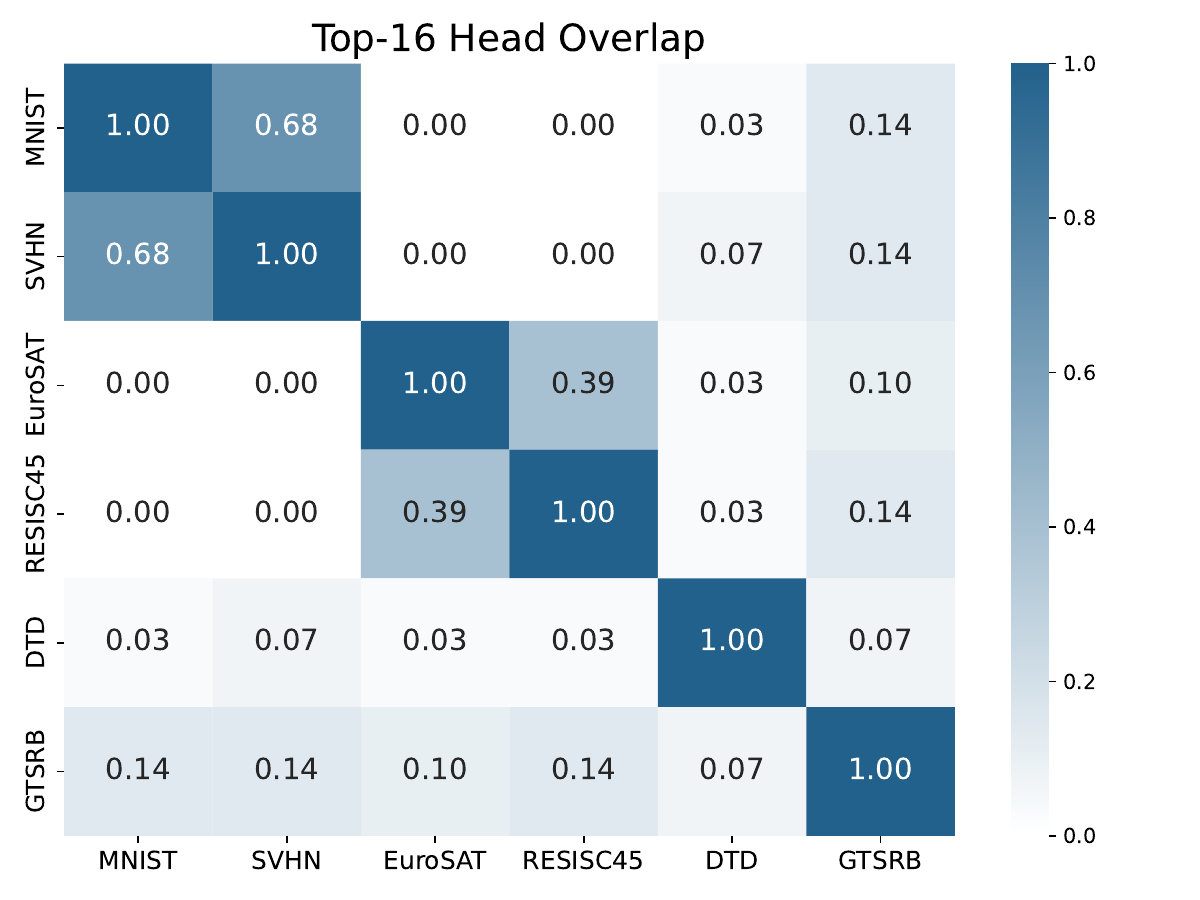}
 \end{minipage}
 \hfill
 \begin{minipage}[t]{0.49\textwidth}
 \centering
 \includegraphics[width=\textwidth]{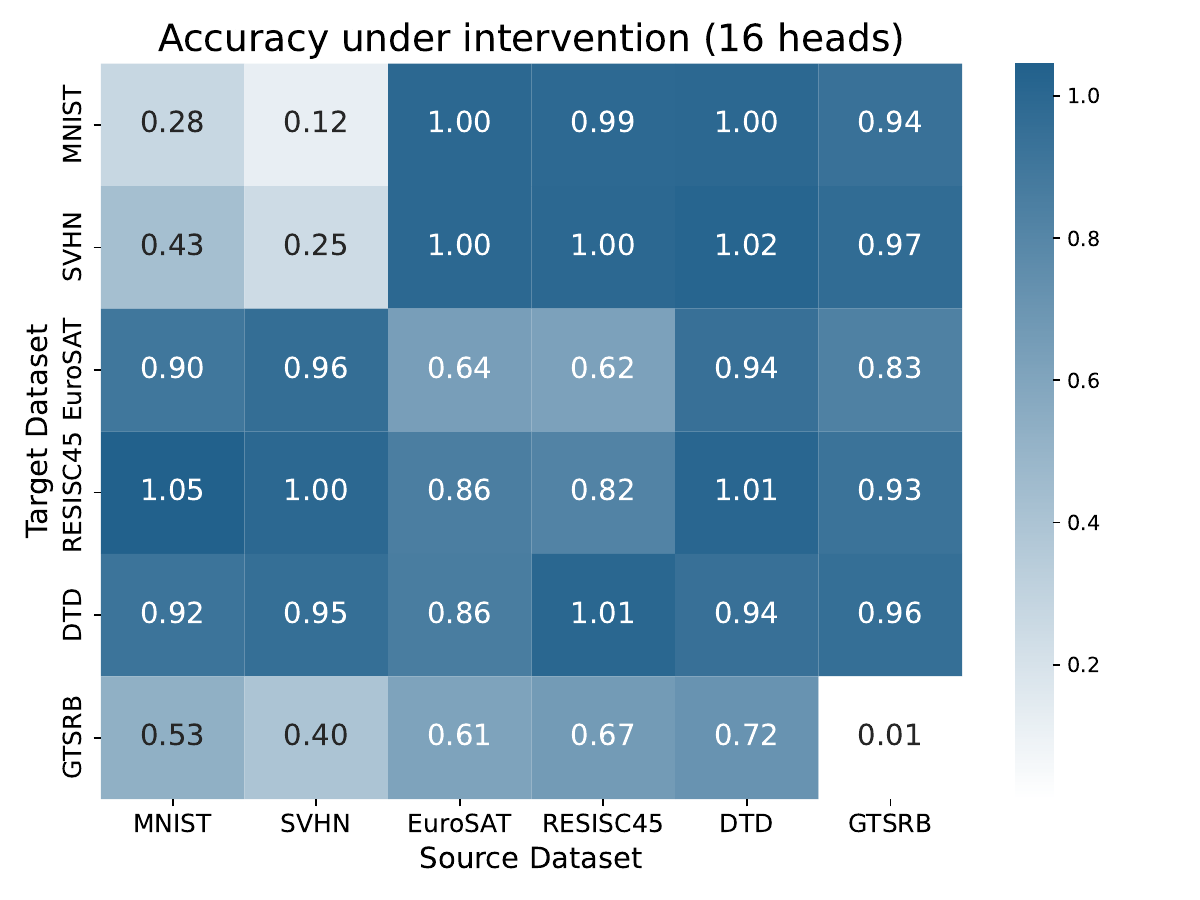}
 \end{minipage}
 \caption{(left) Jaccard similarity between sets of top-$16$ LLaVA heads selected with SOMP over different datasets; (right) Classification accuracy on a target dataset, denoted by row, when the top-$16$ heads are selected with SOMP on a source dataset, denoted by column. Accuracy is normalized w.r.t.\ the base accuracy on target dataset.}
 \label{fig:overlap}
\end{figure}

\paragraph{Result analysis.}
The classification results are reported in \cref{fig:classification}, normalized for each dataset with respect to the accuracy obtained by LLaVA when no intervention is applied to its forward pass. For all datasets, inverting the top $32$ heads identified by the method is sufficient to disrupt the classification performance, while inverting $32$ random heads at equivalent layers has substantially lower to no impact on performance. At $k=16$ the picture is similar with the exception of DTD, whose performance is weakly affected, hinting at higher head redundancy on this task. In \cref{fig:overlap}, the interaction between head choices for different datasets is analyzed. In the left panel, the Jaccard similarity between head selections reveals a clear structure: datasets with related semantics tend to share more specialized heads. For example, \gls{mnist} and \gls{svhn} (both digit recognition tasks) exhibit substantial overlap, as do EuroSAT and RESISC45 (both involving remote sensing imagery). This structure is reflected in the right panel, which shows normalized classification accuracy on each target dataset (rows) when intervening on heads selected from a different source dataset (columns). Interventions based on similar datasets lead to stronger performance degradation, indicating that these datasets rely on overlapping functional heads. Large drops in GTSRB (traffic sign recognition) performance are also observed when intervening with heads selected from \gls{mnist} or \gls{svhn}. Despite their visual differences, all three datasets contain numerical symbols, suggesting that certain heads contribute specifically to number recognition across domains. The same analysis with $k=8$ and $k=32$ heads is shown in \cref{fig:overlap_8,fig:overlap_32}. Logit lens baselines are reported in the appendix of \citet{headpursuit}, where additional experiments on LLaVA-NeXT-13B, Gemma3-12B and Qwen2.5-VL-7B confirm the trends observed here on LLaVA-NeXT-7B.

\begin{figure}[h]
 \centering
 \includegraphics[width=\textwidth]{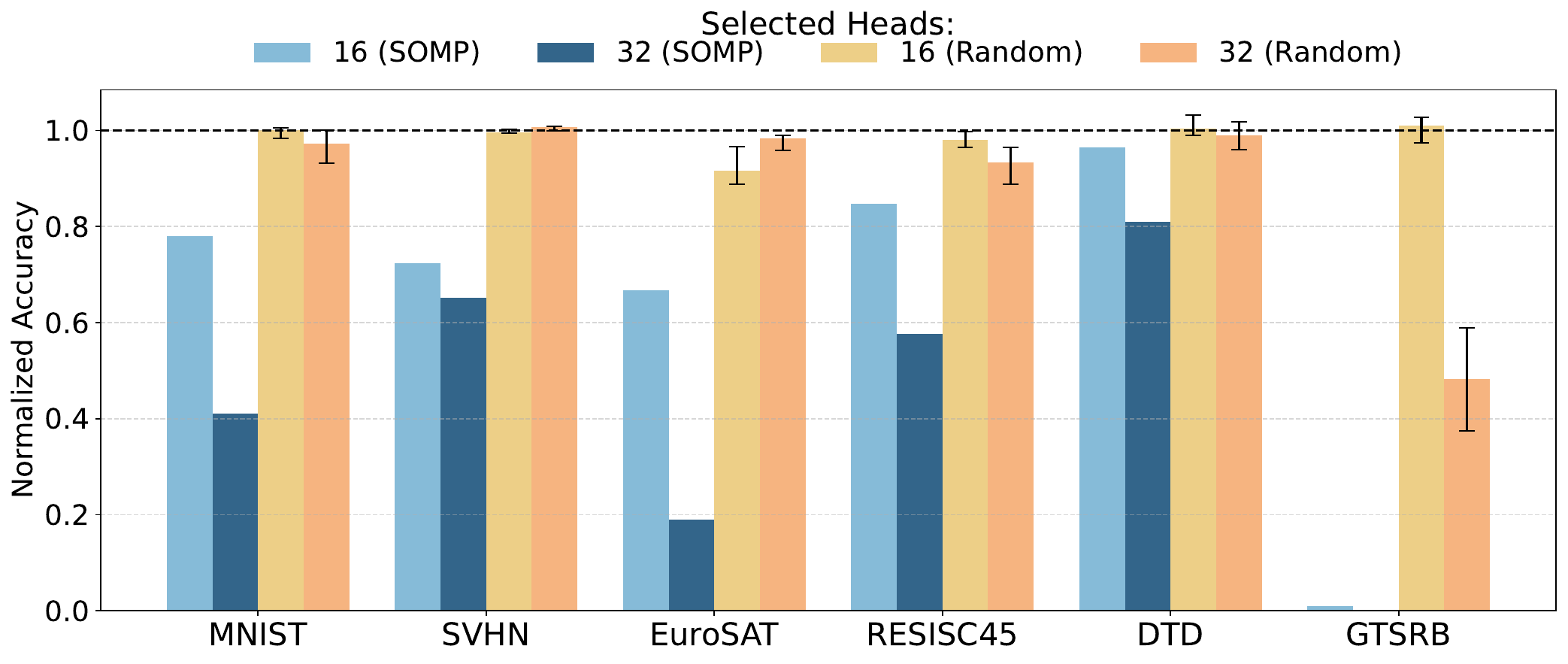}
 \caption{Task-agnostic head selection on image classification benchmarks. Heads were selected using dataset-specific lists of keywords obtained from an external VLM, with no task knowledge. Classification results under different head selection strategies: (light blue) 16 heads with highest variance ratio explained by SOMP; (dark blue) 32 heads with highest explained variance ratio; (yellow) 16 random heads, with the same layer-wise counts of top 16; (orange) 32 random heads, with the same layer-wise counts of top 32.}
 \label{fig:keywords_classification}
\end{figure}

Summing up, intervening on a small set of attention heads selected via SOMP disrupts classification performance across diverse datasets, confirming head-level specialization in LLaVA. Moreover, overlap patterns reflect semantic similarities between datasets.

\begin{figure}[h]
 \centering
 \begin{minipage}[t]{0.49\textwidth}
 \centering
 \includegraphics[width=\textwidth]{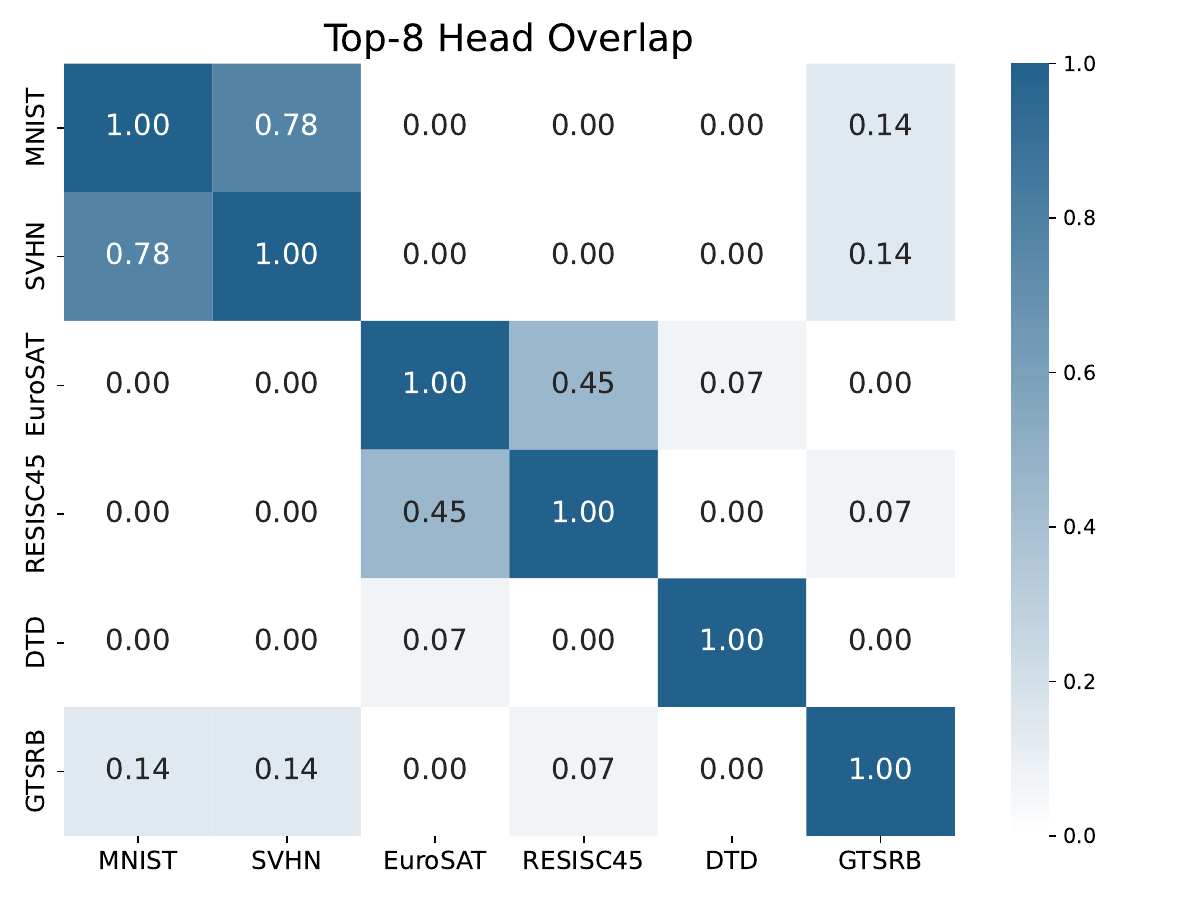}
 \end{minipage}
 \hfill
 \begin{minipage}[t]{0.49\textwidth}
 \centering
 \includegraphics[width=\textwidth]{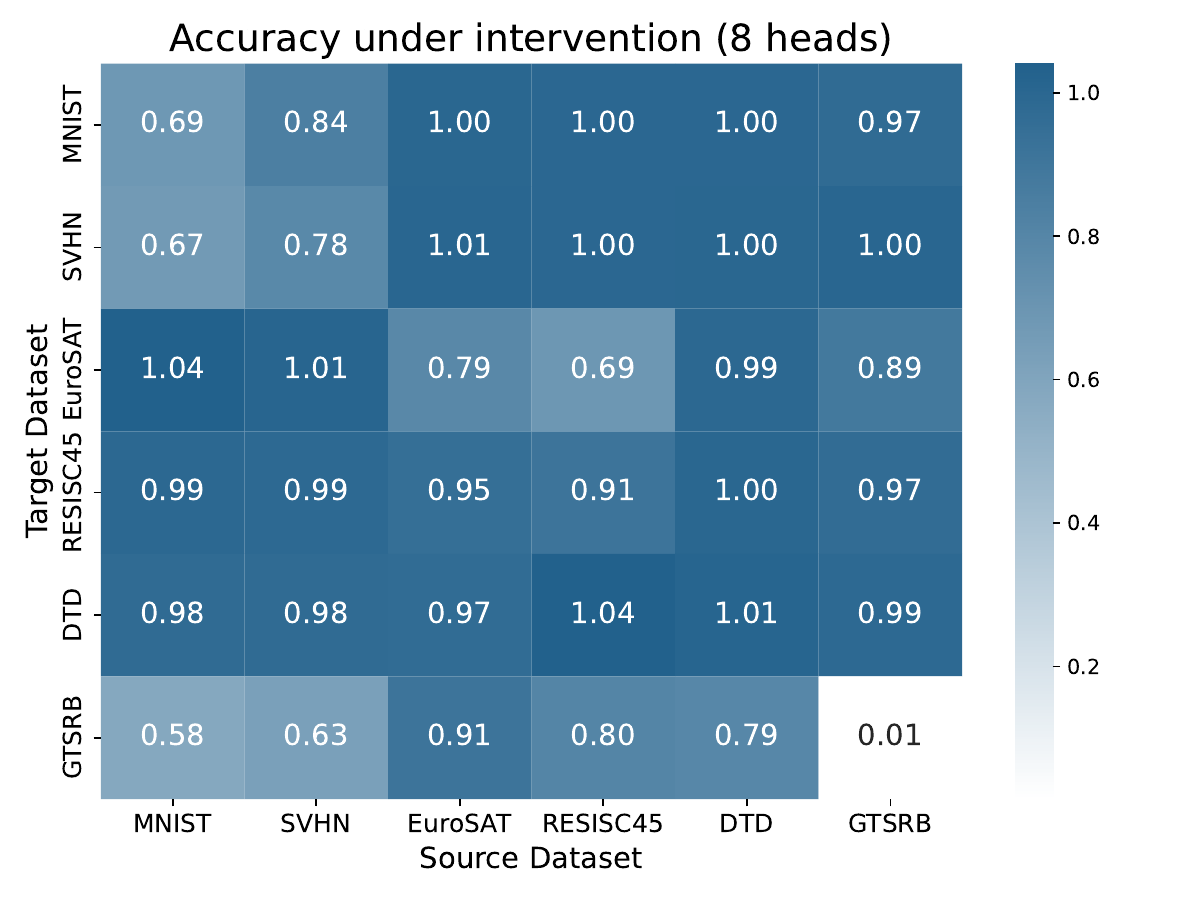}
 \end{minipage}
 \caption{(left) Jaccard similarity between sets of top-$8$ LLaVA heads selected with SOMP over different datasets; (right) Classification accuracy on a target dataset, denoted by row, when the top-$8$ heads are selected with SOMP on a source dataset, denoted by column. Accuracy is normalized w.r.t.\ the base accuracy on target dataset.}
 \label{fig:overlap_8}
\end{figure}

\begin{figure}[h]
 \centering
 \begin{minipage}[t]{0.49\textwidth}
 \centering
 \includegraphics[width=\textwidth]{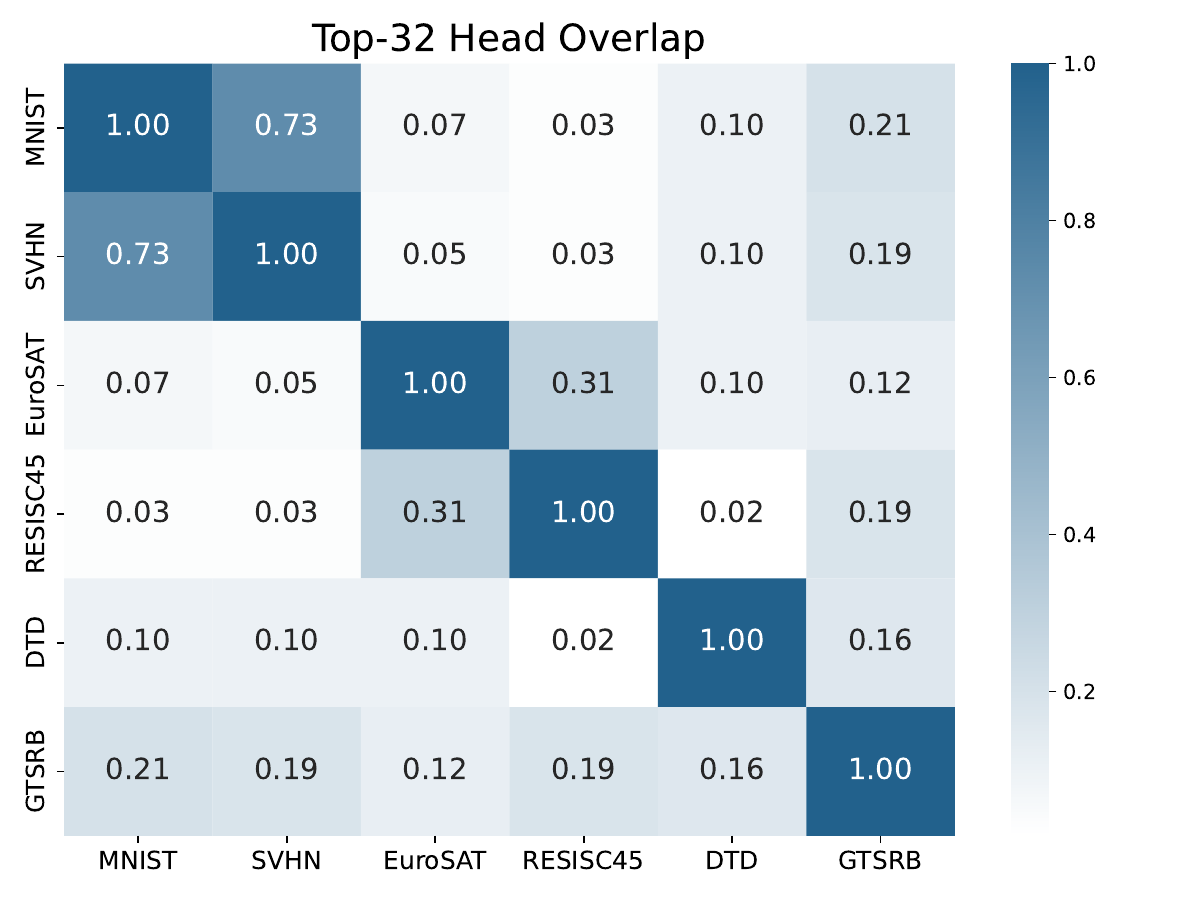}
 \end{minipage}
 \hfill
 \begin{minipage}[t]{0.49\textwidth}
 \centering
 \includegraphics[width=\textwidth]{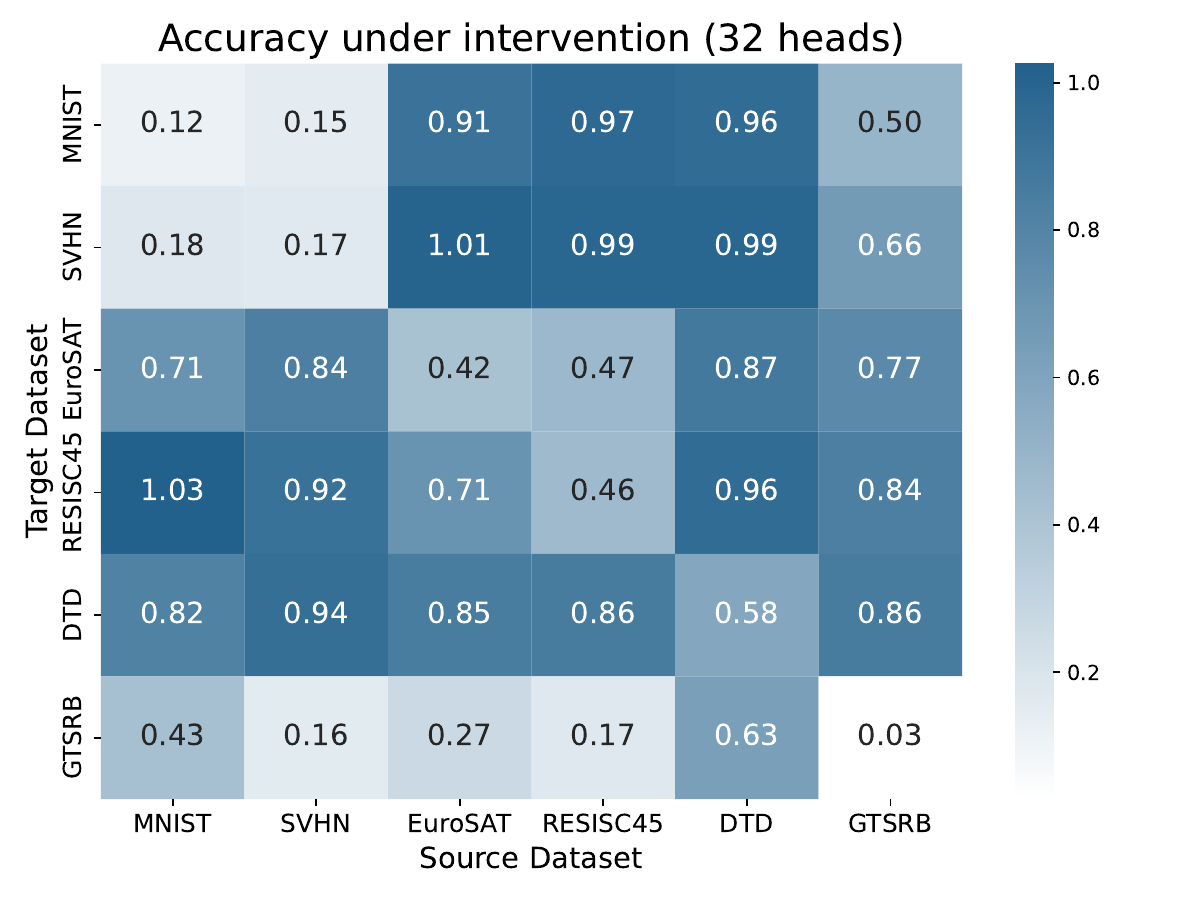}
 \end{minipage}
 \caption{(left) Jaccard similarity between sets of top-$32$ LLaVA heads selected with SOMP over different datasets; (right) Classification accuracy on a target dataset, denoted by row, when the top-$32$ heads are selected with SOMP on a source dataset, denoted by column. Accuracy is normalized w.r.t.\ the base accuracy on target dataset.}
 \label{fig:overlap_32}
\end{figure}

\subsection{Image captioning}
\begin{figure}[h]
 \centering
 \includegraphics[width=\textwidth]{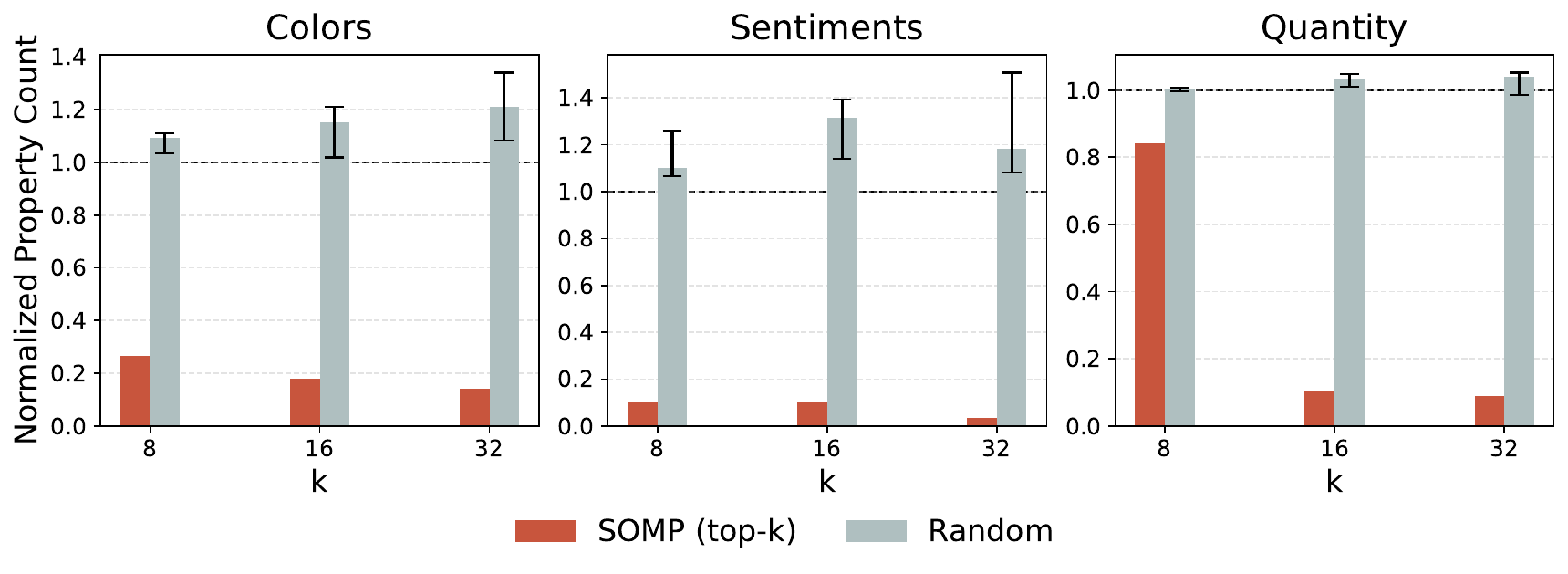}
 \caption{Effect of inhibitory interventions on LLaVA attention heads for Flickr30k captioning. Bars show the normalized presence of target attributes (\textit{colors}, \textit{sentiments}, \textit{quantity}) under different head selection strategies (SOMP and random). Random baselines are reported in terms of medians and interquartile ranges.}
 \label{fig:inhibitory}
\end{figure}

\begin{figure}[h]
 \centering
 \includegraphics[width=\textwidth]{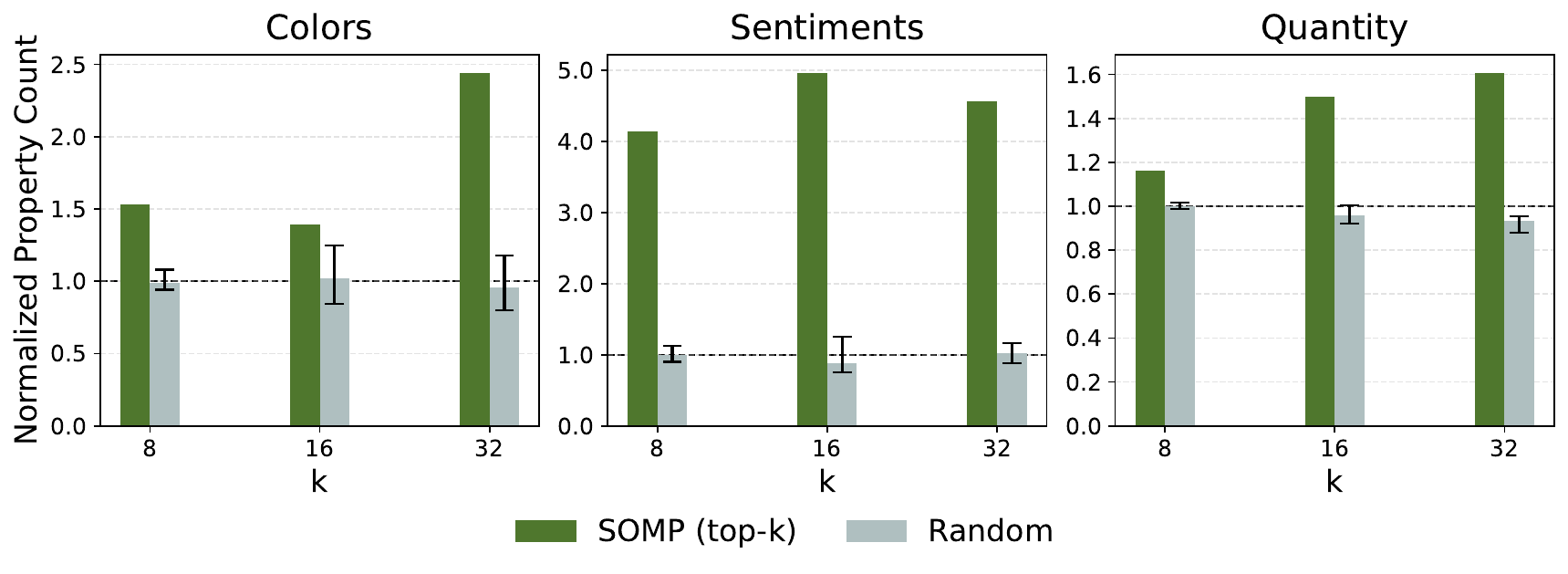}
 \caption{Effect of enhancing interventions on LLaVA attention heads for Flickr30k captioning. Bars show the normalized presence of target attributes (\textit{colors}, \textit{sentiments}, \textit{quantity}) under different head selection strategies (SOMP and random). Random baselines are reported in terms of medians and interquartile ranges.}
 \label{fig:enhancement}
\end{figure}

\begin{figure}[h]
 \centering
 \includegraphics[width=0.8\textwidth]{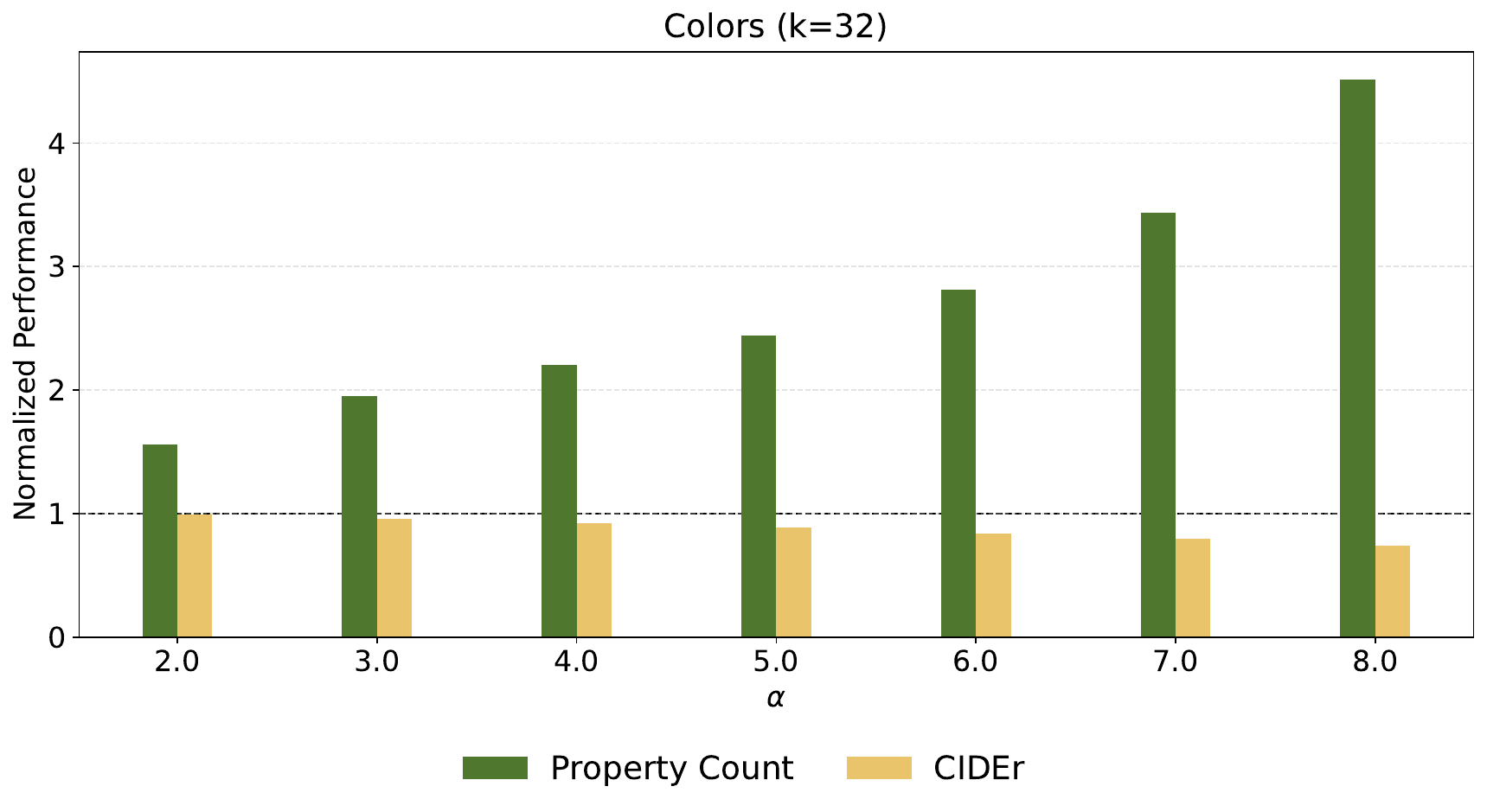}
 \caption{Effect of enhancing interventions on 32 color-specialized heads in image captioning. Increasing the rescaling coefficient $\alpha$ leads to a stronger presence of color-related words in captions, accompanied by a gradual drop in overall caption quality as measured by CIDEr.}
 \label{fig:var_alpha}
\end{figure}

\paragraph{Experimental setting.}
The Flickr30k dataset~\citep{young2014image} is considered, and the possibility of promoting or reducing the presence of words belonging to specific semantic areas in the captions generated by LLaVA is evaluated (other VLMs are evaluated on the same task in the appendix of \citet{headpursuit}). In this experiment, two opposite intervention directions are considered: one is inhibitory, as in previous experiments, and one enhancing. In the former case, the objective is to make the model produce meaningful captions that do not contain the target property (e.g., \textit{colors}), while in the latter the aim is to enhance the target property, while preserving the model's capabilities in generating meaningful descriptions. The two setups reduce to rescaling selected heads by a coefficient that is $\alpha=-1$ in the negative case and $\alpha>1$ in the positive case; the intervention rescales the selected heads by $\alpha$, which realises $\OperatorSameSpace_\Corr$. Heads are selected using SOMP on a dictionary of tokens corresponding to lists of keywords regarding \textit{colors}, \textit{sentiments} and \textit{quantity}, the first two produced by \citet{khayatan2025analyzing} and the last one manually curated. After the intervention, evaluation is carried out in terms of caption quality, measured using the CIDEr metric~\citep{vedantam2015cider}, and effectiveness of the intervention, measured as the average number of target concept keywords present in the captions.
\paragraph{Inhibitory intervention.}
The first setup evaluated is analogous to the previous examples on text generation and image classification tasks. The generation of tokens in a certain semantic domain is inhibited by inverting the signs of a few carefully selected attention heads. The results of this analysis are reported in \cref{fig:inhibitory}, for the three sets of attributes \textit{colors} (left), \textit{sentiments} (center) and \textit{quantity} (right). In all cases, the intervention is able, with as few as 16 heads, to almost completely remove attribute-related keywords from the output captions, while keeping the overall caption quality almost on par with the original, as witnessed by the CIDEr score, which stays at or above $80\%$ of the original. The CIDEr results, together with a comparison with heads selected using the logit lens, are reported in the appendix of \citet{headpursuit}.
\paragraph{Enhancing intervention.}
Across different tasks and data modalities, intervening on selected head activations by inverting their sign is highly effective in disrupting the generation of a target attribute. A different perspective now evaluates whether amplifying those specialized attention heads can incentivize the generation of the target concept. This is accomplished by multiplying the activations of chosen heads by a coefficient $\alpha>1$: $\alpha=5$ is chosen for the experiments as it guarantees a reasonable trade-off between caption quality and attribute enhancement. The results, on the same attributes of the previous experiment (\textit{colors}, \textit{sentiments} and \textit{quantity}), are reported in \cref{fig:enhancement}. As in the previous case, the intervention affects the overall caption quality only marginally (see the appendix of \citet{headpursuit}), while the presence of target concepts increases by more than $60\%$ in all three cases with $32$ heads. The effect of $\alpha$ between 2 and 8 on 32 color-specialized heads is shown in \cref{fig:var_alpha}: the frequency of color-related words increases smoothly with $\alpha$, at the cost of a small decrease in caption quality (CIDEr drops by up to $12\%$ for $\alpha\leq 5$). Captions generated for two sample images after applying interventions in both directions (inhibitory and enhancing) are reported in \cref{tab:example_flickr}.

\begin{table}

 \begin{minipage}{0.99\textwidth}
\centering
 \caption{Examples of captions produced by LLaVA-NeXT on Flickr30k images, before and after inhibiting (top) or enhancing (bottom) 16 heads specialized on \textit{colors} (left) and \textit{sentiments} (right). Analogous examples for \textit{quantity} are reported in the appendix of \citet{headpursuit}.}\label{tab:example_flickr}
\scalebox{0.8}{
\begin{tabular}{l >{\raggedright\arraybackslash}p{5.4cm} @{\hspace{0.5cm}} >{\raggedright\arraybackslash}p{7.1cm}}
\toprule
 \multicolumn{3}{l}{\bf Flickr30k examples:} \\
\midrule
& \makebox[\linewidth]{\includegraphics[height=3.5cm]{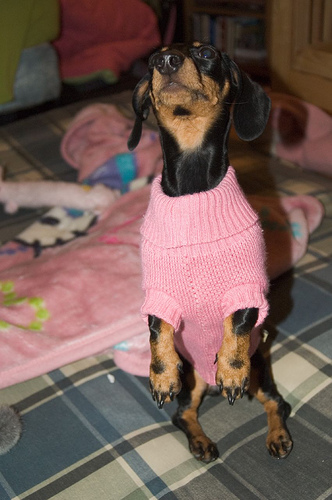}} & \makebox[\linewidth]{\includegraphics[height=3.5cm]{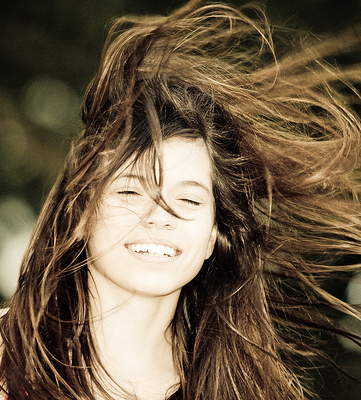}} \\
\midrule
\textbf{Original} & A small dachshund wearing a {\setlength{\fboxsep}{1pt}\colorbox{lightblue}{\strut pink}} sweater. & A young woman with long brown hair and a {\setlength{\fboxsep}{1pt}\colorbox{lightyellow}{\strut smile}}.
\\ \midrule
\midrule
\textbf{Intervention} & {\setlength{\fboxsep}{1pt}\colorbox{lightblue}{\strut \textit{Colors}}} inhibition \mbox{($k=16$, $\alpha=-1$)} & {\setlength{\fboxsep}{1pt}\colorbox{lightyellow}{\strut \textit{Sentiments}}} inhibition \mbox{($k=16$, $\alpha=-1$)}
\\ \midrule
\textbf{Output} & A small dachshund wearing a sweater. & Girl with long brown hair blowing in the wind.
\\ \midrule
\midrule
\textbf{Intervention} & {\setlength{\fboxsep}{1pt}\colorbox{lightblue}{\strut \textit{Colors}}} enhancement \mbox{($k=16$, $\alpha=5$)} & {\setlength{\fboxsep}{1pt}\colorbox{lightyellow}{\strut \textit{Sentiments}}} enhancement \mbox{($k=16$, $\alpha=5$)}
\\ \midrule
\textbf{Output} & A {\setlength{\fboxsep}{1pt}\colorbox{lightblue}{\strut black}} and {\setlength{\fboxsep}{1pt}\colorbox{lightblue}{\strut brown}} dog wearing a {\setlength{\fboxsep}{1pt}\colorbox{lightblue}{\strut pink}} sweater. & A {\setlength{\fboxsep}{1pt}\colorbox{lightyellow}{\strut happy}} girl with long hair and a big {\setlength{\fboxsep}{1pt}\colorbox{lightyellow}{\strut smile}}.
\\ \bottomrule
\end{tabular}
}

 \end{minipage}

\end{table}

Overall, these results show that head-level specialization can be used to control the prevalence of words belonging to a target semantic area in generated image captions. This result holds for both \textit{inhibiting} and \textit{enhancing} the target concept.
\paragraph{Computational resources.}
To perform the experiments, implementations of Mistral-7B~\citep{mistralhf}, LLaVA-NeXT-7B~\citep{llavahf} and the other models listed in the appendix of \citet{headpursuit}, as provided in the HuggingFace transformers library~\citep{wolf2020transformers}, were employed. All experiments were executed on a single NVIDIA H100 GPU equipped with 80GB VRAM.

\chapter[LinEAS]{LinEAS: Linear End-to-end Activation Steering}
\label{ch:lineas}

\chapterpublication[https://github.com/apple/ml-lineas]{lineas}{\textbf{NeurIPS 2025}}{A distributional correspondence between 32 samples is enough to learn SoTA utility-preserving steering vectors for generative models.}

\newcommand{\method}{LinEAS\xspace}

\section{Introduction: Distributional Correspondence Across Layers}

This chapter presents LinEAS, a \gls{correspondence-coupled} same-space operator ($\gls{OperatorSameSpace}_{\gls{Corr}}$, as defined in \cref{ch:methodology}) that operates within a single model through \textit{distributional alignment}. Component-wise interventions (ResiDual's spectral decomposition, \cref{ch:residual}) and sparse selection (HeadPursuit's attention head identification, \cref{ch:headpursuit}) act on individual components. LinEAS establishes correspondence between activation \textit{distributions}.

The correspondence $\Corr$ links source and target distributional properties---for instance, the statistical signatures of representations when processing toxic versus non-toxic prompts, or when generating images with versus without specific concepts. While distributional correspondence represents the weakest form in the taxonomy of geometric structures (\cref{ch:methodology}), requiring only statistical agreement and no instance-level pairing, LinEAS demonstrates that this generality enables learning from very few examples: as few as 32 unpaired samples suffice for effective steering in both language and vision generative models.

LinEAS's defining characteristic is \textit{joint correspondence-guided optimization of transport maps scattered throughout network depth}. Local layer-wise interventions---optimizing each layer independently while freezing others---ignore causal dependencies propagating across layers, causing distributional shifts that degrade model utility. LinEAS addresses this by learning affine transport maps at each layer through end-to-end optimization using a global distributional loss grounded in optimal transport theory. This joint training accounts for layer interactions while enabling sparsity regularization that automatically selects which activations to intervene on, reducing the intervened support by 100$\times$ while preserving alignment quality.

The method demonstrates distributional correspondence on language generation (toxicity mitigation in LLMs) and text-to-image generation (concept removal/induction in generative models), achieving state-of-the-art results among activation steering methods that preserve model utility.

\section{End-to-end Learning of Steering Maps}\label{subsec:end2end}

LinEAS optimizes activation-specific interventions jointly across layers. The hypothesis is that global estimation exploiting causal interdependencies between activations is needed to maintain model utility while achieving the steering goal.

\subsection{Interventions Setup and Distributional Loss}
\label{sec:loss}
Consider a generative model with $L$ intermediate activation layers targeted for intervention. The model can be described as a composition of $L+1$ abstract pretrained functions, where the output, given an input prompt $\obs\in\Obs$, can be written as
$F_\theta(\obs) = \Bf{L+1}\circ \Bf{L} \circ \dots \circ \Bf{1}(\obs).$
For convenience, denote $d_\ell$ as the dimension of activations at layer $\ell$, \ie the size of the activation vector $\Bf{\ell}\circ \dots\circ\Bf{1}(\obs)$. The pretrained computational graph of $L+1$ intermediate frozen layers is interlaced with $L$ comparatively much simpler vector-to-vector maps:
$\widetilde{F}_\theta(\obs) = \Bf{L+1} \circ \RT{L} \circ \Bf{L} \circ \dots \circ \RT{2} \circ \Bf{2} \circ \RT{1} \circ \Bf{1}(\obs),$
where for $1\leq \ell\leq L$, the map $\RT{\ell}:\R^{d_\ell}\to \R^{d_\ell}$ acts on the intermediate activations observed after layer $\ell$, by altering its coordinates and outputting a vector of the same size. The maps $(\RT{\ell})_{\ell \leq L}$ together realise $\OperatorSameSpace_\Corr$.

\textbf{Layerwise Distributional Losses.}
The \gls{correspondence-concept} $\Corr$ links source and target distributions over prompts: $p$ (source behavior) and $q$ (desired behavior). For instance, $p$ and $q$ may represent toxic versus non-toxic sentences (\cref{sec:tox}), or prompts inducing versus suppressing specific visual concepts in image generation. Each sampled prompt traces through the model's $L$ layers, inducing a sequence of activation distributions that LinEAS aligns through joint transport.
In practice, this means having access to a minibatch of samples $(\obs_i)_{i=1}^n\sim p$ and $(\obs'_j)_{j=1}^n\sim q$, tracking the execution trace of their $\ell\leq L$ activations, either modified through interleaved transports for samples from $p$:
\begin{equation}\label{eq:vx}
\vxi_i^{(\ell)}:=\RT{\ell}\circ \Bf{\ell} \circ \RT{\ell-1} \circ\dots \circ \RT{1} \circ \Bf{1}(\obs_i),\quad \ell\leq L,
\end{equation}
or run through the original network for samples of $q$:
\begin{equation}\label{eq:vy}
\veta_j^{(\ell)}:=\Bf{\ell} \circ \dots \circ \Bf{1}(\obs'_j),\quad \ell\leq L.
\end{equation}
The goal is to learn \textit{jointly} all $L$ transport maps so that, for each $\ell\leq L$, the families of vectors $(\vxi_i^{(\ell)})_{i=1}^n$ and $(\veta_j^{(\ell)})_{j=1}^n$ are \textit{similar} with respect to a distributional discrepancy $\Delta$ between probability measures, making the cost below small:
\begin{equation}\label{eq:cost}
\mathcal{L}_{\mathrm{OT}}(\RT{1}, \dots, \RT{L};(\obs_i)_{i=1}^n, (\obs'_j)_{j=1}^n) := \sum_{\ell\leq L} \Delta((\vxi_i^{(\ell)})_{i=1}^n, (\veta_j^{(\ell)})_{j=1}^n).
\end{equation}
\textbf{Sliced Wasserstein Losses.}
To define $\Delta$ at each layer, the approach of~\citet{rodriguez2024controllinglanguagediffusionmodels} is adopted: summing the $d_\ell$ squared univariate Wasserstein distances between the $d_\ell$ marginal distributions at each layer $\ell$. This choice is motivated by the typical setting's high-dimensionality / low sample regime, $d_\ell \gg N$, with $N$ the number of prompts available for each distribution, which would hinder the use of more complex multivariate distributional losses that account more closely for cross-variable effects. Empirically, adding univariate quantities yields more robust loss estimation that translates to better downstream task performance than considering, e.g., Sinkhorn divergences~\citep{genevay2018learning}.

The activations can be arranged as matrices
\[\mU:=[\vxi_1^{(\ell)},\dots,\vxi_n^{(\ell)}]^\top,\quad \mV:=[\veta_1^{(\ell)},\dots,\veta_n^{(\ell)}]^\top\in\R^{n\times d_\ell},\]
with one sample per row; $\Delta$ of the two families in \cref{eq:cost} is $\Delta$ of these matrices. To compute their 1D-Wasserstein distance~\citep[Chap. 2]{santambrogio2015optimal}, these activations must be first sorted in increasing order along the sample axis, \ie each column separately:
\[\widetilde{\mU} := \sort(\mU),\quad \widetilde{\mV} := \sort(\mV)\]
to define the (squared) sliced Wasserstein distance~\citep{rabin2012wasserstein} computed only on the \textit{canonical directions}, namely:
\begin{equation}\label{eq:sw}
\Delta(\mU, \mV):= \sum_{k=1}^{d_\ell} W_2^2(\mU_{\cdot k}, \mV_{\cdot k}) = \frac1n\sum_{k=1}^{d_\ell} \|\widetilde{\mU}_{\cdot k} - \widetilde{\mV}_{\cdot k}\|^2,
\end{equation}
where $W_2$ is the 2-Wasserstein distance between the empirical distributions of the $k$-th coordinates, $\mU_{\cdot k}$ denoting the $k$-th column.

\textbf{Differences with \iti, \linearact and \reft.}
Both \iti~\citep{li2024inference} and \linearact~\citep{rodriguez2024controllinglanguagediffusionmodels} optimize each $\RT{\ell}$ \textit{independently} across layers, minimizing a single distributional difference in closed form (\linearact) or learning a linear classifier (\iti), and assuming all other layers are frozen. While arguably much faster, this also generates causal inconsistencies, which can be partially resolved as in \linearact by using a sequential approach: when $\RT{\ell-1}$ is trained, $\RT{\ell-1}$ is reused to recompute all activation distributions used for $\RT{\ell}$. This suboptimality is hypothesized to limit generalization. This independent approach also precludes trade-offs when choosing which activations to turn on/off across layers, which can easily be surfaced using sparsity regularizers. One fundamental difference with \reft~\citep{wu2024reft} is that \method does not need paired data, \ie sample $\obs'_i$ does not need to be a counterfactual of $\obs_i$ (paired indices are used here for the pairing methods), which is the case for \reft. While counterfactual data is important for applications like translation, there is a vast amount of applications (\eg toxicity) where paired data is not available.

\subsection{Parameterization and Regularization}
Following~\citet{rodriguez2024controllinglanguagediffusionmodels}, each map $\RT{\ell}$ is parameterized as an affine map for each activation $\ell\leq L$, namely for $\vx\in\R^{d_\ell}$, one has
\begin{equation}\label{eq:map}
\RT{\ell}(\vx) := \ROmega{\ell} \odot \vx + \Rb{\ell}, \quad \ROmega{\ell}, \Rb{\ell}\in\R^{d_\ell},
\end{equation}
where $\odot$ is the element-wise product. Write $\allw:=(\ROmega{1}, \dots, \ROmega{L})$ and $\allb:=(\Rb{1}, \dots, \Rb{L})$ for the collections of all scale and intercept parameters.
In what follows, since each map ${\RT{\ell}}$ is entirely parameterized through its scale/intercept parameters $\allw, \allb$, notation is overloaded to define
\[
\mathcal{L}_{\mathrm{OT}}(\allw, \allb ;(\obs_i)_{i=1}^n, (\obs'_j)_{j=1}^n) := \mathcal{L}_{\mathrm{OT}}(\RT{1}, \dots, \RT{L};(\obs_i)_{i=1}^n, (\obs'_j)_{j=1}^n).\]

\textbf{Sparse Regularization.}
A sparsity regularizer carries out both \textit{layer} and \textit{within-layer} selection of activations. This can be achieved by using structured regularization, using either 1-norms or 2-norms:
\begin{align*}
\mathcal{R}_1(\allw, \allb)&:= \sum_\ell\left(\|\ROmega{\ell}-\1\|_1+\|\Rb{\ell}\|_1\right),\\
\mathcal{R}_{\mathrm{G}}(\allw, \allb)&:=\sum_\ell\sqrt{d_\ell}\left(\|\ROmega{\ell} - \1\|_2 + \|\Rb{\ell}\|_2\right),
\end{align*}
resulting in a \textit{sparse group lasso} regularizer~\citep{simon2013sparse,pmlr-v70-yoon17a}, $
\mathcal{R}:= \beta_1 \mathcal{R}_{1} + \beta_{\mathrm{G}} \mathcal{R}_{\mathrm{G}}$, which can be added to the cost to result in:
\begin{equation}
\label{eq:objective}
\begin{aligned}
 \mathcal{L}(\allw, \allb) :={} & \E_{(\obs_i)_{i=1}^n \sim p, (\obs'_j)_{j=1}^n \sim q}\left[ \mathcal{L}_{\mathrm{OT}}(\allw, \allb; (\obs_i)_{i=1}^n, (\obs'_j)_{j=1}^n)\right] \\
 & + \gamma \mathcal{R}(\allw, \allb),
\end{aligned}
\end{equation}
where $\gamma$ controls the amount of sparsity, \ie larger $\gamma$ will result in fewer activations and/or layers being intervened on, with solution such that $\ROmega{\ell} \approx \1$ and $\Rb{\ell} \approx\0$.

\textbf{On the choice of sparsity hyperparameters.}\quad
The regularizer adds two hyperparameters, $\beta_1$ and $\beta_{\mathrm{G}}$, on top of $\gamma$; this is a small number, since the affine maps themselves have none.
Empirically, tuning $\gamma\in[0, 1]$ with $\beta_1 = \beta_{\mathrm{G}} = 1$ provides a range of trade-offs between conditioning and utility as reported in \cref{fig:tox_sparsity_data32}. This also reduces the number of additional hyper-parameters to tune to just one ($\gamma$) and enables automatic layer selection (see the appendix of \citet{lineas}, section ``Group Sparsity Trade-offs'').

\subsection{Optimization}
\textbf{Proximal SGD.}\quad
$\mathcal{L}$ in \cref{eq:objective} is optimized with proximal stochastic gradient descent (PSGD) with a learning rate of 0.1 and cosine decay (see the appendix of \citet{lineas} for the algorithm and the proximal operators). %
The method assumes access to 2 sets of unpaired $N$ prompts $(\obs_i)_{i=1}^N$ and $(\obs'_j)_{j=1}^N$ and runs PSGD on minibatches of activations $(\vxi_i^{(\ell)})_{i=1}^n$ and $(\veta_j^{(\ell)})_{j=1}^n$ of size $n$. Note that activations originating from $\obs'_j$ use the untouched network, and can be pre-computed beforehand.

\section{Experimental Results}
\label{sec:results}

\subsection{Toxicity Mitigation in LLMs}
\label{sec:tox}

This section analyzes LinEAS effectiveness at toxicity mitigation, comparing with prompting, \caa~\citep{rimsky2023steering}, \reft~\citep{wu2024reft}, \iti~\citep{li2024inference}, and \linearact~\citep{rodriguez2024controllinglanguagediffusionmodels} on three LLMs ranging from 1.5B to 7B parameters, by aligning the activations of $N=32$ toxic to $32$ non-toxic sentences sampled from the Jigsaw dataset~\citep{jigsaw-toxic-comment-classification-challenge}.

\textbf{Toxicity Metrics.} Toxicity mitigation is evaluated on the \textit{RealToxicityPrompts} (RTP) dataset~\citep{gehman2020realtoxicityprompts} and the \textit{Thoroughly Engineered Toxicity} (TET) dataset~\citep{luong2024realistic}. For RTP, the evaluation follows \citet{rodriguez2024controllinglanguagediffusionmodels}
by sampling 1000 prompts from the dataset and letting the model (intervened or not) complete them. For TET, the 2546 prompts provided are used. Generations are scored using the open-source Roberta toxicity classifier (\RTC)~\citep{logacheva-etal-2022-paradetox}. \toxclsrtp and \toxclstet denote the respective percentage of generations flagged as toxic on RTP and TET.

\textbf{Utility Metrics.} To measure whether the utility of the model is affected by these interventions, \pplwik (the perplexity obtained on a fixed set of 20k Wikipedia sentences~\citep{wikidump}) is reported, as well as the overall 5-shot accuracy on the MMLU compendium~\citep{hendryckstest2021}.

\textbf{Oracle Baselines.}
Two oracle baselines are introduced that require strong supervision directly from the \RTC~oracle (\ie the classifier used for test-time metrics), giving them a strong advantage. A LoRA adapter and LoFIT~\citep{yin2024lofit} are trained using RTC labels, yielding the oracle methods \lorartc~and \lofitrl. \cref{tab:toxicity_main} reports only \lofitrl~results as it outperformed \lorartc~(see the appendix of \citet{lineas}).

\textbf{Setup.} All methods have access to only 32 toxic and 32 non-toxic sentences (unpaired). LinEAS is optimized for 1K steps using SGD with learning rate 0.1. To focus on the benefit of end-to-end loss, $\gamma=0$ is used for LinEAS (see \cref{sec:tox_sparsity} for $\gamma$'s impact).
For \linearact and \caa, default settings are used with intervention strength $\lambda=1$. For \reft, training runs for 10 epochs (selected via epoch sweep).
For \iti, $\lambda=0.5$ is used (obtained through grid search). Toxicity mitigation is evaluated by intervening on different layers, reporting the best overall layer type per method: {\texttt{.*post.*layernorm}} for LinEAS and \texttt{.*o\_proj} for \iti and \linearact, according to the Huggingface implementation. Both \caa and \reft intervene on residuals of Transformer blocks, as suggested in their original works. For \caa, two baselines are run: intervening on all layers (reported in \cref{tab:toxicity_main}), and on the middle layer (as in the original paper; omitted given its worse performance, see the appendix of \citet{lineas}). Additionally, prompting is included as a conditioning strategy, preprompting with \textit{``Continue the text in a non-toxic way:''}.

\definecolor{Gray}{gray}{0.95}
\begin{table}[th]
 \centering
 \resizebox{0.83\columnwidth}{!}{
\begin{tabular}{llllrrrr}
\toprule
Model & Method & \#Params & \toxclsrtp $(\downarrow)$ & \toxclstet $(\downarrow)$ & \pplwik $(\downarrow)$ & MMLU $(\uparrow)$ \\

\arrayrulecolor{black} \midrule
\rowcolor{Gray} \multirow{8}{*}{\cellcolor{white}\qwenonefiveb} & None & - & $3.00_{\color{gray}{\,0.54}}$ & $23.09_{\color{gray}{\,0.67}}$ & $13.67_{\color{gray}{\,0.00}}$ & $60.95_{\color{gray}{\,0.00}}$ \\
\rowcolor{Gray}\cellcolor{white} & LoFIT-RL & 0.86M & $0.37_{\color{gray}{\,0.06}}$ & $4.36_{\color{gray}{\,0.00}}$ & $14.12_{\color{gray}{\,0.07}}$ & $59.74_{\color{gray}{\,0.14}}$ \\
\arrayrulecolor{gray!50} \cmidrule{2-7}

 & \prompt & - & $4.07_{\color{gray}{\,0.38}}$ & $21.02_{\color{gray}{\,1.44}}$ & $13.65_{\color{gray}{\,0.00}}$ & $60.96_{\color{gray}{\,0.00}}$ \\
 & \caa\hspace{-1mm}${}^\star$ & 0.043M & $1.15_{\color{gray}{\,0.37}}$ & $5.77_{\color{gray}{\,2.14}}$ & $\red{19.30}_{\color{gray}{\,2.76}}$ & $\red{37.67}_{\color{gray}{\,6.95}}$ \\
 & \reft\hspace{-1mm}${}^\star$ & 0.39M & $2.57_{\color{gray}{\,0.60}}$ & $18.17_{\color{gray}{\,3.04}}$ & $15.58_{\color{gray}{\,0.52}}$ & $58.84_{\color{gray}{\,0.23}}$ \\

 & \iti & 0.043M & $1.87_{\color{gray}{\,0.21}}$ & $18.16_{\color{gray}{\,0.62}}$ & $12.39_{\color{gray}{\,0.09}}$ & $60.88_{\color{gray}{\,0.08}}$ \\
 & \linearact & 0.086M & $1.50_{\color{gray}{\,0.35}}$ & $13.88_{\color{gray}{\,1.72}}$ & $13.88_{\color{gray}{\,0.16}}$ & $60.09_{\color{gray}{\,0.25}}$ \\
 & \method & 0.086M & $\textbf{1.07}_{\color{gray}{\,0.46}}$ & $\textbf{12.70}_{\color{gray}{\,0.74}}$ & $14.10_{\color{gray}{\,0.07}}$ & $59.97_{\color{gray}{\,0.16}}$ \\
\arrayrulecolor{black} \midrule

\rowcolor{Gray} \multirow{8}{*}{\cellcolor{white}\gemmatwob} & None & - & $4.00_{\color{gray}{\,0.45}}$ & $13.39_{\color{gray}{\,1.42}}$ & $14.79_{\color{gray}{\,0.00}}$ & $53.03_{\color{gray}{\,0.00}}$ \\
\rowcolor{Gray}\cellcolor{white} & LoFIT-RL & 0.11M & $0.40_{\color{gray}{\,0.20}}$ & $1.76_{\color{gray}{\,0.00}}$ & $15.43_{\color{gray}{\,0.08}}$ & $52.17_{\color{gray}{\,0.17}}$ \\
\arrayrulecolor{gray!50} \cmidrule{2-7}

 & \prompt & - & $4.60_{\color{gray}{\,0.36}}$ & $12.32_{\color{gray}{\,0.67}}$ & $14.81_{\color{gray}{\,0.00}}$ & $53.18_{\color{gray}{\,0.00}}$ \\
 & \caa\hspace{-1mm}${}^\star$ & 0.06M & $0.80_{\color{gray}{\,0.00}}$ & $2.44_{\color{gray}{\,1.99}}$ & $\red{23.52}_{\color{gray}{\,2.67}}$ & $\red{26.86}_{\color{gray}{\,0.08}}$ \\
 & \reft\hspace{-1mm}${}^\star$ & 0.54M & $2.85_{\color{gray}{\,0.49}}$ & $11.15_{\color{gray}{\,1.91}}$ & $\red{19.93}_{\color{gray}{\,0.30}}$ & $48.99_{\color{gray}{\,1.34}}$ \\

& \iti & 0.06M & $1.17_{\color{gray}{\,0.60}}$ & $7.15_{\color{gray}{\,0.92}}$ & $14.00_{\color{gray}{\,0.11}}$ & $52.78_{\color{gray}{\,0.23}}$ \\
& \linearact & 0.12M & $1.60_{\color{gray}{\,0.32}}$ & $7.76_{\color{gray}{\,0.39}}$ & $14.78_{\color{gray}{\,0.12}}$ & $52.43_{\color{gray}{\,0.57}}$ \\
& \method & 0.24M & $\textbf{0.73}_{\color{gray}{\,0.10}}$ & $\textbf{4.02}_{\color{gray}{\,0.68}}$ & $15.46_{\color{gray}{\,0.21}}$ & $52.22_{\color{gray}{\,0.40}}$ \\

\arrayrulecolor{black} \midrule
\rowcolor{Gray} \multirow{8}{*}{\cellcolor{white}\qwensevenb} & None & - & $3.92_{\color{gray}{\,0.59}}$ & $25.16_{\color{gray}{\,0.92}}$ & $10.67_{\color{gray}{\,0.00}}$ & $74.26_{\color{gray}{\,0.00}}$ \\
\rowcolor{Gray}\cellcolor{white} & LoFIT-RL & 0.10M & $1.10_{\color{gray}{\,0.38}}$ & $7.11_{\color{gray}{\,0.30}}$ & $10.91_{\color{gray}{\,0.16}}$ & $73.87_{\color{gray}{\,0.17}}$ \\
\arrayrulecolor{gray!50} \cmidrule{2-7}

 & \prompt & - & $6.80_{\color{gray}{\,0.00}}$ & $21.22_{\color{gray}{\,0.21}}$ & $10.65_{\color{gray}{\,0.00}}$ & $74.23_{\color{gray}{\,0.00}}$ \\
 & \caa\hspace{-1mm}${}^\star$ & 0.10M & $1.20_{\color{gray}{\,0.00}}$ & $9.25_{\color{gray}{\,3.07}}$ & $12.83_{\color{gray}{\,0.00}}$ & $\red{48.58}_{\color{gray}{\,0.00}}$ \\
 & \reft\hspace{-1mm}${}^\star$ & 0.90M & $3.33_{\color{gray}{\,0.96}}$ & $20.38_{\color{gray}{\,2.37}}$ & $13.80_{\color{gray}{\,1.20}}$ & $70.43_{\color{gray}{\,0.60}}$ \\

 & \iti & 0.10M & $2.63_{\color{gray}{\,0.44}}$ & $19.98_{\color{gray}{\,1.24}}$ & $9.63_{\color{gray}{\,0.03}}$ & $74.08_{\color{gray}{\,0.05}}$ \\
 & \linearact & 0.20M & $2.72_{\color{gray}{\,0.46}}$ & $21.64_{\color{gray}{\,2.00}}$ & $11.42_{\color{gray}{\,0.34}}$ & $72.18_{\color{gray}{\,0.16}}$ \\
 & \method & 0.20M & $\textbf{1.95}_{\color{gray}{\,0.48}}$ & $\textbf{14.95}_{\color{gray}{\,0.92}}$ & $10.91_{\color{gray}{\,0.35}}$ & $73.67_{\color{gray}{\,0.05}}$ \\
\arrayrulecolor{black}\bottomrule
\end{tabular}}
\vskip 3mm
\caption{Toxicity mitigation on the RTP and TET datasets using three different models, \qwenonefiveb, \gemmatwob, and \qwensevenb. Strongly degraded utility is marked in \red{red}. Results shown are at low data regime ($N=32$ sentences to estimate the interventions). Results for \method improve on \iti and \linearact with similar impact on utility metrics. The quality of these interventions is often on par with the oracle baseline \lofitrl in terms of utility, although the strong oracle supervision yields better mitigation. \textit{$^\star$\caa and \reft are designed to use paired data, which is not available in the toxicity setting.}}
 \label{tab:toxicity_main}
\vspace{-1em}
\end{table}

\textbf{Results.} \cref{tab:toxicity_main} summarizes the toxicity mitigation experiments averaged over 4 generation seeds (and RTP samplings). \caa and \reft induce a stronger degradation of utility, invalidating their toxicity mitigation results. Note that both methods are designed for paired data, which does not exist in the toxicity setup, so they are not used in their nominal setting, which affects their performance.
Prompting is not effective for the models tested, and even increases \toxclsrtp.
\method achieves a consistent toxicity mitigation, outperforming all steering methods that preserve utility (\iti, \linearact, prompting) at this low data regime. For example, \method reduces \gemmatwob toxicity by $5.5\times$, getting closer to the mitigation obtained with the oracle \lofitrl.
In terms of utility, \method shows a minimal degradation, with values very similar to the utility incurred by the oracle \lofitrl. In absolute terms, \method reduces MMLU by less than 1 point and increases \pplwik by at most 0.7.
\method demonstrates much greater robustness to layer choice than \iti and \linearact (see the appendix of \citet{lineas}). Additionally, \cref{fig:data_ablation} provides an analysis on higher data regimes, up to $N=1024$ sentences, showing that \method remains reliable in such scenarios.
The compute needed to train these methods differs widely:~in the low data regime $N=32$, estimating each method on a Nvidia A100-80Gb GPU and \qwensevenb takes 37s (\iti), 30s (\linearact, which uses closed forms), 500s (LinEAS for 1K steps) and 27300s for \lofitrl.

\textbf{User Study.} A user preference study complements the quantitative results by evaluating perceived quality of continuations generated by different intervention methods. The findings indicate strong preference for LinEAS over three alternatives: \linearact, \iti, and no intervention (identity). Specifically, users preferred LinEAS in $57.70\%$ of cases compared to $18.43\%$ for \linearact and $11.67\%$ for \iti.

\begin{figure*}[t!]
 \centering
 \includegraphics[trim={0 0 0 0},clip,width=\textwidth]{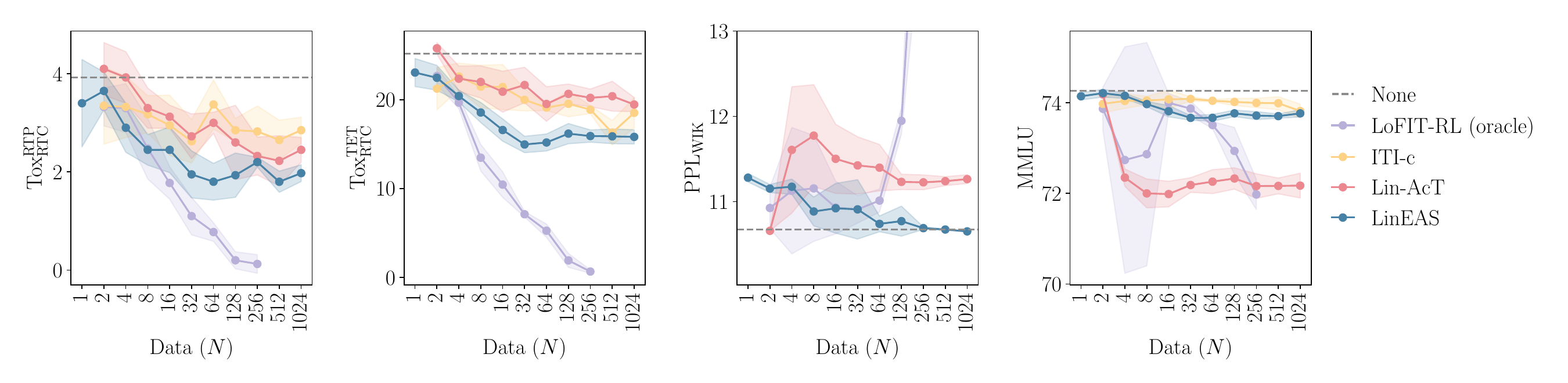}
 \vskip -1mm
 \caption{\textbf{\method is effective at low data regime.} Toxicity mitigation (two left-most plots) and utility (two right-most plots) as a function of the amount of data available to learn interventions. \method shows better performance (low toxicity and utility close to original dashed lines) for low data, and stable performance for $N\geq 32$. ITI-c in the legends denotes \iti.}
 \label{fig:data_ablation}
\end{figure*}
\subsection{Effect of Data on Toxicity Mitigation}
\label{sec:tox_data}

With the same toxicity setting as in \cref{sec:tox}, the amount of data used to estimate interventions is ablated using \qwensevenb (best model studied in terms of MMLU). The sweep ranges from $N$ from 1 to 1024, meaning access to $N$ toxic and $N$ non-toxic sentences. Note that the results in \cref{tab:toxicity_main} correspond to $N=32$.
\Cref{fig:data_ablation} plots the evolution of toxicity (\toxclsrtp and \toxclstet) and utility (\pplwik and MMLU) as a function of $N$, averaged over 4 random sweeps (standard deviation as shaded areas). \method achieves superior toxicity mitigation even at very low data regimes, while maintaining utility close to the original model (horizontal dashed lines) and the \lofitrl oracle. Moreover, \method's performance is more stable for a large range of $N$ (32 to 1024), even more stable than the oracle which diverges for $N>128$. Hyperparameters are fixed for all methods, including the oracle, which shows greater sensitivity to training settings.

\begin{figure*}[t!]
 \centering
 \includegraphics[trim={0 0 0 0},clip,width=\textwidth]{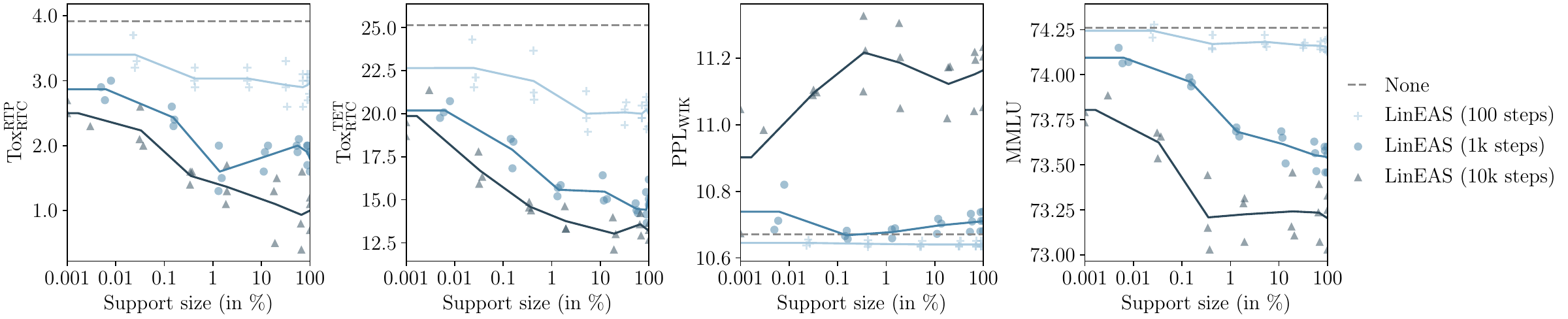}
 \vskip -1mm
 \caption{\textbf{Sparsity improves utility while mitigating toxicity}. Toxicity results on \qwensevenb using only 32 sentences, at different levels of sparsity $\gamma$ that result in different support sizes (x axis). At 1K optimization steps, with a support of about 1\%, similar toxicity is maintained (left, center-left) while \pplwik decreases (center-right) and MMLU increases (right). Note that too long optimizations (10k steps) might harm utility due to overfitting. Similarly, short optimizations (\eg 100 steps) and strong sparsity lead to low conditioning (mild toxicity mitigation).}
 \label{fig:tox_sparsity_data32}
\end{figure*}

\subsection{Effect of Sparsity on Toxicity Mitigation}
\label{sec:tox_sparsity}

Intuitively one should only steer the smallest set of activations needed to achieve a desired goal in order to preserve utility and keep most of the model's inference graph untouched. This section explores how sparsity affects toxicity mitigation in the setup of \cref{sec:tox} as $\gamma$ increases from $0$ to $0.1$. Increasing $\gamma$ results in fewer activations being \textit{transported}, measured as $\mathrm{support} = \|(\allw \neq \1) + (\allb \neq \0)\|_0 \big/ \sum_{\ell=1}^{L} d_\ell$, \ie all activations transported either by rescaling or shifts.

\Cref{fig:tox_sparsity_data32} shows how \toxclsrtp and \toxclstet, as well as the utility \pplwik and MMLU, evolve as the sparse support decreases (x axis), for \qwensevenb in the low ($N=32$) data regime. The figure shows results of 3 sweeps of $\gamma$ with different random seeds (markers), plotting the average at each $\gamma$ level (line). Note that at $\gamma=0$ the support is approximately 100\%. In the case of 1K steps (reported in \cref{tab:toxicity_main}), one can afford reducing the support to about $1\%$ and still maintain the toxicity mitigation values at 100\% support. At these support values, \pplwik and MMLU improve, validating the hypothesis that smaller supports help preserve utility. Additionally, short optimizations (\eg 100 steps) lead to mild conditioning (poor toxicity mitigation) while long optimizations (\eg 10k steps) lead to gradual degradation in utility.

\subsection{Steering Text-to-Image Generation Models}
\label{sec:results_diffusion}

This section explores how \method can be used to remove the presence of concepts in text-to-image (T2I) generation: a task that plays a key role in generation alignment, similar to toxicity mitigation in LLMs. To preserve the focus on low data and compute budget,
\method is applied to DMD2~\citep{yin2024dmd2}, a recent single-step text-to-image generation model distilled from SDXL~\citep{podellsdxl} with a GAN~\citep{goodfellow2014generative} loss. Note that diffusion guidance methods are available to condition these models; however, they typically require multiple denoising steps (even up to 200)~\citep{Ye2024-TFG, Guo2024-GGDM} and, at times, trajectory resampling and evaluation strategies \citep{Singhal2025-FK}. These additional overheads make them suitable for settings with higher compute budget, which differs from the low-compute setting examined here.

\textbf{Setup.} As in \citet{rodriguez2024controllinglanguagediffusionmodels}, the strength of \method applied to all \texttt{layernorm} layers is modulated by introducing a scale $0\leq \lambda\leq 1$ when applying interventions. Intuitively $\lambda=0$ results in no intervention, while $\lambda=1$ carries out a full \method transport, any value in between reflecting a gradual change. Following~\cref{sec:tox}, the focus is on concept mitigation/removal using $32$ samples for each the source and the target distribution. \method is trained for 1000 steps with batch size 4, AdamW, learning rate of $10^{-4}$ and $\gamma=0$. LinEAS is compared with \iti and \linearact, also weakly supervised methods working with \textit{unpaired data}.

\textbf{Data.} An open-source LLM is queried for a diverse set of prompts covering 3 different conditioning categories and 5 different concepts per category with 32 prompts per concept, totalling 480 prompts. \textbf{Styles}: \texttt{vaporwave, lush, low-poly, ukiyo-e, fantasy}; \textbf{objects}: \texttt{robots, axolotl, book, car, hourglass}; and \textbf{perspective}: \texttt{macro, fisheye, bokeh, underwater, and tilt-shift}. The same model is queried to obtain 32 neutral prompts for evaluation.

\textbf{Metrics.} The evaluation measures (1) \textbf{\clipscore}~\citep{hessel2021clipscore}, the cosine similarity between CLIP embeddings of the generated images and the concept description.
(2) \textbf{\imgscore}, the cosine similarity between DINOv2-small~\citep{oquab2024dinov} image embeddings generated with $\lambda>0$ and images generated with $\lambda=0$ using the same prompts.
\clipscore assesses whether the generations are truthful to the desired style, and \imgscore whether they are perceptually similar to those generated without intervention.

\begin{figure}[t]
\captionsetup{margin=0pt}
\begin{minipage}{0.49\linewidth}
\centering
\resizebox{\linewidth}{!}{%
\begin{tabular}{l@{\hskip 3pt}c@{\hskip 3pt}c@{\hskip 3pt}c}
\toprule
Method & $\uparrow$ User Pref. & $\uparrow$ IMGSc.
 & $\downarrow$ CLIPSc. \\
\midrule
\iti & $12.4_{\color{gray}{\,5.5}}\%$ & $0.24_{\color{gray}{\,0.19}}$ & $0.19_{\color{gray}{\,0.02}}$ \\
\linearact & $24.4_{\color{gray}{\,7.0}}\%$ & $0.45_{\color{gray}{\,0.21}}$ & $0.18_{\color{gray}{\,0.03}}$ \\
\method & \textbf{63.3}$_{\color{gray}{\,6.6}}$\% & \textbf{0.66}$_{\color{gray}{\,0.19}}$ & $0.18_{\color{gray}{\,0.03}}$ \\
\bottomrule
\end{tabular}}
\captionof{table}{\method mitigates concepts on DMD2~\citep{yin2024dmd2} while staying perceptually similar to the original image. Users prefer \method 63.3\% of the time (left) since it maintains a higher fidelity to the non-intervened original model when using the same prompt (center), and matches other methods at concept removal (right). Results were obtained with $\lambda=1$ and they are aggregated across all concepts.}
\label{tab:t2i-quantitative}
\end{minipage}\hfill%
\begin{minipage}{0.49\linewidth}
\centering
\includegraphics[trim={0 0 0 0},clip,width=0.8\linewidth]{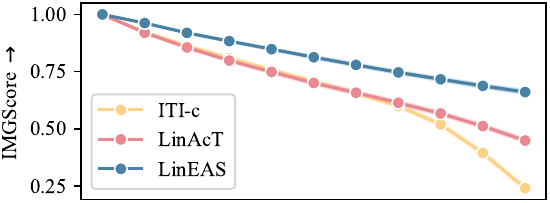}
\includegraphics[trim={0 0 0 0},clip,width=0.8\linewidth]{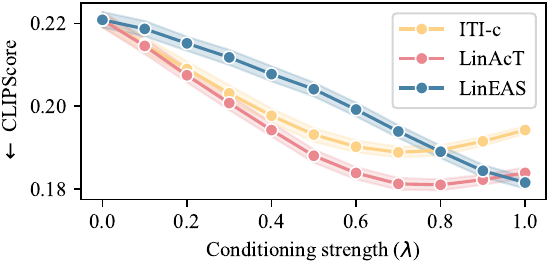}
\vskip -2mm
\captionof{figure}{\imgscore and \clipscore at multiple conditioning strengths ($\lambda$). (top) \method is more faithful to DMD2 with $\lambda=0$. (bottom) \method is more consistent and predictable at mitigation. ITI-c and LinAcT in the legends denote \iti and \linearact.}
\vskip -4mm
\label{fig:t2i-curves}
\end{minipage}
\end{figure}

\textbf{User study.} To complement the quantitative analysis, a user study was conducted with 20 prompts $\times$ 5 conditioning concepts, yielding 100 pairs total. For each pair, an original image is generated and steered using \iti, \linearact, or LinEAS on 5 concepts. A pool of 10 participants selected their preferred output (1000 evaluations total), showing strong preference for LinEAS as reported in \cref{tab:t2i-quantitative} (left). The question posed to participants was \textit{``Which image blends best prompt and style, while remaining faithful to the untouched output?''}.

\textbf{Results.} \cref{tab:t2i-quantitative} summarizes the results for text-to-image evaluation. \method is preferred in 63.3\% of the evaluations (24.4\% for \linearact, and 12.4\% for \iti). These results agree with the automated metrics: LinEAS is more faithful to the images produced by DMD2 with $\lambda=0$ for the same prompts, with an \imgscore of 0.66 compared to 0.45 for \linearact and 0.24 for \iti while all methods achieve similar \clipscore.
\Cref{fig:t2i-curves} explores how \imgscore and \clipscore change with $\lambda$. LinEAS is consistently more faithful to DMD2 with $\lambda=0$ in terms of \imgscore (top) while showing strong linear correlation with \clipscore, making LinEAS more consistent and predictable.

\begin{figure*}[htb]
 \centering
 \includegraphics[width=\linewidth]{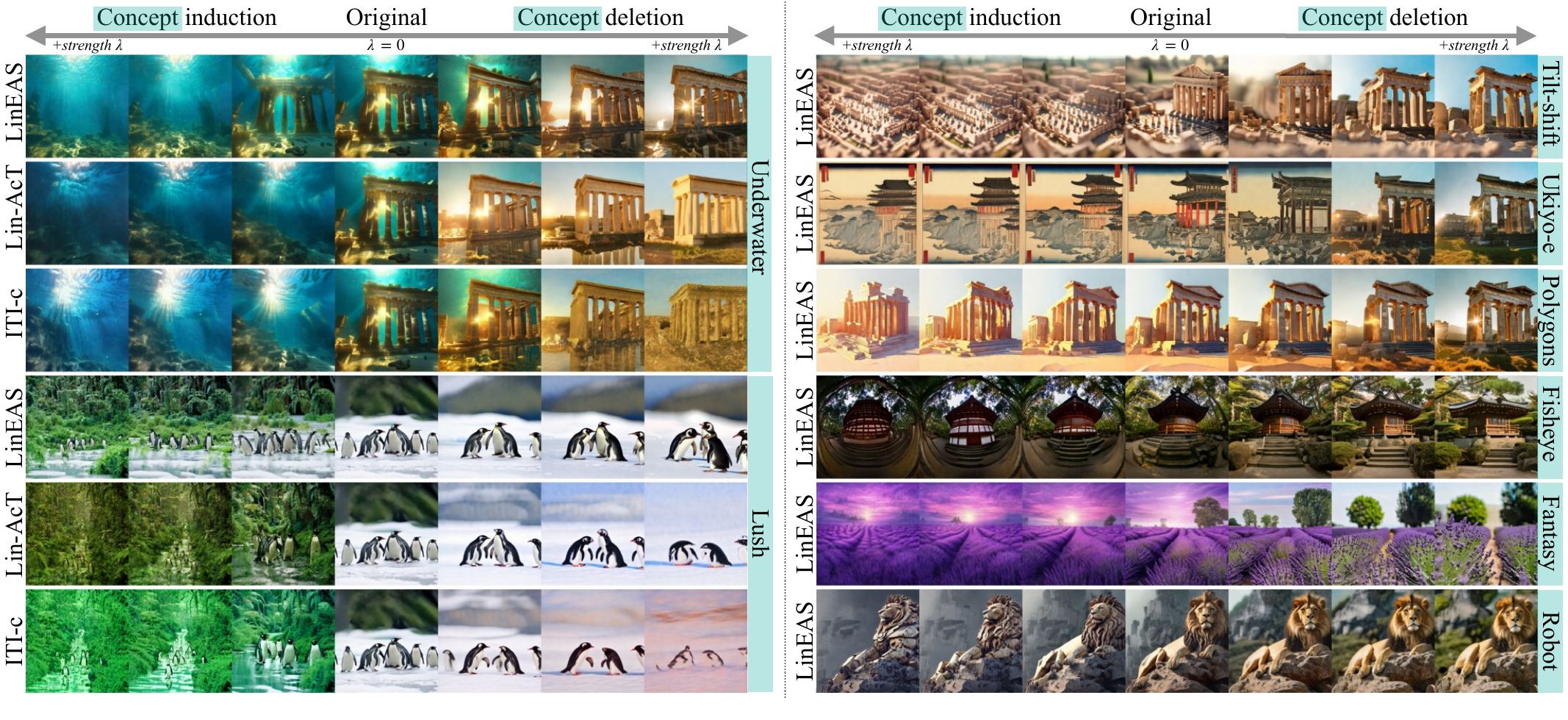}
 \vskip -1mm
 \caption{Generations using DMD2~\citep{yin2024dmd2}. (Left) Concept deletion using \iti, \linearact and \method for two different concepts, starting from prompts that contain the concept. \method shows a more gradual deletion ($\lambda=0.4, 0.7, 1$), and better preservation of the original image ($\lambda=0$). (Right) Qualitative examples of \method on 6 more concepts. Inverting the steering maps results in concept induction, probably due to strong structure in activation space. \method also outperforms the other methods under this setting. ITI-c in the row labels denotes \iti.}
 \label{fig:diffusion_comparison}
\end{figure*}

\textbf{Inverted LinEAS}. Surprisingly, inverting the affine maps in \method,
\[\RT[-1]{\ell}(\vx) := (\vx - \Rb{\ell}) \odot \frac{1}{\ROmega{\ell}}\]
(defined when every entry of $\ROmega{\ell}$ is non-zero), tends to negate the conditioning thus switching from mitigation to induction and vice-versa (see ``concept induction'' in \cref{fig:diffusion_comparison}). We speculate that this behavior is due to a strong structure in the activation space.

\FloatBarrier
\subsection{Layer Selection Analysis}
\label{sec:layer_ablation}

To evaluate LinEAS robustness with respect to intervened layer choice, a sweep over different layer types within the DMD2 UNet was conducted. \cref{tab:layer_ablation} presents results averaged over 15 concepts, measuring image consistency using IMGScore (higher is better) and concept removal using CLIPScore (lower is better).

The findings indicate that while intervening on all layer normalization modules (the setting used in main experiments) provides the best trade-off between metrics, LinEAS demonstrates robust performance across all tested layer configurations. This suggests that activation steering is not overly sensitive to specific layer choices in UNets.

\begin{table}[h!]
\centering
\resizebox{0.83\columnwidth}{!}{\begin{tabular}{l c c c}
\toprule
\textbf{Intervened Layers} & \textbf{\# Modules} & \textbf{IMGScore} $\uparrow$ & \textbf{CLIPScore} $\downarrow$ \\
\midrule
\texttt{unet} - all layer norms & 256 & $0.714 \pm 0.054$ & $0.131 \pm 0.033$ \\
\texttt{unet.transformer} - all FeedForward & 70 & $0.780 \pm 0.053$ & $0.137 \pm 0.031$ \\
\texttt{unet} attentions - K and Q projections & 280 & $0.935 \pm 0.016$ & $0.157 \pm 0.028$ \\
\texttt{unet} attentions - V projections & 140 & $0.818 \pm 0.051$ & $0.140 \pm 0.031$ \\
\texttt{unet} attentions - in projections & 11 & $0.923 \pm 0.017$ & $0.155 \pm 0.028$ \\
\texttt{unet.resnet} - all layer norms & 34 & $0.896 \pm 0.023$ & $0.156 \pm 0.027$ \\
\bottomrule
\end{tabular}}

\caption{Study on the choice of intervened layers for LinEAS in the DMD2 UNet. Image consistency (IMGScore $\uparrow$) and concept removal (CLIPScore $\downarrow$) are reported, averaged over 15 concepts. The method shows robustness, with the default setting (all layer norms) offering the best trade-off.}
\label{tab:layer_ablation}
\end{table}

\makeatletter
\let\old@endpart\@endpart
\def\@endpart{\vspace{3cm}}
\makeatother

\part{Conclusions and Outlook}

\begin{center}
\includegraphics[width=0.65\textwidth]{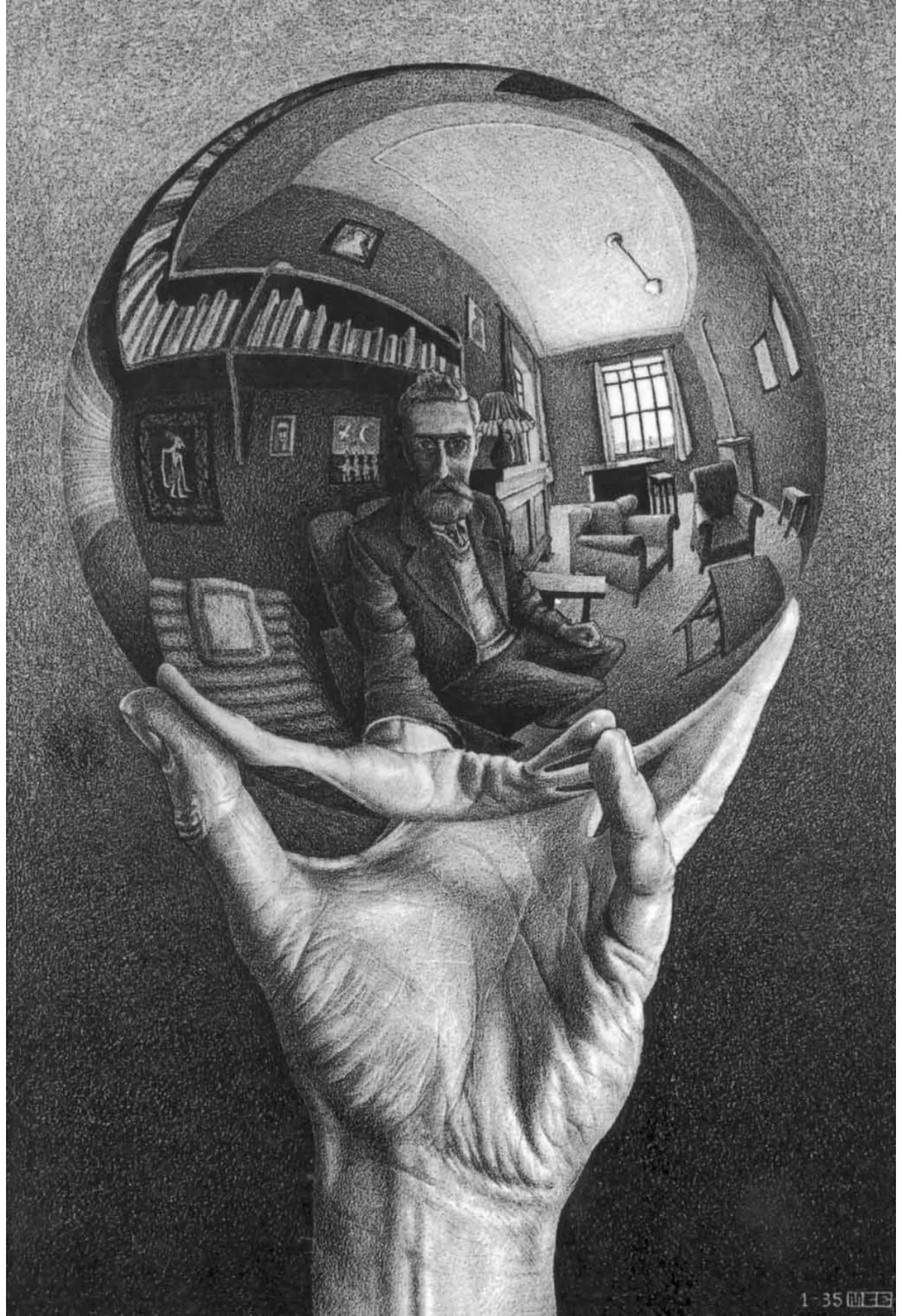}
\captionof*{figure}{\textbf{Hand with Reflecting Sphere} by M.C. Escher (1935). The sphere reflects what no single viewpoint can capture---the world distorted but complete. In reflection, we see where we began, the path we took, where we have arrived, and perhaps where the journey leads next.}
\end{center}
\vfill
\begin{center}\footnotesize\textcopyright{} The M.C. Escher Company, The Netherlands. All rights reserved. Reproduced for academic, non-commercial use.\end{center}

\makeatletter
\let\@endpart\old@endpart
\makeatother

\clearpage

\chapter{Conclusions}
\label{ch:conclusions}

\section{Synthesis and Insights}

This thesis demonstrates that semantic correspondence systematically reveals geometric structure shared across independently trained models, enabling zero-shot transformation between their representations.

The framework decomposes representation alignment into correspondence-based extraction (\cref{ch:relative-representations,ch:product-invariances,ch:asif}), correspondence discovery from few or no paired samples (\cref{ch:bootstrapping,ch:learning_to_align}), and transformation (\cref{ch:semantic_alignment,ch:lfm,ch:inverse-relative-projection}), with applications to neuroscience, uncertainty quantification, and reinforcement learning (\cref{ch:domain-applications}), interpretability (\cref{ch:residual,ch:headpursuit}), and activation steering (\cref{ch:headpursuit,ch:lineas}). Across fourteen methods spanning vision, language, reinforcement learning, generative models, and neuroscience, the principle holds: when models extract overlapping semantic structure from data and objectives, their representations exhibit alignable geometry despite divergent parameterizations. These contributions organize by geometric structure---angular relationships, spectral properties, orthogonal alignment, distributional correspondence, component specialization---as detailed in \cref{ch:introduction}.

\textbf{Operator families.} The framework decomposes into two complementary operator families (\cref{ch:methodology}). \Glspl{correspondence-parallel} extract semantic coordinates independently per space, enabling comparison and composition across arbitrary architectures at the cost of decoder compatibility. \Glspl{correspondence-coupled} construct direct transformations using paired observations, preserving decoder compatibility at the cost of per-pair estimation. Inverse relative projection (\cref{ch:inverse-relative-projection}) bridges the families, demonstrating that modular semantic coordinates can be chained with reconstruction to recover target-native coordinates without joint training.

\textbf{Task objectives and alignability.} The cross-domain empirical evidence reveals a deeper pattern: correspondence-based alignment succeeds when task objectives force models to disentangle semantic structure from nuisance variation. We read discriminative tasks---classification, matching, policy optimization---as imposing this pressure, requiring models to extract invariances that correspondence operators exploit. Subsequent large-scale studies have corroborated our findings: \citet{huh2024platonic} measured mutual $k$-nearest-neighbor alignment among 78 vision models and between vision models and a dozen language models, framing the observation as the ``Platonic Representation Hypothesis''. Later work qualifies the hypothesis, arguing that what converges is local neighborhood structure, and that global geometry does not~\citep{groger2026aristotelian}.

\textbf{Data efficiency.} Beyond task objectives determining alignability, unexpected patterns emerged across methods. Data efficiency exceeded initial expectations: bootstrapping correspondence (\cref{ch:bootstrapping}) achieves comparable results with one order of magnitude fewer anchors than ground truth supervision, and learning to align (\cref{ch:learning_to_align}) needs no paired samples at all, since learned embeddings amortize the Gromov-Wasserstein problem and aligning new samples requires a single optimal transport solve. ASIF (\cref{ch:asif}) assembles zero-shot multimodal models competitive with CLIP using a fraction of the training data, raising questions about the relative roles of parametric learning versus memory-augmented retrieval. The methods scale from minimal supervision (32 unpaired samples per distribution suffice for LinEAS toxicity mitigation) to massive correspondence sets (ASIF handles 1.6 million anchor pairs through sparsification), suggesting that geometric priors constrain alignment more powerfully than raw sample counts alone.

\textbf{Interpretability through geometry.} Interpretability follows from geometric structure and requires no separate analysis machinery. ASIF factorizes perception and interpretation, relieving encoders from attributing meaning while enabling analysis through relative representations. Relative representations themselves identify which anchors govern decisions through angular proximity in semantic coordinates. Latent functional maps expose off-diagonal distortion structure through spectral decomposition, localizing where transformations break down. ResiDual reveals emergent specialization of attention heads through spectral geometry of residual representations, identifying task-relevant principal components in the residual stream. HeadPursuit extends this by decomposing head outputs through simultaneous orthogonal matching pursuit over the unembedding space, recovering semantically specialized heads whose intervention produces targeted generation changes. These geometric views complement prediction-level attribution, offering architectural understanding of which structural components encode semantic distinctions.

\textbf{Cross-domain compatibility.} The framework generalizes beyond artificial model alignment. Cross-domain applications extend the principle to reinforcement learning, ensemble uncertainty, and fMRI data (\cref{ch:domain-applications}): reinforcement learning agents align through \glspl{correspondence-parallel}; in ensembles of networks with different parameterizations, the relative projection separates reparameterization effects from model uncertainty; and human brain fMRI representations align across individuals when a task-informed encoder precedes the relative projection, which suggests that the same correspondence principles extend to biological neural systems (\cref{sec:cosyne}). This breadth suggests the framework captures fundamental properties of semantic organization; these properties are not artifacts of specific training procedures or architectures.

\section{Limitations and Boundaries}

\textbf{Semantic overlap as prerequisite.} The fundamental constraint is semantic overlap: we expect models trained on unrelated domains or objectives with incompatible semantic emphasis not to expose alignable geometry, although no chapter tests this case. If so, this behavior is a feature: it shows that alignment exposes intrinsic structure and does not produce arbitrary coordinate mappings. \Glspl{correspondence-parallel} preserve structure up to the metric used to compute them. This delimits the scope of correspondence-based methods while pointing to the geometric prerequisites for alignability.

\textbf{Stitching and validation strategy.} Recent work~\citep{smith2025functional} cautions that successful stitching does not necessarily imply representational similarity. Stitching evaluates whether representations accommodate downstream components (classifiers, decoders), which constrains only task-relevant geometry. The critical asymmetry: our framework assumes semantic overlap and seeks geometric structure that should follow from shared semantics, but geometric compatibility alone does not guarantee semantic similarity. Two representations can stitch successfully when task-specific structure happens to align, even when broader semantic organization differs. This matters for validation strategy. Methods validated purely through task performance---stitching accuracy, classification transfer---measure functional compatibility, which can mislead when \textit{directly optimized}. However, our experiments primarily rely on \textit{zero-shot} stitching or structural validation: latent similarity (neighborhood overlap in \cref{ch:relative-representations}, \gls{cka} in \cref{ch:product-invariances}), reconstruction fidelity (\cref{ch:inverse-relative-projection}), cross-domain consistency (\cref{ch:semantic_alignment}), and spectral properties (\cref{ch:lfm,ch:residual}), which probe intrinsic geometric structure and do not depend on task-specific compatibility.

\textbf{Projection and dimensionality.} Specific technical limitations emerged across methods. \Glspl{correspondence-parallel} such as the relative projection incur information loss by projecting $\R^d$ to $\R^{\nanch}$, with $\nanch$ the number of anchors, trading fidelity for modularity. Zero-shot stitching through inverse relative projection, which passes through this projection, trails each classification head used on its own encoder (the non-stitch column of \cref{tab:stitching}) by 3--20 accuracy points. \citet{fumero2024latent} report that the performance of latent functional maps can be sensitive to the number of eigenvectors, and that finding the right count without extensive experimentation remains open; latent functional maps also assume full correspondence between domains and are not adapted to partial mappings where only subsets align~\citep{fumero2024latent}.

\textbf{Architecture-specific requirements.} Architecture-specific requirements constrain applicability. ResiDual fundamentally extracts information from residual units already containing task-relevant principal components; when this assumption fails, \citet{basile2025residual} expect ResiDual not to recover the alignment, with a drop in downstream performance. Its focus on the \texttt{[CLS]} token ignores sequence-level information, though preliminary results not included in \cref{ch:residual} suggest high spectral compatibility extends to patch tokens~\citep{basile2025residual}. As \citet{headpursuit} note, HeadPursuit's simultaneous orthogonal matching pursuit imposes linearity assumptions that may miss nonlinear head representation structure, and its interventions apply uniformly across all tokens, without selecting by position or modality.

\textbf{Deployment contexts.} Method-specific trade-offs emerge in deployment contexts. ASIF's performance depends critically on semantic coverage of correspondence anchors~\citep{norelli2023asif}: we expect large gaps between queries and available anchors to force extrapolation beyond the anchor set, a constraint shared by similarity-based methods. As \citet{norelli2023asif} note, when multimodal data is abundant and training cost acceptable, ASIF's non-parametric formulation still falls short of contrastively trained models like CLIP, and the high dimensionality of relative representations, even when sparse, poses challenges for generation tasks like text-to-image synthesis.

\textbf{Computational scaling.} We expect computational scaling to limit deployment at foundation model scales (see the memory and inference costs of ASIF in \cref{ch:asif}, the solver timings in \cref{ch:learning_to_align}, and the training times in \cref{ch:lineas}). \gls{relrep} scales with anchor count and dimensionality, ResiDual requires repeated eigendecomposition across attention heads, and latent functional maps depend on expensive Laplacian spectra for high-resolution structures. These computational challenges are then likely to be engineering bottlenecks: approximate solvers, distributed implementations, and probabilistic correspondence could address them without fundamentally changing the framework.

\textbf{Theoretical foundations.} The core empirical observation---semantic overlap induces geometric overlap---lacks formal theory. Open questions include bounding alignment error in terms of data manifold geometry, deriving sufficient conditions for geometric convergence from optimization dynamics, establishing guarantees relating task complexity to correspondence density, and characterizing information preserved versus discarded by semantic coordinate transformations. Without such theory, the framework remains empirically validated across domains but theoretically incomplete. Addressing these gaps would transform correspondence from observed phenomenon into predictable tool with provable guarantees.

\section{Open Directions and Implications}

Understanding these constraints clarifies directions that would extend the framework's scope while addressing its current limitations.

\textbf{Hierarchical correspondence.} Hierarchical correspondence could organize anchors at multiple abstraction levels---coarse semantic categories for global structure, fine-grained concepts for local neighborhoods---mirroring the hierarchical organization evident in deep networks. High-level correspondence would establish broad alignment (``mammal'' versus ``vehicle''), while low-level correspondence refines distinctions (``sedan'' versus ``truck''), reducing burden on any single operator while improving compositional precision. This extends existing composition strategies without requiring new geometric machinery.

\textbf{Compositional pipelines.} Building on the mixed pipelines of \cref{ch:inverse-relative-projection}, graph-structured compositions---where intermediate models are stepping stones between source and target---would support multi-hop alignment between models that never co-occur in data, enabling transfer across ecosystems that remain otherwise isolated. The open question is whether composing multiple transformations compounds alignment error or whether intermediate semantic coordinates provide error-correcting structure.

\textbf{Streaming and federated settings.} Streaming environments require incremental correspondence updates as new data arrives, with no offline recomputation over static batches---a challenge for methods like latent functional maps that depend on global spectral structure. Federated learning scenarios would benefit from exchanging semantic coordinates: clients would send no raw gradients or model weights and could preserve privacy by transmitting only relative similarities to shared anchors. However, formal privacy guarantees against reconstruction attacks remain open, requiring theoretical analysis of what information semantic coordinates leak about source representations.

\textbf{Correspondence density trade-offs.} Optimal correspondence density varies across methods and tasks. Semantic alignment performance depends on anchor point count (\cref{ch:semantic_alignment}), and inverse relative projection trades the stability of the inverse (its condition number) against the number of anchors (\cref{ch:inverse-relative-projection}). Factors linked to latent space compatibility---intrinsic dimension, manifold curvature, local isotropy---could guide correspondence strategy but lack formal characterization. Future work investigating these trade-offs would provide practitioners with design principles for correspondence-based systems.

\textbf{Architectural and modal extensions.} Extensions to other architectures and modalities remain largely unexplored. ResiDual's formulation based solely on residual decomposition could apply across virtually any transformer architecture, with constraints at the layer level creating opportunities to enhance model inference, and the possibility of disabling specific spectral components opening fairness applications~\citep{basile2025residual}. HeadPursuit's head-level interventions could extend to selective rescaling by input position or modality, enabling targeted control in multimodal settings---disabling heads over image patches while preserving text understanding, for instance~\citep{headpursuit}. Latent functional maps could explore other domains and modalities, defining new ad-hoc constraints and regularizers through their versatile functional representation~\citep{fumero2024latent}. Their performance in fully unsupervised settings, while promising, requires further development, and extending partial mapping techniques from geometry processing would remove the full correspondence assumption noted above~\citep{fumero2024latent}.

\textbf{Standardization and tooling.} Proliferating alignment methods create systematic comparison challenges: each contribution introduces custom evaluation protocols and baselines that resist direct measurement. To address this fragmentation, we are developing \textsc{Latentis}\footnote{\url{https://github.com/Flegyas/latentis}}---a unified API consolidating correspondence-based operators (most methods in this thesis), spectral techniques (CCA, SVCCA, PWCCA), optimal transport (Gromov-Wasserstein), and neural stitching baselines with standardized benchmarks and pretrained model interfaces across vision and language architectures. The library served as computational infrastructure for one research group at LogML 2025 Summer School,\footnote{\url{https://www.logml.ai}} which suggests that standardized interfaces can lower barriers for researchers entering the field. The codebase remains under active development (public beta); we believe that, to be significant, new contributions should build on comparable foundations, reuse shared measurement apparatus, and be reproducible.

\textbf{Modularity and efficiency.} The framework's capabilities create both opportunities and risks extending beyond technical performance metrics. By separating encoders, correspondence operators, and decoders, the framework reduces the need for end-to-end retraining whenever task requirements shift or new models become available. This decomposition can lower the energy cost of adapting models to new tasks---reusing pretrained components that would otherwise be discarded, as in the policy stitching of \cref{ch:domain-applications}---and enables incremental development where individual modules improve independently without breaking downstream dependencies. We expect the representation nexus this creates, where modular latent spaces connect through correspondence, to support collaboration across organizations without sharing training data or raw model parameters.

\textbf{Safety and governance.} These capabilities introduce risks alongside benefits. The ability to translate or steer representations can bypass safety systems (redirecting aligned models toward harmful outputs) or repurpose models without consent (extracting proprietary capabilities through correspondence-based transfer). HeadPursuit and LinEAS show that the same tools can reduce toxic generation (\cref{ch:headpursuit,ch:lineas}), but broader governance and access control remain necessary to prevent malicious alignment from undermining deployment safeguards or intellectual property protections. As \citet{headpursuit} note, the ability to manipulate specific aspects of model behavior aids transparency and alignment but may also conceal or amplify content in ethically concerning ways. Downstream users should carefully evaluate applications, especially in sensitive domains.

Neural network training has progressed by scaling data, compute, and parameters---an engineering approach that treats models as monolithic artifacts optimized in isolation. This thesis demonstrates an alternative: independently trained models compose when correspondence reveals their shared geometric structure, creating a representation nexus where latent spaces meet. Our contributions establish empirical foundations across vision, language, reinforcement learning, and neuroscience, unified by a central principle: semantic overlap induces geometric overlap, making zero-shot transformation possible. The open questions---formalizing this semantic-geometric connection, establishing theoretical guarantees, scaling to foundation models---define a research program extending beyond these methods. Correspondence is a philosopher's stone that extracts the intrinsic geometry of representations beneath arbitrary parameterizations. It leaves the representations themselves unchanged.

\backmatter

\printbibliography[heading=bibintoc]

\cleardoublepage
\printglossary[type=abbreviations,title=List of Abbreviations]
\cleardoublepage
\printglossary[type=concepts,title=List of Concepts]
\cleardoublepage
\printglossary[type=symbols,title=List of Symbols]

\end{document}